%% file: main.tex
\pdfoutput=1
\documentclass{physicalai-survey}
\usepackage{physicalai}

\newcommand{\projectpageurl}{https://physical2rsi.github.io}
\newcommand{\githubrepourl}{https://github.com/Physical2RSI/Awesome-Physical2RSI}

\definecolor{PAIResInk}{HTML}{14233A}
\definecolor{PAIResLink}{HTML}{00A6B2}
\newcommand{\resourcerow}[3]{%
  \raisebox{-.2mm}{\makebox[3.4mm][c]{\color{PAIResInk}#1}} &
  {\sffamily\small\bfseries\color{PAIResInk}#2} &
  {\footnotesize\rmfamily\color{PAIResLink}\expandafter\url\expandafter{#3}}\\}

\title{\texorpdfstring{Toward Recursive Self-Improvement in Physical AI}{Toward Recursive Self-Improvement in Physical AI}}

\newcommand{\affillogo}[3][]{\raisebox{-.5\height}{\includegraphics[height=#2,keepaspectratio,#1]{#3}}}
\titlelogos{\makebox[\linewidth]{\makebox[.9\linewidth]{%
  \affillogo[trim=49.2bp 50.64bp 49.2bp 47.52bp,clip]{11mm}{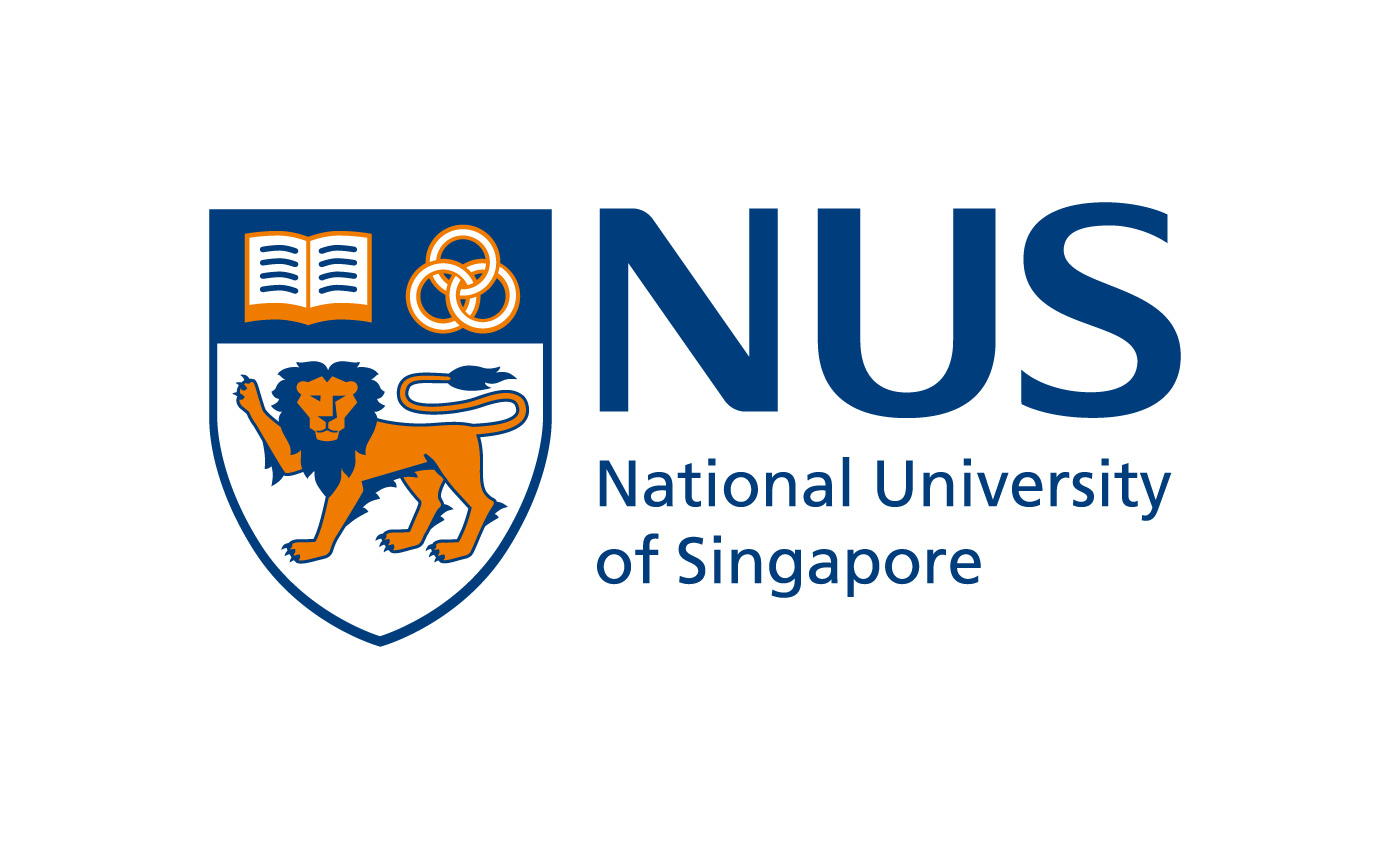}\hfill
  \affillogo{11.5mm}{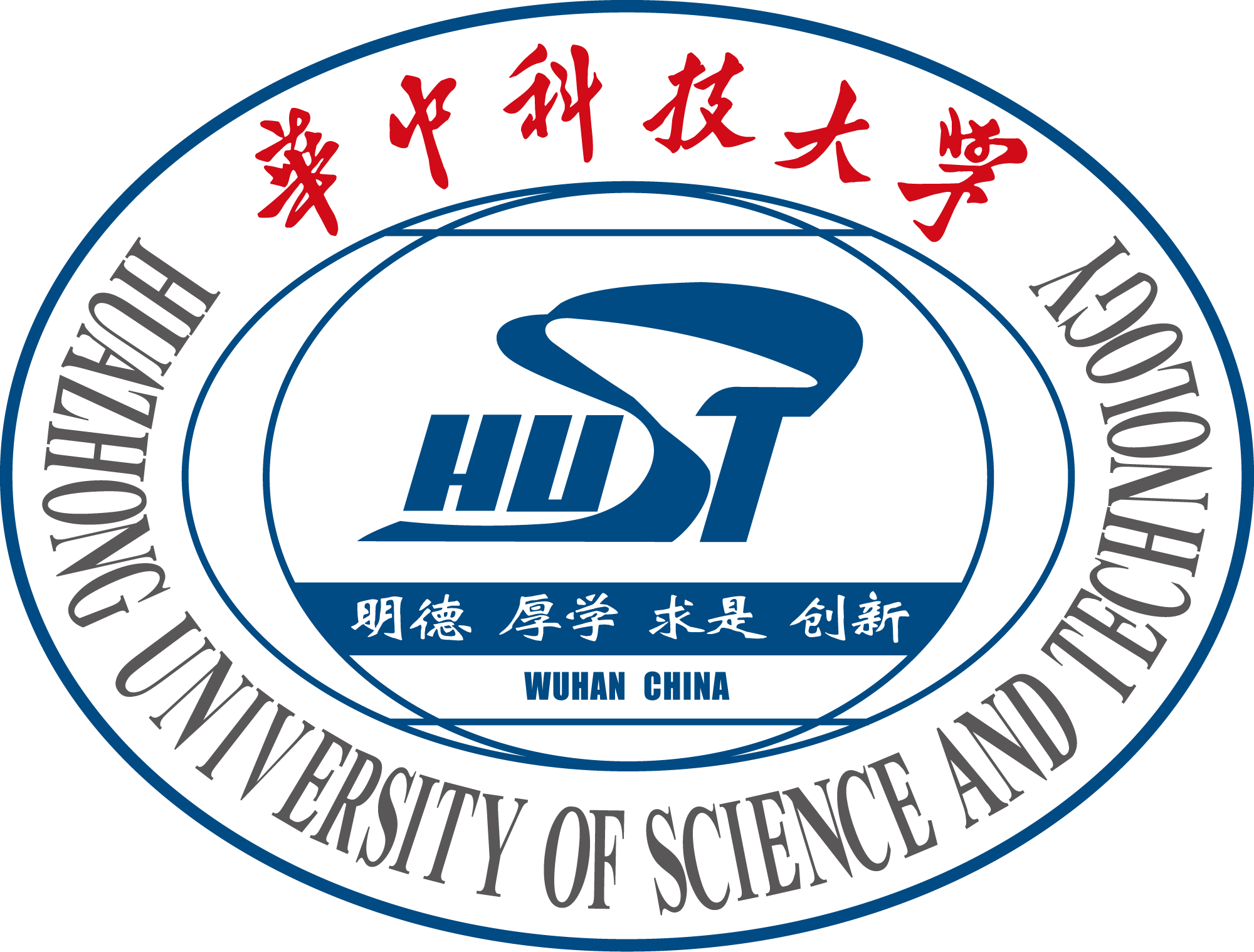}\hfill
  \affillogo[trim=38.4bp 31.44bp 36.48bp 43.68bp,clip]{11.5mm}{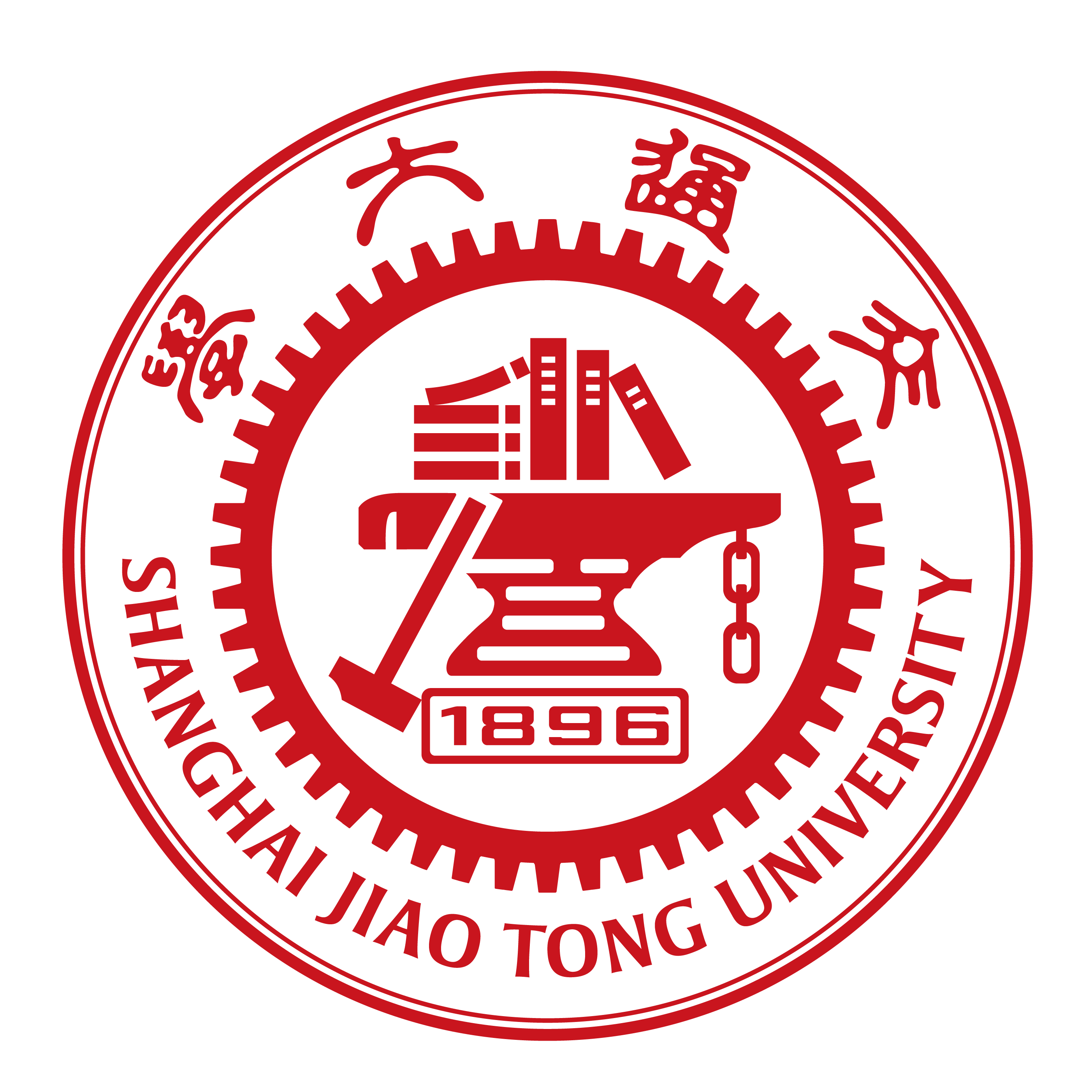}\hfill
  \affillogo{10.2mm}{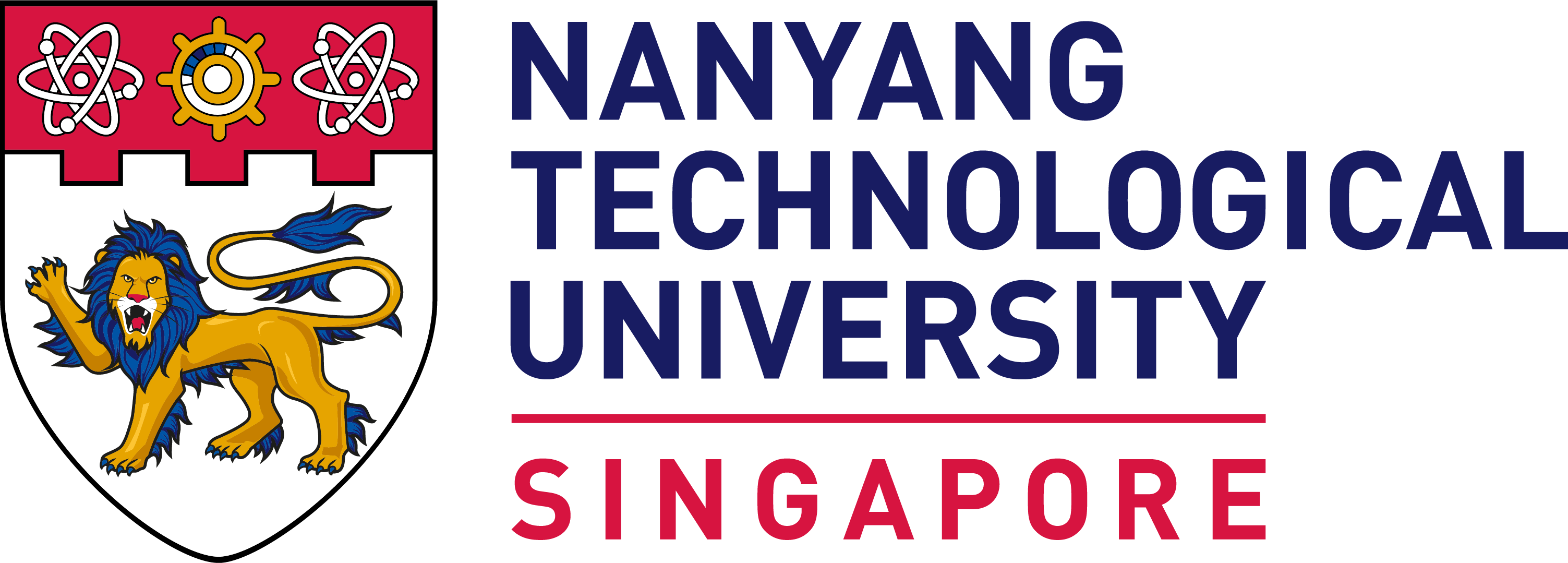}\hfill
  \affillogo{10mm}{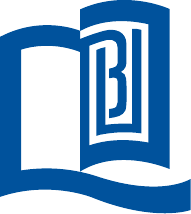}}}}

\newcommand{\physicalaiAuthors}{%
  {\fontfamily{ppl}\bfseries\fontsize{10.5}{13}\selectfont
    \mbox{Meng Luo\textsuperscript{1}}\enspace\allowbreak
    \mbox{Zechen Bai\textsuperscript{1}}\enspace\allowbreak
    \mbox{Shanqing Xu\textsuperscript{2}}\enspace\allowbreak
    \mbox{Yanlin Li\textsuperscript{1}}\enspace\allowbreak
    \mbox{Kaixin Li\textsuperscript{1}}\enspace\allowbreak
    \mbox{Mengchen Qian\textsuperscript{2}}\enspace\allowbreak
    \mbox{Weifeng Lu\textsuperscript{3}}\enspace\allowbreak
    \mbox{Ran Zhang\textsuperscript{4}}\enspace\allowbreak
    \mbox{Yingtian Zou\textsuperscript{3}}\enspace\allowbreak
    \mbox{Ziyang Luo\textsuperscript{5}}\enspace\allowbreak
    \mbox{Wei Chen\textsuperscript{2}}\enspace\allowbreak
    \mbox{Hongzhan Lin\textsuperscript{1*}}\par}
  \vspace{1.5mm}%
  {\fontfamily{ppl}\bfseries\itshape\fontsize{9}{11}\selectfont\color{PAIMuted}
    \mbox{\textsuperscript{1}National University of Singapore}\quad\allowbreak
    \mbox{\textsuperscript{2}Huazhong University of Science and Technology}\quad\allowbreak
    \mbox{\textsuperscript{3}Shanghai Jiao Tong University}\quad\allowbreak
    \mbox{\textsuperscript{4}Nanyang Technological University}\quad\allowbreak
    \mbox{\textsuperscript{5}Hong Kong Baptist University}\par}}
\author{\physicalaiAuthors}

\begin{document}
\newgeometry{top=11mm,bottom=18mm,left=23mm,right=23mm,headheight=16pt,headsep=7mm}
\enlargethispage{8mm} % Reserve room for the corresponding-author footnote.
\begin{physicalaifrontmatter}
{\fontsize{10.5}{13.4}\selectfont\justifying\noindent
\emph{Recursive self-improvement} (RSI) describes a system that improves not only its behavior but also the process by which it improves. Recent digital agents test this idea by rewriting their own code and checking each version with executable tests. Bringing these mechanisms into the physical world requires learning from partial observations through costly trials, while accounting for changing operating conditions and potentially irreversible consequences. We position RSI within robot learning and trace pathways from improving robot capabilities to improving the processes that generate them. We examine how physical experience can strengthen these improvement mechanisms and when the resulting gains can support further self-improvement of the mechanisms themselves. Gains carry forward only when feedback addresses remaining learning needs, the receiving component can use the change, its value survives changes in tasks or hardware, and continued practice remains safe and affordable. Existing studies, concentrated in manipulation, support behavioral improvement, benefits to later learning, and procedure improvement. Some systems also change the improver's knowledge or tools, but the physical studies reviewed here have yet to isolate gains attributable to recursion. We synthesize these findings into evaluation principles and research directions for machines that, trial by trial, safely improve the way they improve.\par}
\vspace{4mm}
\begingroup
\setlength{\tabcolsep}{1.2mm}\setlength{\extrarowheight}{0pt}\hypersetup{urlcolor=PAIResLink}
\renewcommand{\arraystretch}{1.0}
{\centering\begin{tabular}{@{}c l l@{}}
  \resourcerow{\faGlobe}{Project}{\projectpageurl}
  \resourcerow{\faGithub}{GitHub}{\githubrepourl}
  \raisebox{-.2mm}{\makebox[3.4mm][c]{\color{PAIResInk}\faEnvelope}} &
  {\sffamily\small\bfseries\color{PAIResInk}Contact} &
  {\footnotesize\rmfamily\color{PAIResLink}\href{mailto:mluo@u.nus.edu}{\nolinkurl{mluo@u.nus.edu}},\enspace \href{mailto:danielhzlin@nus.edu.sg}{\nolinkurl{danielhzlin@nus.edu.sg}}}\\
\end{tabular}\par}
\endgroup
\end{physicalaifrontmatter}
\begingroup
\renewcommand{\thefootnote}{\fnsymbol{footnote}}
\footnotetext[1]{Corresponding author.}
\endgroup
\vspace{4.5mm}
\begingroup
\captionsetup{skip=1.5pt}
\input{figures/teaser_scene}
\endgroup
\restoregeometry
\paiContents

\input{sections/01_introduction}
\input{sections/02_framework}
\input{sections/07_evidence}
\input{sections/03_policy_improvement}
\input{sections/04_model_mediated}
\input{sections/05_modular_improvement}
\input{sections/06_improvement_process}
\input{sections/06b_reviser}
\input{sections/06c_closing_loop}
\input{sections/08_dependencies}
\input{sections/09_continuation}
\input{sections/11_research_agenda}
\input{sections/12_conclusion}

\FloatBarrier\clearpage\pagestyle{physicalairefs}
\bibliographystyle{plainnat}
\bibliography{references}
\clearpage\pagestyle{physicalai}\appendix
\addtocontents{toc}{\protect\setcounter{tocdepth}{1}}
\input{sections/appendix_landscape}
\input{sections/appendix_reported_results}
\input{sections/appendix_evaluation}
\input{sections/appendix_gallery}
\end{document}

%% file: figures/teaser_scene.tex
\begingroup
\centering
\makebox[\linewidth][c]{\resizebox{1.06\linewidth}{!}{\begin{tikzpicture}[art fonts,x=1cm,y=1cm,line cap=round,line join=round,
  every node/.append style={execute at begin node={\hyphenpenalty=10000\exhyphenpenalty=10000}},
  pmark/.style={circle,fill=#1!85!black,fill opacity=.78,text opacity=1,draw=white,draw opacity=.9,line width=.6pt,text=white,inner sep=0pt,minimum size=.32cm,
    font=\sffamily\bfseries\fontsize{6.2}{6.2}\selectfont,drop shadow={opacity=.22,shadow xshift=.4pt,shadow yshift=-.5pt}},
  lmark/.style={circle,fill=#1!85!black,draw=white,line width=.5pt,text=white,inner sep=0pt,minimum size=.34cm,
    font=\sffamily\bfseries\fontsize{6.6}{6.6}\selectfont},
  ghead/.style={anchor=west,inner sep=0pt,text=#1!85!black,font=\sffamily\bfseries\fontsize{7.8}{9}\selectfont,execute at end node={\vphantom{(g}}},
  gtext/.style={anchor=north west,inner sep=0pt,align=left,text=ArtSlate,font=\sffamily\fontsize{7.0}{8.2}\selectfont}]
\path[use as bounding box] (0,-2.05) rectangle (16.08,6.90);
% The relation, read left to right.
\node[inner sep=0pt,text=PAIInk,font=\sffamily\fontsize{8.8}{11}\selectfont] at (8.04,6.62)
  {\textbf{\textcolor{ArtPurple!85!black}{Digital RSI} supplies the mechanisms}\enspace{\color{ArtSlate!70}\faIcon{long-arrow-alt-right}}\enspace\textbf{\textcolor{ArtAmber!80!black}{Physical RSI}} tests how they use physical experience and hold under physical constraints};
% The painting (1902 x 737 px): digital sky on the left, a doorway in the middle, the robot's desert on the right.
\begin{scope}
  \clip[rounded corners=2.5mm] (0,0) rectangle (16.08,6.231);
  \node[anchor=south west,inner sep=0pt] at (0,0) {\includegraphics[width=16.08cm]{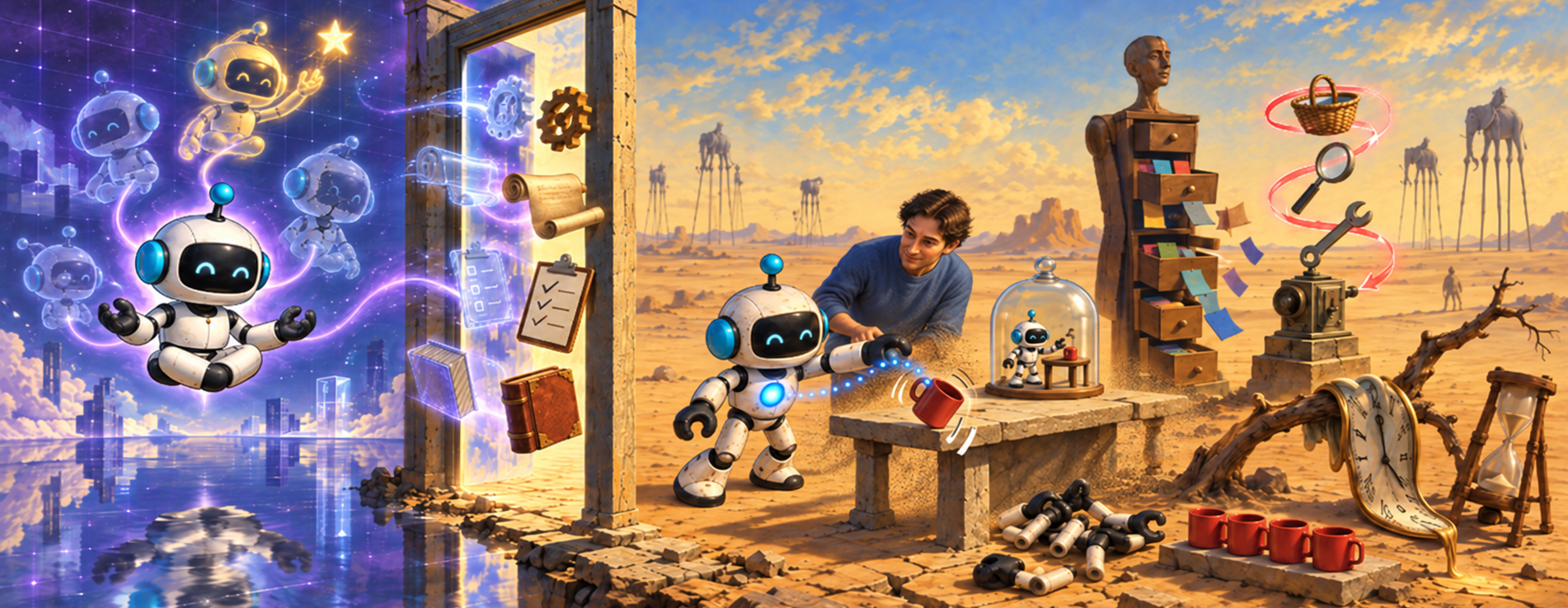}};
\end{scope}
\draw[ArtSlate!35,line width=.5pt,rounded corners=2.5mm] (0,0) rectangle (16.08,6.231);
% Small keys placed beside the objects they name.
\node[pmark=ArtPurple] at (.59,5.47) {A};
\node[pmark=ArtPurple] at (2.11,1.92) {B};
\node[pmark=ArtSlate] at (5.61,5.64) {C};
\node[pmark=ArtBlue] at (9.67,2.70) {1};
\node[pmark=ArtTeal] at (11.00,3.10) {2};
\node[pmark=ArtAmber] at (11.37,4.12) {3};
\node[pmark=ArtRed] at (13.10,4.12) {4};
\node[pmark=ArtSlate,font=\fontsize{5.4}{5.4}\selectfont] at (10.10,4.12) {\faIcon{user-check}};
\node[pmark=ArtSlate,font=\fontsize{5.4}{5.4}\selectfont] at (14.80,1.25) {\faIcon{hourglass-half}};
\node[pmark=ArtSlate,font=\fontsize{5.4}{5.4}\selectfont] at (9.72,.38) {\faIcon{wrench}};
% Legend: each group sits under the part of the painting it describes.
\draw[ArtPurple!85!black,line width=1.6pt] (.10,-.14) -- (4.02,-.14);
\shade[left color=ArtPurple!85!black,right color=ArtSage!85!black] (4.20,-.168) rectangle (6.52,-.112);
\draw[ArtSage!85!black,line width=1.6pt] (6.70,-.14) -- (12.82,-.14);
\draw[ArtSlate!85!black,line width=1.6pt] (13.00,-.14) -- (15.98,-.14);
% Digital RSI.
\node[ghead=ArtPurple] at (.10,-.44) {Digital RSI\,\textperiodcentered\,mechanisms};
\node[lmark=ArtPurple] at (.27,-.825) {A};
\node[gtext] at (.52,-.72) {self-edits scored by tests:\\cheap, parallel, reversible};
\node[lmark=ArtPurple] at (.27,-1.545) {B};
\node[gtext] at (.52,-1.44) {the improver rewrites itself};
% Carried over.
\node[ghead=ArtSlate] at (4.20,-.44) {Carried over};
\node[lmark=ArtSlate] at (4.37,-.825) {C};
\node[gtext] at (4.62,-.72) {tests, code,\\archives, loops};
% Physical side: where a change lands (1--4) and the improver (5).
\node[ghead=ArtSage] at (6.70,-.44) {Self-improvement\,\textperiodcentered\,where a change lands};
\node[lmark=ArtBlue] at (6.87,-.825) {1};
\node[gtext] at (7.12,-.72) {\textbf{\textcolor{ArtBlue!85!black}{Policy}}};
\node[lmark=ArtTeal] at (6.87,-1.185) {2};
\node[gtext] at (7.12,-1.08) {\textbf{\textcolor{ArtTeal!85!black}{World model}}};
\node[lmark=ArtAmber] at (9.92,-.825) {3};
\node[gtext] at (10.17,-.72) {\textbf{\textcolor{ArtAmber!85!black}{External knowledge}}};
\node[lmark=ArtRed] at (9.92,-1.185) {4};
\node[gtext] at (10.17,-1.08) {\textbf{\textcolor{ArtRed!85!black}{Learning procedure}}};
\node[lmark=ArtRed,font=\fontsize{6}{6}\selectfont] at (6.87,-1.545) {\faIcon{sync-alt}};
\node[gtext] at (7.12,-1.44) {\textbf{\textcolor{ArtRed!85!black}{Improver}}: revises the procedure, then itself};
% Physical constraints.
\node[ghead=ArtSlate] at (13.00,-.44) {Physical constraints};
\node[lmark=ArtSlate,font=\fontsize{6}{6}\selectfont] at (13.17,-.825) {\faIcon{user-check}};
\node[gtext] at (13.42,-.72) {outcome checks can err};
\node[lmark=ArtSlate,font=\fontsize{6}{6}\selectfont] at (13.17,-1.185) {\faIcon{hourglass-half}};
\node[gtext] at (13.42,-1.08) {trials cost time, resets};
\node[lmark=ArtSlate,font=\fontsize{6}{6}\selectfont] at (13.17,-1.545) {\faIcon{wrench}};
\node[gtext] at (13.42,-1.44) {wear cannot be undone};
\end{tikzpicture}}}%
\begingroup\captionsetup{hypcap=false}
\captionof{figure}{\textbf{From digital to physical recursive self-improvement.} Numbers mark where experience leaves a change: policy~\paiTeaserMark{ArtBlue}{1}, world model~\paiTeaserMark{ArtTeal}{2}, and external knowledge~\paiTeaserMark{ArtAmber}{3} improve what the robot does; the learning procedure~\paiTeaserMark{ArtRed}{4} changes how it learns; and the improver~\paiTeaserMark{ArtRed}{\faIcon{sync-alt}}, drawn as the loop around the procedure, revises it. RSI further requires that repair knowledge improve the improver and that the improver, once improved, help produce its own next improvement.}
\label{fig:teaser}
\endgroup\par\endgroup

%% file: sections/01_introduction.tex
\sectionaccent{PAIInk}
\section{Introduction}
\label{sec:introduction}

\emph{Recursive self-improvement} (RSI) describes a system that improves the process by which it improves, not only its behavior. The idea goes back to I.~J.~Good's speculation about machines that design better machines~\citep{good1966speculations} and to self-referential learners such as the G\"odel machine~\citep{schmidhuber2003godel}, and recent digital agents have begun to approach it~\citep{arxiv260707663}. STOP~\citep{zelikman2024stop} applies a program that improves code to its own source, and the Darwin G\"odel Machine~\citep{zhang2025dgm} lets improved coding agents modify themselves; both judge each candidate with executable tests. Yet a system can get better at a task, and a fixed procedure can keep producing better versions of it, without anything recursive happening. RSI requires two further links (Section~\ref{sec:key-terms}): what the system retains from earlier improvements must make its next revision better, and the \emph{improver}, the component that revises the learning procedure, must in turn use what it has gained to help produce its own next improvement.

\emph{Physical} RSI asks the same question of systems that learn by acting in the physical world, such as robots. Digital RSI supplies the mechanisms (language models that write reward code or controllers, archives of earlier variants, and automated experiment loops); physical RSI tests how these mechanisms use physical experience and whether they hold under physical constraints (Figure~\ref{fig:teaser}). The physical setting changes the experimental conditions in four ways (Table~\ref{tab:physical-digital}). First, physical outcome checks depend on sensors and task-specific verifiers, people, or learned detectors whose reliability must be established; even an automated grasp-success check can miss what it does not sense, and an error in that judgment can be reinforced and then executed on hardware. Second, every candidate costs robot time, resets, supervision, and wear, so a revision must pay off within few trials. Third, practice changes its own conditions: objects move, grippers wear, calibration drifts, and rolling back software does not repair a damaged object. Fourth, what can be improved extends beyond prompts and code to controllers, dynamics models, and the interface to a particular body. Each difference bears on the central question of RSI, whether what one round of revision teaches the improver makes the next revision better.

\input{tables/physical_digital}

Consider a robot arm that keeps missing the same grasp. The failure can have several causes, and each calls for a change kept in a different place; these places organize Sections~\ref{sec:base-loop} and~\ref{sec:emerging-directions}. The robot's \emph{policy}, the learned controller that turns observations into motor commands, may never produce the recovery motion; an expert correction can teach it, as RECAP~\citep{arxiv251114759} learns from deployment experience and corrections (Section~\ref{sec:policy-improvement}). The robot may mistake a light touch for a stable hold because its model of contact is wrong; a \emph{world model} refitted to contact data can correct the prediction and give the policy a place to practice, as in PlayWorld~\citep{arxiv260309030} (Section~\ref{sec:model-mediated}). The robot may have recovered from this failure before without keeping what worked; \emph{external knowledge}, such as a saved recovery or program, can be called the next time, as Zetta~\citep{arxiv260816590} grows code-based critics and recoveries around a frozen policy in simulation (Section~\ref{sec:modular-improvement}). Finally, the robot may keep choosing the wrong kind of repair; then the \emph{learning procedure} must change: which attempts to collect, how to judge them, and how to turn them into updates (Section~\ref{sec:emerging-directions}). Eureka~\citep{ma2024eureka} rewrites reward code from simulated training results, and ENPIRE~\citep{arxiv260619980} lets coding agents modify the training infrastructure from robot experiments. The first three routes improve what the robot does; the fourth improves how it learns, and it is where physical RSI would begin. Recursion would further require that knowledge of why earlier repairs worked, which we call \emph{repair knowledge}, improves how the next procedure revision is made. Figure~\ref{fig:two-routes} contrasts this route with applying a fixed training recipe to new experience.

\input{figures/teaser}

Our central argument is that physical experience can improve task competence, the learning procedure, and the improver that revises this procedure, and that four recurring conditions, each sharpened by the physical setting, decide whether such gains carry forward. Each can fail: the new signal may not address what the learner still lacks, because sensing shows only part of a failure; the receiving component may be unable to absorb it on its body; its value may not survive later changes in behavior or hardware; or further practice may become unsafe or unaffordable. For the grasping robot, this could mean collecting more of the grasps it already makes well, refining a contact model that its controller cannot use, relying on a fix that stops working once the gripper is replaced, or practicing until the gripper wears out. A change can improve execution and still fail to help later learning: added autonomous data often yield modest gains that saturate and can even regress~\citep{arxiv241101813}.

\paragraph{From learning actions to improving how learning happens.}
Several research traditions supply pieces of this progression. Self-supervised grasping showed that physical outcomes alone can supervise behavior at scale, first by learning from tens of thousands of real grasp attempts~\citep{pinto2016supersizing} and then through large-scale real-robot reinforcement learning (RL) with QT-Opt~\citep{kalashnikov2018qtopt}. Interactive imitation learning, as in DAgger~\citep{ross2011dagger} and HG-DAgger~\citep{kelly2019hgdagger}, showed how an expert can correct the learner in the states that the learner itself reaches. Lifelong robot learning~\citep{thrun1995lifelong} asks how knowledge gained on earlier tasks can reduce the real-world experimentation needed for later ones; model-based RL, from Dyna~\citep{sutton1990dyna} and PILCO~\citep{deisenroth2011pilco} to Dreamer~\citep{hafner2020dreamer}, asks how recorded transitions can support practice inside a learned model; and skill-based approaches, such as options~\citep{sutton1999options} and the shared RoboEarth knowledge base~\citep{waibel2011roboearth}, ask how an extended behavior can be stored and called again. Each tradition widened what one learning episode can pass on to the next.

Recent generalist and foundation-model-based systems widen both what can change and what role the change plays. RoboCat~\citep{arxiv230611706} turns a policy adapted to a new task into a source of training data for the next generalist, and SILVR~\citep{arxiv250606658} revises a video-based planner using its own executions. In RoboCat, Eureka (in simulation), and ENPIRE, experience reaches the machinery that produces future updates (a data collector, a reward function, the training infrastructure) as well as the policy that receives them. This brings the physical problem close to the long-standing aspiration of learning to learn~\citep{thrun1998learning}, and several reviewed systems already contain the components that a demonstration of recursion would need. The open question is how to realize that aspiration with limited trials, imperfect sensing, and a body that changes over time.

\paragraph{Scope and related surveys.}
Our scope is the improvement of policies, world models, memories, skills, and the software procedures that update or use them, whenever their evidence or evaluation is connected to physical action. We discuss hardware change when it affects compatibility, maintenance, or continued experimentation. Autonomous morphology design and physical self-reconstruction are adjacent topics outside our scope.

Recent surveys cover neighboring ground (Table~\ref{tab:related-surveys}). Some focus on agents and embodiment: recursive self-improvement in AI systems~\citep{arxiv260707663}, self-evolving agents~\citep{arxiv250721046}, weights versus executable skills as stores of improvement~\citep{arxiv260801851}, self-evolving embodied AI~\citep{arxiv260204411} and humanoids~\citep{arxiv260913236}, and levels of physical recursive self-improvement~\citep{arxiv261018842}. Others analyze world models for robot learning~\citep{arxiv260500080} and manipulation~\citep{arxiv260600113}, joint world--action models~\citep{arxiv260916074}, actionable simulators~\citep{arxiv260115533}, verifiers for robot policies~\citep{arxiv260909250}, and physical AI broadly~\citep{arxiv251004978}. Our organizing question is different: after a change has been acquired, \emph{who uses it next, and what does that use improve?} Component names alone cannot answer it, because two data collectors with similar task success can yield training data of very different value; a more accurate predictor can still produce less useful training data for a policy; and the same kind of retrieved memory can change either how a program is repaired or how an outcome is judged.

\input{tables/related_surveys}

The review makes three connected contributions. First, it positions RSI within robot learning by separating method paradigms and problem settings (from reinforcement learning and meta-learning to AutoML) from a ladder of claims (behavioral improvement, later-learning benefit, procedure improvement, and the two links of recursion), each defined by the comparison that would establish it. It organizes physical self-improvement along that ladder according to where experience leaves its change and which decision of the learning procedure it revises, showing that data volume and component accuracy are unreliable proxies for progress. Second, it separates mechanisms that change an improver's knowledge or tools from evidence that the improver has improved: existing studies support the first three claims, while recursive gains still need dedicated verification. It reviews digital precedents and physical boundary cases and specifies a two-round experiment that could isolate the recursive relation, together with its controls and an estimate of its hardware cost; Table~\ref{tab:claim-status} records where the reviewed studies stand on each claim. Third, it names the four breakdowns above as recurring conditions (\emph{informative feedback}, \emph{usability}, \emph{robustness to distribution shift}, and \emph{continued practice}) and shows how each fails and which capability would restore it. Four research questions organize the review.

\begin{pairqs}
\pairesearchquestion{RQ1}{What limits a robot's next improvement?}{Is useful experience missing, or is progress limited by supervision, optimization, action generation, interfaces, or retention?}
\pairesearchquestion{RQ2}{Which changes address that limitation?}{How does experience change behavior, reusable models or skills, and the collection, supervision, curriculum, or update procedures used in later learning?}
\pairesearchquestion{RQ3}{When can improvements build on one another?}{When do improvements carry forward into better learning procedures and improvers, and what evidence would establish recursive self-improvement?}
\pairesearchquestion{RQ4}{What makes self-improvement different in the physical world?}{How do sensing, embodiment, reset, safety, cost, and hardware changes constrain the process?}
\end{pairqs}
 Section~\ref{sec:framework} lays the foundations: a primer for newcomers (Section~\ref{sec:primer}), where improvement is kept, the ladder from self-improvement to recursion, the position of RSI relative to neighboring learning paradigms (Section~\ref{sec:positioning}), and the benchmarks and metrics used in the reviewed studies (Section~\ref{sec:benchmarks}); Figure~\ref{fig:overview} maps the review, and Appendix~\ref{app:glossary} collects the terms. Section~\ref{sec:mechanisms} climbs the ladder, from improving the learner (Section~\ref{sec:base-loop}) and the learning procedure (Section~\ref{sec:emerging-directions}) to improving the improver (Section~\ref{sec:reviser}) and closing the recursive loop (Section~\ref{sec:closing-loop}). Section~\ref{sec:challenges} turns to what current evaluation cannot yet measure, why gains stop compounding, how practice can stay safe, and which problems remain open, including the two-round test that recursion would require.

%% file: tables/physical_digital.tex
\begin{table}[!htbp]
\centering
\caption{Experimental conditions of digital and physical self-improvement. Digital examples include STOP, AlphaEvolve, and the Darwin G\"odel Machine.}
\label{tab:physical-digital}
\begingroup\small
\setlength{\tabcolsep}{5pt}\renewcommand{\arraystretch}{1.18}\arrayrulecolor{PAIRule}
\begin{tabularx}{\linewidth}{@{}L{3.2cm}YY@{}}
\paitoprule
\pahead{Issue} & \pahead{\paiBadge{ArtPurple}{\paiGlyphCode}\enspace Digital systems} & \pahead{\envhw\enspace Physical systems} \\
\paimidrule
Outcome checks & Executable tests, scores, model judges & Sensors, human checks, learned detectors \\\paihair
Cost per candidate & Compute, tokens, sandbox runs & Plus robot time, resets, wear, oversight \\\paihair
Reversibility & Snapshots often restorable & Software restorable; damage may persist \\\paihair
Nonstationarity & Tasks, tools, APIs, test sets change & Plus calibration, object, hardware drift \\\paihair
Reset & Often a fresh process or seed & May need a person or reset policy \\\paihair
Mutable objects & Prompts, code, rewards, workflows & Plus controllers, dynamics, body interfaces \\\paihair
Shared-error risk & Candidate may overfit its evaluator & Scoring errors can reinforce faulty actions \\
\paibottomrule
\end{tabularx}\endgroup
\end{table}

%% file: figures/teaser.tex
\begin{figure}[!tp]
\centering
\begin{tikzpicture}[art fonts,x=1cm,y=1cm,line cap=round,line join=round,
  every node/.append style={execute at begin node={\hyphenpenalty=10000\exhyphenpenalty=10000}}]
\path[use as bounding box] (0,-7.60) rectangle (16.08,1.98);
% Left panel: the behavior loop as a closed cycle of four step cards around its retained-change hub.
\artpanel[1]{.02}{-6.80}{8.20}{1.95}{ArtTeal}{Improving what the robot does}
\node[anchor=base east,inner sep=0pt,text=ArtTeal!85!black,font=\artserif\fontsize{8.4}{9.8}\selectfont] at (8.00,1.665) {\faIcon{lock}\enspace fixed recipe\vphantom{Ag}};
% Left panel: a 2x2 cycle whose cards fill the panel; teal beads on the gaps give the direction, and the retained-change hub sits on the crossing.
\tikzset{acell/.style={art card=ArtTeal,execute at end node={\vphantom{g}},text width=3.27cm,minimum width=3.55cm,minimum height=3.525cm},
  bcell/.style={art card=ArtAmber,text width=3.86cm,minimum width=4.10cm,minimum height=2.22cm,execute at end node={\vphantom{g}}}}
\node[acell] (a1) at (1.985,-.4725) {\includegraphics[height=2.15cm]{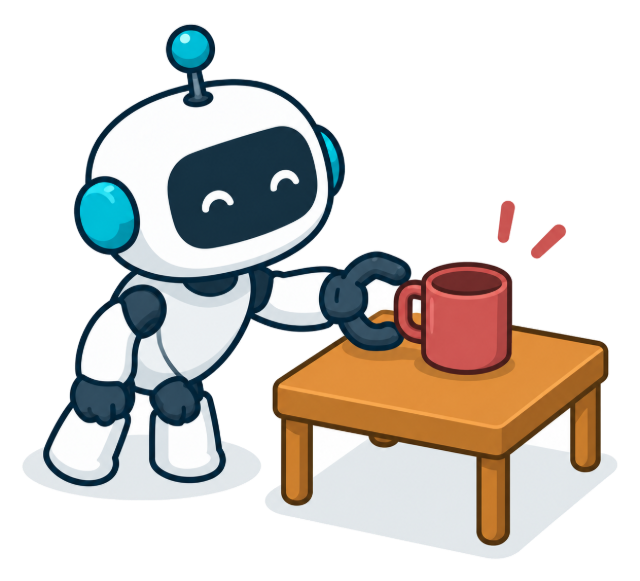}\\[1pt]\textbf{Act and practice}\\Try the task or\\a recovery motion};
\node[acell] (a2) at (6.235,-.4725) {\includegraphics[height=2.15cm]{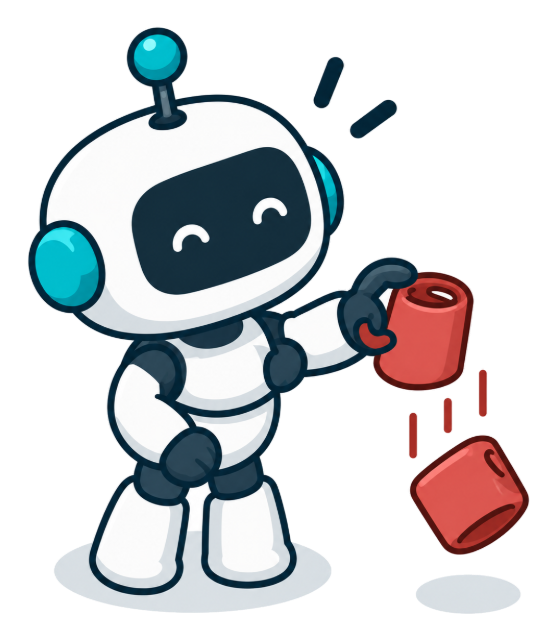}\\[1pt]\textbf{Physical outcome}\\Contact, slip, or\\progress on task};
\node[acell] (a3) at (6.235,-4.7975) {\includegraphics[height=2.15cm]{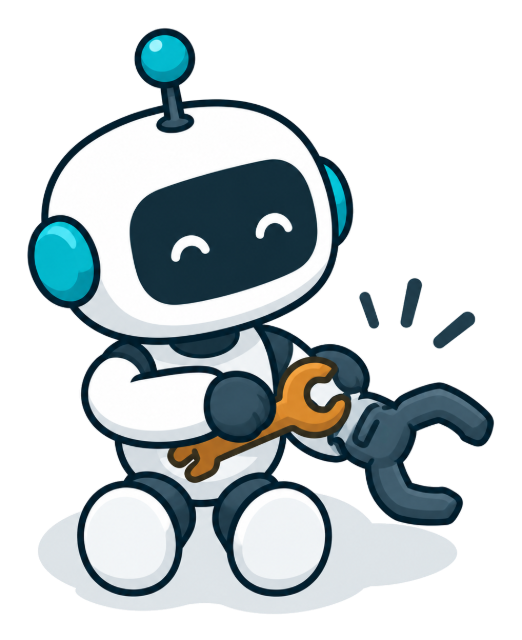}\\[1pt]\textbf{Learn or repair}\\Train, retrieve, or\\edit software};
\node[acell] (a4) at (1.985,-4.7975) {\includegraphics[height=2.15cm]{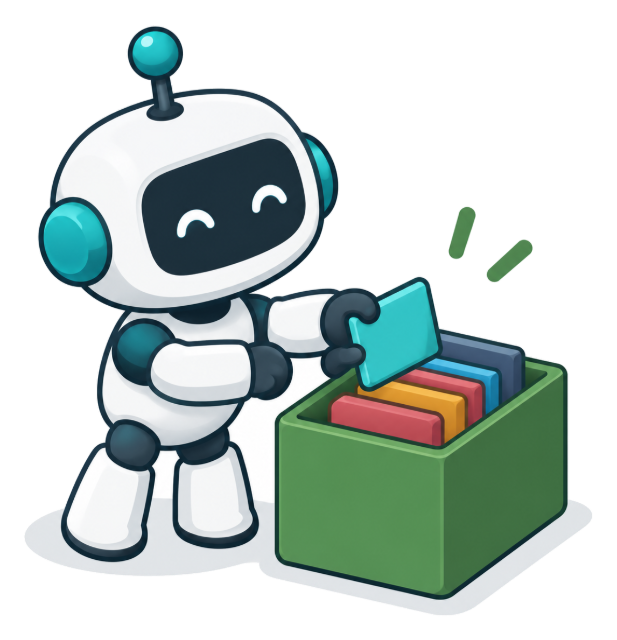}\\[1pt]\textbf{Retain the change}\\Policy, world model,\\or external knowledge};
\node[art bead=ArtTeal,minimum size=.44cm,font=\fontsize{6.6}{7}\selectfont] at (4.11,-.4725) {\faIcon{arrow-right}};
\node[art bead=ArtTeal,minimum size=.44cm,font=\fontsize{6.6}{7}\selectfont] at (6.235,-2.635) {\faIcon{arrow-down}};
\node[art bead=ArtTeal,minimum size=.44cm,font=\fontsize{6.6}{7}\selectfont] at (4.11,-4.7975) {\faIcon{arrow-left}};
\node[art bead=ArtTeal,minimum size=.44cm,font=\fontsize{6.6}{7}\selectfont] at (1.985,-2.635) {\faIcon{arrow-up}};
\node[circle,fill=white,draw=ArtTeal!60,line width=.8pt,minimum size=1.8cm,inner sep=0pt,text width=1.32cm,align=center,text=ArtTeal!85!black,font=\artserif\fontsize{7.6}{8.8}\selectfont,drop shadow={opacity=.18,shadow xshift=.6pt,shadow yshift=-.8pt}] at (4.11,-2.635) {Retained changes guide later practice};
% Right panel: the procedure route as three image-beside-text cards; the question card fills the right column.
\artpanel[2]{8.40}{-6.80}{16.06}{1.95}{ArtAmber}{Improving how it learns}
\node[bcell,anchor=north] (b1) at (10.69,1.29) {\includegraphics[height=1.22cm,width=3.70cm,keepaspectratio]{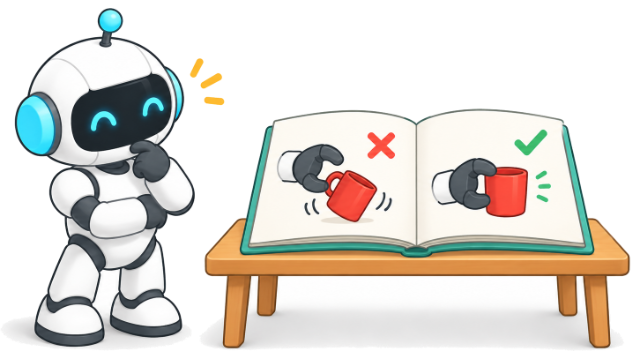}\\[1pt]\textbf{Retain repair knowledge}\\Why a repair worked, and when};
\node[bcell,anchor=south] (b3) at (10.69,-6.56) {\includegraphics[height=1.22cm,width=3.70cm,keepaspectratio]{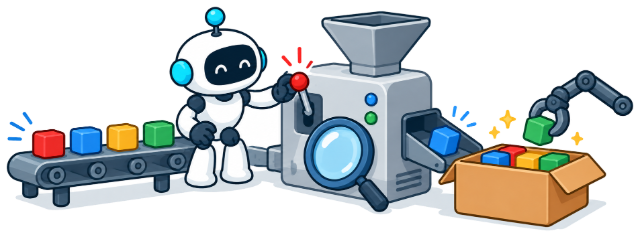}\\[1pt]\textbf{Revised procedure}\\Collector, evaluator, updater};
\path (b1.south) -- (b3.north) coordinate[midway] (bmid);
\node[bcell] (b2) at (bmid) {\includegraphics[height=1.22cm,width=3.70cm,keepaspectratio]{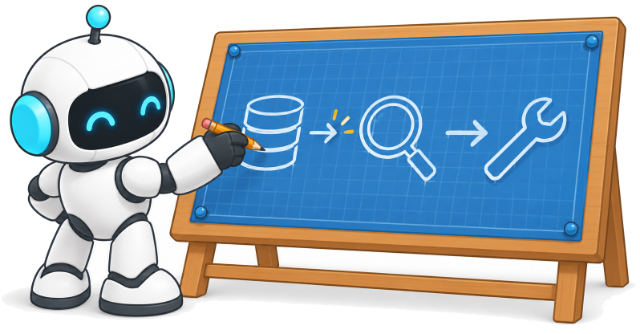}\\[1pt]\textbf{The improver revises}\\Collection, judgment, and updating};
\path (b1.south)--(b2.north) node[midway,art bead=ArtAmber,minimum size=.34cm,font=\fontsize{5.8}{6.5}\selectfont] {\faIcon{arrow-down}};
\path (b2.south)--(b3.north) node[midway,art bead=ArtAmber,minimum size=.34cm,font=\fontsize{5.8}{6.5}\selectfont] {\faIcon{arrow-down}};
\draw[art loop=ArtRed,rounded corners=2mm,-{Stealth[length=2.2mm,width=2.2mm]},dash pattern=on 4pt off 2.5pt] (b3.east) -- ++(.40,0) |- (b2.east);
\coordinate (qsw) at (13.33,0 |- b3.south); \coordinate (qne) at (15.85,0 |- b1.north);
\path[fill=white,drop shadow={opacity=.18,shadow xshift=.6pt,shadow yshift=-.8pt},rounded corners=2mm] (qsw) rectangle (qne);
\begin{scope}
  \clip[rounded corners=2mm] (qsw) rectangle (qne);
  \fill[ArtRed!8] (qne) rectangle ([yshift=-3.10cm]qsw |- qne);
\end{scope}
\draw[ArtRed!40,line width=.55pt,rounded corners=2mm] (qsw) rectangle (qne);
\draw[ArtRed!30,line width=.5pt] ([yshift=-3.10cm]qsw |- qne) -- ([yshift=-3.10cm]qne);
\node[text=ArtRed,font=\fontsize{15}{15}\selectfont,inner sep=0pt] at (14.59,.76) {\faIcon{sync-alt}};
\node[inner sep=0pt,text=ArtRed!85!black,font=\sffamily\bfseries\fontsize{7.8}{9.4}\selectfont] at (14.59,.19) {Recursive question};
\node[anchor=north,inner sep=0pt,text width=2.35cm,align=center,text=ArtRed!85!black,font=\artserif\fontsize{8.2}{9.9}\selectfont] at (14.59,-.14) {Does repair knowledge improve how the next revision is made?};
\node[anchor=north,inner sep=0pt,text=ArtSlate,font=\sffamily\bfseries\fontsize{7.8}{9.4}\selectfont] at (14.59,-2.31) {Two links to test};
\node[circle,fill=ArtRed,inner sep=0pt,minimum size=.36cm,text=white,font=\sffamily\bfseries\fontsize{6.8}{7}\selectfont] at (13.65,-2.86) {R1};
\node[anchor=north west,inner sep=0pt,text width=1.84cm,align=left,text=ArtSlate,font=\sffamily\fontsize{8.0}{9.6}\selectfont] at (13.89,-2.71) {Repair knowledge improves the improver};
\node[circle,fill=ArtRed,inner sep=0pt,minimum size=.36cm,text=white,font=\sffamily\bfseries\fontsize{6.8}{7}\selectfont] at (13.65,-4.51) {R2};
\node[anchor=north west,inner sep=0pt,text width=1.84cm,align=left,text=ArtSlate,font=\sffamily\fontsize{8.0}{9.6}\selectfont] at (13.89,-4.36) {Once improved, the improver helps produce its own next improvement};
% Bridge between the two routes, drawn last so it sits on both panel frames.
\node[art bead=ArtSlate,minimum size=.40cm,font=\fontsize{6.4}{7}\selectfont] at (8.30,-2.635) {\faIcon{arrow-right}};
\node[inner sep=0pt,font=\sffamily\bfseries\fontsize{7.8}{9.4}\selectfont,text=ArtSlate] at (8.04,-7.28)
  {Physical limits:\enspace\chip{\faIcon{eye-slash}\enspace incomplete sensing}\enspace\chip{\faIcon{plug}\enspace body interfaces}\enspace\chip{\faIcon{tools}\enspace hardware and task change}\enspace\chip{\faIcon{shield-alt}\enspace unsafe, costly practice}};
\end{tikzpicture}%
\caption{\textbf{Improving what the robot does and how it learns.} On the left, the learner improves under a fixed recipe and retains the change in its policy, world model, or external knowledge. On the right, the improver revises the learning procedure using retained repair knowledge; the dashed return asks whether that knowledge improves how the next revision is made (R1), and whether the improver, once improved, then helps produce its own next improvement (R2). The bottom row lists the physical limits behind the four conditions that let gains carry forward.}
\label{fig:two-routes}
\end{figure}
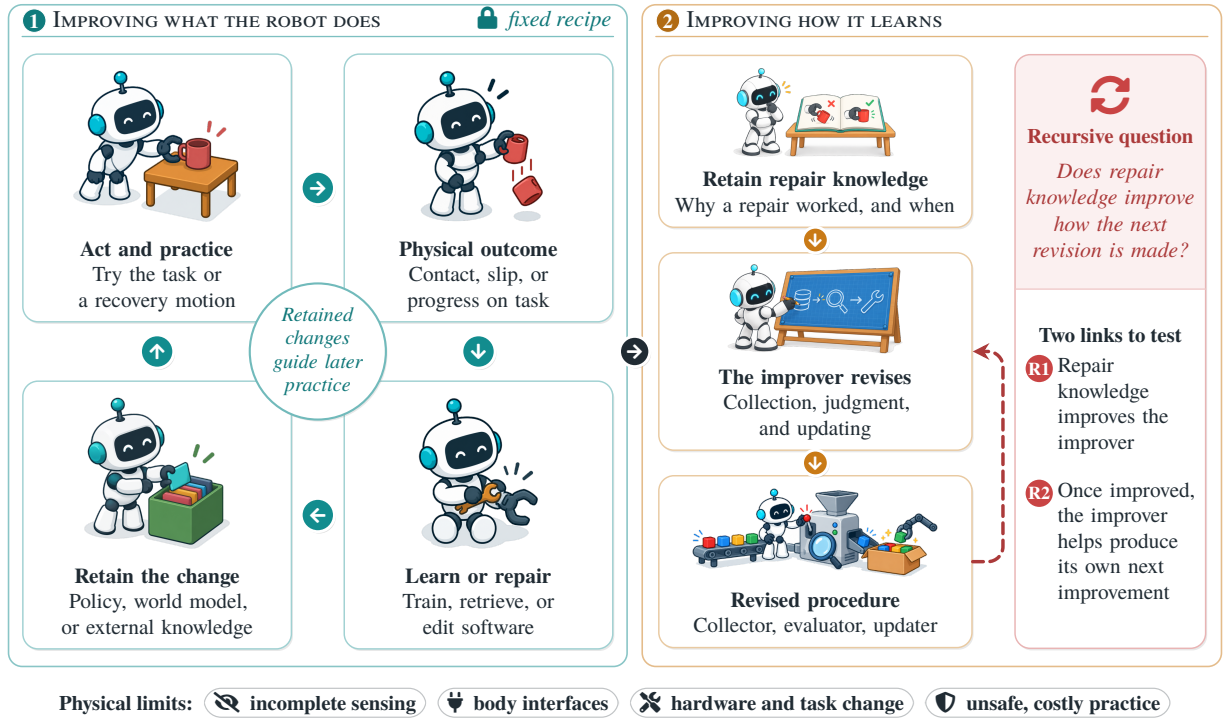

%% file: tables/related_surveys.tex
\begin{table}[!tbp]
\centering
\caption{Closely related reviews and the complementary focus of this review.}
\label{tab:related-surveys}
\begingroup\footnotesize
\setlength{\tabcolsep}{4pt}\renewcommand{\arraystretch}{1.12}\arrayrulecolor{PAIRule}
\begin{tabularx}{\linewidth}{@{}L{4.85cm}L{5.35cm}Y@{}}
\paitoprule
\pahead{Review} & \pahead{Scope and organization} & \pahead{Complementary focus here} \\
\paimidrule
\tableworkref{RSI in AI}{arxiv260707663} & Targets, loop closure, self-evaluation & Physical experience across rounds \\\paihair
\tableworkref{Self-Evolving Agents}{arxiv250721046} & What, when, how, and where to evolve & Physical feedback for procedure revision \\\paihair
\tableworkref{Weights or Skills?}{arxiv260801851} & Weights vs.\ skills; self-improving code & Later-learning benefit of skills \\\paihair
\tableworkref{Self-evolving Embodied AI}{arxiv260204411} & Memory, task, world, body, model & What interaction supplies, who uses it \\\paihair
\tableworkref{Self-Evolving AI for Humanoids}{arxiv260913236} & Four mechanisms, safety, evaluation & Sensing, reset, interfaces, maintenance \\\paihair
\tableworkref{Who Improves the Robot?}{arxiv261018842} & L0--L5 levels of physical RSI & Claims defined by their comparisons \\\paihair
\tableworkref{World Model for Robot Learning}{arxiv260500080} & Policy coupling, simulation, generation & Model-based gains on real robots \\\paihair
\tableworkref{World Models for Manipulation}{arxiv260600113} & Representation, action link, lifecycle & Physically corrected model-based training \\\paihair
\tableworkref{World-Action Models}{arxiv260916074} & Architectures, training, applications & Prediction gains across learning rounds \\\paihair
\tableworkref{Actionable Simulators}{arxiv260115533} & Grounding, self-evolution, closed loop & Model errors that reach the next learner \\\paihair
\tableworkref{Physical AI}{arxiv251004978} & Perception, reasoning, models, action & Enabling updates; why gains stop \\
\paibottomrule
\end{tabularx}\endgroup
\end{table}

%% file: sections/02_framework.tex
\sectionaccent{PAIInk}
\section{Foundations of Physical RSI}
\label{sec:background-scope}
\label{sec:framework}

We follow every system in this review along one path: from experience, to a retained change, to the change's next use. Later sections use this path to work out when a gain survives another round, without assuming a common architecture.

\input{figures/carry_forward_conditions}

\subsection{Learning to Act}
\label{sec:primer}

A few groups of ideas are enough to follow the rest of the review. We introduce them with the running example of the robot that keeps missing a grasp.

\paragraph{Policies.} A \emph{policy} is the controller that maps what the robot currently observes (camera images, joint angles, force readings, and possibly a language instruction) to its next motor command; for the grasping robot, it turns a view of the object into the next movement of the arm and gripper. In the systems reviewed here, policies are usually neural networks. \emph{Generalist} policies are trained on large collections of robot data spanning many tasks, as in RT-1~\citep{brohan2023rt1}, and sometimes many robot bodies, or \emph{embodiments}, as in Octo~\citep{octo2024}. A \emph{vision-language model} (VLM) reads images and text and outputs text; robot systems often use one to propose goals or to judge whether an attempt succeeded. A \emph{vision-language-action} (VLA) model is a generalist policy built on a pretrained VLM, so that it reads images and an instruction and outputs actions; RT-2~\citep{brohan2023rt2}, OpenVLA~\citep{arxiv240609246}, and $\pi_0$~\citep{arxiv241024164} are examples. Many recent policies output a short sequence of future actions at once, called an \emph{action chunk}.

\paragraph{Learning from demonstrations.} The simplest way to obtain a policy is \emph{imitation learning} from demonstrations, which are often recorded while a person teleoperates the robot~\citep{argall2009survey}. In \emph{behavior cloning} (BC), the policy is trained to reproduce the demonstrated action in each recorded situation. BC learns only from recorded situations, so its small errors can compound: if the gripper drifts slightly off the object into a state that no demonstration covers, the policy may have no good action to take. Interactive methods such as DAgger~\citep{ross2011dagger} address this by asking an expert what to do in the situations that the learner itself reaches. In practice, a person often takes over when the robot errs, and these \emph{interventions} supply corrective training data.

\paragraph{Learning from rewards.} In \emph{reinforcement learning} (RL)~\citep{sutton2018rl,kober2013rl}, the robot instead tries actions and receives a \emph{reward} that scores the outcome, for example 1 for a stable grasp and 0 otherwise. On a real robot, rewards do not arrive automatically: a person, a hand-written check, or a learned \emph{reward model}, such as a success detector, must judge each attempt. We call a component that supplies such judgments for learning an \emph{evaluator}. Because rewards are often sparse and delayed, many methods learn a \emph{value function}, or \emph{critic}, that predicts how much future reward to expect; some systems use the word critic more loosely for any component, learned or hand-written, that scores behavior. The \emph{advantage} of an action measures how much better it is than what the policy would typically do in the same situation. \emph{Off-policy} methods can learn from experience collected by earlier versions of the policy or by other sources, which they keep in a \emph{replay buffer}. On hardware, every trial costs time, supervision, wear, and a \emph{reset} that returns the scene to a usable starting state, so the source of the reward and the number of available trials become central resources.

\paragraph{Improving an existing policy.} Large pretrained policies are rarely retrained from scratch. They are \emph{fine-tuned}, or post-trained, on new data. A small \emph{residual} policy can add corrections on top of the actions of a frozen base policy, such as a slight change to the angle at which the gripper approaches the object. \emph{Distillation} trains one policy to imitate another, for example to fold a specialist's behavior back into a generalist. When a policy learns a sequence of tasks, it can forget earlier ones, a problem studied in continual learning~\citep{parisi2019continual}; \emph{replaying} stored data from earlier tasks is a common protection.

\paragraph{Models of the world.} A \emph{world model}, or dynamics model, predicts what will happen after an action, for instance the next camera image or the next object positions~\citep{ha2018worldmodels,hafner2020dreamer}. The grasping robot's contact model is one example: it predicts whether a touch will become a stable hold. A robot can use such a model to \emph{plan}, by imagining the outcomes of candidate actions, or to generate \emph{imagined experience} on which a policy is trained. A physics simulator is a hand-built model of this kind; a learned world model is fitted to recorded robot data and can be wrong in ways that a policy may learn to exploit.

\paragraph{Improvements outside the weights.} Not every improvement is stored in network weights. A system can keep \emph{memories} of earlier episodes, maintain a library of callable \emph{skills}~\citep{sutton1999options}, or write \emph{programs} that compose perception and control functions~\citep{liang2023code}. Large language models are increasingly used to write such code or to revise reward functions~\citep{ma2024eureka}.

\paragraph{Simulation and hardware.} Results come either from simulation benchmarks or from hardware experiments on physical robots, and training in simulation before deployment on hardware is called \emph{sim-to-real} transfer. Section~\ref{sec:physical-consequences} discusses what each setting can establish.

\subsection{Defining Recursive Self-Improvement}
\label{sec:selection}
\label{sec:four-uses}
\label{sec:key-terms}

\paragraph{Where experience leaves a change.}
We consider loops in which experience produces a retained change that affects later attempts or learning. Human demonstration, correction, reset, or approval can remain part of the loop: automation concerns who makes each decision, whereas learning concerns whether experience changes it (Table~\ref{tab:human-system-roles} in Appendix~\ref{app:landscape}). A failed update can also be informative by showing where this connection breaks.

Generic pretraining without a subsequent experience--update--use comparison serves as background, as does adaptation confined to a single attempt. Rapid Motor Adaptation~\citep{arxiv210704034}, for example, infers current conditions from recent history without retaining the adaptation. Zeva's interaction memory is a boundary case: it carries across retries within an episode but not across episodes or learning rounds. Simulation-only and formal studies supply enabling mechanisms, while meta-learning, AutoML, and digital agents supply precedents (Section~\ref{sec:positioning}); direct evidence about physical RSI must come from physical systems.

A retained change can serve four uses (Figure~\ref{fig:taxonomy}). It can choose or execute actions, predict outcomes or supply practice, support reuse (for example, calling a saved recovery or consulting guidance for a new program), or change how later data are collected, judged, and turned into updates. These uses are usually carried by policies, world models, external knowledge, and learning procedures, respectively (Sections~\ref{sec:base-loop} and~\ref{sec:emerging-directions}). They can overlap: a policy can collect training data, a model can judge outcomes, and a memory can guide an updater. Each study is discussed mainly under the use its experiment tests, regardless of whether its authors call it self-evolving or recursively self-improving.

\input{figures/taxonomy}

\paragraph{Learner, learning procedure, and improver.}
The learner acquires task capabilities; the \emph{learning procedure} determines what experience to collect, how to judge it, and how to construct updates. An \emph{improver} proposes or selects revisions of that procedure. It may be a person, a fixed rule, a search process, or a learned component, with knowledge in memory, weights, code, or prompts. Figure~\ref{fig:two-routes} separates the two levels: the learner improves under a fixed recipe, whereas the improver revises the recipe itself.

In the grasp example, adding a corrected grasp to the training data is a \emph{repair}, an update made in response to an observed failure or limitation. The rule choosing which failed grasps need correction belongs to the learning procedure; whoever devises or changes that rule is the improver. The object being changed determines the role: an editor of task skills or data belongs to the procedure, whereas an editor of rewards, evaluators, collection rules, or update-generating training code acts as an improver.

\paragraph{From better behavior to recursion.}
We distinguish four claims (Table~\ref{tab:comparison-questions}), and evidence for an earlier claim does not establish a later one. \emph{Behavioral improvement} means better execution under a common task test. A \emph{later-learning benefit} means that retained data, weights, models, skills, or code help a subsequent learner. \emph{Procedure improvement} means that a revised collector, evaluator, editor, or updater teaches better than its earlier version. This comparison requires \emph{matched learners}, which differ only in the component under test and share starting conditions, resources, and evaluation protocol (Figure~\ref{fig:downstream-comparison}).

We operationalize \emph{recursive self-improvement} through two further links, which this review uses as empirical criteria for attributing a gain to recursion. First (R1), \emph{repair knowledge} retained from earlier repairs must improve the improver: on new failures, its procedures must teach matched learners better than those produced by its earlier version. Second (R2), the improver, once improved, must help produce its own next improvement: given the same new repair histories, a self-revision proposed using its acquired knowledge must yield an improver that outperforms both the improver it revised and a revision of that same improver proposed with the acquired knowledge frozen. A fixed improver can repeatedly improve procedures without satisfying either link; an external designer's growing expertise does not establish system-level recursion either. Section~\ref{sec:reviser-criteria} develops the grasp example. A concurrent survey grades physical RSI by whether an update changes capabilities or the improvement procedure and whether its gain holds on the same or on other tasks, reserving its top level for an independently verified recursive gain~\citep{arxiv261018842}; the claims here are instead defined by the comparison that would establish each. The name is shared by other concurrent work. PhysRSI-eval~\citep{arxiv261026848} applies it to agents with frozen weights that revise code, tools, and notes after each graded attempt, and Physical RSI~1.0~\citep{physicalrsi2026} names a system in which a multimodal foundation model revises the harness through which a robot acts. We read these works, like any other, by the comparisons they report; a name does not establish either link. In symbols, write $L$ for a learner, $J(L)$ for its outcome under a protected success evaluation, and $P=(c,e,d,u)$ for a learning procedure whose components choose what experience to collect, evaluate outcomes, select the training data, and apply the update rule; $P(L)$ is the learner that $P$ produces from $L$ under a fixed budget. Behavioral improvement means $J(L')>J(L)$. Procedures are compared from a common starting learner $L_0$ with matched resources and evaluation, and we write $J(P)=J(P(L_0))$: procedure improvement means $J(P')>J(P)$, and a later-learning benefit from a retained artifact $A$ means $J(P(L_0;A))>J(P(L_0))$, where $P(L_0;A)$ is trained with access to $A$. An improver $I$ maps failure histories $F$ to a procedure $I(F)$. Write $I^{K}$ for the improver with knowledge $K$, $K_0$ and $K_1$ for its knowledge before and after earlier repairs, and $\rho_{K}(I,H)$ for the improver obtained when a self-revision of $I$ is proposed from repair histories $H$ with knowledge $K$ available. On new failures $F$, held out from everything the improver has learned from and from $H$, the two links require
\begin{align*}
\text{R1:}\quad & J\bigl(I^{K_1}(F)\bigr) > J\bigl(I^{K_0}(F)\bigr),\\
\text{R2:}\quad & J\bigl(\rho_{K_1}(I^{K_1},H)(F)\bigr) > \max\bigl\{J\bigl(I^{K_1}(F)\bigr),\ J\bigl(\rho_{K_0}(I^{K_1},H)(F)\bigr)\bigr\}.
\end{align*}
In R1, the right-hand side is the frozen-knowledge control. In R2, the first term requires the successor to improve on its parent, so a revision that changes nothing cannot pass; the second revises the same parent with the acquired knowledge frozen and so isolates what that knowledge contributes to the revision. Withholding the acquired knowledge while keeping the same parent is straightforward when that knowledge resides in a memory snapshot, code, or prompts; when it is stored in the improver's weights, the R2 control needs a design of its own.

\input{tables/comparison_questions}

\paragraph{How far a result supports recursion.}
For R1 and R2, we distinguish \emph{mechanism only}, where machinery could change the improver; an \emph{uncontrolled result}, consistent with a link but without a control that withholds only what was acquired; and a \emph{controlled result}, which tests that contribution with matched learners and budgets (Section~\ref{sec:recursive-test}). The claim and its level of support are separate.

Two controls answer different questions. A frozen-knowledge control withholds acquired repair knowledge while retaining the improver, testing that knowledge's contribution to the procedures produced (R1) or the self-revision proposed (R2). A fixed-improver control keeps the initial improver while the rest of the system improves. A gain over it shows that revising the improver contributes to later improvement, bearing on R2, but does not identify which acquired knowledge caused the revision.

\paragraph{What lets gains carry forward.}
Four conditions recur across these claims. \emph{Informative feedback} distinguishes the remaining learning need. \emph{Usability} means that the receiving component can represent, optimize, and execute the change. \emph{Robustness to distribution shift} asks whether the change stays useful as the policy, model, task, or hardware changes, including changes caused by improvement itself. \emph{Continued practice} asks whether another informative trial remains safe and affordable. These four conditions overlap and serve as diagnostic lenses rather than independently validated constructs. Section~\ref{sec:compounding} examines their failures, and Figure~\ref{fig:overview} connects these conditions to the uses and claims above.

\subsection{Relating RSI to Robot Learning}
\label{sec:positioning}

Reinforcement learning and meta-learning already let robots improve from experience and learn how to learn. Self-training, continual learning, and automated machine learning (AutoML) let systems learn from their own outputs, keep learning across tasks, or search for better training recipes. It is therefore fair to ask what RSI adds. These are method paradigms or problem settings, which say how an update is computed or what it must cope with, whereas the claims of Section~\ref{sec:key-terms} say what an experiment shows to have changed. An RSI system may use any of them to revise any of its components, so the categories overlap rather than exclude one another: RSI is not a further paradigm after meta-learning but a level of improvement that a system built with any of them might reach (Table~\ref{tab:paradigms}).

\input{tables/paradigms}

\paragraph{Reinforcement learning.}\label{sec:rl-loop} Standard RL improves a policy under an objective and update mechanism chosen before training~\citep{sutton2018rl}. Run repeatedly on hardware, as in HIL-SERL~\citep{arxiv241021845} and RECAP~\citep{arxiv251114759}, it yields behavioral improvement, and its data or policy can supply a later-learning benefit. RL loops can also be nested inside an outer optimization over part of the learning procedure. Eureka searches over reward programs using the RL training results they produce~\citep{ma2024eureka}, and meta-gradient RL adapts the return while the agent learns~\citep{xu2018metagradient}. LPG discovers an update rule~\citep{oh2020lpg}, and DiscoRL learns such a rule across many agents and environments and applies it to unseen benchmarks~\citep{oh2025discorl}. These loops target procedure improvement, while the outer search itself stays fixed. DERL~\citep{arxiv251213399} departs from this on text-based and mathematical reasoning tasks: it trains its reward proposer by policy gradient on the validation success of the inner policies, so the search learns from its own trials, while the rule that trains the proposer stays fixed. In the grasp example, RL that tunes the grasp under a fixed slip penalty improves behavior, whereas a search that revises the slip penalty changes the procedure; further rounds of the first kind do not become the second.

\paragraph{Self-training.}\label{sec:self-training} Many systems called self-improving train on their own outputs. In self-training such as Noisy Student~\citep{xie2020noisystudent}, a teacher trained on labeled data labels unlabeled data, and a student trained on both becomes the next teacher. Expert Iteration~\citep{anthony2017expert} and STaR~\citep{zelikman2022star} apply the same pattern to search and reasoning, and RoboCat~\citep{arxiv230611706} applies it to robot data collection. Each round improves behavior and helps the next learner through the data it generates, while the recipe that generates, labels or filters, and trains on the data stays fixed. When the model also judges its own outputs, as in Self-Rewarding Language Models~\citep{yuan2024selfrewarding}, the evaluator changes with training, but the judging prompt and the training recipe do not.

\paragraph{Continual learning.}\label{sec:continual-learning} Continual, or lifelong, learning asks how a learner that meets a sequence of tasks can reuse what it learned on earlier ones~\citep{thrun1995lifelong} without forgetting them~\citep{parisi2019continual}. Backward and forward transfer~\citep{lopezpaz2017gem} measure how learning one task changes performance on earlier and later ones. Backward transfer thus measures retention, whereas forward transfer as defined there is performance on a new task before training on it; LIBERO~\citep{arxiv230603310}, a benchmark of sequential robot manipulation tasks, instead scores forward transfer by success on each new task averaged over its training, which is closer to a later-learning benefit. Replay, regularization, and other protections against forgetting are usually chosen by designers, so continual learning defines a setting that an RSI system must also handle (Section~\ref{sec:accumulation}) rather than a way of improving the improver.

\paragraph{Meta-learning.}\label{sec:meta-learning} Meta-learning trains over many tasks so that new tasks are learned faster~\citep{thrun1998learning}, and what it learns decides which claim it can support. An initialization from which a few gradient steps solve a new task, as in MAML~\citep{finn2017maml}, is an acquired artifact and supports a later-learning benefit. Other methods learn part of the learning procedure: recurrent policies whose activations carry out adaptation~\citep{duan2016rl2,wang2016learningrl}, a dynamics prior that adapts online, as in GrBAL~\citep{arxiv180311347}, learned optimizers~\citep{andrychowicz2016learning}, and evolved or learned losses~\citep{houthooft2018epg,yu2018daml,bechtle2021ml3}. When a learner guided by such a part learns better than a matched learner guided by the original, the comparison supports procedure improvement. LMPC~\citep{arxiv240211450} is a boundary case: it fine-tunes the model that writes reward code on past teaching sessions so that later teaching needs fewer turns, but that code drives a model-predictive controller directly rather than a training run, so the gain lies in how behavior is constructed. The outer meta-training procedure that produces the part (a gradient loop, an evolutionary search, or a fine-tuning recipe) is specified by its designers and is not changed by the revisions it produces, so learning reaches the procedure but not the process that produces it.

\paragraph{Automated machine learning.}\label{sec:automl} AutoML searches over architectures, hyperparameters, and training pipelines~\citep{hutter2019automl}, and AutoML-Zero extends the search to learning algorithms~\citep{real2020automlzero}; a found configuration that teaches matched learners better supports procedure improvement. Some searchers learn from their own record: the controller of neural architecture search~\citep{zoph2017nas} is trained with reinforcement learning on the validation accuracy of the architectures it generates. Because it edits the learner rather than the rules for collecting, judging, or updating, it is a learned part of the procedure, like the skill editor of PRACTICE~\citep{arxiv260830760}, and the rule that trains it stays fixed.

\paragraph{Where recursion adds a requirement.}\label{sec:recursion-requirement} RSI moves the process that produces revisions inside as well, as the two links of Section~\ref{sec:key-terms} require. The requirement is self-referential, as in the G\"odel machine~\citep{schmidhuber2003godel}, STOP~\citep{zelikman2024stop}, the Darwin G\"odel Machine~\citep{zhang2025dgm}, and G\"odel Agent~\citep{arxiv241004444}, which let the improving process act on its own code. Generalized Agent Iteration places iterative policy improvement and RSI in one formal framework, in which whether the improvement mechanism sits inside the agent separates the two, and whether the evaluation standard is grounded outside it decides whether goals can drift~\citep{arxiv260913406}. The criterion is therefore which component changes with the revision history, not whether some outer recipe stays fixed. In the meta-learning and AutoML methods above, the component that produces revisions of the learning procedure does not change; an improver fine-tuned by a fixed recipe on its own repair histories does change and can satisfy R1, which the frozen-knowledge comparison of Section~\ref{sec:recursive-test} tests. Conversely, an improver whose code changes in every round satisfies neither link if the changes come from a designer or from a search that learns nothing from earlier rounds.

\subsection{Facing Physical Limits}
\label{sec:physical-consequences}
\label{sec:paradigms-physical}

Table~\ref{tab:physical-digital} lists the terms we use to judge how closely a study is tied to physical experience. Hardware interaction connects learning directly to physical consequences, but a consequence still needs a diagnosis and a reliable judge. A binary failure label on a grasp may not distinguish pose error from slip, whereas force measurements or contact history may separate the two. Simulation, learned prediction, and formal checks represent selected consequences whose adequacy must be tested at the decision they support. Simulation is cheap and repeatable but omits some physical effects, whereas hardware captures real contact, sensing, and wear but allows few trials, so the two settings support different conclusions about cost, transfer, and the comparisons a study can afford.

Because practice changes its own conditions (Section~\ref{sec:introduction}), returning to a usable starting state may require a reset policy or human assistance; early work on autonomous real-world RL already named resets, onboard perception, and rewards without instrumentation as the ingredients that continued learning needs~\citep{zhu2020ingredients}. Moving knowledge to another robot adds a further gap, because reuse across bodies needs compatible coordinates, timing, sensing, and reachable effects. A study's connection to physical experience is therefore a matter of degree. It depends on the \emph{success evaluation} (who or what judges whether an attempt succeeded), on interaction coverage and recoverability, and on whether successive trials remain comparable. The success evaluation may rely on the same evaluator that supplied the learning signal or on one independent of it, a difference that Section~\ref{sec:cost-ledger} shows to matter. Table~\ref{tab:physical-digital} contrasts these conditions with those of digital self-improvement, Section~\ref{sec:safety} develops their consequences for continued learning, and Table~\ref{tab:domain-scope} in Appendix~\ref{app:landscape} shows what evidence each robot domain contributes.

These conditions weigh on each paradigm of Section~\ref{sec:positioning} in a different place. Meta-learning needs a distribution of training tasks, and on hardware every sampled task costs trials, resets, and supervision; domain-adaptive meta-learning meta-trains on previously collected human and robot demonstrations, and LMPC collected its training data in simulation. Eureka and DrEureka run their reward search in simulation, where many candidates can be trained in parallel, and DrEureka then deploys the resulting policy on hardware~\citep{ma2024dreureka}. For RSI, the bottleneck shifts to revision histories: an improver learns from them, and each entry needs at least one learning round judged by physical trials, so these histories accumulate more slowly than the task samples of meta-learning or the transitions of RL, and no executable test settles whether a revision helped. Simulation, stored traces, and learned models can screen candidate revisions before hardware trials, but their adequacy for that screening is itself a claim to test (Section~\ref{sec:test-controls}).

%% file: figures/carry_forward_conditions.tex
\begin{figure}[!tbp]
\centering
\begin{tikzpicture}[art fonts,x=1cm,y=1cm,line cap=round,line join=round,
  every node/.append style={execute at begin node={\hyphenpenalty=10000\exhyphenpenalty=10000}}]
% (a) Four uses
\artpanel[a]{.02}{2.55}{16.06}{4.01}{ArtBlue}{Where a change is kept, and its use (Sections~\ref*{sec:base-loop} and~\ref*{sec:emerging-directions})}
\foreach \cx/\col/\ic/\ttl/\use in {2.025/ArtBlue/robot/Policy/behavior, 6.035/ArtTeal/cube/World model/prediction and practice, 10.045/ArtAmber/recycle/External knowledge/reuse and recovery, 14.055/ArtRed/cogs/Learning procedure/{collect, judge, update}}
  {\node[art box=\col,text width=3.50cm,inner xsep=3pt,font=\fontfamily{qtm}\bfseries\fontsize{8.6}{10.4}\selectfont] at (\cx,3.07) {\strut{\color{\col}\fontsize{8.0}{9.6}\selectfont\faIcon{\ic}}\hspace{3pt}\ttl\\{\mdseries\itshape\fontsize{7.8}{9.2}\selectfont\use}};}
% (b) Four conditions, illustrated with the grasping robot's breakdowns from the introduction
\artpanel[b]{.02}{-1.15}{16.06}{2.35}{ArtTeal}{What must hold for a change to carry forward: four conditions (Section~\ref*{sec:compounding})}
\foreach \cx/\img/\ih/\ttl/\dsc in {
  2.025/art_signal/1.6cm/Informative feedback/{Does it address the\\remaining need?},
  6.035/art_recipient/1.42cm/Usability/{Can the receiving\\component use it?},
  10.045/art_shift/1.6cm/{Robustness to\\distribution shift}/{Does it stay useful?},
  14.055/art_continuation/1.6cm/Continued practice/{Can informative\\practice continue?}}
  {\node[art img,anchor=south] at (\cx,.21) {\includegraphics[height=\ih]{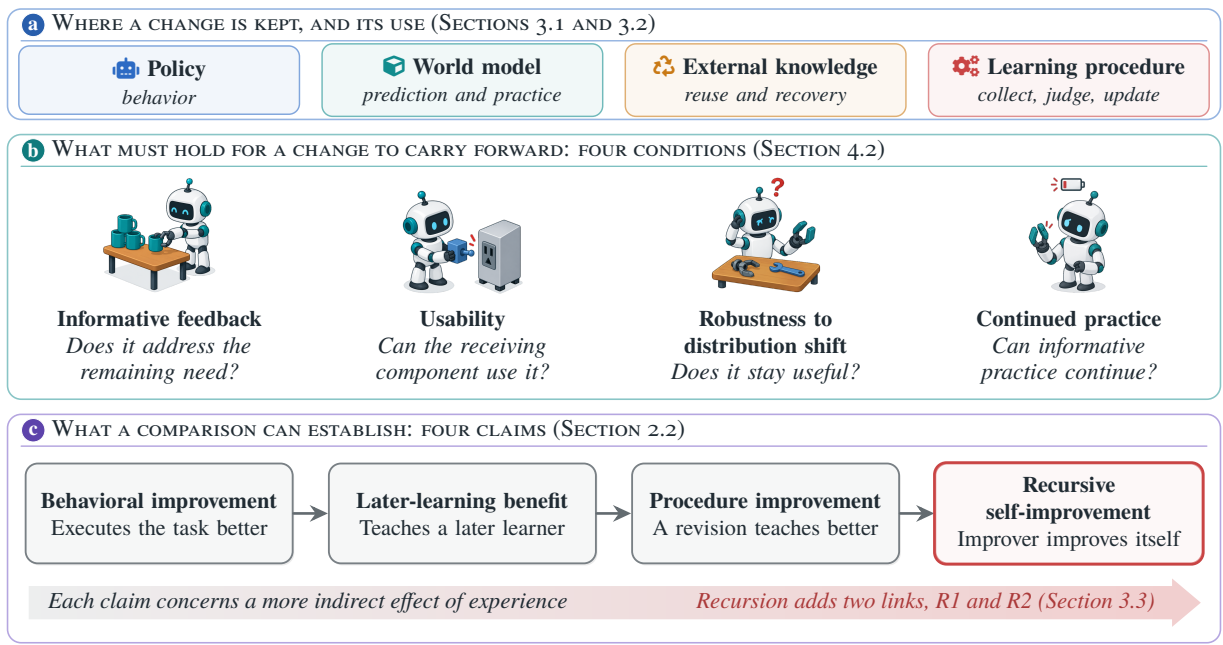}};
   \node[art ctext,text width=3.45cm,font=\artserif\fontsize{8.8}{10.0}\selectfont] at (\cx,-.465) {{\sffamily\upshape\bfseries\fontsize{8.4}{10.0}\selectfont\ttl}\\\dsc};}
% (c) Four claims
\artpanel[c]{.02}{-4.37}{16.06}{-1.35}{ArtPurple}{What a comparison can establish: four claims (Section~\ref*{sec:key-terms})}
\foreach \cx/\col/\ttl/\dsc/\n in {
  2.025/ArtSlate/Behavioral improvement/Executes the task better/k1,
  6.035/ArtSlate/Later-learning benefit/Teaches a later learner/k2,
  10.045/ArtSlate/Procedure improvement/A revision teaches better/k3,
  14.055/ArtRed/Recursive self-improvement/Improver improves itself/k4}
  {\node[fill=\col!4,draw=\col!60,line width=.7pt,rounded corners=1.6mm,text width=3.25cm,minimum height=1.32cm,inner sep=4pt,align=center,text=ArtSlate,font=\sffamily\fontsize{8.4}{10.2}\selectfont,drop shadow={opacity=.16,shadow xshift=.6pt,shadow yshift=-.8pt}] (\n) at (\cx,-2.66) {\textbf{\ttl}\\\dsc};}
\foreach \a/\b in {k1/k2,k2/k3,k3/k4} {\draw[-{Stealth[length=2.4mm,width=2.2mm]},line width=1pt,draw=ArtSlate!70] (\a)--(\b);}
\draw[ArtRed,line width=1pt,rounded corners=1.6mm] (k4.south west) rectangle (k4.north east);
% Direction of the claims: a full-width gradient arrow inside panel (c).
\shade[left color=ArtSlate!8,right color=ArtRed!26] (.30,-3.62) -- (15.42,-3.62) -- (15.42,-3.52) -- (15.80,-3.84) -- (15.42,-4.16) -- (15.42,-4.06) -- (.30,-4.06) -- cycle;
\node[anchor=west,inner sep=0pt,text=ArtSlate,font=\artserif] at (.55,-3.84) {Each claim concerns a more indirect effect of experience};
\node[anchor=east,inner sep=0pt,text=ArtRed!85!black,font=\artserif] at (15.20,-3.84) {Recursion adds two links, R1 and R2 (Section~\ref*{sec:reviser-criteria})};
\end{tikzpicture}%
\caption{\textbf{The review at a glance.} Illustrations in (b) show the four grasp breakdowns from Section~\ref{sec:introduction}.}
\label{fig:overview}
\label{fig:dependencies}
\end{figure}

%% file: figures/taxonomy.tex
\begin{figure}[!tp]
\centering
\begin{tikzpicture}[art fonts,x=1cm,y=1cm,line cap=round,line join=round,
  every node/.append style={execute at begin node={\hyphenpenalty=10000\exhyphenpenalty=10000}},
  cellcap/.style={inner sep=0pt,align=center,text=ArtSlate,font=\sffamily\fontsize{8.2}{9.6}\selectfont,anchor=south,
    execute at end node={\vphantom{g}}},
  minilab/.style={font=\sffamily\fontsize{7.4}{8}\selectfont,text=ArtSlate,inner sep=0pt,anchor=base,
    execute at begin node={\vphantom{Ag}}},
  hand/.style={draw=ArtRed!80!black,dash pattern=on 2pt off 1.5pt,line width=.6pt,rounded corners=1.5mm,fill=white,
    text=ArtRed!85!black,font=\artserif,align=center,inner xsep=3pt,inner ysep=2pt},
  mini/.style={fill=white,draw=#1!45,line width=.5pt,rounded corners=1.4mm,minimum width=1.22cm,minimum height=1.56cm},
  ibox/.style={fill=white,draw=ArtRed!55,line width=.6pt,rounded corners=1.6mm,minimum width=0.98cm,minimum height=1.40cm},
  vig/.style={inner sep=0pt}]
\path[use as bounding box] (0,-11.72) rectangle (16.08,.74);
% Column cards (x0, x1): experience 3.00--7.15, retained change 7.80--11.25, immediate use 11.90--15.86.
% Both 0.65 cm gaps hold a 0.42 cm arrow centered in the gap, so each arrow sits 0.115 cm from both cards.
\def\cAa{3.00}\def\cAb{7.15}\def\cBa{7.80}\def\cBb{11.25}\def\cCa{11.90}\def\cCb{15.86}
% Row frame: tinted panel, darker use column, numbered badge, use name and principal claim, three cards, two arrows.
\newcommand{\userow}[6]{%
  \begin{scope}
    \clip[rounded corners=1.8mm] (.09,#2) rectangle (16.01,#1);
    \fill[white] (.09,#2) rectangle (16.01,#1);
    \fill[#3!5] (.09,#2) rectangle (2.85,#1);
  \end{scope}
  \draw[#3!30,line width=.45pt] (2.85,#2)--(2.85,#1);
  \draw[#3!50,line width=.6pt,rounded corners=1.8mm] (.09,#2) rectangle (16.01,#1);
  \node[circle,fill=#3!88!black,draw=white,line width=.35pt,text=white,inner sep=0pt,minimum size=.36cm,
    font=\sffamily\bfseries\fontsize{8}{8}\selectfont] at (.40,#1-.31) {#4};
  \node[align=center,text width=2.58cm,inner sep=0pt] at (1.47,{(#1+#2)/2}) {%
    {\scshape\fontsize{10.4}{11.8}\selectfont\color{ArtSlate}#5\par}\vspace{3pt}%
    {\centering\artserif\color{#3!75!black}#6\par}};
  \foreach \xa/\xb in {\cAa/\cAb,\cBa/\cBb,\cCa/\cCb}
    \path[fill=white,draw=#3!38,line width=.55pt,rounded corners=1.6mm,
      drop shadow={opacity=.12,shadow xshift=.5pt,shadow yshift=-.6pt}] (\xa,#2+.08) rectangle (\xb,#1-.10);
  \node[art fat=#3,minimum height=.42cm,minimum width=.36cm] at ({(\cAb+\cBa)/2-.008},{(#1+#2)/2+.30}) {};
  \node[art fat=#3,minimum height=.42cm,minimum width=.36cm] at ({(\cBb+\cCa)/2-.010},{(#1+#2)/2+.30}) {};}
% Header
\shade[left color=ArtPurple,middle color=ArtTeal,right color=ArtAmber] (.09,.02) rectangle (16.01,.05);
\node[font=\scshape\fontsize{9.4}{10}\selectfont,text=ArtSlate] at (1.47,.35) {Kept in};
\node[art card=ArtSlate,font=\sffamily\bfseries\fontsize{8.4}{10}\selectfont,inner xsep=2.5pt,inner ysep=1.5pt] at (5.075,.35) {\faIcon{hand-paper}\enspace Physical experience};
\node[art card=ArtSlate,font=\sffamily\bfseries\fontsize{8.4}{10}\selectfont,inner xsep=2.5pt,inner ysep=1.5pt] at (9.525,.35) {\faIcon{archive}\enspace Retained change};
\node[art card=ArtSlate,font=\sffamily\bfseries\fontsize{8.4}{10}\selectfont,inner xsep=2.5pt,inner ysep=1.5pt] at (13.88,.35) {\faIcon{share}\enspace Immediate use};
% (1) Behavior: rows span 2.55 cm; captions sit 0.16 cm above the row bottom, images start 0.60 cm above it.
\userow{-.10}{-2.65}{ArtBlue}{1}{Policy}{Behavior}
\node[vig,anchor=south] at (5.075,-2.05) {\includegraphics[width=3.85cm]{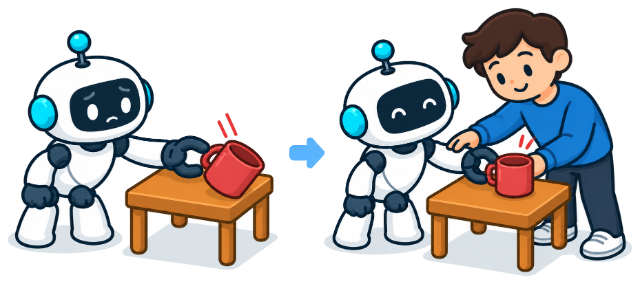}};
\node[cellcap,text width=3.95cm] at (5.075,-2.49) {Corrected grasps};
\node[vig] at (9.525,-1.19) {\includegraphics[width=3.20cm]{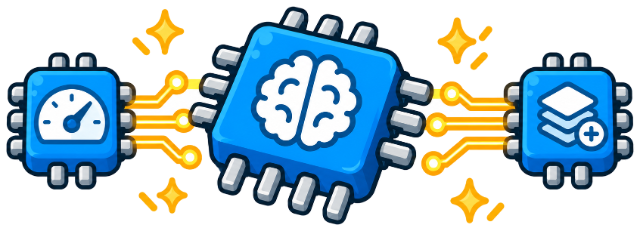}};
\node[cellcap,text width=3.30cm] at (9.525,-2.49) {Policy, value, residual};
\node[vig,anchor=south] at (13.10,-2.05) {\includegraphics[height=1.72cm]{figures/art_act}};
\node[hand] at (14.90,-1.05) {Policy as\\collector\\[-1pt]{\upshape\scriptsize\faIcon{arrow-down}}};
\node[cellcap,text width=3.80cm] at (13.88,-2.49) {Execute or choose actions};
% (2) Prediction and practice
\userow{-2.79}{-5.34}{ArtTeal}{2}{World model}{Prediction and practice}
\node[vig,anchor=south] at (5.075,-4.74) {\includegraphics[width=3.85cm]{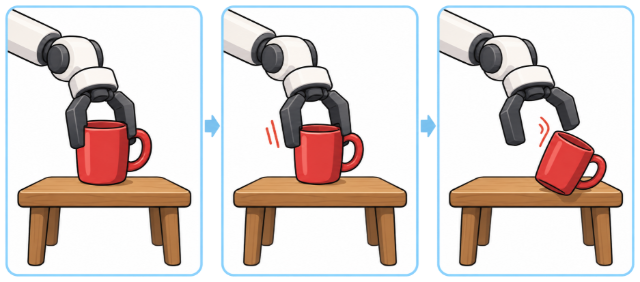}};
\node[cellcap,text width=3.95cm] at (5.075,-5.18) {Contact and slip trajectories};
\node[vig,anchor=south] at (9.525,-4.74) {\includegraphics[width=3.20cm]{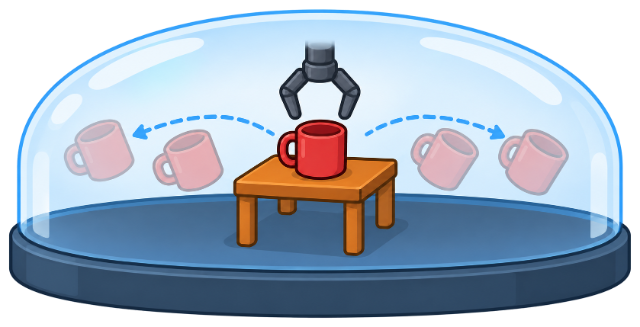}};
\node[cellcap,text width=3.30cm] at (9.525,-5.18) {Dynamics or world model};
\node[vig,anchor=south] at (13.10,-4.42) {\includegraphics[height=1.42cm]{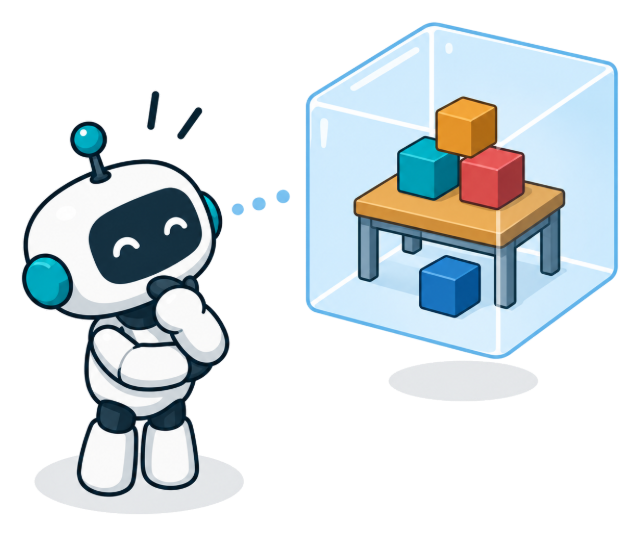}};
\node[hand] at (14.90,-3.74) {Model as\\judge\\[-1pt]{\upshape\scriptsize\faIcon{arrow-down}}};
\node[cellcap,text width=3.80cm] at (13.88,-5.18) {Plan or supply training experience};
% (3) Reuse and recovery
\userow{-5.48}{-8.03}{ArtAmber}{3}{External knowledge}{Reuse and recovery}
\node[vig,anchor=south] at (5.075,-7.43) {\includegraphics[width=3.85cm]{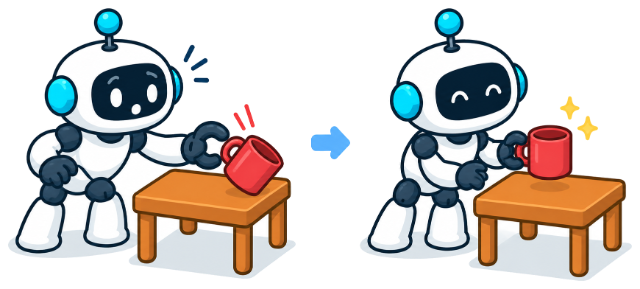}};
\node[cellcap,text width=3.95cm] at (5.075,-7.87) {A successful regrasp};
\node[vig,anchor=south] at (8.66,-7.11) {\includegraphics[height=1.30cm]{figures/art_retain}};
\node[art card=ArtAmber,inner sep=2.2pt] at (10.27,-6.34) {{\ttfamily\fontsize{7.6}{9}\selectfont regrasp()}\\[-1pt]{\rmfamily\itshape\fontsize{7.6}{9}\selectfont recovery skill}};
\node[cellcap,text width=3.30cm] at (9.525,-7.87) {Memory, skill, contract, program};
\node[vig,anchor=south] at (13.10,-7.43) {\includegraphics[height=1.72cm]{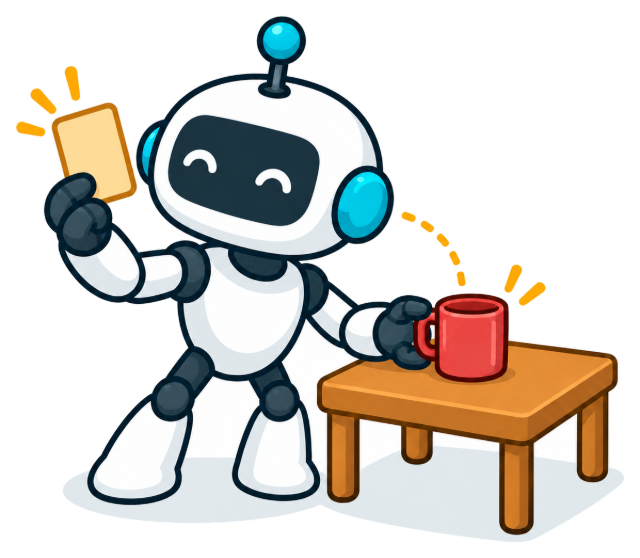}};
\node[hand] at (14.90,-6.43) {Memory as\\guide\\[-1pt]{\upshape\scriptsize\faIcon{arrow-down}}};
\node[cellcap,text width=3.80cm] at (13.88,-7.87) {Retrieve, recover, or synthesize};
% (4) Learning procedure
\userow{-8.17}{-10.72}{ArtRed}{4}{Learning procedure}{Collect, judge, update}
\foreach \x/\img/\lab in {3.775/art_outcome/Failure, 5.075/art_repair/Repair, 6.375/art_act/Effect}{
  \node[mini=ArtRed] (m\lab) at (\x,-9.15) {};
  \node[vig] at (\x,-8.97) {\includegraphics[height=1.06cm]{figures/\img}};
  \node[minilab] at (\x,-9.80) {\lab};}
\draw[-{Stealth[length=1.4mm,width=1.4mm]},line width=.7pt,draw=ArtRed!70] (mFailure.east)--(mRepair.west);
\draw[-{Stealth[length=1.4mm,width=1.4mm]},line width=.7pt,draw=ArtRed!70] (mRepair.east)--(mEffect.west);
\node[cellcap,text width=3.95cm] at (5.075,-10.56) {Failed grasps and their repairs};
\foreach \x/\ic/\lab in {8.455/database/Collect, 9.525/search/Judge, 10.595/wrench/Update}{
  \node[ibox] at (\x,-9.10) {};
  \node[text=ArtRed!85!black,font=\fontsize{17}{17}\selectfont] at (\x,-8.87) {\faIcon{\ic}};
  \node[minilab] at (\x,-9.67) {\lab};}
\node[cellcap,text width=3.30cm] at (9.525,-10.56) {Collector, evaluator, editor, updater};
\node[vig,anchor=south] at (13.88,-9.80) {\includegraphics[width=3.70cm]{figures/art_procedurew}};
\node[cellcap,text width=3.80cm] at (13.88,-10.56) {Choose experience, judge, or construct updates};
% Return to new physical experience
\node[art card=ArtSlate,font=\sffamily\bfseries\fontsize{8.4}{10}\selectfont,inner sep=2.5pt] (next) at (5.075,-11.38) {\faIcon{redo}\enspace New practice or physical experience};
\draw[art flow] (13.88,-10.72)--(13.88,-11.38)--(next.east);
\draw[art flow] (next.north)--(5.075,-10.72);
\end{tikzpicture}%
\caption{\textbf{Where a retained change is kept and how it is used.} Each row follows the grasp example from physical experience to the retained change and its immediate use; the italic line names the use. Dashed tags mark secondary handoffs into the learning procedure.}
\label{fig:taxonomy}
\end{figure}

%% file: tables/comparison_questions.tex
\begin{table}[!htbp]
\centering\footnotesize
\caption{The comparison each claim requires. All comparisons use matched learners and budgets; in R2, the updated improver is revised once with its acquired knowledge and once with that knowledge frozen, and the successor must also beat the updated improver itself (inequalities in Section~\ref{sec:key-terms}).}
\label{tab:comparison-questions}
\setlength{\tabcolsep}{4pt}\renewcommand{\arraystretch}{1.18}
\begin{tabularx}{\linewidth}{@{}L{3.4cm}L{6.0cm}Y@{}}
\paitoprule
\pahead{Claim} & \pahead{Question} & \pahead{Relevant comparison}\\
\paimidrule
Behavioral improvement & Does the system act better? & $J(L')$ vs $J(L)$; common task test\\
\paihair
Later-learning benefit & Does it help a later learner? & $J(P(L_0;A))$ vs $J(P(L_0))$; matched learners\\
\paihair
Procedure improvement & Does the new procedure teach better? & $J(P')$ vs $J(P)$; matched learners\\
\paihair
Recursion, R1 & Does repair knowledge help the improver? & Updated vs earlier improver, new failures\\
\paihair
Recursion, R2 & Does the improved improver improve itself? & Acquired vs frozen knowledge; vs parent\\
\paibottomrule
\end{tabularx}
\end{table}

%% file: tables/paradigms.tex
\begin{table}[!htbp]
\centering
\caption{How learning paradigms and problem settings relate to the claims of Section~\ref{sec:key-terms}. The last four columns show how each approach can support each claim; the R1, R2 column concerns whether the improvement mechanism itself improves.}
\label{tab:paradigms}
\begingroup\footnotesize
\setlength{\tabcolsep}{3pt}\renewcommand{\arraystretch}{1.12}\arrayrulecolor{PAIRule}
\begin{tabularx}{\linewidth}{@{}L{2.15cm}L{2.55cm}L{2.45cm}L{1.6cm}L{2.32cm}L{2.0cm}Y@{}}
\paitoprule
\pahead{Approach} & \pahead{Usually learned} & \pahead{Usually fixed} & \pahead{Behavior} & \pahead{Later learning} & \pahead{Procedure} & \pahead{R1, R2}\\
\paimidrule
Standard RL & Policy, value & Reward, algorithm & Typical & Data, policy & \nores{} & Improver's tool\\\paihair
Outer-loop RL & Reward, update rule & Outer search & Via learner & Reward, rule & Typical & Self-improving search\\\paihair
Self-training & Model, own outputs & Labeling, filtering, recipe & Typical & Generated data & Evaluator, if self-judging & Self-improving recipe\\\paihair
Continual learning & Model, task stream & Anti-forgetting and update rules & Per task; retention & Forward transfer during learning & \nores{} & Setting to handle\\\paihair
Meta-learning & Init., adaptation, loss, rule & Meta-training & Via learner & Initialization & Learned loss or rule & Self-improving meta-training\\\paihair
AutoML & Architecture, settings, algorithm & Search, or its trainer & Via learner & Configuration & Typical & Self-improving search\\
\paibottomrule
\end{tabularx}\endgroup
\end{table}

%% file: sections/07_evidence.tex
\subsection{Benchmarking Self-Improvement}
\label{sec:benchmarks}
\label{sec:metrics}
\begin{table}[!tbp]
\centering
\caption{Benchmarks used or proposed in the reviewed studies. \envhw\ hardware; \envsim\ simulation; entries without an icon use generated videos, a game, or digital tasks.}
\label{tab:benchmarks}
\begingroup\footnotesize
\setlength{\tabcolsep}{4pt}\renewcommand{\arraystretch}{1.12}\arrayrulecolor{PAIRule}
\begin{tabularx}{\linewidth}{@{}L{2.75cm}L{3.35cm}L{1.18cm}Y@{}}
\paitoprule
\pahead{Benchmark} & \pahead{Literature} & \pahead{Setting} & \pahead{What it tests}\\
\paimidrule
\addlinespace[3pt]\multicolumn{4}{@{}l@{}}{\tablegroup{Task suites}}\\\paihair
\textbf{LIBERO} & \citealp{arxiv230603310} & \envsim & Task sequences; transfer and forgetting\\\paihair
\textbf{LIBERO-PRO} & \citealp{arxiv251003827} & \envsim & Perturbed objects, initial states, instructions, environments\\\paihair
\textbf{LIBERO-Plus} & \citealp{arxiv251013626} & \envsim & Seven perturbations, e.g., camera, initial state\\\paihair
\textbf{RoboCasa} & \citealp{arxiv240602523} & \envsim & Everyday tasks, mainly kitchens\\\paihair
\textbf{RoboTwin 2.0} & \citealp{chen2025robotwin2} & \envsim & Bimanual manipulation\\\paihair
\textbf{RoboDojo} & \citealp{arxiv260704434} & \envsim\,\envhw & 42 bimanual sim, 18 real tasks\\\paihair
\textbf{VLABench} & \citealp{zhang2024vlabench} & \envsim & Language-conditioned manipulation suite\\\paihair
\textbf{Meta-World} & \citealp{yu2019metaworld} & \envsim & Multi-task and meta-RL manipulation\\\paihair
\textbf{SIMPLER} & \citealp{arxiv240505941} & \envsim & Sim counterparts of real setups\\\paihair
\textbf{Habitat, R2R-CE} & \citealp{savva2019habitat}; \citealp{krantz2020vlnce} & \envsim & Navigation\\\paihair
\textbf{EmbodiedBench} & \citealp{yang2025embodiedbench} & \envsim & MLLM agents, household to manipulation\\
\addlinespace[3pt]\multicolumn{4}{@{}l@{}}{\tablegroup{Component and protocol benchmarks}}\\\paihair
\textbf{FailBench} & \citealp{arxiv260903611} & \envsim\,\envhw & Success detectors, 2,197 attempts\\\paihair
\textbf{CaliBench} & \citealp{arxiv260816829} &  & Outcome calibration of video world models\\\paihair
\textbf{RLE-Bench} & \citealp{arxiv260934210} & \envsim & Coding agents; hidden perturbations\\\paihair
\textbf{RoboCoach-Bench} & \citealp{arxiv260121570} & \envsim & 32 tasks for training-code agents\\\paihair
\textbf{MC-EvoBench} & \citealp{arxiv261013829} &  & Transfer controls, Minecraft\\\paihair
\textbf{SEAR} & \citealp{arxiv261018792} & \envsim\,\envhw & Controls for retries, familiarity, privileged state, budget, chance\\\paihair
\textbf{Household} & \citealp{arxiv261022471} & \envsim & Prototype: no in-domain demos; running world clock\\\paihair
\textbf{PhysRSI-eval} & \citealp{arxiv261026848} & \envsim & Five axes of improvement; fixed interaction budget\\
\addlinespace[3pt]\multicolumn{4}{@{}l@{}}{\tablegroup{Digital benchmarks of recursive self-improvement}}\\\paihair
\textbf{PAST-Bench} & \citealp{arxiv260804003} &  & Retained experience on vs off\\\paihair
\textbf{RSIBench-Data} & \citealp{arxiv260725886} &  & Agent-revised data for a fixed model\\\paihair
\textbf{RSIGym} & \citealp{arxiv261010310} &  & Share of remaining gap closed\\
\paibottomrule
\end{tabularx}\endgroup
\end{table}

Most comparisons in this review run in simulation, and the benchmarks differ in what they stress (Table~\ref{tab:benchmarks}). LIBERO~\citep{arxiv230603310} supplies sequential manipulation suites, and LIBERO-PRO~\citep{arxiv251003827} perturbs manipulated objects, initial states, task instructions, and environments to test whether behavior depends on the original cues; LIBERO-Plus~\citep{arxiv251013626} varies seven factors, including camera viewpoint, initial robot state, and sensor noise. RoboCasa~\citep{arxiv240602523} broadens household task coverage, and RoboTwin~2.0~\citep{chen2025robotwin2} and RoboDojo~\citep{arxiv260704434} cover bimanual manipulation. VLABench~\citep{zhang2024vlabench} and Meta-World~\citep{yu2019metaworld} supply further manipulation suites, and SIMPLER~\citep{arxiv240505941} mirrors real robot setups for policy evaluation. Navigation studies use Habitat~\citep{savva2019habitat} and R2R-CE~\citep{krantz2020vlnce}, and household agents use EmbodiedBench~\citep{yang2025embodiedbench}. Except for LIBERO's lifelong protocol, which scores transfer and forgetting along a task sequence, these suites test a system at one point in time, and a claim about improvement across rounds still needs a protocol on top that fixes what carries over between rounds and what stays held out (Section~\ref{sec:accumulation}). A recent benchmark prototype moves in this direction: it provides no in-domain demonstrations, carries experience over between requests within a deployment, and names how quickly an agent becomes competent in a deployment as a quantity of interest~\citep{arxiv261022471}. PhysRSI-eval~\citep{arxiv261026848} fixes what carries over between rounds, in simulation. Six model-and-configuration systems revise a persistent workspace of code, tools and notes with frozen weights, for up to 17 graded rounds on each of 18 RoboDojo tasks, and each system is scored on five axes: first-attempt score, how early gains appear, attained score, physical budget saved, and the share of first-round failures later solved. The axes need not agree: Claude Opus 5.5 attains a higher score than GPT-6.1 Sol but scores lower on gain timing and budget saved. In a six-task ablation with GPT-6 Astra, withholding all information from earlier rounds lowers success over rounds 4--6 from 15 of 36 attempts to 3 of 36.

A second group of benchmarks targets the components on which self-improvement relies. FailBench~\citep{arxiv260903611} tests success detectors on 2,197 real and simulated manipulation attempts, where the best of 13 VLM detectors reaches 0.77 balanced accuracy. CaliBench~\citep{arxiv260816829} tests whether the outcome frequencies of video world models are calibrated. MC-EvoBench~\citep{arxiv261013829} fixes the controls that a transfer claim needs, and RLE-Bench~\citep{arxiv260934210} holds out robustness perturbations that coding agents cannot query. RoboCoach-Bench~\citep{arxiv260121570} collects 32 simulated tasks for agents that write and revise training code. Digital benchmarks of recursive self-improvement isolate related questions. PAST-Bench~\citep{arxiv260804003} turns retained experience on and off across ordered task sequences, and RSIBench-Data~\citep{arxiv260725886} has agents revise training-data strategies for a fixed model. RSIGym~\citep{arxiv261010310} measures the fraction of the remaining performance gap that agents close across benchmarks.

Hardware evaluations instead use task suites built by each study, and their budgets vary widely, from a handful of trials per task, as in PhysEvo's 25 trials over five real tasks~\citep{arxiv261008995}, to hundreds of evaluation episodes, as in the more than 600 pairwise episodes of RoboArena~\citep{arxiv250618123}. Most controlled downstream comparisons concern manipulation; navigation, locomotion, flight, and robotic laboratories support particular mechanisms without testing each condition across robot bodies (Table~\ref{tab:domain-scope}). Simulation results support mechanism comparisons at scale, whereas hardware results show what survives real contact, sensing, and resets (Section~\ref{sec:physical-consequences}). Because tasks, trial counts, and metrics differ, results are compared within each study rather than across studies; Appendix~\ref{app:reported-results} records the settings behind selected numbers.

Most studies report success rate, the fraction of trials that reach a task-defined goal as judged by a person, a scripted check, or a learned detector. Some give partial credit for progress through task stages, which needs a different statistical model than binary success. Others add efficiency during execution, such as cycle time in HIL-SERL~\citep{arxiv241021845} or throughput in RECAP~\citep{arxiv251114759}.

Learning efficiency is reported as the time, episodes, or supervision needed to reach a level of performance. ConRFT~\citep{arxiv250205450} reaches 96.3\% average success on eight real tasks after 45--90 minutes of online fine-tuning, and VLAC~\citep{arxiv250915937} raises real success from about 30\% to 90\% within 200 episodes. ENPIRE~\citep{arxiv260619980} reports the time its robot fleet needs to reach a target, and LMPC~\citep{arxiv240211450} counts chat turns per successful teaching session.

\emph{Retention} asks whether earlier competence survives later learning or deployment, and a later-learning benefit on a new task is \emph{forward transfer}. ACCRUE~\citep{arxiv261018596} reports transfer as a cut in real-robot training time for later tasks, CIDER~\citep{arxiv260821899} reports forward transfer of an accumulated policy, and MC-EvoBench sets out the controls that such a claim needs.

Resource use is counted in robot hours, demonstrations, interventions, compute, or model tokens. CARD~\citep{arxiv241014660} needs about 14K tokens per task against about 663K for Eureka, and ENPIRE's fleet scaling shortens the time to target at lower per-robot utilization and faster-than-linear token growth; because studies count different parts of this work, their costs are not directly comparable (Section~\ref{sec:cost-ledger}).

%% file: sections/03_policy_improvement.tex
\sectionaccent{PAIBlue}
\section{Mechanisms toward Physical RSI}
\label{sec:mechanisms}

This section reviews physical self-improvement by what a system actually changes, following the ladder of Section~\ref{sec:key-terms}. Section~\ref{sec:base-loop} covers changes to the learner $L$, kept in policies, world models, or external knowledge; a policy trained with relabeled targets, a reward, or an optimization algorithm belongs there, because those rules stay fixed. Section~\ref{sec:emerging-directions} covers changes to the rules themselves, the learning procedure $P$ that collects, judges, and updates, and Section~\ref{sec:reviser} covers changes to the improver $I^{K}$ that revises that procedure (notation of Section~\ref{sec:key-terms}). Section~\ref{sec:closing-loop} adds no fourth component: it asks whether these layers close into a recursive relation, in which the improver, once improved, helps produce its own next improvement. A system that serves several roles is discussed mainly where its experiment tests it, with cross-references elsewhere; Table~\ref{tab:rsi-systems} collects the systems that describe themselves as recursive self-improvement.

\subsection{Improving Policies, World Models, and Knowledge}
\label{sec:base-loop}

This section reviews the base loop of physical self-improvement: experience improves what the robot does, and the data, models, and skills it leaves behind can help later learners. These are the first two claims of Section~\ref{sec:key-terms}, behavioral improvement and later-learning benefit. We follow the change to where it is kept, in a policy (Section~\ref{sec:policy-improvement}), a world model (Section~\ref{sec:model-mediated}), or memories, skills, and code (Section~\ref{sec:modular-improvement}). The learning procedure that produces these changes stays fixed here; where a study also changes how experience is collected, judged, or turned into updates, that part appears in Section~\ref{sec:emerging-directions}.

\subsubsection{Policy Updates}
\label{sec:policy-improvement}

The most direct way for experience to help is to change the policy itself, and on the path to RSI the policy is also the most common target of a revised learning procedure. A deployed policy determines both what the robot does and which situations later appear in its training data, so the question to ask of new data is what learning signal they contain. Four kinds of signal recur. \emph{Relabeling} reinterprets behavior that has already happened: an attempt that pushed the cup aside instead of grasping it is still a correct example of how to push the cup aside. A \emph{reward} scores attempted outcomes so that optimization can prefer better ones. A \emph{correction} supplies an action that the policy did not take, typically from a person who intervenes. \emph{Replay} reuses earlier experience to preserve or recover previously learned behavior (Figure~\ref{fig:teaching-signals}). Each signal fits a different failure.

\input{figures/teaching_signals}

\paragraph{From practice to training targets.}
\label{sec:targets}

A recorded trajectory becomes useful for learning only once a goal, reward, or action label is attached to it. Hindsight experience replay~\citep{arxiv170701495} relabels an attempt with the goal it actually achieved, and SOAR~\citep{arxiv240720635} carries this achieved-goal relabeling to goal-conditioned imitation. Earlier large-scale systems such as MT-Opt~\citep{kalashnikov2022mtopt} shared physical outcomes across twelve tasks; pretrained generalists change the starting point, because practice can refine behavior that is already useful, and value-based fine-tuning must then keep offline estimates calibrated, as Cal-QL~\citep{nakamoto2023calql} does in simulation. RoboCat~\citep{arxiv230611706} closes the resulting cycle: a generalist adapted to a target task from demonstrations collects trajectories that, mixed with retained multi-task data, retrain the generalist offline. RoboCat also relabels unsuccessful episodes with the goals they reached, whereas SOAR retrains only on attempts that its fixed VLM judges successful (Appendix~\ref{app:reported-results}).

Adding more of the robot's own data does not necessarily improve the targets it learns from. In their study of mostly single-task policies on five simulated and two real manipulation tasks, \citet{arxiv241101813} find that added autonomous data yield modest gains that saturate and, over two to four collection rounds, can regress. On two of the tasks, human demonstrations started from conditions where rollouts had failed help more than randomly chosen ones, whereas autonomous collection from the same conditions offers no comparable advantage. More state diversity in demonstration data does not always help, whereas inconsistent actions often hurt~\citep{belkhale2023quality}. These results separate two needs, \emph{finding} the difficult conditions and \emph{supplying useful actions} there; repeating the policy's familiar successes addresses neither.

\paragraph{From feedback to better actions.}
\label{sec:corrections}

Relabeling cannot supply an action the policy never produced, but an incomplete attempt can still reveal partial progress, and a person can replace part of it with a corrective motion. SI-EFM~\citep{arxiv250915155} takes the first route: a frozen model of the steps remaining to a goal rewards each step that brings the robot closer, and policy gradients then learn from incomplete real-robot attempts without new demonstrations, though only among variations that the policy actually tries. FAR~\citep{arxiv260701111} uses failed attempts in a different way: it turns them into preference data that steers retries away from the failed behavior and adds successful recoveries to later training, with gains in simulation and on real manipulation tasks.

Corrections take the second route. HIL-SERL~\citep{arxiv241021845} builds on SERL~\citep{luo2024serl}, which combines off-policy learning, learned reward classifiers, and reset policies, and adds online human interventions so that corrective transitions inform the critic that guides the actor. RECAP~\citep{arxiv251114759} conditions the policy on the advantage of each action, fitted from deployment trajectories and expert corrections. Max-Q selective imitation~\citep{arxiv260815088} regresses the actor toward whichever action a Monte Carlo chunk critic scores higher: its own action or a replay-buffer action, which is often an intervention. The target therefore shifts from corrections to the policy's own behavior as it improves. On real USB insertion, it reaches 99\% success after half an hour, whereas HIL-SERL needs about five hours to reach its 100\% ceiling (each at its best checkpoint). SiLRI~\citep{arxiv251224288} treats human interventions as possibly suboptimal: it poses real-world RL as a constrained problem whose bound at each state is set by the uncertainty of the interventions, solved with a state-wise Lagrange multiplier, and reaches 90\% success on real manipulation tasks in at most half the time HIL-SERL needs. The same recorded transition can become a value target, a weighted action target, or a conditioning variable, as RAPolicy~\citep{arxiv260922888}, SI-EFM, and RECAP illustrate, and which of these the learner receives changes what it learns.

Two studies keep a pretrained VLA frozen during online learning and train an added module alongside it. BEE~\citep{arxiv260927450} uses the consistency of human corrections to constrain RL updates to a residual policy around a frozen VLA. On three real tasks and one simulated task, it reaches 91.2\% average success against 57.5\% for an RL baseline at a matched online-data budget. RL Token~\citep{arxiv260423073} instead adapts a VLA to expose a compact readout, keeps it frozen, and trains a small actor--critic on that readout online. On four real tasks, minutes to a few hours of practice speed up the hardest phase by up to 3$\times$ and raise success as judged by a human operator, for example from 20\% to 65\% on screw installation. Outside manipulation, SELFI~\citep{arxiv240300991} fine-tunes a pretrained navigation policy with online model-free RL whose Q-values keep the model-based objective of its offline pretraining, improving collision avoidance and, in a user study, social compliance in real environments.

Reward-guided post-training of VLAs has scaled mainly in simulation, while its most reliable results come from smaller hardware studies. VLA-RL~\citep{arxiv250518719} on LIBERO~\citep{arxiv230603310} and $\pi_{\texttt{RL}}$~\citep{arxiv251025889} run in simulation, and FLaRe~\citep{arxiv240916578} and SimpleVLA-RL~\citep{arxiv250909674} add real-robot results. In eight-task hardware studies, ConRFT~\citep{arxiv250205450} reaches 96.3\% average success after 45--90 minutes of online fine-tuning and RL-100~\citep{arxiv251014830} succeeds in 1,000 of 1,000 evaluation episodes. Task-specific hardware studies such as GR-RL~\citep{arxiv251201801} for shoe lacing and VLA-Precision~\citep{arxiv260904355} for chemistry tasks push precision further (Appendix~\ref{app:reported-results}). Autonomous real-world RL also reaches beyond tabletop arms: on Spot mobile manipulators, task-relevant autonomy, behavior priors, and generic rewards reach 80\% average success on four tasks~\citep{mendonca2024continuously}.

Without per-task success checks, such training needs rewards that generalize, and the stricter test of a learned reward is the behavior it teaches rather than its prediction accuracy. Pretrained vision and language models supply them as zero-shot similarities in VLM-RM~\citep{rocamonde2024vlmrm} and RoboCLIP~\citep{sontakke2023roboclip}, both evaluated in simulation, or as embedding distances learned from human video in VIP~\citep{ma2023vip} and LIV~\citep{ma2023liv}, both with real-robot evaluations. ReWiND~\citep{arxiv250510911}, RoboReward~\citep{arxiv260100675}, and VLAC~\citep{arxiv250915937} judge learned rewards by how they transfer and by the behavior they teach; VLAC raises real success on four tasks from about 30\% to 90\% within 200 episodes. Finer progress distinctions preserve more of what incomplete attempts teach: SARM2~\citep{arxiv260610305} conditions its reward model on the estimated task stage, adapts it once, and holds it fixed while the SPIRAL loop learns residual corrections to a frozen VLA. On crumpled garments, three SPIRAL rounds raise success from 4/12 to 8/12, while the matched loop with only a sparse success reward falls from 4/12 to 2/12.

\paragraph{From proposals to policies.}
\label{sec:selection-expansion}

Feedback that can rank behavior leaves a second choice: where the improvement should live. Choosing among proposals can improve execution without retraining, as V-GPS~\citep{arxiv241013816} does by reranking a frozen policy's proposals with an offline value function; PA-RL~\citep{arxiv241206685} also improves sampled actions under a critic and trains the policy on them with a supervised loss. Learned judges can steer a frozen policy in the same way. UF-OPS~\citep{arxiv260310282} trains success and time-to-success verifiers once on the policy's labeled rollouts and selects among sampled action chunks at run time, with reported gains of 25 to 80 percentage points on five real Aloha tasks. World-Coherent Decoding~\citep{arxiv260902159} learns during deployment to rank a frozen world action model's sampled futures by the mismatch between predicted and observed frames. In RoboTwin~2.0 simulation, under the authors' limited-randomization protocol, it raises Hard-split success from 55.80\% to 60.90\%. Keeping scoring, improvement, and policy change separate avoids the need to adapt policy gradients to each action representation, whether iteratively denoised chunks in Diffusion Policy~\citep{arxiv230304137}, discretized tokens in OpenVLA~\citep{arxiv240609246}, or flow matching in $\pi_0$~\citep{arxiv241024164}. Selection gains can stay in the selector or be absorbed into the generator. VERITAS~\citep{arxiv260618247} behavior-clones trajectories steered toward agreement with a visual path proposed by a VLM. Q-Planning~\citep{arxiv260821204} instead leaves the cloned policy unchanged and updates a Q-function from successful and failed rollouts. On its two real-robot tasks, terminal rewards come from human episode labels and success is measured on the collection episodes themselves.

Both routes stay bounded by their proposals: if the generator never produces a grasp that avoids the slip, no selector can recover one. A \emph{residual} policy modifies the proposed motion while keeping the base policy as a prior. ResiP~\citep{arxiv240716677} trains one in simulation and then distills it into a vision policy co-trained with real demonstrations. DSRL~\citep{arxiv250615799} instead steers the input noise of a frozen diffusion policy, searching only within its learned distribution. DPPO~\citep{ren2025dppo} fine-tunes the diffusion policy itself and transfers a simulation-trained assembly policy to hardware. Expanding proposals helps only if the resulting experience teaches the policy where it needs help, which PLD~\citep{arxiv251100091} tests directly. A residual-RL specialist takes over after the generalist has run for a sampled number of steps, and the hybrid trajectories are distilled back. In real cube pickup, a PLD specialist and an RLPD~\citep{ball2023rlpd} specialist both reach full success within two hours from the same initialization, and each supplies 200 successful trajectories to the same distillation. The resulting generalists nevertheless differ, succeeding in 30 and 16 of 30 trials, respectively, because only the PLD data include recoveries from the corner states that the learner itself visits; collector success alone therefore does not reveal training value. Delay creates a further mismatch: Real-Time EXPO-FT~\citep{arxiv260918207} edits the action chunks of a slower VLA with the latest observation and selects among candidates with a critic, because an action can become wrong between prediction and execution. Selection suffices when useful behavior already lies in the proposal set; otherwise the generator must be optimized, corrected, steered, or fine-tuned, and distillation retains corrections that should become habitual (Table~\ref{tab:policy-systems-comparison}).

\input{tables/policy_systems_comparison}

\paragraph{From single runs to lifelong practice.}
\label{sec:sharing}

At scale, RoboNet~\citep{dasari2020robonet} pools trajectories across laboratories and embodiments, and Learning While Deploying~\citep{arxiv260500416} repeatedly updates one generalist from rollouts and interventions across sixteen robots. SOP~\citep{arxiv260103044} streams on-policy experience and human interventions from a fleet to a cloud learner and reports real-world performance that scales near-linearly with the number of robots. OpenBot-Fleet~\citep{arxiv240507515} sends learned navigation policies back to user-operated robots. More sources widen coverage only where observations, actions, and task definitions align. Practice also needs a way back to a useful starting state: Leave No Trace~\citep{arxiv171106782} learns forward and reset policies in simulation, ReLMM~\citep{sun2022relmm} learns reset-free on a physical mobile manipulator, and MEDAL++~\citep{sharma2023selfimproving} and RoboFuME~\citep{yang2023robofume} need only occasional manual resets on real robots.

Against forgetting, capacity can be allocated or routed, as AtomicVLA~\citep{arxiv260307648} grows expert modules for atomic skills and CLARE~\citep{arxiv260109512} expands and selects adapters without task labels, or adaptation can be combined with rehearsal, as in LifelongVLA~\citep{arxiv260714852} and iManip~\citep{arxiv250307087}, or with targeted replay, as in Memory Anchors~\citep{arxiv260826545}. Separation costs capacity and routing; retaining experience, as in gradient episodic memory~\citep{lopezpaz2017gem}, costs storage but can also speed later tasks~\citep{xie2022lifelong}. CIDER~\citep{arxiv260821899} freezes the accumulated policy as a teacher before each new task and alternates current-task RL with distillation on archived states; over six real tasks and three runs each, the retained policy needs no more training steps to reach the success criterion than the same RL agent from the common initialization, and fewer on some tasks. Replay and distillation assume a fixed action interface: FAN~\citep{arxiv260921358} shows that updating action-normalization statistics between tasks shifts the coordinates shared by training and execution; statistics frozen after a small task-independent calibration give the highest average score across four real task streams.

\begin{paiinsight}[PAISyn1]{Different failures call for different teaching signals}
\begin{insightpoints}
\item \textbf{Match the signal to the failure.} More successes supply neither a missing action nor a finer distinction.
\item \textbf{The signal must reach an executable action.} An optimizer, selector, or action interface must absorb it, and changed behavior calls for rechecking coverage, retention, and the interface.
\item \textbf{Behavioral gains are well established; later-learning benefits are rarer.} Hardware comparisons establish behavioral gains; RoboCat, PLD, and CIDER also report later-learning benefits.
\item \textbf{Choose the signal from recorded failures.} Compare a procedure that picks relabeling, a better reward, a correction, or replay from recorded failures with a fixed choice under the same budget (Sections~\ref{sec:select-experience} and~\ref{sec:open-challenges}).
\end{insightpoints}
\end{paiinsight}

%% file: figures/teaching_signals.tex
\begin{figure}[!tbp]
\centering
\begin{tikzpicture}[art fonts,x=1cm,y=1cm,line cap=round,line join=round,
  every node/.append style={execute at begin node={\hyphenpenalty=10000\exhyphenpenalty=10000}},
  cellcap/.style={inner sep=0pt,align=center,text=ArtSlate,font=\sffamily\fontsize{8.2}{9.6}\selectfont,anchor=south,
    execute at end node={\vphantom{g}}},
  effect/.style={inner sep=0pt,align=center,text=ArtBlue!85!black,anchor=north,
    font=\sffamily\bfseries\fontsize{8.8}{10.4}\selectfont,execute at end node={\vphantom{g}}},
  ex/.style={inner sep=0pt,align=center,text=ArtSlate!85,anchor=north,font=\rmfamily\itshape\fontsize{8.0}{9.4}\selectfont},
  vig/.style={inner sep=0pt}]
\path[use as bounding box] (0,-5.24) rectangle (16.08,.02);
% Four column panels in the section color; each holds a white card (experience and caption),
% the change the signal makes to the training data, and representative systems from the section.
\foreach \n/\xa/\ttl/\img/\ih/\cap/\eff/\exs in {
  1/.02/Relabeling/art_relabel/2.28cm/{Pushed aside, not grasped}/{Same behavior, new target}/{e.g., hindsight replay, SOAR},
  2/4.06/Reward/art_reward/2.28cm/{A lift that slips partway}/{Score attempted progress}/{e.g., SI-EFM, SARM2},
  3/8.10/Correction/art_correct/2.28cm/{A person adds the recovery}/{Supply a missing action}/{e.g., HIL-SERL, Max-Q},
  4/12.14/Replay/art_replay/1.98cm/{Earlier and new experience}/{Retain earlier behavior}/{e.g., Memory Anchors, CIDER}}
  {\pgfmathsetmacro{\xb}{\xa+3.92}\pgfmathsetmacro{\cx}{\xa+1.96}
   \artpanel[\n]{\xa}{-4.48}{\xb}{0}{ArtBlue}{\ttl}
   \path[fill=white,draw=ArtBlue!38,line width=.55pt,rounded corners=1.6mm,
     drop shadow={opacity=.12,shadow xshift=.5pt,shadow yshift=-.6pt}] (\xa+.12,-3.40) rectangle (\xb-.12,-.54);
   \node[vig,anchor=south] at (\cx,-2.92) {\includegraphics[height=\ih,width=3.46cm,keepaspectratio]{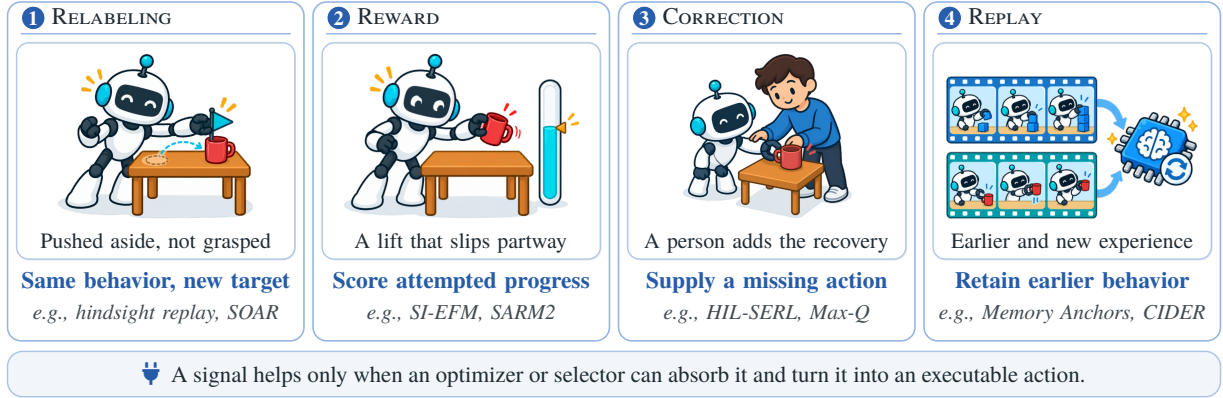}};
   \node[cellcap,text width=3.50cm] at (\cx,-3.32) {\cap};
   \node[effect,text width=3.70cm] at (\cx,-3.56) {\eff};
   \node[ex,text width=3.70cm] at (\cx,-4.00) {\exs};}
% Shared condition, from the opening of the section.
\path[fill=ArtBlue!6,draw=ArtBlue!40,line width=.55pt,rounded corners=2mm] (.02,-5.22) rectangle (16.06,-4.62);
\node[inner sep=0pt,text=ArtSlate,font=\sffamily\fontsize{8.6}{10}\selectfont] at (8.04,-4.92)
  {\textcolor{ArtBlue!85!black}{\faIcon{plug}}\enspace A signal helps only when an optimizer or selector can absorb it and turn it into an executable action.};
\end{tikzpicture}%
\caption{\textbf{Four teaching signals from physical experience.} Each column follows the grasp example: what was experienced, what the signal changes in the training data, and representative systems from this section.}
\label{fig:teaching-signals}
\end{figure}

%% file: tables/policy_systems_comparison.tex
\begin{table}[!htbp]
\centering
\caption{Policy updates and selected within-study comparisons. Outcome checks give the setting, result, and scorer where identified; unresolved scorers are listed in Section~\ref{sec:outcome-channels}. \updmark\ updated; \fixmark\ held fixed.}
\label{tab:policy-systems-comparison}
\begingroup\footnotesize
\setlength{\tabcolsep}{4pt}\renewcommand{\arraystretch}{1.20}\arrayrulecolor{PAIRule}
\begin{tabularx}{\linewidth}{@{}L{2.55cm}L{2.5cm}L{2.4cm}L{3.0cm}Y@{}}
\paitoprule
\pahead{System} & \pahead{Components} & \pahead{Update signal} & \pahead{Outcome check} & \pahead{Comparison}\\
\paimidrule
\systemcite{RoboCat}{arxiv230611706} & \updmark Generalist policy \fixmarkn training recipe & Reward model; hindsight goals & \envhw\ Human counts; learned filter separate & Demos vs added self-generated data \\
\paihair
\systemcite{SOAR}{arxiv240720635} & \updmark Policy \fixmarkn proposer, detector, editor & VLM filtering; hindsight goals & \envhw\ Task success & Pretrained vs autonomous-data co-training \\
\paihair
\systemcite{SI-EFM}{arxiv250915155} & \updmark Policy \fixmarkn progress model & Steps-to-go reward & \envhw\ Task success & Behavior cloning vs reward-guided learning \\
\paihair
\systemcite{HIL-SERL}{arxiv241021845} & \updmark Actor, critics \fixmarkn success classifier & Reward; human corrections & \envhw\ Task success & RL vs corrective imitation; intervention ablations \\
\paihair
\systemcite{RECAP}{arxiv251114759} & \updmark VLA, value model \fixmarkn recipe & Outcomes, fitted value & \envhw\ Rater judgments as success labels & Advantage conditioning vs weighted or policy-gradient fitting \\
\paihair
\systemcite{RAPolicy}{arxiv260922888} & \updmark Actor, critic \fixmarkn RL recipe & Values; interventions & \envhw\ Unassisted collection, final trials & Replay-anchored vs predicted-action targets \\
\paihair
\systemcite{SARM2/\allowbreak SPIRAL}{arxiv260610305} & \updmark Residual policy \fixmarkn reward, base VLA & Stage-aware reward & \envhw\ Success/progress rubric & Stage-aware vs sparse rewards \\
\paihair
\systemcite{VERITAS}{arxiv260618247} & \updmark Policy \fixmarkn visual critic & Steering, filtering & \envsim\,\envhw\ Task success & Verifier-curated vs human data, matched size \\
\paihair
\systemcite{Q-Planning}{arxiv260821204} & \updmark Q-function \fixmarkn base policy & Env. success; human labels & \envsim\,\envhw\ Env. success; human-labeled collection episodes & Base vs fine-tuning on successes vs Q-guided; 10 sim, 5 real rounds \\
\paihair
\systemcite{PLD}{arxiv251100091} & \updmark Specialists, generalist \fixmarkn distillation & Task reward; learned success classifiers & \envsim\,\envhw\ Task success & Specialist collectors; matched distillation \\
\paihair
\systemcite{CIDER}{arxiv260821899} & \updmark Actor, critics, adapter \fixmarkn RL/distillation & Classifier; interventions & \envhw\ Old/new-task trials & Continual vs single-task start; old-task retests \\
\paibottomrule
\end{tabularx}\endgroup
\end{table}

%% file: sections/04_model_mediated.tex
\subsubsection{World Models for Prediction and Practice}
\label{sec:model-mediated}

Physical trials are slow and costly, so a natural way to stretch them is to learn a model of the world and practice inside it. A world model predicts how a physical state may change after an action; for a grasp, it might predict whether the object will stay between the fingers when the gripper closes. The prediction can serve as a \emph{plan} that must be translated into motion, as \emph{imagined experience} from which a policy is trained, or as \emph{predicted features} supplied to a policy. Each role tests a different form of \emph{usability}: success in one role says little about the others. Generative video models offer all three interfaces: a generated video serves as a plan in UniPi~\citep{du2023unipi}, as a controllable simulator in UniSim~\citep{yang2024unisim} and, learned without action labels, in Genie~\citep{bruce2024genie}, and as policy-training data in DreamGen~\citep{jang2025dreamgen}. Cosmos~\citep{nvidia2025cosmos} develops world foundation models for physical applications, WISA~\citep{arxiv250308153} and PISA~\citep{arxiv250309595} target physical consistency, and SPIRAL~\citep{arxiv260308403} (unrelated to the SPIRAL policy loop of Section~\ref{sec:corrections}) updates its generator with a critic that checks action adherence and physical realism. A world model can also enter the learning procedure by judging candidate policies or changes before they reach hardware (Sections~\ref{sec:revise-supervision} and~\ref{sec:recursion-routes}).

\paragraph{Using models to plan and train.}
\label{sec:plan-or-train}

Classical model-based methods supplied the main ingredients. Dyna~\citep{sutton1990dyna} interleaves real and modeled experience, PILCO~\citep{deisenroth2011pilco} and PETS~\citep{arxiv180512114} account for model uncertainty, MBPO~\citep{arxiv190608253} keeps imagined rollouts short to limit compounding error, self-supervised visual planning~\citep{ebert2017visualplanning} plans through learned video prediction, and latent world models~\citep{ha2018worldmodels,hafner2020dreamer} train policies in compact dynamics, as DayDreamer~\citep{wu2023daydreamer} did on physical robots. A model's value for control need not track its prediction accuracy~\citep{lambert2020mismatch,farahmand2017value,grimm2020equivalence}.

Planning first exposes an interface problem: a predicted video of the gripper closing is not yet a robot command. SILVR~\citep{arxiv250606658} translates video plans through a separately trained inverse-dynamics model, and on hardware its successful executions update only the task video planner. In its simulated Meta-World~\citep{yu2019metaworld} ablation over twelve unseen tasks, updating both planner and translator reaches 44.2\% success, against 26.8\% with the translator frozen: an improving planner can be held back by a translator that cannot express its new motions. The hardware counterpart of this comparison remains to be run.

A model used for training shifts the burden to policy optimization, and its gain depends on how far the policy may optimize against the model. RISE~\citep{arxiv260211075} freezes its learned dynamics and value models, limits imagined interactions to two steps, and updates an advantage-conditioned VLA; its hardware comparison reimplements RECAP~\citep{arxiv251114759} with the same $\pi_{0.5}$~\citep{black2025pi05} initialization, because the $\pi_{0.6}$ model used by RECAP is not available. PlayWorld~\citep{arxiv260309030} trains only the small DSRL~\citep{arxiv250615799} policy that chooses its generator's input noise against the fitted world model, rewarded by a progress estimator trained on demonstrations; with twenty hardware trials per checkpoint, its best reported task improves by 65 percentage points over the base policy, yet later optimization inside the model can reduce real success. WMPO~\citep{arxiv251109515}, Ctrl-World~\citep{arxiv251010125}, ProphRL~\citep{arxiv251120633}, and World-Gymnast~\citep{arxiv260202454} likewise improve VLAs with imagined rollouts. Fine-tuning on imagined successes raises Ctrl-World's policy success by 44.7 percentage points, and ProphRL, which runs flow-policy GRPO inside an action-conditioned video model pretrained on more than 31M trajectories, raises overall success on four real UR30e tasks by 24.6 to 30.0 percentage points over the starting policies of three VLA backbones (Appendix~\ref{app:reported-results}). Whether new physical experience should also repair the practice model is a further choice, one that changes what later updates learn from; Section~\ref{sec:reuse-across-rounds} follows VLAW's loop of this kind. Figure~\ref{fig:model-learning-loops} separates these decisions, and Table~\ref{tab:model-mediated-comparison} records what the main model-assisted studies update, hold fixed, and compare.

\input{tables/model_mediated_comparison}
\input{figures/model_learning_loops}

\paragraph{Feeding models the right data.}
\label{sec:coverage}

Whether new rollouts can repair a model depends on what they cover for the component that consumes them. A contact model needs partial contacts, slips, and recoveries, not only successes: PlayWorld collects such evidence through autonomous play, SILVR through task-directed rollouts from its current planner, and Motus2~\citep{arxiv260830237} routes curated successes to action generation and failed or suboptimal interactions to dynamics or value prediction, freezing its video backbone and evaluator during later policy optimization. When natural failures miss the needed corrections, XPACE~\citep{arxiv260917372} constructs them: its world model renders deviations from expert demonstrations followed by recoveries, and fine-tuning on a mixture containing 8\% of these examples yields hardware gains across three tasks evaluated with twenty trials per method. A failed contact may be valuable for predicting consequences, while only a validated recovery should supervise the action policy.

Dynamics can also change. Roughly 100 seconds of real quadrotor flight support online identification of a residual dynamics model and in-flight policy updates~\citep{arxiv260210111}; Variational Neural Dynamics~\citep{arxiv260627353} instead infers which recurring operating condition applies from recent interaction; and Self-adapting Robotic Agents~\citep{arxiv260304029} detect out-of-distribution events from prediction residuals and trigger fine-tuning. D-World~\citep{arxiv261017114}, which its authors describe as recursive self-improvement, keeps an explicit digital twin that a frozen language model revises after each executed action chunk and accepts a revision only when replayed residuals do not grow; on a real dual-arm robot, one revision reduced a rendered-depth residual from 22.4 to 1.5\,mm. After damage, \citet{bongard2006selfmodel} revise a model of the robot's body and search it for a new gait, Intelligent Trial and Error~\citep{cully2015adaptation} searches a precomputed repertoire of behaviors, and GrBAL~\citep{arxiv180311347} adapts a meta-learned dynamics prior (Section~\ref{sec:meta-learning}). Which route saves physical trials depends on what useful structure survives the change.

\paragraph{Shaping what models predict.}
\label{sec:preserve}

For control, a representation must distinguish futures that imply different actions, which pixel-accurate prediction neither guarantees nor requires. Joint-embedding predictive architectures predict in learned feature spaces and plan toward goals there, as V-JEPA~2-AC~\citep{arxiv250609985} and DINO-WM~\citep{zhou2025dinowm} do, and D-JEPA~\citep{arxiv260924749} aligns competing predicted futures with the candidate that executes best. Physical properties must remain readable: in PokeWorld~\citep{arxiv260727017}, predicting touch makes stiffness decodable whereas merely supplying touch as an input does not, and PSG-JEPA~\citep{arxiv260806799} adds proprioceptive objectives where forward prediction alone omits such structure. DreamerV3~\citep{hafner2025dreamerv3}, which reconstructs observations, and TD-MPC2~\citep{hansen2024tdmpc2}, which does not, bracket the design space. Retrieved history in Mem-World~\citep{arxiv260618960}, force--torque prediction in MuSe~\citep{arxiv260630988}, and predicted fingertip wrenches in TORL-VLA~\citep{arxiv260609337} supply distinctions that camera data alone can miss.

The information must also reach the policy through a usable interface, and the model must respond correctly to commands. GigaBrain-0.5M*~\citep{arxiv260212099} conditions action learning on predicted future-state tokens, whereas VLAW turns predicted futures into reward-filtered training data; a conditioned policy may ignore or over-rely on auxiliary tokens, whereas synthetic data may amplify prediction errors. WorldEcho~\citep{arxiv260824885} shows that models trained only on expert trajectories, given commands outside that distribution, either ignore the commanded motion or produce visually invalid rollouts, and CausalWorld~\citep{ahmed2021causalworld} lets physical and visual variables be changed independently. Across embodiments, particle-based action representations~\citep{arxiv251101177} offer a shared geometric interface, while XEWorld~\citep{arxiv260805799} finds that current models bind joint commands more readily to familiar appearance than to transferable kinematics. A visually plausible model may still lack decision-relevant state, a usable interface, or correct responses to intervention.

\paragraph{Judging models by their use.}
\label{sec:test-model}

Failure can enter at four links between a model and its policy: transition prediction, outcome scoring, policy optimization, or translation into executable motion. End-to-end hardware results, as in RISE and PlayWorld, measure joint usefulness but do not say which link broke, so evaluation needs both final performance and comparisons that change one link at a time. DexTouch-WM~\citep{arxiv260920649} shows why. Its world model trained on mixed human and robot data improves held-out prediction and, on average, agreement between policy rankings in imagination and on hardware, yet replacing 100 of 200 real training trajectories with synthetic counterparts generated by this model lowers the mean physical partial-credit score for each of three policy families (FTP-1, $\pi_{0.5}$, and X-VLA) across four tasks. A model can become a better surrogate evaluator and still produce a worse training mixture.

Other methods change how predictions are consumed. WISE~\citep{arxiv260903681} calls a frozen world model only at states where interaction matters and bounds the imagined horizon, and HaWMPO~\citep{arxiv260909941} trains a hallucination estimator on discrepancies between generated and observed outcomes and uses it to soften rewards from unreliable action chunks. How prediction is queried also matters: in CaliBench~\citep{arxiv260816829}, individually plausible videos can have poorly calibrated outcome frequencies, and in a simulated differential-drive study~\citep{arxiv260902811}, a 20-step rollout without intermediate measurements picks a different state estimator than closed-loop tracking in 18 of 24 sensing conditions, whereas replay error under the recorded schedule disagrees in five. Evaluation should reproduce the actions, horizon, and observation schedule of the policy's use.

\begin{paiinsight}[PAISyn1]{A useful model preserves distinctions the controller can exploit}
\begin{insightpoints}
\item \textbf{Cover and preserve what the decision requires.} Physical data must reach the transitions, contacts, and recoveries the consumer lacks, and prediction must preserve distinctions that change outcomes.
\item \textbf{Prediction helps only through its consumer.} Predictive gains help control only through a suitable evaluator, learner, and action interface.
\item \textbf{Value belongs to the full chain.} Hardware gains such as VLAW's belong to combined pipelines, and DexTouch-WM's better surrogate evaluator yields worse training data.
\item \textbf{Let outcomes select the repair.} Refitting the model, restricting imagination, and downweighting unreliable predictions each sit in a different system; Section~\ref{sec:agenda-diagnosis} asks how physical outcomes could choose among them.
\end{insightpoints}
\end{paiinsight}

%% file: tables/model_mediated_comparison.tex
\begin{table}[!htbp]
\centering
\caption{Model-assisted learning. Columns follow Table~\ref{tab:policy-systems-comparison}.}
\label{tab:model-mediated-comparison}
\begingroup\footnotesize
\setlength{\tabcolsep}{4pt}\renewcommand{\arraystretch}{1.20}\arrayrulecolor{PAIRule}
\begin{tabularx}{\linewidth}{@{}L{2.25cm}L{3.35cm}L{3.1cm}L{2.55cm}Y@{}}
\paitoprule
\pahead{System} & \pahead{Components} & \pahead{Update signal} & \pahead{Outcome check} & \pahead{Comparison}\\
\paimidrule
\systemcite{SILVR}{arxiv250606658} & \updmark Video planner \fixmarkn prior, inverse dynamics & Filtered successes & \envhw\ Human judge, also filters rollouts & \envhw\ Planner rounds; \envsim\ inverse model frozen vs updated \\
\paihair
\systemcite{RISE}{arxiv260211075} & \updmark Policy \fixmarkn dynamics, value model & Imagined advantage & \envhw\ Task success & vs RECAP reimplementation; matched $\pi_{0.5}$ \\
\paihair
\systemcite{PlayWorld}{arxiv260309030} & \updmark Noise policy \fixmarkn reward, base policy & Imagined progress & \envhw\ Task success per checkpoint & vs demo-trained world model; later checkpoints can regress \\
\paihair
\systemcite{VLAW}{arxiv260212063} & \updmark World model, reward model, policy & Real outcomes; filtered imagined successes & \envhw\ Final-state task criteria & Full loop vs Filtered BC, DSRL; matched rollouts \\
\paihair
\systemcite{DexTouch-WM}{arxiv260920649} & \updmark Model, then policy \fixmarkn training recipes & Generated observations & \envhw\ Partial-credit task scores & All-real vs half-generated; same data size; three learners \\
\paibottomrule
\end{tabularx}\endgroup
\end{table}

%% file: figures/model_learning_loops.tex
\begin{figure}[!tbp]
\centering
\begin{tikzpicture}[art fonts,x=1cm,y=1cm,line cap=round,line join=round,
  every node/.append style={execute at begin node={\hyphenpenalty=10000\exhyphenpenalty=10000}},
  card/.style={fill=white,execute at end node={\vphantom{g}},rounded corners=1.6mm,text width=2.63cm,minimum height=.95cm,align=center,inner sep=4pt,text=ArtSlate,
    font=\sffamily\fontsize{8.2}{9.9}\selectfont,drop shadow={opacity=.16,shadow xshift=.6pt,shadow yshift=-.8pt}},
  tall/.style={minimum height=1.25cm},
  updated/.style={card,draw=ArtTeal,line width=1pt},
  fixed/.style={card,draw=ArtSlate!40,line width=.6pt,fill=ArtGray!30},
  physical/.style={card,draw=ArtBlue!70,line width=.6pt},
  role/.style={circle,fill=#1,draw=white,line width=.8pt,minimum size=.30cm,inner sep=0pt,text=white,font=\fontsize{5.2}{5.2}\selectfont},
  flow/.style={-{Stealth[length=2.4mm,width=2.2mm]},line width=1pt,draw=ArtSlate!70},
  back/.style={draw=#1!85!black,line width=1.1pt,rounded corners=2mm,-{Stealth[length=2.4mm,width=2.2mm]}},
  tag/.style={art note,fill=white},
  vig/.style={inner sep=0pt}]
% Role badges on each card's corner: updated (sync), fixed (lock), physical data or tests (robot).
\def\roleU#1{\paiBadgeAt{([xshift=.03cm,yshift=-.03cm]#1.north west)}{ArtTeal}{\paiGlyphCycle}}
\def\roleF#1{\paiBadgeAt{([xshift=.03cm,yshift=-.03cm]#1.north west)}{ArtSlate!55}{\paiGlyphPin}}
\def\roleP#1{\paiBadgeAt{([xshift=.03cm,yshift=-.03cm]#1.north west)}{ArtAmber}{\paiGlyphArm}}
\path[use as bounding box] (0,-11.55) rectangle (16.06,1.0);
% (a) SILVR
\artpanel[a]{0}{-1.75}{16.04}{.86}{ArtBlue}{SILVR: collect on the real robot, then revise the planner}
\node[vig] at (1.32,-.655) {\includegraphics[width=2.33cm,height=1.98cm,keepaspectratio]{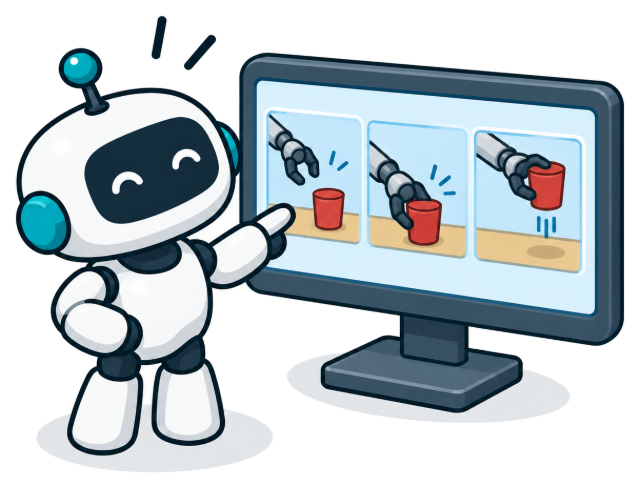}};
\node[fixed] (sp) at (4.205,-.30) {Generic video prior\\fixed guidance};
\node[updated] (sv) at (7.615,-.30) {\textbf{Task video planner}\\fine-tuned each round};
\node[fixed] (si) at (11.025,-.30) {Inverse dynamics\\trained separately};
\node[physical] (sr) at (14.435,-.30) {Robot execution\\observed trajectories};
\draw[flow] (sp)--(sv); \draw[flow] (sv)--(si); \draw[flow] (si)--(sr);
\draw[back=ArtBlue] (sr.south)--++(0,-.62)-| (sv.south);
\node[tag,fill=ArtBlue!4] at (11.025,-1.395) {Successful executions};
\roleF{sp}\roleU{sv}\roleF{si}\roleP{sr}
% (b) RISE
\artpanel[b]{0}{-4.85}{16.04}{-2.09}{ArtTeal}{RISE: improve a policy inside fixed learned models}
\node[vig] at (1.32,-3.68) {\includegraphics[width=2.33cm,height=1.98cm,keepaspectratio]{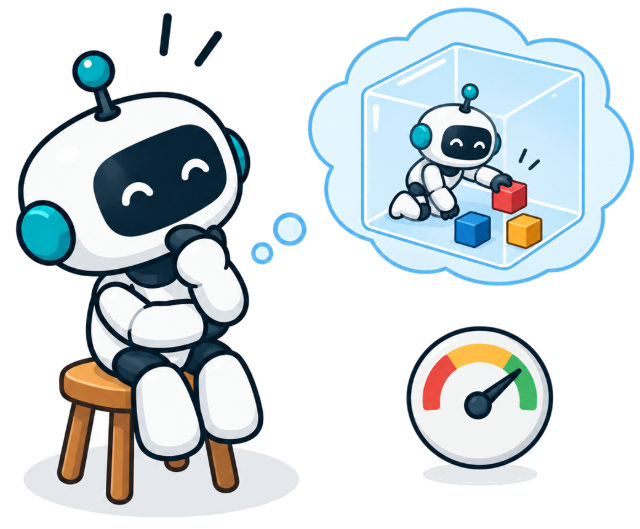}};
\node[physical,tall] (rd) at (4.205,-3.40) {Recorded data\\before policy training};
\node[fixed,tall] (rm) at (7.615,-3.40) {Dynamics model\\+ value model\\both fixed};
\node[updated,tall] (rp) at (11.025,-3.40) {\textbf{Policy}\\advantage conditioning};
\node[physical,tall] (rt) at (14.435,-3.40) {Real-robot tests\\20 trials per task};
\draw[flow] (rd)--node[art note,above=1.5pt,inner xsep=0pt,font=\artserif\fontsize{7.4}{8.6}\selectfont]{Fit}(rm); \draw[flow] (rm)--(rp); \draw[flow] (rp)--(rt);
\draw[back=ArtTeal] (rp.south)--++(0,-.50)-| (rm.south);
\node[art note,anchor=west] at ([xshift=.15cm]rp.south |- 0,-4.35) {Actions to fixed models};
\roleP{rd}\roleF{rm}\roleU{rp}\roleP{rt}
% (c) PlayWorld
\artpanel[c]{0}{-7.95}{16.04}{-5.19}{ArtAmber}{PlayWorld: acquire play data, fit a model, then optimize a noise policy}
\node[vig] at (1.32,-6.78) {\includegraphics[width=2.33cm,height=1.98cm,keepaspectratio]{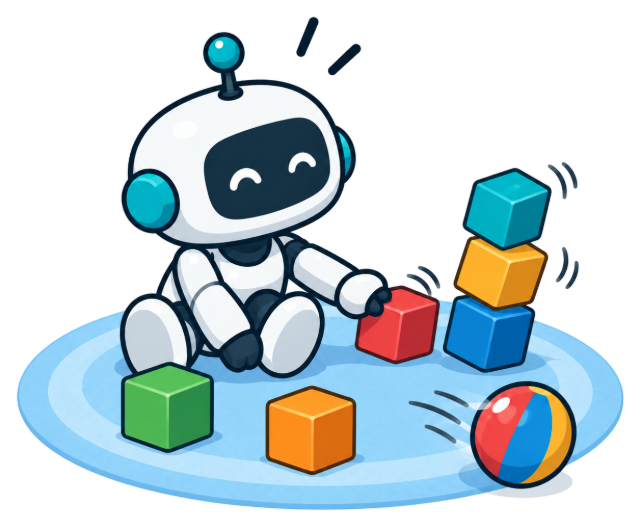}};
\node[physical,tall] (pd) at (4.205,-6.50) {Autonomous play\\30 hours on hardware};
\node[fixed,tall] (pm) at (7.615,-6.50) {Play-trained model\\+ reward from demos\\fit before training};
\node[updated,tall] (pa) at (11.025,-6.50) {\textbf{Trainable noise policy}\\frozen base policy};
\node[physical,tall] (pt) at (14.435,-6.50) {Checkpoint tests\\20 trials per checkpoint};
\draw[flow] (pd)--node[art note,above=1.5pt,inner xsep=0pt,font=\artserif\fontsize{7.4}{8.6}\selectfont]{Fit}(pm); \draw[flow] (pm)--(pa); \draw[flow] (pa)--(pt);
\draw[back=ArtAmber] (pa.south)--++(0,-.50)-| (pm.south);
\node[art note,anchor=west] at ([xshift=.15cm]pa.south |- 0,-7.45) {Actions to fixed model};
\roleP{pd}\roleF{pm}\roleU{pa}\roleP{pt}
% (d) VLAW
\artpanel[d]{0}{-11.05}{16.04}{-8.29}{ArtPurple}{VLAW: update the world model, then the policy; repeat with new physical data}
\node[vig] at (1.32,-9.88) {\includegraphics[width=2.33cm,height=1.98cm,keepaspectratio]{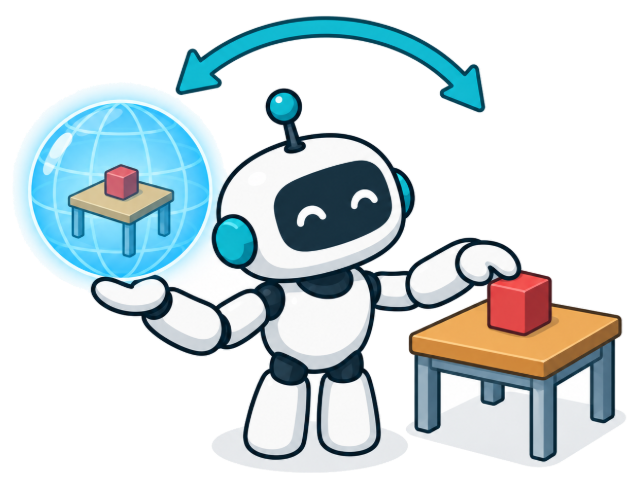}};
\node[physical,tall] (vd) at (4.205,-9.60) {Real trajectories\\successes and failures};
\node[updated,tall] (vm) at (7.615,-9.60) {\textbf{World model}\\new + retained data};
\node[updated,tall] (vp) at (11.025,-9.60) {\textbf{VLA policy}\\successful real +\\imagined trajectories};
\node[physical,tall] (vr) at (14.435,-9.60) {Robot execution\\next collection round};
\draw[flow] (vd)--(vm); \draw[flow] (vm)--(vp); \draw[flow] (vp)--(vr);
\draw[back=ArtPurple] (vr.south)--++(0,-.50)-| (vd.south);
\node[tag,fill=ArtPurple!4] at (9.32,-10.725) {New real data; two rounds; reward filter fit in round 1};
\roleP{vd}\roleU{vm}\roleU{vp}\roleP{vr}
% Legend
\node[circle,minimum size=.30cm,inner sep=0pt] (l1) at (.25,-11.36) {};\paiBadgeAt{(l1.center)}{ArtTeal}{\paiGlyphCycle}
\node[art text,anchor=west,inner xsep=3pt] (t1) at (l1.east) {updated in the indicated loop};
\node[circle,minimum size=.30cm,inner sep=0pt,anchor=west] (l2) at ([xshift=.45cm]t1.east) {};\paiBadgeAt{(l2.center)}{ArtSlate!55}{\paiGlyphPin}
\node[art text,anchor=west,inner xsep=3pt] (t2) at (l2.east) {fixed in that phase};
\node[circle,minimum size=.30cm,inner sep=0pt,anchor=west] (l3) at ([xshift=.45cm]t2.east) {};\paiBadgeAt{(l3.center)}{ArtAmber}{\paiGlyphArm}
\node[art text,anchor=west,inner xsep=3pt] at (l3.east) {physical data or tests};
\end{tikzpicture}%
\caption{\textbf{Where models, policies, and physical data change.} Learning loops of SILVR, RISE, PlayWorld, and VLAW~\citep{arxiv250606658,arxiv260211075,arxiv260309030,arxiv260212063}.}
\label{fig:model-learning-loops}
\end{figure}

%% file: sections/05_modular_improvement.tex
\subsubsection{Memories, Skills, Code, and Runtimes}
\label{sec:modular-improvement}

Not every useful change needs to enter a policy's weights or a world model. A robot that misses grasps might remember that a particular object tends to slip, save a regrasp routine that worked, or add a monitor that interrupts the policy when slip is detected. We use \emph{memory} for stored information that must be interpreted when it is used, \emph{skill} for a callable behavior with an execution interface, and \emph{runtime} or \emph{software} for the logic that retrieves, composes, monitors, or repairs those behaviors (Figure~\ref{fig:external-knowledge}). The same artifact can cross these boundaries (a recovery written down as text becomes a skill once a controller can execute it), but the distinction identifies who uses the change next; for all three, the question is whether a change kept outside model weights is \emph{usable}. On the path to RSI, external knowledge is also where a system can keep the repair knowledge that recursion requires (Section~\ref{sec:key-terms}).

\input{figures/external_knowledge}

Earlier systems kept different kinds of artifacts. Options~\citep{sutton1999options} encode temporally extended behaviors that a higher-level controller can call, and Horde~\citep{sutton2011horde} learns many sensorimotor predictions in parallel. RoboEarth~\citep{waibel2011roboearth} shares models and action recipes among robots. Code as Policies~\citep{liang2023code} writes programs against perception and control APIs, and Inner Monologue~\citep{huang2023inner} feeds execution feedback back into planning. In digital environments, including games~\citep{arxiv260916679}, Voyager~\citep{arxiv230516291} accumulates executable skills and Reflexion~\citep{arxiv230311366} stores written reflections on earlier failures, showing how a frozen model can still take part in an agent that improves; the \emph{Weights or Skills?} survey~\citep{arxiv260801851} reviews this broader choice. Physical reuse additionally needs valid sensing, control, and conditions of applicability.

\paragraph{Remembering experience.}
\label{sec:memory}

Memory systems differ in what they store, how strongly they abstract it, through which interface it is read, and how long it persists, and a memory helps only if these choices preserve a relation that a later decision can exploit. Some systems keep the scene and past interaction themselves: RoboMemory~\citep{arxiv250801415} and AllDayNav~\citep{arxiv260610927} preserve continuity of the environment and of past interaction, and PhyAgentOS~\citep{arxiv260716636} keeps sessions, evidence, recovered lessons, and capability descriptions as versioned records inside a runtime that also checks compatibility, supervises execution, and invokes recovery, so its gains belong to the integrated package.

Others distill experience into reusable guidance, which must then be checked against later conditions. Robo-Cortex~\citep{arxiv260518729} distills progress and failure histories into principles and induced navigation heuristics, evaluated mainly in unseen settings of the Habitat simulator~\citep{savva2019habitat} with a preliminary physical study of ten navigation goals. RoboHarn-Evo~\citep{arxiv260937583} also revises what it stores: physical feedback updates task- and action-level entries together with their stated applicability, and knowledge acquired in simulation reaches 41.7\% mean zero-shot success on real tasks, rising to 62.5\% after updates from real interaction. PhysMem~\citep{arxiv260220323} makes the check explicit: a frozen VLM planner commits a hypothesis about physical properties to memory only after targeted interaction verifies it. On simulated medium-difficulty brick insertion this reaches 76\% success against 53\% without memory and 23\% when raw experience is retrieved directly (100 episodes per condition, three seeds), so retrieving raw experience can do worse than having no memory; its real xArm6 experiments report plan quality over 30-minute deployment sessions. PragmaBot~\citep{arxiv250716713} summarizes the successful episodes of a real legged manipulator into a retrievable memory, raising single-trial success over twelve scenarios, eight of them new but structurally similar, from 22\% for a planner without memory to 80\%. Storage is only half of reuse: a retriever must select the record, and a recipient must assign it the right role.

How long a memory lives also limits what it can show. With the foundation policy of Zeva~\citep{arxiv260830880} fixed, removing its interaction memory lowers success by 10--20 percentage points on five real ChemLab-Evo tasks, but the store is cleared before each episode, so the benefit is adaptation across retries within an episode.

\paragraph{Reusing skills and code.}
\label{sec:skills}

A one-off recovery becomes a reusable skill only when it is detached from the incident that produced it, with an implementation, a callable interface, and stated conditions and effects that let a planner place it in a new sequence. Implementations come from demonstrations in LOTUS~\citep{arxiv231102058} and Uni-Skill~\citep{arxiv260302623}, and from a person when the library cannot carry out the plan in HaLP~\citep{arxiv230914321}. The robot's own execution supplies them in MobiAgent~\citep{arxiv261003476} and ReSYNC~\citep{arxiv260618328}. MobiAgent clusters verified deployment rollouts into atomic skills and raises real success on an Astribot S1 from 32.5\% to 57.5\%, and ReSYNC learns recovery policies together with the symbolic predicates and operators that let a planner compose them.

Code-writing agents often call two artifacts ``skills'': a callable routine and guidance for constructing one. LRLL~\citep{arxiv240618746} distills simulated instruction--program pairs into callable Python APIs under a curriculum. ASPIRE~\citep{arxiv260700272} retrieves guidance distilled from earlier program repairs when debugging a new controller; on three physical tasks whose embodiment and API differ from the simulation source, retrieval reduces the reasoning tokens spent on the real robot, and held-out success is unchanged on one task and improves on two. Reconstruct, Practice, Go Real~\citep{arxiv261002204} revises symbolic skills and its system prompt during simulated practice, keeps a change only after a cross-task check, and reaches 95.0\% against 75.5\% for ASPIRE on held-out initializations of 22 simulated tasks; the frozen system succeeds in all 30 physical trials.

Libraries can also be built in advance and frozen, and the clearest results come from passing a frozen library to fresh or weaker agents. RATs~\citep{arxiv260619419} distills successful play into a code-skill library for simulated LIBERO-PRO tasks. EmbodiedEvo~\citep{arxiv261018867} keeps code skills and written lessons that raise fresh agents' success on unseen layouts of the RoboDojo simulation benchmark~\citep{arxiv260704434} from 8.50\% to 30.17\%, with similar success at 53.0\% lower execution cost than the raw interaction history. Its frozen library also raises a weaker executor that failed to acquire it by 8.86 percentage points over no experience. PhysRSI-eval~\citep{arxiv261026848} freezes GPT-6 Astra workspaces after six rounds of revision on six RoboDojo tasks and tests each on five unseen layouts with no further revision; across 12 workspaces, the stored files alone succeed in 25 of 60 tests. Workspaces pooled from the other tasks also bring the first success earlier in three cases, for example from round 14 to round 7 for Claude Opus 5.5 on the Organize Table task. RoboSkill~\citep{arxiv260937810} reports real-robot gains on its hard tasks, and Agent as Policy~\citep{arxiv260912541} shortens execution across repeated real trials. Embodied Turing Machines~\citep{arxiv261012369} make every decision in code on simulated bimanual tasks (Table~\ref{tab:rsi-systems}). A frozen library can also be written from demonstrations rather than practice. In an evaluation of GPT-6-Astra~\citep{arxiv261022471}, the model studies a task's training demonstrations and writes task guidance, motion references, executable helpers, and recovery procedures into a package that is frozen before testing and given to every fresh test session. On four simulated humanoid tasks of the SIMPLE benchmark, the package raises success from 70.8\% to 90.8\% (120 episodes per condition), with gains that depend on how closely the demonstrations cover the test tasks.

Studies that vary the retained artifact while holding its user fixed show most clearly what the artifact must encode. ReTrace~\citep{arxiv261015814} lets a model controlling a real Franka arm trace a late failure to an earlier decision, test one correction per attempt, and keep supported corrections as a text skill. With skills that one model learned in a few attempts, a second executing model raises plug insertion and ball pouring from 0\% to 80\% and 100\%, respectively, over ten real trials each. In a code-transfer study~\citep{arxiv260919906}, with the coding model, cases, interfaces, and per-session limits held fixed, mean success on nine new tasks in the simulated RoboCasa environment~\citep{arxiv240602523} is 45.2\% without references, 41.5\% with initial source programs, and 57.0\% with execution-optimized ones: a retained artifact helps only if it encodes distinctions that the next synthesizer can exploit. Table~\ref{tab:modular-systems-comparison} summarizes selected external-artifact updates of this section and their outcome checks.

\input{tables/modular_systems_comparison}

Even a correct skill fails if it is called under the wrong conditions, and some systems revise those conditions directly. VASO~\citep{arxiv260605395} model-checks skill-based plans against temporal-logic specifications and turns counterexample traces into updates of each skill's semantic contract. Across tasks for a Jackal ground robot and the PX4 autopilot, eleven specifications, and 400 generated plans, the optimized skills reach 97.2\% (handcrafted alignment) or 96.8\% (automatic) specification compliance; compliance holds relative to the encoded propositions, and unmodeled sensing or contact errors lie outside it. The conditions can also concern the robot itself. KnowBody~\citep{arxiv260928530} keeps an explicit, revisable model of how the robot's body produces effects, with the VLM frozen, and rechecks knowledge that depends on a revised body estimate before reusing it. Over 32 fixed-budget trials on four real-robot tasks, it completes 75\% against 25\% for the native harness. With persistent updates, its planner rounds fall by 29--53\% from the first to the fifth recorded success. Composition also creates handoffs, as when a placed mug hides the next object from the camera, so a skill's stated effects should include the state it hands to its successor.

\paragraph{Revising the runtime.}\label{sec:runtime}
The runtime selects context, monitors progress, interrupts, routes control to recoveries, and decides when the nominal policy may resume; revising it can change the effective agent while the weights of planner and controller stay fixed. The reported comparisons show that runtime components interact, and most of them are measured in simulation. SHAPER~\citep{arxiv260811350} edits textual procedural guidance and a Python harness around a frozen planner and executor, and on VLABench~\citep{zhang2024vlabench} the two edits are not additive: the full configuration scores 34.50\% against 28.25\% for the seed, less than the skill-only and harness-only gains combined. Zetta~\citep{arxiv260816590} grows code-based critics that detect local failure signatures such as a slipping grasp and promotes a critic--recovery patch only after diagnostic replay and a fresh closed-loop rollout. Harness VLA~\citep{arxiv260708448} keeps a VLA frozen as a retryable contact primitive and learns from execution traces when it and a few analytic primitives apply. SHAPER and Zetta are evaluated in simulation; Harness VLA measures its gains in simulation, with demonstrations on two dual-arm Franka setups. Baton~\citep{arxiv260816889} explores each subtask separately around a frozen VLA and stores memories of the transitions between subtasks; with exploration on one scene seed and evaluation on ten held-out seeds of 26 simulated RoboMemArena tasks, it reaches 63.5\% task success against 46.1\% for the strongest prior method, and the system is also run on a real PiPER-X arm.

Several systems revise programs and tools around a frozen model and gate what they admit. DynaHarness~\citep{arxiv260940306} admits a revision of the capabilities around a frozen $\pi_{0.5}$ only after paired regression checks, and its frozen final system reaches 75.2\% on 800 newly sampled simulated LIBERO-PRO initial states against 17.5\% for the bare policy. HarnessPAI~\citep{arxiv260929166}, evaluated only in simulation, has a coding agent revise a task program around a frozen action model and reaches 96.5\% on simulated LIBERO-PRO tasks, on evaluation seeds that include its development seeds; fine-tuning $\pi_{0.5}$ on the harness's successful rollouts raises held-out-seed success from 34.9\% to 73.7\%. URAI~\citep{arxiv260939018} has a frozen agent call validated tools one complete motion at a time, raising pooled success on simulated RoboDojo tasks from 18.0\% under direct fingertip control to 53.0\%; tool revision by its programming agent appears in an author-steered case study. RoboRSI~\citep{arxiv261012424}, named for recursive self-improvement, confines each patch to the skill branch where a failure began and releases it only after review, tests, and replay; on three of its four simulation benchmarks, its gains are measured while it improves online on the evaluation tasks (Table~\ref{tab:rsi-systems}). RACaP~\citep{arxiv260929394} replaces its parent system only when a candidate achieves strictly higher success under identical tasks, seeds, and budgets: over 596 simulated development episodes, nine of the 23 tested candidates were promoted, and the frozen result reaches 54.4\% on the 90 LIBERO tasks it evolved on and 45.0\% zero-shot on LIBERO-PRO. In DynaHarness, HarnessPAI, URAI, and RoboRSI, any agent that writes revisions is a fixed model, although HarnessPAI's growing library of failure-to-repair notes changes what its program writer knows.

A different route trains the program writer itself. PyRUA-Lean~\citep{arxiv261001939} fine-tunes a 4B model that composes robot primitives, including VLA policies, into Python cells on 45 simulated LIBERO-PRO task instances, raising its success on unseen instances from 18.8\% to 44.1\%.

\paragraph{Transferring across robots.}
\label{sec:cross-robot}

Moving knowledge to another robot can fail at each level of description. Saved as a task, the regrasp transfers only if the receiving robot can ground it in skills of its own, and saved as a named skill, it transfers only if that robot implements the same effect. Saved as an end-effector motion, it transfers only if both robots interpret the motion alike; saved as joint commands, it transfers only to an arm of the same kind. SayCan~\citep{ahn2023saycan} grounds skills by combining their usefulness for the task with their estimated chance of success, while Open X-Embodiment~\citep{arxiv231008864} standardizes data from twenty-two embodiments but maps actions to a coarse common end-effector format that leaves coordinate frames unaligned, so the same numerical action can induce different motions on different robots. RoboFoundry~\citep{arxiv260932862} separates embodiment-invariant decisions from embodiment-specific execution so that evolved memories and recovery skills can transfer, and PhyAgentOS checks observations, action semantics, timing, capabilities, and safety configuration before building adapters; its validation on Franka, PIPER, and Dobot arms tests whether the interface carries over, not whether a saved lesson does. ASPIRE tests a lesson on a single receiving robot: guidance from simulated Franka repairs helps a coding agent debug programs for a bimanual robot with a different API, on which no simulated program runs unchanged. SHAPER, Zetta, and the code-transfer study move artifacts across simulated tasks but not across robots. Evaluation should hold the artifact or the adapter fixed while varying the other.

\begin{paiinsight}[PAISyn1]{Reuse depends on the conditions saved with a repair}
\begin{insightpoints}
\item \textbf{Avoid rediscovering execution knowledge.} Saved experience helps when the limitation is a recurring failure, a missing recovery or applicability condition, or runtime logic that defeats an adequate skill.
\item \textbf{Reuse requires retrieval and realization.} A reusable artifact should therefore state its required observations, action semantics, preconditions, effects, and validation evidence, which later changes can make stale.
\item \textbf{Reuse by later learners is tested only in part.} Most reported gains are behavioral improvements, often of integrated packages or in simulation, and Zeva's memory benefit is episode-local. ASPIRE and the code-transfer study report benefits to later learning: a retained artifact helps a coding agent construct controllers for new tasks, on a second robot in ASPIRE's case; RATs and EmbodiedEvo pass frozen libraries to other agents.
\item \textbf{Aging and revalidation.} The open comparisons concern when a stored repair becomes stale, how to revalidate it after a change in object, tool, or robot, and whether accumulated repair knowledge improves the rule used on a later failure (Sections~\ref{sec:repair-histories} and~\ref{sec:revision-knowledge}).
\end{insightpoints}
\end{paiinsight}

%% file: figures/external_knowledge.tex
\begin{figure}[!tbp]
\centering
\begin{tikzpicture}[art fonts,x=1cm,y=1cm,line cap=round,line join=round,
  every node/.append style={execute at begin node={\hyphenpenalty=10000\exhyphenpenalty=10000}},
  cellcap/.style={inner sep=0pt,align=center,text=ArtSlate,font=\sffamily\fontsize{8.2}{9.6}\selectfont,anchor=south,
    execute at end node={\vphantom{g}}},
  effect/.style={inner sep=0pt,align=center,text=ArtPurple!85!black,anchor=north,
    font=\sffamily\bfseries\fontsize{8.8}{10.4}\selectfont,execute at end node={\vphantom{g}}},
  ex/.style={inner sep=0pt,align=center,text=ArtSlate!85,anchor=north,font=\rmfamily\itshape\fontsize{8.0}{9.4}\selectfont},
  wcard/.style={fill=white,draw=ArtPurple!38,line width=.55pt,rounded corners=1.6mm,
    drop shadow={opacity=.12,shadow xshift=.5pt,shadow yshift=-.6pt}},
  rstep/.style={fill=white,draw=ArtPurple!45,line width=.5pt,rounded corners=1.3mm,minimum width=1.80cm,minimum height=.52cm,
    inner xsep=2pt,inner ysep=1.5pt,text=ArtSlate,font=\sffamily\fontsize{8.0}{9.4}\selectfont,execute at begin node={\vphantom{Ag}}},
  chk/.style={fill=white,draw=ArtPurple!45,line width=.5pt,rounded corners=1.3mm,minimum height=.50cm,inner xsep=3pt,
    text=ArtSlate,font=\sffamily\fontsize{8.0}{9.4}\selectfont,execute at begin node={\vphantom{Ag}}},
  flowa/.style={-{Stealth[length=1.4mm,width=1.4mm]},line width=.7pt,draw=ArtPurple!70},
  vig/.style={inner sep=0pt}]
\path[use as bounding box] (0,-6.96) rectangle (16.08,.02);
% Top row: memory, skill, runtime. Each panel: white card (illustration + caption), effect, examples.
\foreach \n/\xa/\ttl/\eff/\exs in {
  1/.02/Memory/{Stored information, read when used}/{e.g., RoboMemory, Zeva},
  2/5.41/Skill/{Callable behavior with an interface}/{e.g., ReSYNC, VASO},
  3/10.80/Runtime/{Logic that monitors and repairs}/{e.g., SHAPER, Zetta}}
  {\pgfmathsetmacro{\xb}{\xa+5.26}\pgfmathsetmacro{\cx}{\xa+2.63}
   \artpanel[\n]{\xa}{-4.48}{\xb}{0}{ArtPurple}{\ttl}
   \path[wcard] (\xa+.12,-3.40) rectangle (\xb-.12,-.54);
   \node[effect,text width=5.0cm] at (\cx,-3.56) {\eff};
   \node[ex,text width=5.0cm] at (\cx,-4.00) {\exs};}
% (1) Memory: a note that this mug may slip, reread before it is used.
\node[vig,anchor=south] at (2.65,-2.92) {\includegraphics[height=2.28cm,width=4.70cm,keepaspectratio]{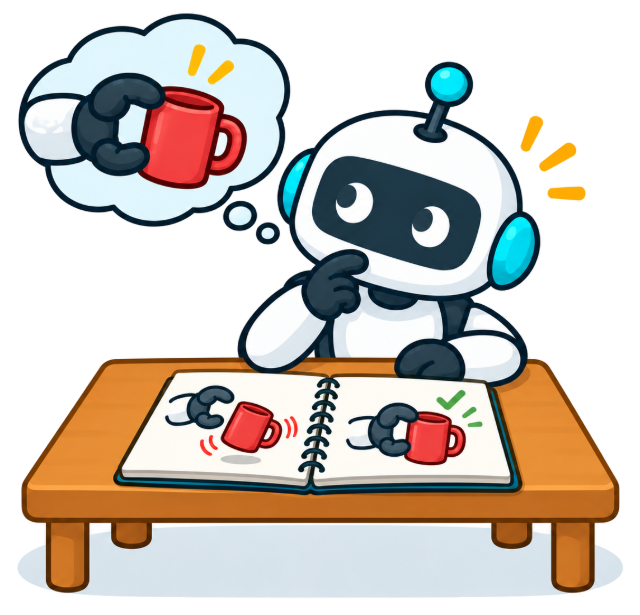}};
\node[cellcap,text width=4.9cm] at (2.65,-3.32) {A note: this mug may slip};
% (2) Skill: the saved regrasp with its stated conditions and effect.
\node[vig,anchor=south] at (6.78,-2.92) {\includegraphics[height=2.00cm]{figures/art_invoke}};
\node[fill=white,draw=ArtPurple!60,line width=.6pt,rounded corners=1.4mm,inner sep=0pt,minimum width=2.34cm,minimum height=2.10cm] (sk) at (9.22,-1.87) {};
\node[anchor=north,inner sep=0pt,font=\ttfamily\fontsize{8.4}{9.6}\selectfont,text=ArtPurple!85!black] at (9.22,-.94) {regrasp()};
\draw[ArtPurple!35,line width=.45pt] (8.17,-1.30)--(10.27,-1.30);
\node[anchor=north,inner sep=0pt,align=center,text width=2.10cm,font=\sffamily\fontsize{7.6}{9.0}\selectfont,text=ArtSlate] at (9.22,-1.42)
  {{\bfseries Condition}\\slip detected\\[2pt]{\bfseries Effect}\\stable grasp};
\node[cellcap,text width=4.9cm] at (8.04,-3.32) {A saved regrasp routine};
% (3) Runtime: monitor, interrupt, recover, resume around a fixed policy.
% A 2x2 loop: monitor -> interrupt -> recover -> resume -> monitor.
% The fixed policy sits inside the loop; the steps around it are the runtime that changes.
\foreach \x/\y/\ic/\lab/\nm in {12.08/-1.16/eye/Monitor/s1, 14.78/-1.16/pause-circle/Interrupt/s2, 14.78/-2.18/redo/Recover/s3, 12.08/-2.18/check-circle/Resume/s4}
  {\node[rstep] (\nm) at (\x,\y) {\textcolor{ArtPurple!85!black}{\smash{\faIcon{\ic}}}\enspace\lab};}
\draw[flowa] (s1.east)--(s2.west); \draw[flowa] (s2.south)--(s3.north);
\draw[flowa] (s3.west)--(s4.east); \draw[flowa] (s4.north)--(s1.south);
\node[inner sep=0pt,text=ArtSlate!75,font=\fontsize{18}{18}\selectfont] (pl) at (13.43,-1.67) {\faIcon{microchip}};
\node[inner sep=0pt,text=white,font=\fontsize{6.5}{6.5}\selectfont] at (13.43,-1.67) {\faIcon{lock}};
\node[anchor=north,inner sep=0pt,font=\rmfamily\itshape\fontsize{7.8}{9.2}\selectfont,text=ArtSlate!85] at (13.43,-2.56) {the policy inside the loop stays fixed};
\node[cellcap,text width=4.9cm] at (13.43,-3.32) {A monitor that interrupts on slip};
% (4) Reuse on another robot: the saved repair passes a preflight check before it runs elsewhere.
\artpanel[4]{.02}{-6.94}{16.06}{-4.62}{ArtPurple}{Reuse on another robot}
\path[wcard] (.14,-6.82) rectangle (15.94,-5.14);
\node[vig] at (1.30,-5.98) {\includegraphics[height=1.52cm]{figures/art_retain}};
\node[vig] at (14.78,-5.98) {\includegraphics[height=1.52cm,width=2.10cm,keepaspectratio]{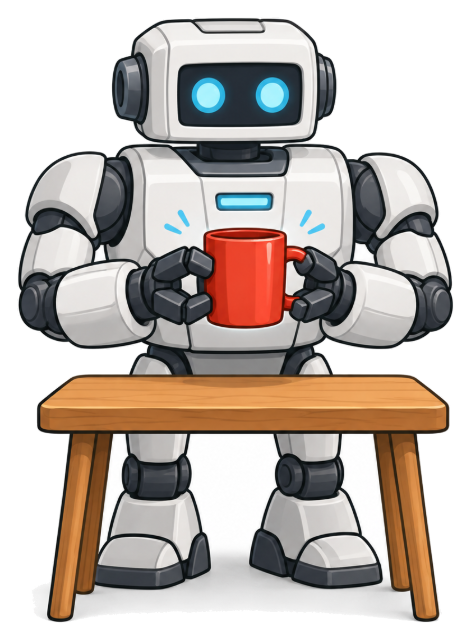}};
\node[inner sep=0pt,font=\sffamily\bfseries\fontsize{8.4}{10}\selectfont,text=ArtPurple!85!black] at (8.04,-5.42)
  {\faIcon{clipboard-check}\enspace Preflight check before building adapters};
\foreach \x/\lab in {3.70/Observations, 5.87/Action meaning, 8.04/Timing, 10.21/Capabilities, 12.38/Safety}
  {\node[chk,text width=1.84cm,align=center,minimum width=2.02cm] at (\x,-5.97) {\lab};}
\draw[flowa] (2.25,-5.97)--(2.62,-5.97);
\draw[flowa] (13.46,-5.97)--(13.83,-5.97);
\node[inner sep=0pt,font=\rmfamily\itshape\fontsize{8.0}{9.4}\selectfont,text=ArtSlate!85] at (8.04,-6.52)
  {Retrieve the repair, adapt the interface, test the result \quad e.g., ASPIRE, PhyAgentOS};
\end{tikzpicture}%
\caption{\textbf{Three forms of external knowledge in the grasp example.} Each panel shows one saved form of the slip repair and representative systems from this section; reuse on another robot first checks the interface.}
\label{fig:external-knowledge}
\end{figure}

%% file: tables/modular_systems_comparison.tex
\begin{table}[!htbp]
\centering
\caption{External-knowledge updates.}
\label{tab:modular-systems-comparison}
\begingroup\footnotesize
\setlength{\tabcolsep}{4pt}\renewcommand{\arraystretch}{1.20}\arrayrulecolor{PAIRule}
\begin{tabularx}{\linewidth}{@{}L{2.75cm}L{3.0cm}L{2.7cm}L{3.1cm}Y@{}}
\paitoprule
\pahead{System} & \pahead{Components} & \pahead{Update signal} & \pahead{Outcome check} & \pahead{Comparison}\\
\paimidrule
\systemcite{Robo-Cortex}{arxiv260518729} & \updmark Heuristic memory \fixmarkn planner model & Simulated outcomes; reflection & \envsim\ Habitat task metrics; \envhw\ Preliminary test & Transferred heuristics vs basic pipeline \\
\paihair
\systemcite{PhyAgentOS}{arxiv260716636} & \updmark Records, runtime \fixmarkn base policies & Platform-specific outcome checks & \envsim\ Criteria shared with verifier & Integrated runtime comparison \\
\paihair
\systemcite{Zeva}{arxiv260830880} & \updmark Episode memory \fixmarkn foundation policy & Action-effect interpretation & \envhw\ Task outcomes & Episode-local gain; memory resets \\
\paihair
\systemcite{RoboHarn-Evo}{arxiv260937583} & \updmark Task/action knowledge \fixmarkn VLM, executor & Physical feedback across episodes & \envsim\ Benchmark success; \envhw\ Real tasks & Knowledge-error repair; sim-to-real transfer \\
\paihair
\systemcite{ReSYNC}{arxiv260618328} & \updmark Recoveries, predicates \fixmarkn planner & Simulated recovery execution & \envsim\ Task execution; \envhw\ Real-robot transfer & Baselines on unseen tasks \\
\paihair
\systemcite{ASPIRE}{arxiv260700272} & \updmark Repair guidance \fixmarkn coding model & Simulation; hardware debugging & \envhw\ Task completion & Later learning; retrieval contrast; three tasks \\
\paihair
\systemcite{Uni-Skill}{arxiv260302623} & \updmark Implementation \fixmarkn video repository & Retrieved examples; task feedback & \envsim\,\envhw\ Task success & Enabling synthesis evidence \\
\paihair
\systemcite{VASO}{arxiv260605395} & \updmark Contracts \fixmarkn checker, controllers & Formal counterexamples & Checker: specification compliance; \envsim\ Task completion & Simulated gain; shared formal abstraction \\
\paihair
\systemcite{SHAPER}{arxiv260811350} & \updmark Guidance, harness code \fixmarkn model weights & Simulated execution feedback & \envsim\ Success; shared task definition & Held-out splits; component controls \\
\paihair
\systemcite{Zetta}{arxiv260816590} & \updmark Critics, recovery code \fixmarkn base policies & Simulated validation and task signals & \envsim\ Milestone checks; shared semantics & Repeated repair; zero-shot task transfer \\
\paihair
\systemcite{RATs}{arxiv260619419} & \updmark Code-skill library \fixmarkn coding model & Self-proposed play tasks & \envsim\ Held-out tasks; \envhw\ Preliminary transfer & Play vs no-play; other agents \\
\paihair
\systemcite{EmbodiedEvo}{arxiv261018867} & \updmark Skills, lessons \fixmarkn coding models & One exploration scene per task & \envsim\ RoboDojo score and success & Fresh agents; other models; raw history \\
\paihair
\systemcite{PhysRSI-eval}{arxiv261026848} & \updmark Code, tools, notes \fixmarkn model weights & Graded attempt each round & \envsim\ RoboDojo native score and success & Six systems; history ablation; frozen and cross-task tests \\
\paihair
\systemcite{Embodied Turing Machines}{arxiv261012369} & \updmark Shared state and decision code \fixmarkn coding agents & Closed-loop library development & \envsim\ RoboDojo task success & One library across tasks; no test-time model \\
\paihair
\systemcite{Code transfer}{arxiv260919906} & \updmark Program archive \fixmarkn synthesis rules & Simulated task execution & \envsim\ Common benchmark checks & Nine new tasks; three synthesis runs \\
\paibottomrule
\end{tabularx}\endgroup
\end{table}

%% file: sections/06_improvement_process.tex
\subsection{Improving Learning Procedures}
\label{sec:emerging-directions}

The changes reviewed so far also alter what the next round of learning starts from. An improved policy reaches states that were previously inaccessible, so the data available for the next update change too. The resulting experience can expose distinctions missing from supervision; revised supervision changes what an update teaches; and the repair that follows can become evidence for constructing later updates. Along this chain, a change of any form can take on a new role in the \emph{learning procedure} (Figure~\ref{fig:taxonomy}). Improving that procedure is where RSI would begin, because it changes how the robot learns rather than only what it does.

Reuse and procedure change can overlap, but they answer different questions: reuse asks whether a change is carried forward, whereas procedure change asks whether it alters collection, judgment, or updating for a later learner. We take the procedure's decisions in turn: where the next experience comes from ($c$, Section~\ref{sec:select-experience}), how outcomes are judged ($e$, Section~\ref{sec:revise-supervision}), what an update learns from ($d$, Section~\ref{sec:reuse-across-rounds}), and how the update is made ($u$, Section~\ref{sec:construct-update}). Section~\ref{sec:automate-experiments} then turns to how whole experimental loops are run, which can change several of these decisions at once, and Table~\ref{tab:learning-procedures} summarizes representative systems. A changed procedure is not yet an improved one: improvement requires that a learner guided by the revised procedure learn better than a matched learner guided by the earlier one (Section~\ref{sec:key-terms}; Figure~\ref{fig:downstream-comparison}).

\input{figures/downstream_comparison}

\subsubsection{Choosing What to Practice}
\label{sec:select-experience}

Once a recovery removes a coarse grasp failure, the remaining errors may concentrate on subtle contact events. More demonstrations of the old kind may then be less useful than tactile checks, a finer reward, or a new rule for editing skills. Better execution has changed the learning problem, and the system now has to find the experience that teaches what is still missing. The most useful experience for an improved policy need not come from the distribution that served its predecessor, and choosing it involves three decisions: which tasks or states to practice, what supervision an attempt needs, and when collection is safe. Active learning~\citep{cohn1996active} selects observations for their expected contribution to learning; in robotics, however, selecting an observation often means choosing a behavior, task, environment, or trajectory that must actually be executed. SAC-X~\citep{riedmiller2018sacx} schedules auxiliary behaviors that help the robot reach sparse rewards; automatic domain randomization~\citep{openai2019rubik} widens the range of simulated conditions as the policy becomes competent; and Enhanced POET~\citep{wang2020poet} co-evolves environments together with the policies that solve them. In each case, what the learner can already do helps determine what it encounters next.

At the task level, the frontier worth targeting is learnable challenge: tasks hard enough to teach something and easy enough to be learned. Methods based on intrinsic motivation~\citep{oudeyer2007intrinsic}, automatic curriculum learning~\citep{portelas2020curriculum}, and autotelic agents that set their own goals~\citep{colas2022autotelic} track learning progress across goals or environment parameters. Systems built on language models enlarge the space of candidate lessons by generating tasks and environments, as in GenSim~\citep{wang2024gensim} and RoboGen~\citep{wang2024robogen}; OMNI~\citep{zhang2024omni} combines estimated learning progress with a model of what humans find interesting, while Eurekaverse~\citep{liang2025eurekaverse} generates environments that become progressively more challenging while remaining diverse and learnable. AutoRT~\citep{ahn2024autort} proposes tasks with language models, filters them against a written robot constitution, and over seven months collected about 77,000 episodes on 53 robots. Generation supplies possibilities, but a signal that depends on the learner's current capability is still needed to decide which possibility should become the next lesson.

FIND~\citep{arxiv260932069} supplies such a signal on hardware: a VLM agent selects feasible tasks with low recent success and judges each attempt by comparing observations before and after execution, and residual RL on a frozen $\pi_{0.5}$ raises success on eight real tasks from 55\% to 71.9\%, while people still reset scenes when needed. InSight~\citep{arxiv260624884} lets a VLM name the primitives a steerable VLA lacks for a new task, practices them with VLM-parameterized single-axis control, and retrains the VLA on successful segments; on a real xArm, twisting and pouring reach 92\% and 96\% against 32\% and 16\% for a zero-shot code-as-policy baseline with the same VLM (25 trials each). Agentic Skill Discovery~\citep{arxiv240515019} has a language model propose the next task from its skill library and a pool of failed tasks, and keeps a skill only after a VLM confirms it; in simulation, RL reaches at least 90\% success on 17 of 22 tabulated tasks. EmbodiedRSI~\citep{arxiv261010498} applies a related signal to an agent harness: it keeps competing code and skill hypotheses, selects the experiments expected to distinguish them, and learns which memories to keep according to their value for later improvement. It is evolved entirely in simulation (Table~\ref{tab:rsi-systems}).

Goal exploration shows what such a signal should track. SAGG-RIAC~\citep{baranes2013goals} selects regions of the goal space where competence is changing, rather than repeatedly sampling the goals with the lowest success. Its experiments on learning inverse models move practice away from both mastered and unreachable regions. Failure prediction and learning-progress estimation answer different questions: where is the policy weak, and where is another attempt likely to change what it can do? ALAN~\citep{arxiv230206604} applies a related signal on hardware, exploring two real-world play-kitchen setups by favoring changes it can cause to objects and its uncertainty about predicted change. Practice Makes Perfect~\citep{arxiv240215025} makes the choice explicit for parameterized skills. It estimates each skill's competence, extrapolates the gain from another attempt, and weighs that gain by planned task success. Over one to three hours of reset-free practice on a Spot arm, the fraction of evaluation tasks solved rises from 0.0 to 0.8 in one environment and from 0.53 to 0.93 in the other.

At the level of states and datasets, failure becomes useful when it can be converted into appropriate supervision, and the clearest comparisons fix the demonstration budget while varying where it is spent. A criticality model trained on policy outcomes~\citep{arxiv260728251} can bias collection toward scenarios that are likely to end in failure, such as hard initial states. The method uses importance-sampled experience for RL and criticality-selected expert demonstrations for imitation learning. In its two multi-round simulated runs the predictor is retrained on the current policy's rollouts each round. Its evaluations span simulated locomotion and manipulation, VLA benchmarks, and a physical banana-placement task. A high criticality score still leaves open whether a state calls for autonomous exploration, expert correction, or avoidance. A world model can rank the gaps instead: RoboCoach~\citep{arxiv260939685} imagines where each skill expert first times out and requests demonstrations for the highest-ranked subtask--expert pairs, and with matched demonstration budgets success rises from 13.3\% to 75.0\% on a Franka and from 40.0\% to 83.8\% on an AgileX platform, against 30.0\% and 47.5\% for uniform collection into one adapter. SAVE~\citep{arxiv260618043} sends each expert demonstration to the task and initial state where an ensemble's velocity fields disagree most, reaching 47\% average success on three real Franka tasks at a fixed budget of 150 demonstrations, 8 to 25 percentage points above the baselines.

Failures can also be rebuilt or searched for in simulation. F4R~\citep{arxiv260935575} rebuilds each diagnosed failure as a simulated scene for targeted practice and gains 18.75 percentage points out of distribution on four real tasks over a budget-matched targeted imitation baseline, and EmbodiRSI~\citep{arxiv260938905} builds task-specific simulations from deployment scenes for corrections and demonstrations aimed at current weaknesses (Table~\ref{tab:rsi-systems}). SEES~\citep{arxiv260932698} restores, in simulation, the states before the sub-skill that most often stalls a long-horizon task and trains that skill family with PPO against an LLM-written success check, gaining over three rounds on RoboCasa365 (Appendix~\ref{app:reported-results}). Q-DEG~\citep{arxiv261014415} searches over object placement, camera, and lighting for diverse simulated scenes in which a VLA sometimes fails and fine-tunes it on its own successes there, raising average success on held-out adversarial LIBERO-Spatial scenes from 67.2\% to 80.9\% for $\pi_{0.5}$, with a small drop on the default scenes.

Supervision is the second decision: when experience must be acquired physically, selection must also determine when autonomy should yield to an expert. SAGE~\citep{arxiv260829772} estimates short-horizon cost and uncertainty with an ensemble of world models and combines them into a ``fear'' signal, and world-model prediction error adjusts the threshold at which an expert takes over. The resulting interventions provide targeted data for imitation and risk-constrained learning. Most of its evidence comes from driving simulation and replay of Waymo driving data, while the hardware evaluation covers four controlled indoor settings for a ground robot with twenty trials per setting. The mechanism times expert takeover by predicted risk; long-term, unrestricted deployment remains untested. Human-in-the-loop deployment concentrates human attention in the same way. Sirius~\citep{arxiv221108416} lets people intervene during deployment and learns from those interventions, whereas Fleet-DAgger~\citep{arxiv220614349} allocates a limited number of human supervisors across multiple robots. HIL-UMI~\citep{arxiv260920659} carries the same idea into demonstrations collected with the handheld Universal Manipulation Interface~\citep{chi2024umi}: it queries the current policy without executing it, collects human actions where policy--human discrepancy indicates unfamiliar behavior, and separately refines a progress-based advantage estimator. It takes the robot out of demonstration collection, while human action supervision and final hardware evaluation remain necessary. Genie Centurion~\citep{arxiv250518793} also updates the component that decides when to ask. Its task sentinel flags steps not completed in time, after which the robot rewinds for an operator's correction, and the sentinel is fine-tuned together with the policy each round. Across four real tasks at matched trajectory counts, the final-round step-completion score is 0.93, against 0.53 for adversarial data collection and 0.38 for teleoperation.

Some desired experience cannot be demonstrated at first. GLIDE~\citep{arxiv260924996} converts anticipated and observed teleoperation failures into executable guardrails on commands and reuses the revised guardrails during both data collection and policy execution. Across three physical tasks, its ten-trial collection tests rise from zero or one successes without guardrails to seven to nine with the selected refined guardrails. Crossed policy experiments separate the contribution of guarded training data from that of guarded execution. Here, selection extends to constructing the conditions under which a demonstration can be produced at all.

Safety is the third decision. A learned runtime monitor~\citep{arxiv231017552} can request human supervision before a predicted failure, and safe online system identification~\citep{arxiv250421486} keeps informative trajectories within certified constraints. No single score resolves all three decisions: states with similar failure probabilities may represent a missing skill, an unreachable target, or an unsafe transition, so a useful rule for physical data acquisition must combine expected learning benefit with recoverability, the available source of supervision, and the costs of time, resets, wear, and risk.

\subsubsection{Judging Outcomes}
\label{sec:revise-supervision}

A policy that reaches new states also strains its supervision: a reward or evaluator learned on earlier behavior may not score those states correctly. The least invasive response is to update the judge and rescore existing data. PEBBLE~\citep{arxiv210605091} learns a reward model from preferences between pairs of behaviors (supplied by a scripted teacher in its quantitative experiments) and relabels the entire replay buffer whenever that model changes, so that off-policy learning in its simulated locomotion and manipulation tasks always sees one current interpretation of stored experience. RL-VLM-F~\citep{wang2024rlvlmf} keeps this preference-based loop but takes the labels from a vision-language model that compares pairs of images against a text goal, so that its simulated tasks need no human or scripted teacher.

FIERCE~\citep{arxiv260918651} applies this response on hardware: it fits and calibrates an evaluator of progress and failure on local successes and failures. Each refinement round freezes a versioned snapshot of the evaluator, consistently relabels the replay buffer, updates the policy, and later refreshes the evaluator with newer rollouts. On physical peg-in-hole and cup-stacking tasks, the reported single-seed comparisons favor periodic refresh over a frozen task evaluator, although total acquisition cost is not equalized. Feedback thus stays stable during each policy update and can be revised between updates. RLER~\citep{arxiv260933653} refits a success-probability reward model on the policy's own rollouts between rounds of offline RL; on two real Franka tasks, two rounds raise success from 64\% to 84\% and from 60\% to 86\% over the starting policy, 8 to 12 percentage points above keeping the same reward model fixed. EvoHIL~\citep{arxiv260803872} retrains its reward classifier during human-in-the-loop RL from operator-confirmed successes and deploys each version only after a held-out accuracy gate; under a lighting shift on six real tasks, success reaches 0.73 to 1.00 against 0.13 to 0.50 for HIL-SERL, in a comparison the authors describe as not resource-matched. VeriFine~\citep{arxiv261008761} revises the judge itself when policy gains plateau: an agent edits the judge's rubric and asks human experts to settle disagreements, which on driving-reasoning data raises the judge's correlation with human ratings from 0.55 to 0.85 and the policy's reasoning score from 68.05 to 83.13 under matched training budgets.

A learned world model can also act as a judge, and it inherits the blind spots of its training data. On a physical arm, a scene-adapted Ctrl-World model fine-tuned on successes scored 90\% of held-out failing sequences as success-like under a latent-distance diagnostic, and fine-tuning on execution-checked counterfactual failures lowered this to 33\%~\citep{arxiv261009134}. Judges that only rank a frozen policy's candidate actions at run time, without supervising an update, appear with action selection in Section~\ref{sec:selection-expansion}.

A larger change is to revise the reward specification itself. Language models first entered this role as proxy reward functions. \citet{kwon2023reward} prompt a language model with a task description and an episode outcome and use its answer as the reward in negotiation and matrix games. Language to Rewards~\citep{yu2023l2r} has a language model translate instructions into reward parameters that a model-predictive controller optimizes for a simulated quadruped, a simulated dexterous manipulator, and a real robot arm. Text2Reward~\citep{xie2024text2reward} turns a natural-language goal and an abstract description of the environment into executable dense-reward code; after users inspect policy rollouts, their feedback can prompt another version of the code. Its main evaluations use simulated manipulation and locomotion, with policies trained in simulation demonstrated on two tasks with a Franka arm. Eureka~\citep{ma2024eureka} removes the human from this editing loop by training policies under candidate rewards and returning task fitness, together with statistics of each reward component, to the language model. In Text2Reward and Eureka, the language model's weights remain fixed; improvement comes from new context and externally supplied evaluation.

Later systems vary this loop mainly to cut its cost or to check candidates before training. STRIDE~\citep{arxiv250204692} refines humanoid-locomotion rewards from training statistics and reports a best success score about 2.5 times Eureka's on random terrain in simulation, although the two methods use different language models. CARD~\citep{arxiv241014660} first checks whether a candidate reward ranks recorded successes above failures and skips RL training when it does not. It matches or exceeds expert rewards on 10 of 12 simulated tasks with about 14K tokens per task against about 663K for Eureka. MEMENTO~\citep{arxiv260722832} evolves the evaluator before the policy, so that code-policy search in simulation solves Tower of Hanoi in 0.97 of episodes, whereas the same search under the unevolved evaluator solves none.

These loops also need a criterion that stays outside the quantity being optimized. A reward program can score demonstrations sensibly yet induce shortcut behavior during training; a quicker lift can coexist with excessive grip force, jerky motion, or a fragile final state. Text2Reward uses human interpretation as an anchor, while Eureka relies on a task-fitness function specified by researchers. Without such an independent reference, iterating on the reward can amplify a misspecification. Revising supervision has to ask two questions: does the new signal accelerate learning, and does the resulting behavior still satisfy the original objective?

Training conditions can be revised as well as rewards. DrEureka~\citep{ma2024dreureka} runs Eureka to write a reward, evaluates the resulting policy under perturbed simulator parameters to find the ranges in which it still performs well, and then asks a language model to choose a domain-randomization configuration (the ranges of physical parameters varied during simulated training) within those ranges. The resulting policies transfer to physical quadruped and dexterous-manipulation platforms without further real-world training. Here the policy probes its own later training distribution, but only in simulation; deployment experience does not feed back into that distribution. Broadening variation can increase robustness, yet overly wide ranges consume rollouts and can make the task unnecessarily hard to learn.

Physical verification can instead become memory for later judgments. REVOLVE~\citep{arxiv260914633} stores cases in which a VLM's judgment of an outcome disagrees with human verification and retrieves similar mismatches as context for future judgments. In parallel, it updates the policy from successful rollouts and from failure--correction data. Across four physical tasks, policy success rises over five iterations and judgment accuracy over four memory updates, supported by task-specific automated correction and reset routines. Because mismatch memory, policy data, and behavior change together, the study demonstrates a functioning dual loop without separating the contribution of each loop.

Retrieval can alter the effective procedure even when model weights stay fixed: ASPIRE uses earlier repairs to construct controllers, whereas REVOLVE uses earlier judgment errors to interpret later outcomes. FailBench~\citep{arxiv260903611} measures how far such judges can be trusted: across 2,197 real and simulated manipulation attempts, the best of 13 VLM success detectors reaches 0.77 balanced accuracy, and none exceeds 0.60 on contact-intensive assembly. Every loop above that revises behavior against a learned judge is exposed to errors of this kind, so part of the success evaluation must stay outside the loop (Section~\ref{sec:evaluator-integrity}).

\subsubsection{Shaping Training Data}
\label{sec:reuse-across-rounds}

Once experience is collected and judged, the procedure must decide what the next update learns from: which retained data, corrections, demonstrations, or teacher outputs, and from which model. Several examples below run a fixed procedure and report a later-learning benefit; they appear here because they change where the procedure's training data come from. Acquired competence first changes who can supply the next learner's evidence. In RoboCat~\citep{arxiv230611706} (Section~\ref{sec:targets}), an adapted specialist collects data for the next generalist; the study demonstrates one completed cycle of this transfer. PLD~\citep{arxiv251100091} (Section~\ref{sec:selection-expansion}) instead distills a specialist's recoveries from the generalist's own deployment states back into the generalist. The useful output in both cases is data that cover a task, a state, or a correction that the next learner would otherwise lack.

Acquired competence can also lower the cost of later learning in roles other than physical collection: VLAW~\citep{arxiv260212063} amplifies scarce real rollouts through a refreshed world model (below), ASPIRE~\citep{arxiv260700272} avoids rediscovering a working program on a new robot API (Section~\ref{sec:cross-robot}), and CIDER~\citep{arxiv260821899} keeps earlier skills by reusing the accumulated policy as a teacher (Section~\ref{sec:sharing}). Verified solutions of coding agents can also become training data: EmbodiedSWE~\citep{arxiv260927308} expands one simulated solution into diverse demonstrations for a VLA, which then completes a long-horizon task on a real robot, and HexaAnything~\citep{arxiv260935432} trains a model on the verified execution traces of its own agent. Table~\ref{tab:experience-connections} adds related cases from Sections~\ref{sec:base-loop} and~\ref{sec:emerging-directions} and lists, for each retained artifact, what is held fixed, the reported comparison and effort, and the scope of the result.

\input{tables/experience_connections}

A model that supplies training experience can itself be refreshed from new physical experience. VLAW~\citep{arxiv260212063} returns both successful and failed rollouts to world-model training, alongside retained data from the DROID dataset~\citep{arxiv240312945}, and fine-tunes a VLM-based reward model on first-round outcomes. The updated world model then generates closed-loop trajectories, and the VLA is trained on reward-filtered imagined successes together with real successes. Over two rounds and five real task categories, this full loop outperforms behavior cloning on filtered successes (Filtered BC) and DSRL while holding the number of physical rollouts fixed at fifty per category per round. Because model refresh, reward calibration, and synthetic data change together, the gain belongs to the pipeline as a whole.

Other systems refresh the practice model in the same way, and some isolate parts of the effect. World-VLA-Loop~\citep{arxiv260206508} refits its video model and reward head on success and near-success rollouts of each new policy and restarts the policy from its fine-tuned starting point every round; on two real Franka tasks, two rounds raise success from 13.3\% to 50.0\% and from 26.7\% to 53.3\%, the second round adding 13.3 percentage points on both. WoVR~\citep{arxiv260213977} refits its world model on rollouts of the already improved policy and, at the same 2,500-trajectory budget in LIBERO, beats adding base-policy data by 5.0 percentage points; its real-robot gains were obtained without this refresh. WorldSample~\citep{arxiv260702431} refits a video model on accumulating real rollouts and admits its synthetic transitions into critic training by value balance and policy entropy, reaching 82\% average success on five real tasks against 56\% for HIL-SERL with the same demonstrations. MeRL~\citep{arxiv261016517} measures how far its updated world model diverges from stored real continuations and uses that error to set how much imagined data enters each update, gaining 3.5 percentage points on LIBERO-PRO over the same simulator updated without this trust layer. RIWANav~\citep{arxiv261008640}, which calls itself recursive self-improvement, lets the improved navigation policy choose which recorded action--video pairs refit its world model, adding 7.94 percentage points of success over a frozen world model on an offline urban-navigation benchmark.

Corrective actions can also come from outside the robot: dynamics models learned from human video propose them at failed states, and training on them raises success on seven real tasks across two robots from 40\% to 81\%~\citep{arxiv260621406}. A fixed multimodal agent can serve as that outside teacher in a recurring loop: Local Recovery Supervision~\citep{arxiv261005151} audits each round's policy for unresolved failures, has the agent generate recovery demonstrations at those states, and trains the policy that drives the next round of collection. In a preliminary simulated study on LIBERO-Goal, recovery-augmented post-training of a $\pi_0$ checkpoint succeeds in 88 of 100 validation scenes, against 78 for continuing on the original data with the same additional optimization steps.

Selection also applies to experience that has already been retained, as when Memory Anchors~\citep{arxiv260826545} (Section~\ref{sec:sharing}) enriches replay with selected old transitions. Criticality-guided sampling (Section~\ref{sec:select-experience}) asks where new learning is most needed; Memory Anchors asks which earlier distinctions must remain visible while that learning occurs. Together they expose a trade-off between learning new behavior and keeping old behavior, one that the simpler instruction to ``collect hard cases'' hides.

Which gap matters also depends on the component that receives the data, as a world-model example shows. ConfAL-WM~\citep{arxiv260825572} uses a confidence probe to allocate the post-training data budget of a world model across tasks, sampling episodes at random within each task, and gives larger training weight to frames and image patches where the model is least reliable. Under a fixed budget of selected data on the RoboTwin~2.0 benchmark~\citep{chen2025robotwin2}, confidence-guided selection improves scene consistency and trajectory metrics, and confidence weighting improves reconstruction, relative to alternative acquisition scores. Whether the same samples most improve control is a different question, since a transition that is informative about visual prediction may differ from one that supplies the policy with a missing action.

Rollouts can also decide which demonstrations to keep. Demo-SCORE~\citep{arxiv250303707} trains a success classifier on the robot's rollouts and removes demonstrations that resemble failures. On real ALOHA tasks, this improves success by about 30 percentage points on average over training on all demonstrations. CUPID~\citep{arxiv250619121} attributes closed-loop outcomes to individual demonstrations with influence functions and raises success on a real Franka task from 44\% to 84\% with 25 curation rollouts. Both curate once rather than as the policy changes. A model can also generate or vet its own training data. LACY~\citep{arxiv251102239} uses one VLM to map language to actions, describe actions back, and judge their agreement, and filters its own generated data with that judge. RSI-Nav~\citep{arxiv261015782}, described as data-centric recursive self-improvement, writes instructions for unlabeled trajectories, keeps those it can follow to the goal, and retrains before labeling the next shard. With about 2.5M generated samples, it raises simulated R2R-CE success from 52.9\% to 68.6\%. RE-0~\citep{arxiv260932416} admits a stronger teacher's code patches into a student's training data only when paired rollouts from the same simulator snapshot show a gain, which lifts robosuite restacking from 4/50 to 31/50. Its ablations credit the admitted patches rather than the weighting of its distillation loss: on restacking, success-filtered teacher demonstrations stay at the base level, and on a lifting task the paired admission rule beats unfiltered and sign-only selection in every trainer seed.

These later-learning benefits do not require the robot to learn how to revise its learning procedure; often a fixed collection, filtering, or distillation recipe performs the transfer.

\subsubsection{Constructing Updates}
\label{sec:construct-update}

Once the data and supervision for an update are chosen, the update itself is usually still computed by a fixed rule. A further step is to change the procedure that produces the update. We use \emph{updater} for a procedure that maps the learner's current state, experience, and objective to an operation such as a parameter step, a loss function, an edit to a skill library, or a code patch. Section~\ref{sec:meta-learning} reviews precedents that learn initializations, losses, and update rules, mostly in computation. Robot data have already trained part of an update mechanism. \citet{meier2018online} learn online a memory of learning rates that scales the gradient steps of inverse-dynamics models on a real Kuka LWR~IV arm. They reuse it as the payload changes from none to light and heavy and reach lower prediction error early in each task. The memory's own learning rule stays fixed, however, and the gain is measured in model error rather than in control. Here we follow updaters that act on robot learners and their artifacts. AdaRSI~\citep{arxiv261020474}, named for on-demand recursive self-improvement, decides where an update should go. A coding agent first refines its harness and fine-tunes the executor only on sub-steps that stay unreliable after retries from saved states. In RoboCasa365 simulation, this raises unseen composite-task success from 4.6\% to 13.8\%, slightly above fine-tuning every sub-step (12.6\%) at 28\% of its GPU time. Its agent also writes, from pilot tasks, an experience document and a training budget that guide later tasks.

PRACTICE~\citep{arxiv260830760} instantiates a related idea for an external skill library. Given trajectories and the current library, its learned editor can add, revise, merge, or remove skill entries, while a frozen executor uses the resulting library. The editor's training progresses from successful oracle trajectories, through paired success and failure examples, to on-policy distillation from a stronger frozen teacher on the student's own editing contexts. On the simulated EB-ALFRED household tasks from EmbodiedBench~\citep{yang2025embodiedbench}, the failure-aware stage reaches 45.3\% average success, compared with 42.7\% for a control trained only on successes, supporting the value of failures for training the editor. The broader gains across stages are harder to attribute, because the editor, its training data, and the accumulated library change together. What has improved is the editor; the executor, the external teacher, and the staged procedure that trains the editor all remain fixed.

When the value of a proposed change can only be established through robot interaction, constructing updates becomes inseparable from safe execution, resets, measurement, and recovery.

\subsubsection{Automating Experiments}
\label{sec:automate-experiments}

\input{tables/learning_procedures}

An updater can propose a new policy, loss, or training routine, but a physical system can judge that proposal only by running another experiment. Each candidate must begin from a known state, remain within safety limits, be deployed under recorded conditions, and receive an outcome that is sufficiently independent of the code being tested. Resets make successive trials comparable; verification supplies the selection signal; and logging connects a measured change to the hypothesis that produced it. These services belong to the learning loop itself.

ENPIRE~\citep{arxiv260619980} instantiates this loop for research on robot policies. In a human-guided construction stage, coding agents build task-specific services for safety, resets, and verification. The finalized services are then exposed as interfaces that cannot be modified, while coding agents revise the permitted training infrastructure or policy code, launch robot rollouts, inspect trajectories and videos, and retain or abandon hypotheses based on the resulting measurements. This covers more of the research process than executing a fixed fine-tuning script, while keeping the main outcome check outside the policy code that can be changed. Its autonomy still depends on the supplied task objective, demonstrations, hardware, and interface boundaries, and a misspecified verifier could consistently select the wrong revision.

ENPIRE studies two ways to extend a single experimental loop. Running more agent--robot pairs in parallel buys throughput at a resource cost; whether it also improves the updater is a separate question. In a separate transfer, agents distill their search on a pin-insertion task into a written summary and append it to the instructions for a GPU-insertion task, while removing prior trajectories, logs, and checkpoints. The result shows that procedural knowledge can be carried between related experiments; measuring how much that knowledge improves the later search would need a matched run without the summary.

Agents increasingly run the training pipeline itself. In simulation, a coding agent ran more than 70 quadruped-locomotion experiments in fourteen waves under human high-level direction~\citep{arxiv260327416}. HARBOR~\citep{arxiv260608610} stages simulator setup, reward design, and tuning through gated sub-agents and appends distilled lessons to an experience memory, reaching 0.885 success on three-cube stacking in IsaacLab where Eureka and REvolve reach 0.0 under the same model and wall-clock budget. AURA~\citep{arxiv250602507} has language-model agents write staged curricula that specify rewards, domain randomization, and PPO settings, trains each stage in MJX simulation, and refines the next stage from its analysis, retrieving earlier curricula and feedback from a database; its hardware deployment on a custom humanoid is reported qualitatively. RoboClaw~\citep{arxiv260311558} pairs each forward manipulation with an inverse recovery behavior so that a dual-arm mobile manipulator can reset itself and keep collecting data, while one VLM agent coordinates collection, policy learning, and execution. It reports about 25\% higher long-horizon success than its baselines, without stating whether the gain is absolute. It also reports that a manual baseline needs about 2.16 times its human time for the same amount of data. The RoboCoach system~\citep{arxiv260121570}, unrelated to the world-model coach of Section~\ref{sec:select-experience}, pairs a coding agent with a video-reading feedback agent: on RoboCoach-Bench, a set of 32 simulated tasks, the seven-model mean success reaches 0.730 against 0.602 for expert codebases, and on four real tasks in two laboratories the resulting policies reach 0.68 against 0.51 for matched human configurations.

Two studies measure what such loops still need. An agentic loop with evidence-gated promotion~\citep{arxiv260807555} cuts false promotions in LIBERO-10 simulation to 0.001 per run. Yet on its one measured lineage, its selections trail simply taking the final checkpoint (about 50--54\% against 60\% held-out success). Coding agents in RLE-Bench~\citep{arxiv260934210} likewise gain on the development evaluations they can query but lose verifier score on hidden robustness perturbations in simulation.

EndoRSI~\citep{arxiv261013376} brings a human-governed coding agent to an endoscopic robot: benchtop phantom trials rejected an insertion strategy that offline data had supported, and benchtop insertion success rose from 2/21 to 17/21 over four generations of revised policies.

Neighboring fields have automated other parts of the experimental loop. The AI Scientist~\citep{lu2024aiscientist} automates proposal, execution, analysis, and reporting in digital experiments. Robotic chemistry closes a physical loop within more structured apparatus: a mobile robotic chemist~\citep{burger2020chemist} and the A-Lab~\citep{szymanski2023alab} select experiments, manipulate materials, and incorporate measured synthesis or characterization outcomes, while Coscientist~\citep{boiko2023coscientist} adds planning and tool use by a language model. Across these settings, a reusable finding requires the candidate version, initial conditions, operating configuration, measurements, and failed trials to remain attributable.

\FloatBarrier
\begin{paiinsight}[PAISyn2]{Improving a learner and improving its learning procedure are distinct}
\begin{insightpoints}
\item \textbf{Practice should be selected for what it teaches.} A reachable failure, a useful correction, and a learnable task differ, and the best target shifts as the policy improves.
\item \textbf{Judgment and updating can themselves change.} Refreshed evaluators, retrieved guidance, and learned editors change later training; safety, reset, and verification services keep practice possible.
\item \textbf{The revised decision needs a downstream test.} Component contrasts such as FIERCE's refreshed evaluator, RLER's refit reward model, and PRACTICE's failure-aware editor indicate local \emph{procedure improvements}; coupled pipeline gains improve whole loops without isolating the revision responsible.
\item \textbf{From procedure to improver.} Whether an improver that learns from its own revision history produces better procedures on new physical failures is the first link of recursion (Section~\ref{sec:reviser}).
\end{insightpoints}
\end{paiinsight}

%% file: figures/downstream_comparison.tex
\begin{figure}[!htbp]
\centering
\begin{tikzpicture}[art fonts,x=1cm,y=1cm,line cap=round,line join=round,
  every node/.append style={execute at begin node={\hyphenpenalty=10000\exhyphenpenalty=10000}},
  cellcap/.style={inner sep=0pt,align=center,text=ArtSlate,font=\sffamily\fontsize{8.2}{9.6}\selectfont,anchor=south,
    execute at end node={\vphantom{g}}},
  wcard/.style={fill=white,draw=#1!38,line width=.55pt,rounded corners=1.6mm,
    drop shadow={opacity=.12,shadow xshift=.5pt,shadow yshift=-.6pt}},
  rcard/.style={wcard=#1,draw=#1!70,line width=.7pt},
  hchip/.style={inner sep=0pt,text=ArtSlate,font=\sffamily\bfseries\fontsize{9}{10}\selectfont},
  bitem/.style={inner sep=0pt,text=ArtSlate,font=\sffamily\fontsize{8.4}{10}\selectfont,
    execute at begin node={\textcolor{ArtAmber!85!black}{\textbullet}\enspace\vphantom{Ag}}},
  vig/.style={inner sep=0pt}]
\colorlet{ArtOld}{ArtSlate!55}
\path[use as bounding box] (0,-4.52) rectangle (16.08,.52);
% Columns: changed component .09--3.75 | learner card 4.40--7.85 | test card 8.50--12.10 | records 12.82--15.94.
% Each 0.65 cm gap holds a 0.36 cm arrow at the row center.
\newcommand{\cmprow}[7]{% top, bottom, color, title, version, clipboard image, learner number
  \begin{scope}
    \clip[rounded corners=1.8mm] (.09,#2) rectangle (12.25,#1);
    \fill[#3!4] (.09,#2) rectangle (12.25,#1);
    \fill[#3!14] (.09,#2) rectangle (3.75,#1);
  \end{scope}
  \draw[#3!40,line width=.45pt] (3.75,#2)--(3.75,#1);
  \draw[#3!50,line width=.6pt,rounded corners=1.8mm] (.09,#2) rectangle (12.25,#1);
  \node[vig] at (.95,{(#1+#2)/2}) {\includegraphics[height=1.56cm,width=1.32cm,keepaspectratio]{figures/#6}};
  \node[align=center,text width=1.90cm,inner sep=0pt] at (2.68,{(#1+#2)/2}) {%
    {\sffamily\bfseries\fontsize{9.6}{11.2}\selectfont\color{#3!85!black}#4\par}\vspace{3pt}%
    {\artserif\color{#3!75!black}#5\par}};
  \path[rcard=ArtBlue] (4.40,#2+.08) rectangle (7.85,#1-.08);
  \path[rcard=ArtSage] (8.50,#2+.08) rectangle (12.10,#1-.08);
  \node[circle,fill=ArtBlue!85!black,text=white,inner sep=0pt,minimum size=.32cm,
    font=\sffamily\bfseries\fontsize{8}{8}\selectfont] at (4.71,#1-.38) {#7};
  \node[circle,fill=ArtSage!90!black,text=white,inner sep=0pt,minimum size=.32cm,
    font=\fontsize{5.6}{5.6}\selectfont] at (8.81,#1-.38) {\faIcon{clipboard-check}};
  \node[vig,anchor=south] at (6.125,#2+.56) {\includegraphics[height=1.12cm,width=3.10cm,keepaspectratio]{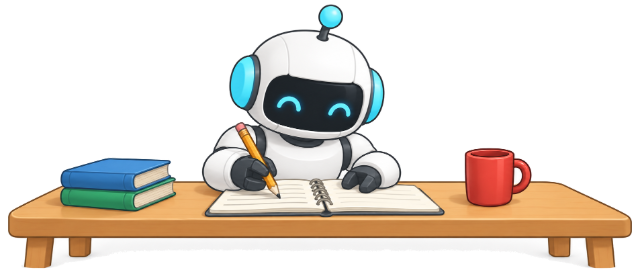}};
  \node[cellcap,text width=3.30cm] at (6.125,#2+.15) {Same start, recipe, budget};
  \node[vig,anchor=south] at (10.30,#2+.56) {\includegraphics[height=1.12cm,width=3.30cm,keepaspectratio]{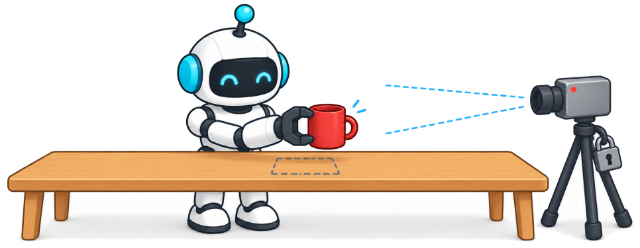}};
  \node[cellcap,text width=3.45cm] at (10.30,#2+.15) {Separate from evaluator};
  \foreach \xc in {4.075,8.175,12.535}
    \node[art fat=#3,minimum height=.42cm,minimum width=.36cm] at (\xc,{(#1+#2)/2}) {};
  \path[wcard=#3] (12.82,{(#1+#2)/2-.66}) rectangle (15.94,{(#1+#2)/2+.66});
  \node[anchor=north,inner sep=0pt,font=\sffamily\bfseries\fontsize{8.2}{9.6}\selectfont,text=#3!85!black] at (14.38,{(#1+#2)/2+.52}) {\vphantom{Ag}Learner #7 outcomes};
  \draw[#3!35,line width=.45pt] (13.02,{(#1+#2)/2+.17})--(15.74,{(#1+#2)/2+.17});
  \node[anchor=north west,inner sep=0pt,align=left,font=\sffamily\fontsize{7.6}{8.8}\selectfont,text=ArtSlate] at (13.10,{(#1+#2)/2+.08})
    {\textcolor{#3!85!black}{\smash{\faIcon{redo}}}\enspace Trials and successes\\\textcolor{#3!85!black}{\smash{\faIcon{layer-group}}}\enspace Independent runs};}
% Header
\shade[left color=ArtPurple,middle color=ArtTeal,right color=ArtAmber] (.09,-.04) rectangle (16.01,-.01);
\node[hchip,anchor=base] at (1.92,.175) {Changed component};
\node[hchip,anchor=base] at (6.125,.175) {\textcolor{ArtBlue!85!black}{\faIcon{user-graduate}}\enspace Matched learner};
\node[hchip,anchor=base] at (10.30,.175) {\textcolor{ArtSage!90!black}{\faIcon{clipboard-check}}\enspace Common outcome check};
\node[hchip,anchor=base] at (14.38,.175) {\textcolor{ArtRed!85!black}{\faIcon{balance-scale}}\enspace Separate records};
% Two matched pipelines: only the evaluator version differs.
\cmprow{-.06}{-1.91}{ArtOld}{Earlier evaluator}{Before revision}{art_evalold}{1}
\cmprow{-2.03}{-3.88}{ArtAmber}{Revised evaluator}{After revision}{art_evalnew}{2}
\draw[{Stealth[length=2mm,width=1.8mm]}-{Stealth[length=2mm,width=1.8mm]},line width=.8pt,draw=ArtRed] (13.08,-1.68)--(13.08,-2.26);
\node[art pill=ArtRed,anchor=west,inner ysep=1.0pt,font=\artserif\fontsize{8.2}{9.6}\selectfont] at (13.28,-1.97) {Compare outcomes};
% Further tests a stronger claim needs.
\path[fill=ArtAmber!6,draw=ArtAmber!40,line width=.55pt,rounded corners=2mm] (.09,-4.50) rectangle (16.01,-3.98);
\node[anchor=base west,inner sep=0pt,text=ArtAmber!85!black,font=\sffamily\bfseries\fontsize{8.6}{10}\selectfont] at (.30,-4.3085)
  {\faIcon{redo}\enspace Also test durability and cost\vphantom{Ag}};
\node[bitem] at (6.62,-4.24) {{\bfseries New tasks}\enspace forward transfer};
\node[bitem] at (9.98,-4.24) {{\bfseries Old tasks}\enspace retention};
\node[bitem] at (13.64,-4.24) {{\bfseries Total cost}\enspace including the revision};
\end{tikzpicture}%
\caption{\textbf{Does a revised component help the next learner?} Two matched learners differ only in the component under test, here an evaluator. Their common outcome check stays fixed and separate from that evaluator; comparing the learners' outcomes tests procedure improvement.}
\label{fig:downstream-comparison}
\end{figure}

%% file: tables/experience_connections.tex
\begin{table}[!htbp]
\centering
\caption{Reported later-learning comparisons, with each study's reported effort and the scope of its result. Reported effort covers different portions of acquisition and use across studies.}
\label{tab:experience-connections}
\begingroup\footnotesize
\setlength{\tabcolsep}{3.5pt}\renewcommand{\arraystretch}{1.20}\arrayrulecolor{PAIRule}
\begin{tabularx}{\linewidth}{@{}L{2.95cm}L{1.95cm}L{1.95cm}L{2.75cm}L{2.7cm}Y@{}}
\paitoprule
\pahead{Case} & \pahead{\retainedmark\,Retained} & \pahead{\paiBadge{ArtSlate!55}{\paiGlyphPin}\,Held fixed} & \pahead{Comparison} & \pahead{Reported effort} & \pahead{Scope} \\
\paimidrule
\systemcite{RoboCat}{arxiv230611706} & Generated data & Training recipe & \envhw\ Demos vs added self-data & Target demos; robot rollouts & One retraining cycle \\
\paihair
\systemcite{Collection}{arxiv241101813} & Filtered data & Collection and BC recipe & \envsim\,\envhw\ Successive data mixtures & $>100$ h collection (study-wide) & Saturation; detector and hardware effects \\
\paihair
\systemcite{PLD}{arxiv251100091} & Recovery trajectories & Distillation recipe, data volume & \envsim\,\envhw\ Specialist collection routes & Specialist RL; collection; distillation & Recovery coverage differs \\
\paihair
\systemcite{CIDER}{arxiv260821899} & Prior policy & RL setup & \envhw\ Continual vs single-task start & Robot steps; human interventions & Six supplied tasks \\
\paihair
\systemcite{VLAW}{arxiv260212063} & Model/data pipeline & Physical rollout count & \envhw\ Full loop vs Filtered BC, DSRL & 50 rollouts per category per round & Combined pipeline effect \\
\paihair
\systemcite{DexTouch-WM}{arxiv260920649} & Generated observations & Policy recipe, data count & \envhw\ All-real vs half-generated data & Data counts; prior model fitting & Effects vary by task and policy \\
\paihair
\systemcite{ASPIRE}{arxiv260700272} & Repair guidance & Coding setup & \envhw\ Retrieval vs none & Tokens to first working program & Three tasks; guidance from sim \\
\paihair
\systemcite{Code transfer}{arxiv260919906} & Source programs & Coding model, cases, interfaces, limits & \envsim\ None vs initial vs optimized archive & Common limits; realized calls/tokens vary & Nine new tasks \\
\paihair
\systemcite{FIERCE}{arxiv260918651} & Refreshed evaluator & Actor architecture, rollout budget & \envhw\ Refreshed vs frozen evaluator & 300 rollouts per condition; extra fitting & One seed; total cost unmatched \\
\paihair
\systemcite{Eureka}{ma2024eureka} & Revised reward & RL algorithm, task fitness & \envsim\ Reflection and iteration ablations & Candidate policy training runs & Fixed reward generator \\
\paibottomrule
\end{tabularx}\endgroup
\end{table}

%% file: tables/learning_procedures.tex
\begin{table}[!htbp]
\centering
\caption{Changes to collection, supervision, update, and experimental procedures.}
\label{tab:learning-procedures}
\begingroup\footnotesize
\setlength{\tabcolsep}{4pt}\renewcommand{\arraystretch}{1.20}\arrayrulecolor{PAIRule}
\begin{tabularx}{\linewidth}{@{}L{2.6cm}L{3.15cm}L{2.65cm}L{3.3cm}Y@{}}
\paitoprule
\pahead{System} & \pahead{Components} & \pahead{Update signal} & \pahead{Outcome check} & \pahead{Comparison}\\
\paimidrule
\systemcite{Criticality model}{arxiv260728251} & \updmark Failure selector, policy \fixmarkn scenario pool & Simulation; physical task outcomes & \envsim\ Failure rate; \envhw\ Banana placement & Adaptive collection pipeline \\
\paihair
\systemcite{GLIDE}{arxiv260924996} & \updmark Guardrails, policy \fixmarkn operator/learner setup & Human teleoperation; observed failures & \envhw\ Task success & Crossed data/runtime controls \\
\paihair
\systemcite{FIERCE}{arxiv260918651} & \updmark Evaluator, policy \fixmarkn actor architecture & Refreshed evaluator & \envhw\ Terminal check, reported as independent & Refreshed vs frozen evaluator \\
\paihair
\systemcite{Text2Reward}{xie2024text2reward} & \updmark Reward code \fixmarkn RL algorithm & Human revision; simulated training & \envsim\ Task criteria; \envhw\ Selected deployment & Reward-program comparisons \\
\paihair
\systemcite{Eureka}{ma2024eureka} & \updmark Reward \fixmarkn fitness, RL algorithm & Simulated task fitness & \envsim\ Supplied fitness, also selects candidates & Reflection/iteration controls \\
\paihair
\systemcite{DrEureka}{ma2024dreureka} & \updmark Reward, randomization \fixmarkn learner & Simulated robustness sweeps & \envhw\ Locomotion/\allowbreak manipulation task measures & Simulation-to-hardware transfer \\
\paihair
\systemcite{REVOLVE}{arxiv260914633} & \updmark Mismatch memory, policy \fixmarkn VLM weights & Audits; judgments & \envhw\ Task success; human-referenced judge accuracy & Judgment/policy trajectory \\
\paihair
\systemcite{PRACTICE}{arxiv260830760} & \updmark Editor, library \fixmarkn executor & Edit distillation; simulated outcomes & \envsim\ EB-ALFRED, EB-Habitat success & Staged editor/library training \\
\paihair
\systemcite{ENPIRE}{arxiv260619980} & \updmark Policy/training code \fixmarkn environment services & Robot rollouts; verification logs & \envhw\ Fixed task-verification services & Agent/harness ablations; 1/4/8-robot scaling \\
\paibottomrule
\end{tabularx}\endgroup
\end{table}

%% file: sections/06b_reviser.tex
\subsection{Improving the Improver}
\label{sec:reviser}

This section discusses mechanisms through which an improver could improve and the evidence offered for them. An improver's knowledge may reside in weights, retrieved memory, code, or prompts (Section~\ref{sec:key-terms}). After stating what would count as improvement, we group mechanisms by where that knowledge lives.

\paragraph{What would count as a better improver.}
\label{sec:reviser-criteria}

As Section~\ref{sec:key-terms} notes, successively better procedures can come from a fixed outer searcher. One example is a loop that its authors call policy-level recursive self-improvement: a criticality model updated from each round's outcomes guides the next collection, but the update rule stays fixed~\citep{arxiv260728251}. An improver can also revise its own selection rules, such as how strict its admission gate is or which failures it debugs next; such a change alters the procedure that improves the system, yet it counts toward recursion only if the revised rules produce better later revisions than the earlier ones.

The missed grasp illustrates the difference between procedure improvement and recursion. Suppose a rollout reveals a visual signature of slip in an attempt that the reward program scored as a success. Using that signature to correct the reward and training a better policy under it would demonstrate procedure improvement. Here the reward program belongs to the learning procedure, the editor that writes it is the improver, and the records it keeps from such edits are its repair knowledge. The recursive route must also retain which failure distinction motivated the edit, under what conditions the edit was valid, and whether the downstream learner benefited. Those records could revise the reward editor's parameters, retrieval memory, or rule-selection policy. The first link is present if the revised editor's rewards for new failures teach matched learners better than the earlier editor's rewards do. The second link is present if, given the same new repair histories, a revision of the editor proposed with what it has learned outperforms both the unrevised editor and a revision of the same editor proposed with that knowledge frozen.

The test leaves room for people and fixed machinery: human-specified goals, a fixed experiment runner, and an independently controlled safety boundary may remain in place. What must be traced is the source of the useful revision. If the gains came only from more trials, compute, or human assistance, a frozen-knowledge control given the same resources would match the updated version.

Against this standard, the physical systems covered by this review offer mechanisms and uncontrolled results for R1. We have not found sufficient evidence that freezes the acquired revision knowledge, matches the conditions of later learning, and isolates an R1 gain on new failures (Sections~\ref{sec:rsi-evidence} and~\ref{sec:recursive-test}); the closest results, described next, come from simulation and do not yet meet all three conditions.

\paragraph{Learning from repair histories.}
\label{sec:repair-histories}

Several systems already keep records of earlier revisions and consult them when revising again, which places knowledge in the improver's memory. AdaRSI's~\citep{arxiv261020474} coding agent writes, after pilot tasks, an experience document recording which sub-steps benefited from training and how much data sufficed, together with a training budget, and both guide later tasks (Section~\ref{sec:construct-update}). HARBOR~\citep{arxiv260608610} distills recurring patterns from tuning rounds into an experience memory that later reward design and tuning retrieve (Section~\ref{sec:automate-experiments}). Drawing on its accumulated experience in a separate run, it redesigns the reward for simulated three-cube stacking in 30 minutes rather than 4 hours, but on the same task and without a frozen-memory control; its ablation without experience removes templates, references, and heuristics, so the contribution of the lessons it accumulates on its own is not separated. AURA~\citep{arxiv250602507} stores earlier curricula, metrics, and human feedback in a database that conditions later curriculum, reward, and training-configuration generation (Section~\ref{sec:automate-experiments}); without the database, the share of its generated curricula that launch training in simulation falls from 99\% to 38\% with GPT-4.1, but what AURA acquired itself is not separated from the human feedback. EmbodiedRSI~\citep{arxiv261010498} learns which memories to keep according to their value for later improvement (Section~\ref{sec:select-experience}); on unseen RoboCasa365 composite tasks in simulation, this learned rule reaches 71.3\% success against 61.7\% for retrieval-based selection, a comparison of memory rules rather than of acquired knowledge. ENPIRE~\citep{arxiv260619980} carries a written summary of one search into the instructions for a related task (Section~\ref{sec:automate-experiments}). In each case the improver's knowledge changes; what remains to be shown is that this change, rather than added trials, compute, or human input, makes the next procedure better.

Two kinds of reuse sit close to R1 without reaching it. ASPIRE~\citep{arxiv260700272} retrieves guidance from earlier program repairs, and the code-transfer study~\citep{arxiv260919906} supplies optimized programs as references; both improve how a fixed coding model constructs later task controllers (Section~\ref{sec:skills}), which is evidence about constructing behavior rather than about revising collection, supervision, or updates. LMPC~\citep{arxiv240211450} trains the model that writes reward code for a model-predictive controller on past teaching sessions, so that teaching needs fewer turns. In simulation, a successful session on unseen tasks takes 1.9 rather than 2.4 chat turns (Section~\ref{sec:meta-learning}). That reward code drives real-time control rather than an RL training run, and the recipe that trains the model stays fixed. HarnessPAI's~\citep{arxiv260929166} growing library of failure-to-repair notes plays the role of ASPIRE's guidance for its program writer: in simulation, three LIBERO-Object tasks reach full success within two rounds with the library rather than several without it, but no frozen-library comparison on new failures separates what the writer has gained (Section~\ref{sec:runtime}).

\paragraph{Revising the tools that revise.}
\label{sec:revise-tools}

PhysEvo~\citep{arxiv261008995}, which its authors describe as physical recursive self-improvement, keeps a single frozen model but lets a meta-agent revise the task agent's tools and skills and also its own diagnostic tools. In simulation it reaches 62.0\% on held-out RoboDojo layouts, an equal-weight average over five capability dimensions, against 47.2\% computed from public evaluation records for the strongest published comparison; on a real arm it succeeds in 84\% of 25 trials while skill revision continues. It is the clearest physical case of an improver that edits its own instruments; the comparison that would attribute part of the gain to those edits, the same meta-agent with its diagnostic tools frozen, remains to be run.

Admission rules are another instrument an improver uses. When an optimizer agent revises a robot agent's harness in simulated LIBERO tasks, a champion--challenger rule avoids the degradation that accepting every revision causes~\citep{arxiv260939304}. RegenHarness~\citep{arxiv260927612} forbids its updater to weaken its own acceptance rule. In both cases the designers set the admission rule, so these results show which instrument an improver should use rather than an improver that improves it. MEMENTO~\citep{arxiv260722832} evolves its rollout evaluator once before policy search, and VeriFine~\citep{arxiv261008761} revises its judge's rubric with human calibration (Section~\ref{sec:revise-supervision}); both change an instrument through a fixed process, which places them at procedure improvement unless the process that revises the instrument itself improves. STOP~\citep{zelikman2024stop} is the digital reference point for such self-application. It applies a language-model-driven improver program to itself; with GPT-4, the self-improved improver raises held-out performance over a few rounds and transfers to new tasks without further optimization, whereas weaker models degrade, and the authors note that the unchanged language model makes this less than full recursive self-improvement. SICA~\citep{arxiv250415228} removes the split between meta-agent and target agent: the best agent so far edits its own codebase, and in one fifteen-iteration run with Claude 3.5 Sonnet, peak accuracy on a fixed 50-question SWE-bench Verified subset rose from 17\% to 53\%, with the same benchmarks used for selection and for reporting. A fixed base agent that makes every edit is the control that would separate better task tools from a better editor. VERSE~\citep{arxiv261002616} lets a harness optimizer revise its own prompts, skills, tools, and notes as well as the executor's harness, with model weights fixed. In its controlled study on held-out SWE-rebench tasks (one run per condition), self-revision without execution-based checks produced no harness that beat the initial one on validation, whereas with such checks it reached 41.05\% against 39.20\% for a fixed optimizer and 39.81\% for checks alone; revising the reviser helped only when its edits could be tested. Predict Before You Modify~\citep{arxiv261013991} adds to a self-modifying agent an editable model that predicts the outcome of each proposed modification and is revised from earlier modifications and their observed results; its robotics domain is quadruped reward design in the Genesis simulator, and its control with the predictor frozen, run on coding tasks, has bootstrap intervals that include zero.

\paragraph{Carrying repair knowledge across tasks and robots.}
\label{sec:carry-repair}

R1 needs repair knowledge that transfers beyond the failures that produced it (Section~\ref{sec:revision-knowledge}), and the evidence so far concerns mostly task-level artifacts. ASPIRE carries guidance from simulated Franka repairs to a bimanual robot with a different API (Section~\ref{sec:cross-robot}), and ENPIRE's written summary moves from pin insertion to GPU insertion on real robots without a matched control. A simulated study of 123 rounds of agentic skill discovery~\citep{arxiv260931760} shows how accumulated revisions can stall: deployed changes kept passing their tests while the target task never succeeded. MC-EvoBench~\citep{arxiv261013829} sets out the controls a transfer claim needs: the target world and inventory are reset, only a declared artifact carries over, and related experience is compared with no-prior, unrelated, and shuffled sources at matched environment steps. In Minecraft, over five seeds, a memory-based agent showed no stable transfer, and related experience lowered stone-tool stage success from 0.150 to 0.050. Applying the frozen-knowledge comparison of Section~\ref{sec:recursive-test} across failure families or robots remains to be run.

%% file: sections/06c_closing_loop.tex
\subsection{Closing the Recursive Loop}
\label{sec:closing-loop}

R2 asks for the last step: the improver, once improved and using what it has acquired, must help produce a next version that is better than itself (Section~\ref{sec:key-terms}). Digital systems supply the closest precedents, because executable tests make each candidate cheap to judge, while physical systems supply boundary cases; the gap between the two points to the paths ahead. In the physical systems covered by this review, we have not found a controlled comparison that isolates an R2 contribution; the closest controlled evidence comes from a simulated reward-design task in a digital system, discussed below.

\paragraph{Digital precedents.}
\label{sec:digital-precedents}

Digital systems also separate applying a fixed update recipe from making the updater itself an object of search. Expert Iteration~\citep{anthony2017expert} alternates tree search with training a network to imitate the search, then uses that network to guide the next search; STaR~\citep{zelikman2022star} repeatedly generates rationales, keeps those that lead to correct answers, and fine-tunes on them. Both repeat a fixed loop. Promptbreeder~\citep{fernando2024promptbreeder} evolves both task prompts and the mutation prompts that change them. ADAS~\citep{hu2025adas} searches over agent designs written as code, and ALMA~\citep{arxiv260207755} searches over memory designs written as code and finds designs that outperform human-crafted ones in four sequential decision-making domains. AlphaEvolve~\citep{arxiv250613131} couples code proposals from language models to executable evaluators.

The Darwin G\"odel Machine (DGM)~\citep{zhang2025dgm}, which comes closest to the second link, modifies its own agent code and keeps an archive of variants. In an ablation in which a fixed base agent performs every modification, gains taper off quickly, whereas letting improved agents modify themselves sustains improvement on its coding benchmarks. That ablation is the closest analogue of the second link that we found. It differs from the physical case in three ways: the improver and the task solver are the same agent; the benchmark that selects each modification also measures it, although the discovered improvements also transfer to a held-out benchmark and to other foundation models; and executable tests reject candidates cheaply.

Other digital systems extend the pattern. G\"odel Agent~\citep{arxiv241004444} lets an agent rewrite the code that performs its own improvement (Section~\ref{sec:recursion-requirement}). SelfSearch~\citep{arxiv260937968} has agents modify themselves using records of earlier self-modification episodes, without downstream reward during the search, and improves population-mean success over the initial agent in all six model--benchmark settings it reports. HyperAgents~\citep{arxiv260319461} make the meta agent itself editable and run the fixed-improver control. With the initial meta agent held fixed for the same number of iterations, the median test score in simulated quadruped reward design reaches 0.213 against 0.372 when the meta agent improves itself (five runs each). Hyperagents transferred to Olympiad-level math grading and run with their meta agent fixed yield an improvement of 0.630 where the initial agent yields none. In AIDE$^2$~\citep{arxiv260926457}, a fixed outer-loop agent rewrites a research agent's code, and the seven accepted improvements transfer to four held-out benchmarks. When the evolved agent instead drives the outer loop, its final grade is close to that of the human-designed driver (0.780 against 0.782 over three seeds), a result its authors call inconclusive.

Other routes make the improver a learned component. Randomly initialized learned optimizers can train one another and, once better, speed up their own training~\citep{metz2021training}, and a self-referential weight matrix modifies its own weights, including the transformations that generate later modifications, although its authors leave a dedicated measurement of self-improvement to future work~\citep{irie2022modern}. MASS~\citep{arxiv261012176} fine-tunes one model that proposes, executes, and judges multi-agent workflows on its own traces; when workflows from two generations are run by the same base executor, those of the newer generation win 64\% of comparisons, although discovery still used each generation's own executor and judge, and its robotics tasks are simulated. SpecRSI~\citep{arxiv261016491} trains its improver on completed searches with the outer update rule fixed and runs the original and the updated improver on the same released artifact, where the updated improver yields a gain of 6.9 against 5.7 percentage points; some of its joint-update conditions have unequal training cost.

Two recent systems sharpen what an R2 test must measure. The Huxley-G\"odel Machine~\citep{arxiv251021614} reports that an agent's own benchmark score only weakly predicts how good its self-modified descendants will be (weighted correlations of 0.27 to 0.44), whereas a score pooled over its descendants predicts it better (0.63 to 0.78). Selecting by that pooled score, with the starting agent and an 800-evaluation budget held fixed, its best agent reaches 56.7\% on a 60-task SWE-bench Verified subset against 53.3\% for DGM and 50.0\% for SICA in single runs, and improves from 34.8\% to 40.1\% on 207 held-out SWE-bench Lite tasks. The Red Queen G\"odel Machine~\citep{arxiv260626294} lets learned evaluators evolve alongside the agents they score and promotes an evaluator only when it beats the incumbent on a fixed ground-truth anchor. Against frozen evaluators, in one run per arm, its coder passes 82.1\% against 75.0\% of 28 held-out tasks (a winner found with more search tokens); in a simulated Unitree Go2 reward-design task, its mean held-out jump height is 0.365 against 0.306\,m over five runs per arm. Its comparison isolates evaluator co-evolution while the meta-agent is editable in both arms, so, unlike DGM and HyperAgents, neither the Huxley-G\"odel Machine nor the Red Queen G\"odel Machine compares improved agents modifying themselves with a fixed base agent making every modification at a matched budget.

\paragraph{Physical boundary cases.}
\label{sec:physical-boundary}

Several robot systems, some evaluated only in simulation, are described by their authors as recursive self-improvement or recursive improvement: the criticality model's policy-level RSI, EndoRSI, EmbodiRSI, Local Recovery Supervision, MobiAgent, RegenHarness, Embodied Turing Machines, RE-0, AdaRSI, PhysEvo, D-World, RIWANav, and RSI-Nav; EmbodiedRSI and RoboRSI use the term in their names. Table~\ref{tab:rsi-systems} lists what each changes, where it was tested, and where this review discusses it. Judged by the claims of Section~\ref{sec:key-terms}, they revise policies, models, skills, data, or procedures, and several also accumulate knowledge that guides their next revision (Section~\ref{sec:repair-histories}). PhysEvo, whose meta-agent edits its own diagnostic tools, comes closest in mechanism (Section~\ref{sec:revise-tools}). Turning any of them into evidence for R2 would require the R2 comparison of Section~\ref{sec:key-terms}, run on held-out failures with matched learners.

\begin{table}[!htbp]
\centering
\caption{Robot self-improvement systems using RSI terminology in papers, system names, or project materials, assessed against the claims defined in Section~\ref{sec:key-terms}; no row establishes R1 or R2. \rsiicon\ and sand shading mark systems whose improver's knowledge or tools change, although no comparison isolates a recursive gain. Dashes mark settings a study does not report.}
\label{tab:rsi-systems}
\begingroup\footnotesize
\setlength{\tabcolsep}{3.5pt}\renewcommand{\arraystretch}{1.12}\arrayrulecolor{PAIRule}
\begin{tabularx}{\linewidth}{@{}L{2.8cm}L{3.25cm}YL{3.05cm}L{2.55cm}@{}}
\paitoprule
\pahead{System} & \pahead{What changes} & \pahead{\envsim\ Simulation / offline} & \pahead{\envhw\ Hardware} & \pahead{Claim / status}\\
\paimidrule
\addlinespace[3pt]\multicolumn{5}{@{}l@{}}{\tablegroup{Improving policies, world models, and knowledge (Section~\ref{sec:base-loop})}}\\\paihair
\tableworkref{D-World}{arxiv261017114} & Digital twin revised by a frozen LLM; residual-gated & \nores{} & \setres{Dual-arm}{one revision: depth residual 22.4 $\to$ 1.5\,mm} & Model accuracy\\\paihair
\tableworkref{MobiAgent}{arxiv261003476} & Skills from verified rollouts & \setres{RoboCasa}{7.5\% $\to$ 27.5\%} & \setres{Astribot S1}{32.5\% $\to$ 57.5\%} & Behavioral\\\paihair
\tableworkref{Embodied Turing Machines}{arxiv261012369} & Stateful decision code & \setres{RoboDojo, 42 bimanual tasks}{70.24\%} & \nores{} & Behavioral\\\paihair
\tableworkref{RoboRSI}{arxiv261012424} & Branch-local patches after review, tests, replay & \setres{Four benchmarks}{+2.7 to 11.0 points over a fixed agent, adapting online on 3 of 4} & \setres{WheelSingleArm M1}{104 case-study runs} & Behavioral\\
\addlinespace[3pt]\multicolumn{5}{@{}l@{}}{\tablegroup{Improving learning procedures (Section~\ref{sec:emerging-directions})}}\\\paihair
\tableworkref{Criticality model}{arxiv260728251} & Collection rule; predictor retrained by a fixed rule & \setonly{Locomotion, manipulation, VLA} & \setonly{Banana placement} & Behavioral\clnote{fixed-rule procedure}\\\paihair
\tableworkref{EmbodiRSI}{arxiv260938905} & Demos and corrections aimed at weaknesses & \setres{From deployment scenes}{50.4\% $\to$ 83.5\% over two rounds} & \setres{ALOHA, 14 subtasks}{83.1\% vs 75.0\%; 10 vs 200 real trajectories per subtask} & Behavioral\\\paihair
\rowcolor{PAISand}
\tableworkref{EmbodiedRSI}{arxiv261010498} & Code and skill hypotheses, experiments, which memories to keep & \setres{RoboCasa365 unseen tasks}{71.3\% vs 40.1\%} & \setres{SO-101, zero-shot}{46.0\% $\to$ 71.3\%} & Behavioral\rsimechanism\\\paihair
\tableworkref{Local Recovery Supervision}{arxiv261005151} & Recovery demos at audited failures & \setres{LIBERO-Goal, preliminary}{88 vs 78 of 100, matched steps} & \nores{} & Behavioral\\\paihair
\tableworkref{RE-0}{arxiv260932416} & Admitting teacher patches into student data & \setres{robosuite restacking}{4/50 $\to$ 31/50; admission rule ablated on lift} & \setres{AgileX PiPER, zero-shot}{13/20 restack} & Behavioral\clnote{procedure component}\\\paihair
\tableworkref{RIWANav}{arxiv261008640} & Which action--video pairs refit the world model & \setres{Offline urban navigation}{+7.94 points vs frozen world model} & \setres{Custom mobile robot}{Qualitative only} & Behavioral\\\paihair
\tableworkref{RSI-Nav}{arxiv261015782} & Filtered self-written instructions; shard-wise retraining & \setres{R2R-CE}{52.9\% $\to$ 68.6\%} & \setres{DEEP Robotics M20}{Qualitative only} & Behavioral\\\paihair
\rowcolor{PAISand}
\tableworkref{AdaRSI}{arxiv261020474} & Update placement; experience document & \setres{RoboCasa365 unseen tasks}{4.6\% $\to$ 13.8\%} & \nores{} & Behavioral\rsimechanism\\\paihair
\tableworkref{EndoRSI}{arxiv261013376} & Policies from a human-governed coding agent & \nores{} & \setres{Benchtop phantom}{2/21 $\to$ 17/21 in four generations} & Behavioral\\
\addlinespace[3pt]\multicolumn{5}{@{}l@{}}{\tablegroup{Improving the improver (Section~\ref{sec:reviser})}}\\\paihair
\rowcolor{PAISand}
\tableworkref{PhysEvo}{arxiv261008995} & Task tools and skills; meta-agent's own tools & \setres{RoboDojo held-out layouts}{62.0\% (dimension mean) vs 47.2\% published reference} & \setres{AgileX PiPER, five tasks}{84\% of 25 trials while revising} & Behavioral\rsimechanism\\\paihair
\tableworkref{RegenHarness}{arxiv260927612} & Harness configuration; cannot weaken its commit gate & \nores{} & \setonly{Quadruped deployment} & Revision gain not measured\\
\paibottomrule
\end{tabularx}\endgroup
\end{table}

\paragraph{Paths toward physical RSI.}
\label{sec:rsi-paths}

Four design requirements follow from the physical constraints discussed above. First, a protected success evaluation, with an outcome check that stays outside what the improver can change (Section~\ref{sec:evaluator-integrity}). Second, staged evaluation: candidate revisions are screened in simulation or on stored traces, and hardware trials go to matched learners for the few that survive (Section~\ref{sec:test-controls}). Third, repair records that carry their conditions of applicability, so that the improver can tell when an earlier lesson no longer holds (Section~\ref{sec:revision-knowledge}). Fourth, a budget that counts trials, resets, wear, and supervision, because revision histories accumulate slowly when each entry costs physical rounds (Section~\ref{sec:paradigms-physical}). A roadmap of autonomy levels~\citep{arxiv260911873} separates structural recursion, in which a revised improver, verifier, or successor generator is merely inherited, from effective recursion, in which it produces better later improvements. In its own scheme of levels, the roadmap places embodied intelligence at the frontier of choosing improvement strategies and acquiring experience, with deployment adaptation shown mainly in simulation. Reaching the top level of the concurrent scheme in Section~\ref{sec:key-terms}~\citep{arxiv261018842} would require comparisons such as those designed in Section~\ref{sec:recursive-revision}.

\begin{paiinsight}[PAISyn3]{Robots have the machinery of recursion, not yet controlled evidence}
\begin{insightpoints}
\item \textbf{The improver's knowledge already changes.} Repair notes, experience documents, and a meta-agent's own diagnostic tools change what robot improvers know and use.
\item \textbf{No physical comparison yet isolates a recursive gain.} The closest results, from HARBOR, AURA, and EmbodiedRSI, are uncontrolled and come from simulation.
\item \textbf{Digital systems show that the control can be run.} The Darwin G\"odel Machine and HyperAgents compare improved agents modifying themselves with a fixed base agent, where executable tests make each candidate cheap to judge; physical revision histories accumulate slowly because each entry costs physical rounds.
\item \textbf{From mechanism to test.} The two-round test of Section~\ref{sec:recursive-test} would isolate either link.
\end{insightpoints}
\end{paiinsight}

%% file: sections/08_dependencies.tex
\sectionaccent{PAICopper}
\section{Challenges and Future Directions for Physical RSI}
\label{sec:challenges}
\label{sec:open-challenges}

This section turns to what stands in the way of physical RSI. Section~\ref{sec:cost-ledger} asks what current evaluation cannot yet measure and how judges can stay aligned as behavior improves; Section~\ref{sec:compounding} explains why observed gains saturate, regress, or fail to carry forward, and how retained improvements can survive system change and finite resources; and Section~\ref{sec:safety} asks how practice can stay safe across rounds. Section~\ref{sec:agenda-diagnosis} sets out what stands between current results and verified recursion, from locating the bottleneck and choosing a usable experiment to learning a transferable repair principle and testing whether it makes improvement recursive. Table~\ref{tab:agenda-overview} lists, for each open problem, the conditions it serves, its central question, and a starting benchmark.

\begin{table}[!htbp]
\centering\small
\caption{The open problems of this section at a glance, in the order of the subsections that discuss them. \condyes\ marks the main focus of each direction among the four conditions: informative feedback (IF), usability (U), robustness to distribution shift (DS), and continued practice (CP). Starting benchmarks are proposals from the corresponding discussions.}
\label{tab:agenda-overview}
\footnotesize\setlength{\tabcolsep}{4pt}\renewcommand{\arraystretch}{1.18}
\begin{tabularx}{\linewidth}{@{}L{3.3cm}C{1.12cm}*{4}{C{0.62cm}}Y Y@{}}
\paitoprule
\pahead{Direction} & \pahead{Section} & \pahead{IF} & \pahead{U} & \pahead{DS} & \pahead{CP} & \pahead{Central question} & \pahead{Starting benchmark}\\
\paimidrule
Keeping Judges Aligned & \ref{sec:agenda-judges} & & & \condyes & & Refresh the judge, keep the task & Vary only the evaluator; protected outcomes\\
\paihair
Surviving System Changes & \ref{sec:maintenance} & & \condyes & \condyes & & Reuse, adapt, revalidate, or retire? & Declared changes plus gradual drift\\
\paihair
Budgeting Physical Resources & \ref{sec:agenda-budget} & & & & \condyes & Spending robot time, attention, compute, wear & Task sequences with upkeep; Pareto reports\\
\paihair
Finding the Bottleneck & \ref{sec:agenda-diagnosis} & \condyes & & & & Which component limits improvement? & Injected faults; held-out natural failures\\
\paihair
Choosing the Next Experiment & \ref{sec:agenda-experiment} & \condyes & & & & Which probe yields usable evidence? & Joint vs fixed, information-gain, sequential\\
\paihair
Transferring Repair Knowledge & \ref{sec:revision-knowledge} & & \condyes & \condyes & & Do repair lessons transfer to new failures? & Two-round test (Figure~\ref{fig:recursive-test})\\
\paibottomrule
\end{tabularx}
\end{table}

\subsection{Measuring Improvement}
\label{sec:cost-ledger}
\label{sec:evidence-approaches}

Section~\ref{sec:benchmarks} described the benchmarks and metrics in use; this subsection asks what they miss. Benchmarks test a system at one point in time, whereas self-improvement unfolds over rounds, so the test tasks must stay outside what the revising system can tune against. HarnessPAI reports 96.5\% on LIBERO-PRO evaluation seeds that include its development seeds, and RoboRSI improves online on its evaluation tasks while its baselines do not. In RLE-Bench~\citep{arxiv260934210}, coding agents improve on the development evaluations they can query but lose 19 to 41 points of verifier score, out of 100, on hidden robustness perturbations in simulation. Held-out seeds, tasks, and perturbations that the revising system never queries are part of a protected success evaluation. A held-out set also loses its validity when its results are reused repeatedly to select candidates, because the selection can overfit the held-out set itself~\citep{dwork2015holdout}.

Comparisons across rounds are rarer than single-round gains.\label{sec:rsi-evidence} Behavioral improvement is widely established (Tables~\ref{tab:policy-systems-comparison}--\ref{tab:modular-systems-comparison}), but later-learning and procedure claims need matched learners (Section~\ref{sec:key-terms}). Few studies supply them. PLD does so for the training value of collected data on hardware, and FIERCE, RLER, WoVR, SAVE, and the demonstration-requesting RoboCoach do so for single procedure components. Integrated pipelines such as VLAW and the RoboCoach system~\citep{arxiv260121570}, whose coding agent changes architecture, data augmentation, optimization, and validation code together, improve as a whole, and most results cover only the few rounds reported. For recursion, three judgments must be kept apart: a component of the procedure changes; the improver's knowledge or tools change, as in AdaRSI, HARBOR, AURA, or PhysEvo; and a recursive gain is isolated by a frozen-knowledge or fixed-improver control with matched learners and budgets. The physical systems reviewed here reach the second judgment at most, while digital systems such as the Darwin G\"odel Machine and HyperAgents run fixed-improver controls with executable tests or in simulation rather than through physical trials. Table~\ref{tab:claim-status} summarizes where the reviewed studies stand on each claim.
\begin{table}[!tbp]
\centering
\caption{Where the reviewed studies stand on each claim of Section~\ref{sec:key-terms}. Systems are representative and follow the assessments in the sections that discuss them; those without an icon are digital.}
\label{tab:claim-status}
\begingroup\footnotesize
\setlength{\tabcolsep}{4pt}\renewcommand{\arraystretch}{1.18}\arrayrulecolor{PAIRule}
\begin{tabularx}{\linewidth}{@{}L{2.75cm}L{4.1cm}Y@{}}
\paitoprule
\pahead{Claim} & \pahead{Strongest support found} & \pahead{Representative systems}\\
\paimidrule
Behavioral improvement & Controlled, hardware & \envhw\ HIL-SERL, RECAP, ConRFT, RL Token; \envsim\ DynaHarness, URAI\\\paihair
Later-learning benefit & Matched learners, hardware & \envhw\ RoboCat, PLD, CIDER, ACCRUE; \envsim\ HarnessPAI\\\paihair
Procedure improvement & Matched learners, hardware, single components & \envhw\ FIERCE, RLER, SAVE, RoboCoach (demo requests); \envsim\ WoVR\\\paihair
R1 & Uncontrolled, simulation & \envsim\ HARBOR; \envsim\,\envhw\ AURA, EmbodiedRSI; mechanism only: \envsim\ AdaRSI; \envsim\,\envhw\ PhysEvo\\\paihair
R2 & Fixed-improver controls, digital only & DGM, HyperAgents, VERSE; no physical system\\
\paibottomrule
\end{tabularx}\endgroup
\end{table}

\label{sec:component-tests}When retained knowledge enters a decision rather than weights, the corresponding measure is a component test that holds the decision-maker fixed and varies only the retained resource. ASPIRE~\citep{arxiv260700272} compares debugging a real robot's controller with and without transferred repair guidance, although whether starting programs and debugging budgets were matched is not stated; REVOLVE's~\citep{arxiv260914633} memory instead enters outcome judgments, so its test would compare judgments of the same observations under old and updated memory, a pathway that its reported judgment--policy trajectories do not isolate.

Even a well-designed comparison establishes improvement only relative to the success judgments made by a person, sensor, or model. Hardware studies differ in whether that channel is independent of the signal used for learning. RoboCat~\citep{arxiv230611706} uses learned reward models for autonomous success detection and relabeling, but counts final evaluation outcomes manually. In SILVR's real-robot experiments~\citep{arxiv250606658}, human judgments supply both the filtering signal and the evaluation outcome. Q-Planning~\citep{arxiv260821204} goes further: a human-provided label supplies the terminal reward on the real robot, and its per-iteration success rates are computed from the same collection episodes. Each protocol supports a different claim. A shared judge can measure improvement against its own criterion but cannot reveal a systematic error in that criterion; an independent outcome audit tests that additional link. The same caution applies to an agent's own report of completion: in the GPT-6-Astra evaluation~\citep{arxiv261022471}, self-reported progress diverged from the actual completion conditions on simulated tasks, and its proposed benchmark keeps judging the trajectory after the agent declares that it is done.

Tables~\ref{tab:policy-systems-comparison}--\ref{tab:learning-procedures} separate execution from outcome judgment for this reason.\label{sec:outcome-channels} Final scorers, or their separation from training, remain unclear for SOAR, SI-EFM, RAPolicy, SARM2, VERITAS's hardware tests, CIDER, RISE, PlayWorld, HIL-SERL, PLD's hardware tests, and VLAW. SI-EFM's collection labels and VLAW's human filter audit do not establish independent final checks. DexTouch-WM specifies five human raters for its model-evaluation comparison; their role in the later mixed-data policy test is not separately stated.

Success-evaluation details are also unresolved for Robo-Cortex, Zeva, ReSYNC, ASPIRE, Uni-Skill, Criticality model, and GLIDE. REVOLVE provides human-referenced judgment accuracy without specifying the final policy judge. FIERCE reports independent terminal checks but insufficient detail to assess separation. These are reporting gaps, not evidence of incorrect judgments.

A success rate should be read with the number of trials and independent training runs behind it.\label{sec:reporting} A few trials leave a wide interval even at perfect success, and trials that share scenes, resets, checkpoints, or training runs are less independent than a binomial calculation assumes. Reports should therefore state trial and run counts, how tasks, initial states, and checkpoints were selected, and whether judges agreed or were blinded~\citep{arxiv240909491}. Differences between methods should carry their own uncertainty. Sequential designs lower the cost: AnyRank~\citep{arxiv261014603} assigns more trials to policies that are hard to separate and needs up to 50\% fewer hardware evaluations than earlier rigorous rankings. RoboArena~\citep{arxiv250618123} instead ranks policies from double-blind pairwise comparisons crowd-sourced across seven institutions on the DROID platform, and AutoEval~\citep{arxiv250324278} supplies automatic success detection and scene resets so that real-world evaluation runs with little human involvement.

A cost comparison is meaningful only once the outcome and its uncertainty are specified. Cost accounting should separate setup and signal construction, acquisition of the revision, downstream training and evaluation, deployment, and later maintenance. ASPIRE reports model tokens to the first working program or to budget exhaustion, a measure that excludes robot time and total system cost. VLAW matches 50 physical rollouts per task category per round against its downstream baselines, while also fitting models and generating synthetic data. FIERCE reports 300 refinement rollouts per condition on each real task and explicitly excludes initialization, evaluator adaptation, calibration, evaluation, resets, and interventions from that count. These scopes answer useful but different questions, and ranking costs across studies would mix them. Table~\ref{tab:experience-connections} preserves the reported scope; a complete comparison of the cost to reach a target would additionally count robot episodes and hours, resets and human time, compute or tokens, failed runs, and revalidation after later updates. Elapsed time belongs in this count too, because simulated benchmarks can pause while a model deliberates: in the GPT-6-Astra evaluation~\citep{arxiv261022471}, the standard RoboDojo budgets correspond to 8--76~s of commanded motion against an average of 55 minutes of computation, so a larger step budget raises simulated success at no cost in elapsed time. On a real Unitree G1 humanoid, about 90\% of each run went to deliberation and one manipulation trial took over an hour, and protective stops and actuator overheating ended trials on other robots. Its proposed benchmark therefore charges deliberation on a world clock that keeps running.

\paragraph{Keeping judges aligned.}\label{sec:agenda-judges}

Measurement inside the learning loop raises a related problem: an experiment becomes misleading if the supervision used to judge it no longer resolves the failures produced by the improved policy. The measurement of the task may then need to evolve without silently changing the task itself. An early grasping policy may mostly miss the object, so a coarse lift detector is informative. A stronger policy may lift the object faster while squeezing hard enough to damage it. Adding force-sensitive evidence can refine the measurement of the original intent (a secure lift without damage), whereas redefining success to favor speed despite damage changes the intended outcome. An evaluator update should state which task requirements remain unchanged and which previously missed failure modes it now detects.

Because behavior shifts toward subtler failures, an evaluator refreshed from examples it selected itself can reproduce its own blind spots, and one refreshed only on recent failures can forget distinctions needed for older behaviors. The supervision loop needs anchors that stay fixed across evaluator versions: outcome-linked examples spanning old and new failure modes, selectively requested physical or human judgments, and a protected success evaluation for disputed cases (Section~\ref{sec:evaluator-integrity}). In the freeze--test--refresh--relabel sequence of Section~\ref{sec:outgrowing}, these anchors serve as the test. Versioning must connect each score to the evaluator, evidence, and task definition that produced it.

A benchmark for evaluator refresh should therefore separate better measurement from better supervision. With the learner, data, and update budget fixed, it varies only the evaluator and its refresh rule, tests judgments on independently selected old and new outcomes, and then trains the same learner from each version's labels and evaluates the policies through protected physical outcomes. An evaluator counts as progress only if it preserves the intended ordering of outcomes across rounds and helps produce a better policy; fitting its refreshed labels more accurately is not enough.

\begin{paiinsight}[PAISyn4]{Different claims need different comparisons}
\begin{insightpoints}
\item \textbf{Recursion is the next claim to test.} Beyond a frozen-knowledge or fixed-improver control, a recursive claim needs independent success evaluation, evidence of durability, and an account of total cost.
\item \textbf{Success evaluation and cost qualify every claim.} Who judges success, how many independent runs stand behind a rate, and what was left out of the cost count determine how far a result generalizes.
\item \textbf{Open question: a shared reporting standard.} The reviewed hardware studies report success evaluation, independent runs, and cost accounting in different ways. Agreeing on a minimal standard built from the reporting items of Section~\ref{sec:reporting} is among the most immediately achievable steps for the field, and it would make self-improvement results comparable across laboratories.
\end{insightpoints}
\end{paiinsight}

\subsection{Accumulating Improvement}
\label{sec:compounding}

The gains reviewed in Section~\ref{sec:mechanisms} are often real but local. Consider the missed grasp once more. After the robot learns to grasp reliably, its new successes may teach little that it does not already know, and its contact model may not cover the faster motions it now makes. A repaired grasp may expose a weakness in the controller that executes it. A growing library of grasp skills may become hard to search, and maintaining all of this machinery may cost more than it saves. These failures map onto the conditions of Section~\ref{sec:four-uses}. Narrowing data strains \emph{informative feedback}, outgrown models and rewards reflect a lack of \emph{robustness to distribution shift}, and a moving bottleneck is a problem of \emph{usability}. Accumulation strains both usability and robustness to distribution shift, and infrastructure cost bears on \emph{continued practice}. Table~\ref{tab:condition-evidence} lists the comparisons that bear most directly on each condition. Carrying a benefit forward means that it remains useful for later behavior or learning; performance need not grow at an accelerating rate. Because recursion depends on gains that carry into the next round, each of these stalls is also an obstacle to RSI.

\begin{table}[!htbp]
\centering\small
\caption{Comparisons that bear most directly on each condition.}
\label{tab:condition-evidence}
\footnotesize\setlength{\tabcolsep}{4pt}\renewcommand{\arraystretch}{1.16}
\begin{tabularx}{\linewidth}{@{}L{2.95cm}L{5.6cm}Y@{}}
\paitoprule
\pahead{Study} & \pahead{Comparison} & \pahead{What it shows}\\
\paimidrule
\addlinespace[3pt]\multicolumn{3}{@{}l@{}}{\tablegroup{Informative feedback}}\\\paihair
\systemcite{Collection}{arxiv241101813} & \envsim\,\envhw\ More autonomous data; targeted vs random demos & Autonomous data saturate; failure-start demos help more\\\paihair
\systemcite{PLD}{arxiv251100091} & \envhw\ Two specialists, equal success and data & Recovery coverage, not collector success, sets training value\\\paihair
\systemcite{SARM2/\allowbreak SPIRAL}{arxiv260610305} & \envhw\ Stage-aware vs sparse reward, matched loops & Stage-aware helps, sparse hurts (crumpled garments)\\\paihair
\systemcite{XPACE}{arxiv260917372} & \envhw\ Mixture with 8\% constructed recoveries & Filtered recoveries can supervise; extra training uncontrolled\\
\addlinespace[3pt]\multicolumn{3}{@{}l@{}}{\tablegroup{Usability}}\\\paihair
\systemcite{SILVR}{arxiv250606658} & \envsim\ Translator frozen vs updated & Frozen translator limits planner gains\\\paihair
\systemcite{DexTouch-WM}{arxiv260920649} & \envhw\ All-real vs half-generated data & Better prediction, lower policy scores\\\paihair
\systemcite{FAN}{arxiv260921358} & \envhw\ Frozen vs updated normalization & Shifted coordinates can break retention\\\paihair
\systemcite{SHAPER}{arxiv260811350} & \envsim\ Skill-only, harness-only, full & Skill and harness edits interact\\
\addlinespace[3pt]\multicolumn{3}{@{}l@{}}{\tablegroup{Robustness to distribution shift}}\\\paihair
\systemcite{FIERCE}{arxiv260918651} & \envhw\ Refreshed vs frozen evaluator (one seed) & Feedback that tracks the policy helps\\\paihair
\systemcite{PlayWorld}{arxiv260309030} & \envhw\ Successive checkpoints, fixed model & Hardware success falls, imagined reward rises\\\paihair
\systemcite{VLAW}{arxiv260212063} & \envhw\ Full loop vs Filtered BC, DSRL & Loop wins at matched rollouts; refresh not isolated\\\paihair
\systemcite{CIDER}{arxiv260821899} & \envhw\ Continual vs single-task start & New tasks learned at least as fast\\
\addlinespace[3pt]\multicolumn{3}{@{}l@{}}{\tablegroup{Continued practice}}\\\paihair
\systemcite{MEDAL++}{sharma2023selfimproving} & \envhw\ Learned reset beside the task policy & Only occasional manual resets\\\paihair
\systemcite{ENPIRE}{arxiv260619980} & \envhw\ One, four, or eight agent--robot pairs & Faster wall-clock, higher total cost\\\paihair
\systemcite{Gate study}{arxiv260910873} & Conservative, paired-binomial, no gate (synthetic) & Error control vs learning opportunity\\\paihair
\systemcite{GLIDE}{arxiv260924996} & \envhw\ Unguarded vs refined-guardrail collection & With guardrails most attempts succeed; without, few\\
\paibottomrule
\end{tabularx}
\end{table}

\paragraph{When data stop teaching.}
\label{sec:narrowing}

When autonomous imitation keeps only successful rollouts (filtered behavior cloning), the next dataset increasingly reflects behavior that the current policy can already produce. A higher success rate can then raise the yield of accepted trajectories while each new trajectory teaches less. The plateaus reported by \citet{arxiv241101813} are consistent with this pattern. PLD~\citep{arxiv251100091} shows the complementary case: collectors with equal success can differ in training value.

When success and training value diverge, choosing what to practice next (a curriculum) must answer two questions. The first is \emph{where} another trial is informative, which failure prediction and competence progress answer differently. The second is \emph{what supervision} the selected state needs. If an adequate recovery already lies within the range of behavior the learner can produce, autonomous practice may expose it. If no useful action is available, repeated failure adds little, and an expert correction, a residual controller, or a new skill may be required.

The choice must also move with the policy. A state that once exposed a useful error can become routine after correction, while better recovery and reset mechanisms can make previously costly states reachable enough to practice. Conversely, an irreversible or unreliable reset can make a theoretically informative failure a poor use of a physical budget. Collection, correction, and recovery should be reassessed together, judged by the downstream learner's gain and the physical cost.

\paragraph{When models and rewards fall behind.}
\label{sec:outgrowing}

A moving curriculum changes the distribution of situations in which the next policy acts, and predictors and evaluators fitted to earlier rounds can lose their footing. Two distinct forms of drift result. First, a world model may have little \emph{support} (few comparable training examples) for the improved policy's actions and contacts, such as the faster grasping motions it now makes, so optimization can favor trajectories that look successful only because the prediction is wrong. Second, an evaluator trained on earlier behavior may remain accurate on old failures yet miss the subtler ones produced by a stronger policy. Agreement between prediction and evaluation is weak evidence when both are computed from the same imagined trajectory and share the same blind spot about the physical world.

Existing methods respond at three different points. WISE~\citep{arxiv260903681} restricts imagination to selected states and short horizons, HaWMPO~\citep{arxiv260909941} downweights rewards from chunks it scores as hallucinated, and VLAW~\citep{arxiv260212063} refits the world model on new physical rollouts. Restriction and downweighting protect an update within the model's current support; refreshing attempts to enlarge that support. The trade-off is temporal: a conservative imagination region can be reliable now yet exclude the transitions required for the next improvement, whereas expanding it consumes physical trials and helps only if collection reaches the new policy's relevant errors.

In PlayWorld~\citep{arxiv260309030}, prolonged fine-tuning inside the model can keep improving the imagined reward while reducing success on hardware, as the policy specializes to small prediction mismatches, so even a comparatively accurate model remains open to exploitation by optimization. Fresh physical rollouts act as an external anchor for deciding when to stop, refresh, or distrust imagination. Studies of generative models trained repeatedly on their own outputs provide an adjacent warning: without enough real data, such loops can progressively lose the tails of the data distribution~\citep{shumailov2024collapse} or lose diversity~\citep{alemohammad2024mad}, a coverage failure that any mixture of real and imagined data must monitor.

Evaluator updates require a complementary timing rule. Combining FIERCE's per-round refresh, which already freezes a versioned evaluator and relabels replay~\citep{arxiv260918651}, with PEBBLE's relabeling~\citep{arxiv210605091} suggests a freeze--test--refresh--relabel sequence: keep feedback stable during an update, test it against current outcomes, revise it when behavior shifts, and make retained data consistent with the new version.

\paragraph{When bottlenecks move.}
\label{sec:bottleneck}

Every repair, including the repairs of prediction and feedback just described, also changes what the next component receives. A bottleneck moves when a previously adequate recipient becomes the dominant source of end-to-end error. Two ablations expose such a recipient: in SILVR's simulated ablation~\citep{arxiv250606658}, a frozen action translator limits what an updated video planner gains, and in SHAPER~\citep{arxiv260811350}, skill guidance and the harness that delivers it have non-additive effects. Neither study follows the bottleneck across rounds, so the consequences below remain hypotheses that a multi-round test would need to check.

When an execution module admits a wider range of behavior, the predictor and evaluator should be retested on the newly reached states. A runtime that rescues failures may hide from the learner the raw mistakes that motivated the rescue, unless trajectories from before the recovery and the corrections themselves are retained. A growing skill set forces the harness to decide which repair applies and which evidence should be preserved. A simulated study of 123 rounds of agentic skill discovery~\citep{arxiv260931760} shows the resulting stall: changes kept passing their tests, yet the target task never succeeded, because perception could not represent the needed relation, learning concentrated on the first step of long skill chains, and the agent optimized the evaluator's errors. A convincing test should evaluate the full pipeline on the distribution reached after the repair, ablate the receiving component, and inspect the next learning update, in addition to rechecking the repaired component.

\paragraph{When knowledge is lost or misused.}
\label{sec:accumulation}

As repairs accumulate, the usability problem becomes one of retention and retrieval. \emph{Retention} asks whether an earlier behavior remains available after the next update; \emph{retrieval} asks whether the right behavior or record is selected under the present conditions. A system can satisfy one without the other. An old skill may survive in the weights yet be routed incorrectly, while a relevant trajectory may be retrieved but decoded through an action interface whose coordinates have changed. Lasting improvement must preserve content together with the distinctions, interfaces, and applicability conditions that make it usable.

Existing methods protect different parts of this path. Three methods rehearse earlier behavior: Memory Anchors~\citep{arxiv260826545} through targeted replay, CIDER~\citep{arxiv260821899} through distillation, and Dream to Remember~\citep{arxiv261013816} inside a frozen world model. Pretraining changes how much protection is needed. On LIBERO suites in simulation, pretrained VLAs forget far less than small policies trained from scratch, simple experience replay sometimes removes forgetting, and skills that appear forgotten recover quickly with finetuning~\citep{arxiv260303818}. Interpolating finetuned weights with the pretrained ones retains generalist abilities while learning a new task, in simulated and real experiments~\citep{arxiv251208333}. Dream to Remember also learns new tasks in imagination, but only where that model is accurate enough. ACCRUE~\citep{arxiv261018596} instead carries lightweight actor--critic networks across a task sequence, cutting real-robot training time for later tasks by 31.0--68.5\%, and FAN~\citep{arxiv260921358} keeps the numerical meaning of retained actions fixed. These mechanisms complement one another: better rehearsal cannot repair coordinate drift, and stable coordinates do not decide which memory or skill applies. Their evaluations also use bounded task streams, so retrieval from an indefinitely growing repertoire is still to be tested.

External memories make that final problem explicit. Robo-Cortex~\citep{arxiv260518729} consolidates episode histories into retrievable principles and heuristics, whereas AllDayNav~\citep{arxiv260610927} stores temporal context and can replace or extend visual memories as navigation experience accumulates. Their reported system-level gains support the usefulness of maintained memory; a general rule for detecting when a once-valid record has become stale has yet to be developed.

Evaluation should expose each burden separately. The task suites of Section~\ref{sec:benchmarks} provide breadth, but each still needs a continual-learning protocol on top. After every update, the informative record is a matrix: performance on the new task, on earlier tasks, on deliberately conflicting states, and under shifted conditions in which stored knowledge may no longer apply. The matrix separates forgetting, retrieval error, and stale reuse, which a single final success rate would collapse together.

\subparagraph{Surviving system changes.}\label{sec:maintenance}
Repair knowledge remains useful only while its applicability conditions hold. A system revision can break these conditions at several interfaces: a sensor or calibration change alters observations, a controller or API change alters action semantics, and a new gripper changes contact dynamics and reachable effects. Whether the change is labeled ``hardware'' or ``software'' matters less than which relation the retained artifact assumed. Replacing a gripper may invalidate a force threshold or joint-space recovery while preserving the higher-level strategy of releasing, retreating, and regrasping; renaming an API call may break code while leaving the intended behavior realizable through a new wrapper.

Maintenance has to operate on artifact-specific conditions. Policy weights depend on observation and action representations and on dynamics, and a learned evaluator depends on sensing and calibration. A skill depends on its preconditions, effects, and available controller, and code depends on APIs, schemas, timing, and permissions. A gate, shield, or monitor depends on the policies and dynamics against which it was validated. A single timestamp or confidence score for a whole memory cannot show which assumption has become stale. Each retained change should instead state its validity conditions: its required inputs, action meanings, physical envelope, software dependencies, and the evidence under which it last worked.

Those conditions support four different responses. An artifact can be reused when its conditions still match, adapted when the task meaning is preserved but an interface can be translated, selectively revalidated when support is uncertain, or retired and rolled back when a required effect is no longer realizable. The distinction also guards against a misleading transfer claim. If a semantic recovery strategy succeeds on a new gripper only after a newly engineered adapter reconstructs its actions, the result tests the strategy and adapter together; the original commands were not what transferred. Conversely, rejecting the strategy because its old joint commands fail would overlook knowledge that remains valid at a higher level of abstraction.

A benchmark could introduce declared changes to sensing, embodiment, controller timing, APIs, and memory schemas, followed by gradual drift in calibration or wear, and compare unconditional reuse, relearning from scratch, and condition-aware adaptation or revalidation. Freezing the retained artifact while varying the adapter, and then the reverse, would separate semantic transfer from interface repair; the evaluation asks whether the system identifies the invalidated condition and restores held-out behavior without unsafe execution.

\paragraph{When costs outrun gains.}
\label{sec:investment}

Building a world model, evaluator, reset routine, or repair library consumes physical trials, engineering, supervision, and compute; reusing it can save later trials but also requires storage, calibration, and validation. Such an investment is worthwhile only if the artifact remains useful long enough to repay its cost. If setup costs $C_0$ and reuse $i$ saves $s_i$ while incurring maintenance cost $m_i$, all counted in one designated resource such as robot hours, the investment pays back only when $\sum_i(s_i-m_i)>C_0$ before a change in task, scene, or robot makes it obsolete (the payback period). When several resources are involved, the criterion applies to each separately rather than to their sum. This is a design criterion that the surveyed studies do not yet estimate; reporting it alongside a comparison would be a natural next measurement.

ENPIRE~\citep{arxiv260619980} provides a concrete but bounded example: its safety, reset, and verification interfaces are built once and reused during later policy search. In its fleet experiments, scaling from one to eight agent--robot pairs reduces the time to reach the target from roughly five to two hours on Push-T and from more than 1.5 hours to about 40 minutes on pin insertion. The same scaling study reports lower mean utilization per robot and faster-than-linear growth in token consumption at eight agents; neither a cost-optimal fleet size nor the rate at which costs are amortized across tasks has been measured yet.

The appropriate repair depends on both expected reuse and the risk of invalidation. A stable embodiment and a recurring family of tasks can justify expensive retraining or reusable infrastructure, whereas an imminent change in hardware or control interface favors a smaller, reversible repair whose assumptions are easier to revalidate. The resources involved are not interchangeable: more compute cannot undo hardware wear, a faster fleet cannot replace a success evaluation that no longer detects damage, and recovery can preserve access to trials without making those trials free. Comparisons should hold the target capability fixed and extend the cost accounting of Section~\ref{sec:cost-ledger} with robot utilization, the period over which the acquired capability is reused, and retained performance after later updates.

\subparagraph{Budgeting physical resources.}\label{sec:agenda-budget}
Beyond explaining why costs can outrun gains, the open problem is how to allocate robot time, human attention, computation, reset effort, and hardware wear over a deployment horizon. Simulation can replace some hardware interaction but creates model-fitting and physical-validation costs, and safety exposure should remain a constraint rather than be traded against savings in time or computation.

An allocation decision should combine expected downstream learning, frequency of reuse, maintenance and revalidation costs, and the probability that a task or system revision invalidates the artifact. It should also preserve option value: a reset capability, spare hardware capacity, or a human escalation path can be worthwhile because it keeps a future informative trial possible, even when it does not immediately raise task success. Three efficiency indicators can complement the causal tests: whether each round yields a gain, whether the gain arrives faster than human-engineered iteration, and whether the cost of each further gain, counted in robot time, attention, and wear, falls. A falling cost can also come from reused infrastructure or easier tasks, so it bears on the second recursive link only after the controlled comparison of Section~\ref{sec:recursive-test} has established that link.

A deployment-horizon benchmark could present a sequence of tasks, failures, maintenance events, and system changes under finite resource capacities, let a method choose at each stage whether to collect, simulate, request supervision, build reusable infrastructure, revalidate an artifact, or retire it, and compare a horizon-aware allocator with myopic and fixed-budget schedules. Because no universal conversion rate relates robot time, human time, compute, and wear, results should report their trade-offs and their sensitivity to the assumed horizon rather than a single cost.

\begin{paiinsight}[PAISyn5]{Diagnose the next broken connection, not only the current score}
\begin{insightpoints}
\item \textbf{Each fix changes the next round's problem.} Once kept data reflect what the policy already does, successes stop teaching and the next signal must distinguish a remaining failure; a repair changes what the next component receives, so the receiver needs retesting too.
\item \textbf{Models and judges fall behind.} Freeze feedback during an update, then test, refresh, and relabel as behavior shifts.
\item \textbf{Infrastructure must repay its cost.} Savings net of maintenance must exceed setup cost before the task, scene, or robot changes.
\item \textbf{Each failure appears in bounded tests.} Table~\ref{tab:condition-evidence}'s comparisons expose each failure within a few rounds, tasks, or configurations; none yet tracks interacting failures over a long deployment.
\item \textbf{Open question: can this diagnosis become part of learning?} In the reviewed studies, researchers diagnose these failures after the fact. A direct test would inject one failure of each kind during a long deployment and score whether the system identifies it and schedules the right next experiment (Section~\ref{sec:agenda-diagnosis}).
\end{insightpoints}
\end{paiinsight}

%% file: sections/09_continuation.tex
\subsection{Keeping Improvement Safe}
\label{sec:safety}

For a robot that improves itself, safety preserves \emph{continued practice}, the opportunity on which another learning round depends: a robot that damages its gripper or the object while practicing loses not only that attempt but also the trials that would have followed. Each update can also invalidate safety machinery built for the earlier system: an updated controller may exceed the assumptions of a safety guard, and a revised reset routine may no longer return the robot to a safe starting state. A recursive system faces this at two levels, because a revised learning procedure can also change what its safeguards must cover. Surveys of safe reinforcement learning~\citep{garcia2015safe} and of safe learning in control~\citep{brunke2022safe} catalog safeguards for collection, judgment, update, and execution. Some mechanisms below have been tested in simulation, formally, or on hardware; others are design requirements inferred from the ways self-modification can fail.

\subsubsection{Keeping Practice Possible}
\label{sec:practice-available}
\label{sec:risks}

Resetting is itself a capability that determines which experiments remain possible, as Leave No Trace~\citep{arxiv171106782} and ReLMM~\citep{sun2022relmm} show. Returning to an accessible starting state preserves the next experiment; a human reset preserves it too, but by moving part of the work outside the autonomous system. The resource that matters is safe access to an informative state, which a count of rollouts measures poorly.

Keeping practice available also means maintaining the interfaces that make another trial interpretable after wear, calibration drift, or software change.

Self-modification creates two windows of safety exposure. During improvement, harm can come from exploration, data collection, resets, and candidate testing before an update is accepted. After an update is accepted, the revised system can visit states or issue commands outside the assumptions of its existing monitors and recovery routines. The two windows can compound. A robot might first crush an object during exploration. A revised reset might then leave the object where the gripper cannot reach it, and an outcome detector that misses the damage can retain that transition as useful experience. Safety is a property of the whole loop of collection, judgment, update, and execution, not of the deployed policy alone.

Existing methods protect different parts of this loop. Safe exploration based on recoverability~\citep{moldovan2012safe} formalizes safety as keeping, with a chosen probability, a route back to a home state. Constrained policy optimization instead keeps policy search near constraint satisfaction at every iteration~\citep{achiam2017cpo}. Recovery RL~\citep{thananjeyan2021recovery} learns this judgment from data, in simulation and on a surgical robot: a safety critic estimates the chance of constraint violation, and a learned recovery policy takes over when that chance crosses a threshold. The Neural Simplex Architecture~\citep{phan2020nsa} switches to a baseline controller when a learned controller issues a potentially unsafe input, retrains the learned controller online, and returns control to it, in case studies on a ground rover and an artificial pancreas. Learning-based control~\citep{brunke2022safe} instead constrains behavior relative to a model of the dynamics and its uncertainty, for example through control barrier functions~\citep{ames2019cbf} that keep the state in a certified safe set; such guarantees remain conditional on assumptions about state estimation, actuation, latency, and the operating envelope. Each notion covers part of physical safety. Task success establishes that an intended outcome occurred; failure detection covers only the failure modes that a detector represents; and constraint satisfaction is meaningful only relative to the specified constraint and model. A self-modifying system can pass one of these tests while failing another.

The scope of a proposed modification determines which assumptions must be revalidated. Policy learning changes which actions are taken, and model-based practice can misestimate contact. A revision of memory, runtime, or resets can alter recovery, and a revision of feedback can change which failures are visible. For each candidate change, a record of assumptions should note what changed, which safeguard depends on it, what was rechecked, and which physical hazards remain unresolved. Table~\ref{tab:safety-changes} applies this analysis to the objects that the four uses of retained change can modify.

A recent survey of self-evolving humanoid robots~\citep{arxiv260913236} similarly organizes safety around admissible changes, verification gates, and boundaries for human oversight, proposed as design principles for later testing; a survey of safety in self-evolving language-model agents~\citep{arxiv261000093} finds the evidence strongest for admission and local containment and weakest for repairing what later versions inherit. A runtime guard can reduce risk while one candidate executes, yet the update process is safe only if admitting an update also revalidates the guard's assumptions after the change.

\input{tables/safety_changes}

\subsubsection{Protecting the Judge}
\label{sec:evaluator-integrity}

A learning system can increase its recorded reward without improving the intended physical outcome. Several mechanisms produce this gap, and each needs its own defense. A misspecified objective rewards the wrong behavior; an ordinary evaluator error mismeasures an otherwise relevant outcome; specification gaming exploits an existing mismatch; and tampering changes the process that generates the learning signal. \citet{amodei2016safety} separate problems such as side effects, reward hacking, and costly supervision, while \citet{everitt2021tampering} distinguish modifying the reward function from modifying its inputs. Moving the camera so that a slipping grasp is hidden affects the evaluator's input, whereas editing the grasp-success checker affects the evaluator itself.

Even ordinary evaluator errors matter once judged outcomes determine which experience is kept. \citet{arxiv241101813} identify accurate success detection as part of the burden of designing environments for autonomous robot data collection. A false positive can place a failed contact in a dataset filtered for successes, after which training can reproduce the pattern that was mislabeled as success. No deliberate manipulation is required: a detector error at judgment time becomes a data error at update time and can return as a behavioral error in the next round. Versioning the evaluator together with the data it labeled makes this propagation traceable, though finding the wrong labels still requires an independent check. In SEAR's simulated self-exploration~\citep{arxiv261018792}, one agent run saved a success verifier that was wrong on 91\% of the 699 later rollouts it declared successful, and the run ended at 4.6\% success, against 15.5\% and 23.6\% for sibling runs without it.

Once feedback itself can be revised, testing a candidate evaluator only on outcomes or labels that it helped select becomes circular. Its admission test needs a \emph{protected success evaluation}: a source of outcome judgments that the candidate update cannot also rewrite and that remains informative about the relevant physical failure. Depending on the hazard, this channel can combine held-out physical trials, write-protected raw measurements, dedicated instrumentation, and human review. ENPIRE's verification services, which its coding agents cannot modify, are one physical instance. A report should state which failure modes the channel can reveal, which components it trusts, and which blind spots remain.

\subsubsection{Admitting and Undoing Changes}
\label{sec:admission}

An admission rule, or gate, turns protected evidence into a decision about whether a candidate update should run, trading the risk of accepting a harmful update against the cost of rejecting a useful one. In the notation of Section~\ref{sec:key-terms}, a confidence-based gate admits a candidate learner $L'$ only when a lower confidence bound on $J(L')-J(L)$ is positive. Batch RL formalized such gates early: high-confidence policy improvement guarantees, at a chosen confidence level, that a returned policy performs above a chosen lower bound~\citep{thomas2015hcpi}, and SPIBB falls back on the baseline policy wherever the data leave the estimate uncertain~\citep{laroche2019spibb}. In a constructed one-step pushing diagnostic, \citet{arxiv260910873} hold the stream of candidate updates fixed and find that a gate based on a conservative range-based confidence bound admits none of them, whereas a paired-binomial gate admits 31.6\% of them at 2,000 episodes per stage; in separate closed-loop runs, always adopting the replay-trained update yields better learning than either gated alternative. The evidence is analytical and synthetic; a hardware safety comparison has yet to be run. Its broader lesson is that admission should report both how well it controls errors and how many learning opportunities it preserves at a stated interaction budget.

Robot-agent studies show the effect of the gate itself. When an optimizer agent revises a robot agent's harness, accepting every revision degrades it within a few rounds, whereas a champion--challenger rule raises held-out success on simulated LIBERO tasks from 51\% to 67\% with 100 rollouts per round~\citep{arxiv260939304}. In simulation, Robo-COP~\citep{arxiv261009228} reaches 73.8\% when it adopts a fine-tuned policy only after verifying an improvement, against 65.8\% for fine-tuning on a fixed schedule. A gate should also report what a candidate loses: a net gain can hide cases that the previous version solved, so a per-task regression check belongs beside the aggregate test. RegenHarness~\citep{arxiv260927612} makes such a protocol explicit for a robot-agent harness: a candidate change to context rules, task templates, routing, or recovery policies is accepted only if it preserves safety and evidence contracts, passes designated regression checks, improves a predeclared metric, and receives release authorization. The updater cannot weaken its own acceptance rule. Its real quadruped deployment documents the runtime rather than measuring gains from revision.

Once a candidate is admitted, runtime mechanisms limit what each trial can do, and they intervene at different points. The shield of \citet{alshiekh2018shielding} corrects actions that would violate a property stated in temporal logic, a formal language for requirements over time such as ``never enter this region''. The learned monitor of \citet{arxiv231017552} instead predicts states that are out of distribution or likely to fail and requests human oversight. The first is conditional on a specification and a model of the environment; the second is conditional on learned dynamics, failure labels, and the coverage of deployment. Neither protects against a hazard absent from its representation.

Other mechanisms move the check earlier or broaden the boundary around execution. VASO~\citep{arxiv260605395} checks whole plans, model-checking skill-based plans against temporal specifications and using counterexamples to revise reusable skill contracts; GLIDE~\citep{arxiv260924996} checks single commands, filtering teleoperation and policy commands through task-specific executable guardrails; and PhyAgentOS~\citep{arxiv260716636} wraps whole supervised sessions in compatibility checks before execution, verification, monitoring, and evidence collection. Safe online system identification~\citep{arxiv250421486} shows the value of stating assumptions precisely: its informative manipulator trajectories respect modeled input, state, and collision constraints, while its bounds on estimated parameters assume bounded sensor noise. Formal compliance, learned failure prediction, command filtering, and runtime mediation provide different kinds of evidence, and none translates directly into an unconditional probability of physical safety.

Systems that change their own software require a further boundary around authority. An updater should receive only the tool, file, network, and actuation permissions needed for the current experiment, while an independently controlled deployment service retains emergency stops, workspace limits, safety-critical calibration, and the authority to approve promotion. Each proposed change needs a provenance record linking the generating model, the prompt or memory used, the code diff, the data and evaluator versions, the test results, and the decision. A correct outcome does not show that each action rested on evidence gathered first; \citet{arxiv261007753} make this point for tool-using software agents by auditing trajectories against a provenance-bound evidence ledger. The checker can still err, but no single permission path can then silently rewrite both the candidate and every mechanism intended to assess it.

Promotion can then proceed from sandboxed tests of interfaces and logic to guarded physical trials within a declared operating envelope. Each stage should record which assumptions it tested and which physical unknowns remain; success in simulation alone cannot establish contact safety or hardware compatibility. Recovery also has to respect dependencies among components. A previous grasp policy may no longer work with a revised model, memory schema, or runtime, so rollback should restore a compatible versioned bundle rather than an isolated checkpoint. Even such a rollback restores only computational state. Trial limits and a stopping mechanism are still required before an update is admitted, because no rollback can undo a crushed object or a damaged gripper.

\subsubsection{Placing Human Oversight}
\label{sec:human-supervision}

Gates and safeguards face a practical dilemma: continuous human monitoring limits scale, but removing supervision leaves unfamiliar failures without a reliable response. Human attention is most informative where automation lacks either a safe action or trustworthy evidence: before an irreversible contact, at a state the policy has not been trained on, when outcome signals disagree, or when a safety-relevant component is proposed for revision. These situations assign the human different roles: taking control prevents harm, a corrective action supplies a learning target, an audit adjudicates an outcome, and an approval authorizes a change. Recording these roles separately matters, because one intervention can otherwise be counted simultaneously as safety, as training data, and as evidence of autonomy. Interruptions can also bias what is learned: an agent rewarded for finishing a task may learn to avoid being stopped, and safe interruptibility requires that learning neither seek nor avoid interruptions, a property that off-policy learners such as Q-learning already have and Sarsa can be given~\citep{orseau2016interruptible}.

Timing determines whether an intervention supplies that information. HIL-SERL~\citep{arxiv241021845} uses demonstrations and human corrections during real-world RL, intervening often early in training and less as performance improves. ThriftyDAgger~\citep{hoque2021thriftydagger} requests help at states that are novel to a policy ensemble or risky under a learned estimate of task success, with thresholds set from an intervention budget. SAGE~\citep{arxiv260829772} makes the threshold itself adaptive, so that predicted risk, uncertainty, and novelty decide when an expert policy takes over. These illustrate two allocation strategies, operator judgment and learned triage; neither is shown to detect every unsafe state. A late intervention may rescue the object without preventing a dangerous contact, whereas an overly conservative trigger can replace so much autonomy that the learner learns little about the limits of its competence.

A rescue also changes how the episode should be learned from. If an autonomous grasp starts to slip and a person then completes the lift, crediting the final success to the entire trajectory can reinforce the very action that required correction. TORL-VLA~\citep{arxiv260609337} addresses this case in real-robot experiments with an intervention-censored critic, which prevents success achieved after an intervention from being credited to the policy's actions before the takeover. The broader requirement is a record of who was in control: each transition should identify who acted, why control changed, and which later outcomes may legitimately supervise the preceding decision.

Fewer takeovers can coexist with more oversight overall, because control effort may fall while learning effort (demonstrations, corrections, and outcome labels) or assurance effort (audits, calibration, checker maintenance, and approvals) rises. Reports should separate these costs and state the trigger, who has authority to respond, and how quickly the supervisor can act. The broader survey of recursive self-improvement by \citet{arxiv260707663} likewise treats the closing of the loop as a continuum and identifies evaluation and the setting of research direction as persistent human bottlenecks. For physical systems, a meaningful claim about autonomy names which decisions became reliably automated and where in the loop supervision moved.

Proposing a change, admitting it, executing it, and judging its outcome can be run by the same people, as long as the four steps do not all depend on the same mutable component or unprotected signal and each stage retains evidence or control capable of exposing and containing failure in the preceding one.

\subsubsection{Revalidating Safety Across Generations}
\label{sec:safety-generations}

Admission and rollback act on one candidate at a time; a system that revises its own learning procedure must also ask whether the evidence behind its safeguards survives the revision. Three questions recur. First, does earlier safety evidence still apply? A guard validated for one controller, action normalization, or reset routine holds only for what it was validated on. FAN shows that updating action-normalization statistics shifts the coordinates shared by training and execution, so a bound stated in the old coordinates would need rechecking, and Table~\ref{tab:safety-changes} lists how other revisions can invalidate safety assumptions. Each generation therefore needs a record of which checks were run against which component versions, and a rule that marks dependent checks as stale when those versions change.

Second, who may change the safety boundary? ENPIRE builds its safety, reset, and verification services in a human-guided stage and exposes them as interfaces that its coding agents cannot modify, and RegenHarness forbids its updater to weaken its own acceptance rule; the same separation should extend to the improver, so that a revision of the learning procedure cannot also relax the checks that judge it.

Third, how are regressions across versions and bodies caught? A net gain can hide cases that the previous version solved, and a hardware change such as a replaced gripper can invalidate behavior that no software rollback restores; regression suites carried across generations and rerun after hardware changes keep earlier competence and safety margins visible. These requirements are inferred from the reviewed failures; testing them over several generations of a self-revising physical system is part of the open problem of surviving system changes (Section~\ref{sec:maintenance}).

\paragraph{Beyond safety.} Self-improving robots also raise questions that safety engineering does not settle. Fleet learning gathers experience in homes and workplaces, as when OpenBot-Fleet~\citep{arxiv240507515} uploads data from user-operated robots, so consent, retention, and access to recorded scenes become part of the design of the learning loop. When a robot revises its own learning procedure, responsibility for a harmful update must also stay traceable; the provenance records and separated authority described above serve accountability as well as safety.

\begin{paiinsight}[PAISyn6]{Keep the ability to learn safe, not only the current behavior}
\begin{insightpoints}
\item \textbf{Safety spans the whole loop.} Exploration, resets, candidate testing, and execution after an update each create their own risk; securing the deployed policy covers only the last of them.
\item \textbf{Protect the evidence that judges change.} One saved verifier in SEAR was wrong on 91\% of the later rollouts it passed; a revisable evaluator needs a success evaluation that the candidate cannot rewrite.
\item \textbf{Separate authority and plan for irreversibility.} No mutable component should both propose and judge a change, and because rollback restores only computational state, trial limits must precede admission.
\item \textbf{Safeguards that keep pace with self-modification.} The reviewed gates, shields, and monitors are each validated in narrow settings, and ENPIRE's safety and verification services are built once and then held fixed; the next step revalidates such safeguards across generations of a robot that revises its own learning procedure.
\end{insightpoints}
\end{paiinsight}

%% file: tables/safety_changes.tex
\begin{table}[!htbp]
\centering\small
\caption{How a revision can invalidate a safety assumption.}
\label{tab:safety-changes}
\footnotesize\setlength{\tabcolsep}{4pt}\renewcommand{\arraystretch}{1.18}
\begin{tabularx}{\linewidth}{@{}L{2.75cm}Y Y Y@{}}
\paitoprule
\pahead{Modified object} & \pahead{Assumption at risk} & \pahead{Targeted safeguard} & \pahead{Residual risk}\\
\paimidrule
Policy or state estimator & Controller's bounds, latency, uncertainty & Recheck limits, timing; monitored trials & Hidden contact or stale uncertainty\\
\paihair
World model or training data & Imagined contact and failure rates match reality & Held-out probes; validated regions only & Rare damage escapes limited probes\\
\paihair
Reward, judge, or selector & High scores mean intended, safe practice & Protected tests; audits of disagreements & Incomplete rubric; shared sensing errors\\
\paihair
Memory, skills, runtime, or reset & Compatible preconditions; recoverable end state & Contract checks; separate recovery tests; versioning & Rollback cannot undo damage\\
\paihair
Training or research code & Access limits, test isolation, budgets & Restricted writes; immutable tests; external stops & Permitted experiments, unmodeled hazards\\
\paibottomrule
\end{tabularx}
\end{table}

%% file: sections/11_research_agenda.tex
\subsection{Making Improvement Recursive}
\label{sec:agenda-diagnosis}
\label{sec:mechanistic-view}

A recursive improver must decide what to repair and what to try before it can learn from its repairs. This subsection takes these capabilities in turn, from diagnosis and experiment choice to transferable repair knowledge, and then designs the test that would show the loop closing.

\paragraph{Finding the bottleneck.}
The agenda begins with diagnosis because changing the wrong component consumes physical trials and can obscure the evidence needed for the next attempt. The same observed failure can have several causes, as the missed grasp of Section~\ref{sec:introduction} shows, so a failure label alone cannot determine which repair to choose. Calling a component the bottleneck therefore states a hypothesis to be tested against other, interacting explanations; logs can narrow those explanations but rarely determine which change will improve downstream behavior, and Table~\ref{tab:mechanism-diagnostics} separates evidence that supports a diagnosis before intervention from comparisons that test a candidate repair after intervention.

\input{tables/mechanism_diagnostics}

Physical traces make this inference harder because they are partial and often censored by safety mechanisms. An intervention may prevent the failure that would have revealed the weakness, while the final rescued success may conceal which autonomous transition required correction. A diagnostic trace should align observations, proposed actions, executed actions, any shield or human modification, feedback versions, and intervention timing. Unobserved signals should stay marked as unknown, never counted as evidence that a component worked. The \emph{bottleneck} is the component hypothesized to limit the next improvement, given the available repairs; diagnosing it does not presuppose a single independent root cause.

Failure explanation is a natural starting point. REFLECT~\citep{liu2023reflect} summarizes multisensory robot experience hierarchically and asks a language model to explain failures and plan corrections, and AHA~\citep{duan2025aha} fine-tunes a vision-language model on procedurally perturbed demonstrations to detect and explain manipulation failures. Both explain a failure from the recorded episode alone. A diagnostic system should go further, keeping competing explanations ranked and choosing a small, safe probe expected to separate them before attempting a costly repair. A concrete benchmark could inject matched policy, model, evaluator, and interface faults that produce similar terminal outcomes within the same task family. Single faults would test basic identification; unseen combinations and intervention-censored traces would test whether the method preserves uncertainty when causes interact or evidence is missing. Fault injection supplies controlled causal labels but not full realism, so the resulting repair choices should also be tested on held-out, naturally occurring hardware failures. Evaluation should measure calibrated cause rankings, information gained per probe, and, most importantly, whether the selected repair improves downstream behavior from matched starting points and physical budgets. This last test separates useful diagnosis from classification that merely names the injected fault.

\paragraph{Choosing the next experiment.}\label{sec:agenda-experiment}
While several causes remain plausible, the most diagnostic experiment and the most useful learning experience can differ. Suppose a grasp failure could reflect either a missing recovery action or an incorrect contact model. Repeating the contact with additional force sensing may separate those explanations, but evidence of a contact-model error helps only if a model update can represent the discrepancy and a later learner can use the revised prediction. An expert correction supplies an immediately usable action target, yet may leave the contact-model error unresolved. An experiment is useful for learning only if an available update method can use the evidence it provides, so its value depends on which uncertainty it reduces and whether the resulting update improves behavior.

An experiment also incurs reset, wear, intervention, and safety costs immediately, whereas its learning benefit appears only after updating and redeployment. Some informative states cannot be revisited safely, yet a promising repair may need one. A staged selector could first use stored traces, counterfactual replay, or short model-based probes, then request new contact or correction only while the remaining explanations would lead to different usable updates. Cheap probes narrow the question but cannot resolve a physical distinction that their data or model omits.

A benchmark for this problem should compare joint experiment--update selection with fixed collection, information-gain-only selection, and a sequential baseline that diagnoses first and chooses an update afterward, scoring held-out behavioral improvement rather than predicted information gain under the reporting rules of Section~\ref{sec:reporting}. Ablating either the predicted information gain or the predicted usability of the evidence would show which part of the joint decision matters.

\paragraph{Transferring repair knowledge.}\label{sec:revision-knowledge}
Recursion requires retaining what was learned from a repair, not only the repaired policy. A stored edit without its applicability conditions supports patch reuse, and a searcher that keeps rediscovering edits repeats a search rather than learning from earlier repairs.

Learning such knowledge requires histories that expose both success and failure. The same slip penalty can help one grasping policy and harm another because their exploration differs; a code repair can depend on an API, timing convention, sensor, or control rate. Each record should connect the proposed explanation and unresolved alternatives, attempted change, starting competence, component and interface versions, data and feedback, physical conditions, resources spent, and downstream successes or regressions. Failed revisions do more than serve as negative examples: they identify missing prerequisites or incorrect causal explanations. Selectively recording only accepted patches would make an editor appear more reliable while removing the evidence needed to decide when not to reuse them.

The main open problem is how to use accumulating real-robot revision histories to learn a procedure improver that transfers across failure types, and to verify that it also helps produce its own next improvement. If what is learned does not transfer beyond near-duplicates of past failures, repair knowledge amounts to a library of patches, and the two-round test below reduces to a test of reuse. The improver must separate transferable gains from those due to extra data, a stronger starting policy, or a changed apparatus, and recognize when to reuse an earlier repair, adapt it, or seek a different explanation. Benchmark tasks should vary starting competence, feedback versions, interfaces, and physical conditions while keeping tools, opportunities, and budgets matched. Direct retrieval of an old patch should be reported as reuse; a transferable repair principle should pass both comparisons of Figure~\ref{fig:recursive-test} on held-out failure families.

\paragraph{Testing recursive improvement.}\label{sec:recursive-revision} Can a system learn from attempted repairs to improve the way it produces further improvements? The two links of Section~\ref{sec:key-terms} describe a causal dependence across revisions, which no count of mutable components, checkpoints, or autonomous steps can establish. Section~\ref{sec:rsi-evidence} sets out the judgments that a recursive claim must keep apart; the two-round test below would isolate either link for a learning procedure.

\paragraph{Where recursion could start.}
\label{sec:recursion-routes}

Each of the four routes can supply the object of a recursive test: a rule for changing that route, together with records of how earlier changes fared.

For \emph{policies}, the natural rule to revise is the one that chooses which experience to collect. A criticality model already steers collection toward likely failures and, in two simulated multi-round runs, is retrained on the current policy's rollouts each round; a recursive version would also learn, from which collections taught later learners, how to revise that criterion.

For \emph{world models}, the rule decides when a model may judge. A learned model is the closest physical counterpart of the executable tests that make digital RSI cheap, because it can screen candidate changes before hardware trials, but its judgments must themselves be checked. On a physical arm, a world-model judge fine-tuned only on successes scores most held-out failing sequences as success-like, and fine-tuning on execution-checked counterfactual failures reduces this sharply~\citep{arxiv261009134}. A recursive version would learn from such mismatches how to refresh the judge, so that each refresh leaves fewer errors for the next.

For \emph{external knowledge}, the rule decides what to keep from a repair. Memories, skills, and code can store repair knowledge in inspectable form, with the applicability conditions that the preceding paragraphs ask each record to carry, and a frozen memory snapshot directly supplies the control described below.

For the \emph{learning procedure}, the rule is the improver itself, such as a reward editor or the procedure that trains a skill-library editor. The test below starts there, because Eureka and PRACTICE already supply fixed-improver baselines. Across the four routes, what differs is where repair knowledge lives and how it can be frozen, not what the test must show.

\paragraph{How to test recursion.}
\label{sec:recursive-test}

A bounded test can focus on reward construction, as in the missed-grasp example of Section~\ref{sec:reviser-criteria}, or on skill-library editing (Figure~\ref{fig:recursive-test}), for which existing systems already supply baselines. Eureka is a reward improver whose knowledge does not change, and PRACTICE is a learned skill editor whose own improver, the staged procedure that trains it, is fixed. The proposed test extends the matched downstream comparison of Figure~\ref{fig:downstream-comparison} from a revised procedure to the improver that produces it by comparing a learned improver before and after it learns from earlier repairs:
\input{figures/recursive_revision_test}

\begin{insightpoints}
\item \textbf{Hold the new problem common.} Give both versions the same new failure histories, tools, objectives, and revision budget. Let them generate or select candidate learning procedures, rather than only task controllers.
\item \textbf{Test what those procedures teach.} Apply the candidates to matched learners and evaluate common physical outcomes. Report the cost of updating the improver separately.
\item \textbf{Follow the improver's next revision.} Given common new repair histories, let the updated improver propose a revision of itself and compare the result both with the unrevised improver and with a control revision of the same improver proposed with its earlier knowledge frozen. Repeating the first two tests on the resulting improvers links further gains to what was learned.
\end{insightpoints}

The first two steps test the first link and the third tests the second, as stated by the R1 and R2 inequalities of Section~\ref{sec:key-terms} (Figure~\ref{fig:recursive-test}); retention, performance under changed conditions, and sustainable resource use decide whether the relation is useful beyond a local controlled test. SEAR~\citep{arxiv261018792} already supplies benchmark controls for such tests, separating retained experience from repeated attempts, task familiarity, privileged state, larger budgets, and chance.

\paragraph{What the test must control.}
\label{sec:test-controls}

A successor can score higher because of accumulated experience or a stronger starting system rather than a better improvement mechanism, as a concurrent survey also notes~\citep{arxiv261018842}. Three details decide whether the test is informative. First, freezing must match the form of the improver's knowledge: a parametric improver is compared with its earlier checkpoint, a retrieval-based improver with its earlier memory snapshot, and an improver whose knowledge lives in code or prompts with their earlier versions, while the base model, tools, and budget stay the same. This suffices for R1, but R2 also requires the control revision to start from the same parent. That holds when the acquired knowledge can be withheld without changing the rest of the improver, as with a memory snapshot or separable code or prompts; when the knowledge lives in the improver's weights, restoring the earlier checkpoint also changes the parent, so the R2 control for a parametric improver needs a design of its own. Second, the new failure families must be held out from everything the improver has learned from; otherwise the comparison measures memorization. Third, the physical cost should be staged: candidate procedures can be screened in simulation or on stored traces, so that hardware trials go to matched learners for the few candidates that survive. With one survivor per improver, the two comparisons need four hardware-trained learners per seed, because the updated improver's learner from the first comparison also serves as the parent baseline of the second: about 1,200 refinement rollouts at FIERCE's per-condition budget, excluding evaluation, resets, and repetition across failure families. Digital precedents run close analogues of both comparisons when evaluation is cheap; the physical version adds protected outcomes and a matched downstream learner. SpecRSI's same-artifact comparison, which runs the original and the updated improver on one released artifact, is a digital instance of this control.

Three near-term programs follow from the agenda. First, combine cross-component diagnosis with joint experiment--update selection under matched multi-resource budgets, so that the correct repair and the cost of identifying it both determine success. Second, test versioned evaluators and repair knowledge on held-out failures and declared hardware or software changes, using protected outcome checks to separate transfer from drift. Third, connect these tests across at least two causally isolated procedure-revision rounds and report their full acquisition, maintenance, and revalidation costs.

\begin{paiinsight}[PAISyn7]{The recursive target is knowledge about improving the learning procedure}
\begin{insightpoints}
\item \textbf{Test procedures, not controllers.} Both versions of the improver must produce learning procedures, judged through matched learners.
\item \textbf{Freeze the knowledge, not the resources.} The control keeps model, tools, and budget and withholds only the acquired repair knowledge, on held-out failure families.
\item \textbf{Where to start.} Reward construction and skill-library editing yield inspectable revisions that simulation can screen, leaving four hardware-trained learners per seed when one candidate per improver survives.
\end{insightpoints}
\end{paiinsight}

%% file: tables/mechanism_diagnostics.tex
\begin{table}[!htbp]
\centering\small
\caption{Checks before repair and comparisons after intervention.}
\label{tab:mechanism-diagnostics}
\footnotesize\setlength{\tabcolsep}{4pt}\renewcommand{\arraystretch}{1.18}
\begin{tabularx}{\linewidth}{@{}L{2.7cm}Y Y Y@{}}
\paitoprule
\pahead{Observed problem} & \pahead{Before repair} & \pahead{Repair comparison} & \pahead{Limits on benefit}\\
\paimidrule
More successes add little & Do logs contain recovery actions? & Matched data with vs without targeted corrections & Correction feasible; failure state recoverable\\
\paihair
Reward misses remaining failures & Rescore stored attempts against outcomes & Frozen vs revised feedback, common start & Needs useful action alternatives\\
\paihair
Useful examples, little policy change & Target fit; action representation & Data and feedback fixed; fitting changed & Targets may be wrong or infeasible\\
\paihair
Better prediction, no policy gain & Action ranking, contact error, command coverage & Real vs generated mixtures, same learner & Predictor may omit needed distinctions\\
\paihair
Plans fail in execution & Commanded vs realized motion; reach, timing & Common plans, alternative controllers & Infeasible plans stay infeasible\\
\paihair
Saved repairs help only locally & Retrieved context, prerequisites, control calls & Retrieval or runtime changed alone; new contexts & Needs available motor competence\\
\paihair
New learning erases old behavior & Old-task retests; coordinate changes & Replay, distillation, or stable coordinates & Hardware changes can still break it\\
\paibottomrule
\end{tabularx}
\end{table}

%% file: figures/recursive_revision_test.tex
\begin{figure}[!htbp]
\centering
\begin{tikzpicture}[art fonts,x=1cm,y=1cm,line cap=round,line join=round,
  every node/.append style={execute at begin node={\hyphenpenalty=10000\exhyphenpenalty=10000}},
  wcard/.style={fill=white,draw=#1!38,line width=.55pt,rounded corners=1.6mm,
    drop shadow={opacity=.12,shadow xshift=.5pt,shadow yshift=-.6pt}},
  rcard/.style={wcard=#1,draw=#1!70,line width=.7pt},
  hlab/.style={inner sep=0pt,anchor=base,text=ArtSlate,font=\sffamily\bfseries\fontsize{8.8}{10}\selectfont},
  ctext/.style={inner sep=0pt,align=center,text width=1.32cm,font=\sffamily\bfseries\fontsize{8.0}{9.4}\selectfont,execute at end node={\vphantom{g}}},
  bitem/.style={inner sep=0pt,anchor=base west,text=ArtSlate,font=\sffamily\fontsize{8.4}{10}\selectfont,
    execute at begin node={\textcolor{ArtRed!85!black}{\textbullet}\enspace\vphantom{Ag}}},
  vig/.style={inner sep=0pt}]
\colorlet{ArtOld}{ArtSlate!55}
\path[use as bounding box] (0,-7.76) rectangle (16.08,.56);
% Five columns shared by both links: knowledge .19--2.69 | improver 3.23--5.73 | procedure 6.28--9.13 | learner 9.67--12.52 | test 13.06--15.91.
% Each 0.54 cm gap holds a 0.30 cm arrow at the row center.
\newcommand{\lrow}[8]{% top, bottom, color, knowledge image, knowledge label, improver label, learner number, (unused)
  \begin{scope}
    \clip[rounded corners=1.8mm] (.09,#2) rectangle (16.01,#1);
    \fill[#3!4] (.09,#2) rectangle (16.01,#1);
  \end{scope}
  \draw[#3!50,line width=.6pt,rounded corners=1.8mm] (.09,#2) rectangle (16.01,#1);
  \foreach \xa/\xb in {.19/2.69,3.2325/5.7325}
    \path[wcard=#3] (\xa,#2+.08) rectangle (\xb,#1-.08);
  \path[rcard=ArtAmber] (6.275,#2+.08) rectangle (9.125,#1-.08);
  \path[rcard=ArtBlue] (9.6675,#2+.08) rectangle (12.5175,#1-.08);
  \path[rcard=ArtSage] (13.06,#2+.08) rectangle (15.91,#1-.08);
  \node[circle,fill=ArtAmber!90!black,text=white,inner sep=0pt,minimum size=.32cm,font=\fontsize{5.6}{5.6}\selectfont] at (6.585,#1-.38) {\faIcon{scroll}};
  \node[circle,fill=ArtSage!90!black,text=white,inner sep=0pt,minimum size=.32cm,font=\fontsize{5.6}{5.6}\selectfont] at (13.37,#1-.38) {\faIcon{clipboard-check}};
  \node[vig,anchor=west] at (.27,{(#1+#2)/2}) {\includegraphics[height=1.00cm,width=.86cm,keepaspectratio]{figures/#4}};
  \node[ctext,text=#3!85!black] at (1.94,{(#1+#2)/2}) {#5};
  \node[vig,anchor=west] at (3.3125,{(#1+#2)/2}) {\includegraphics[height=1.00cm,width=.86cm,keepaspectratio]{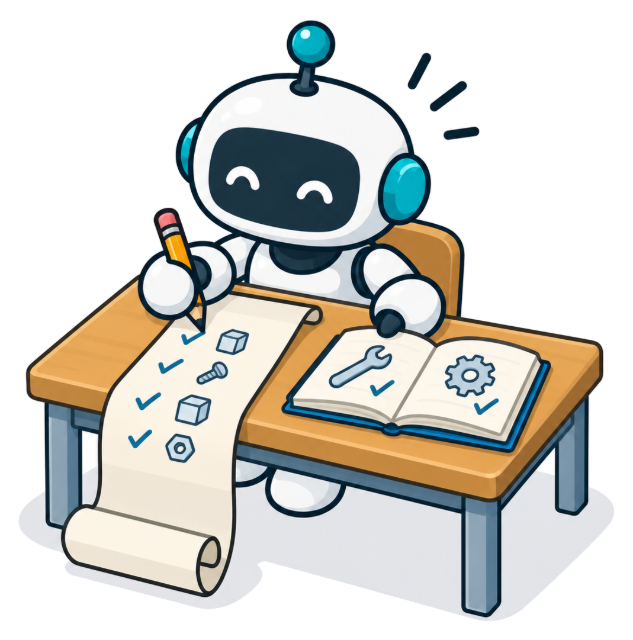}};
  \node[ctext,text=#3!85!black] at (4.955,{(#1+#2)/2}) {#6};
  \node[vig] at (7.70,{(#1+#2)/2}) {\includegraphics[height=.95cm,width=2.60cm,keepaspectratio]{figures/art_procedurew}};
  \node[vig] at (11.0925,{(#1+#2)/2}) {\includegraphics[height=.95cm,width=2.60cm,keepaspectratio]{figures/art_learner}};
  \node[circle,fill=ArtBlue!85!black,text=white,inner sep=0pt,minimum size=.32cm,
    font=\sffamily\bfseries\fontsize{8}{8}\selectfont] at (9.9775,#1-.38) {#7};
  \node[vig] at (14.485,{(#1+#2)/2}) {\includegraphics[height=.95cm,width=2.60cm,keepaspectratio]{figures/art_trial}};
  \foreach \xc in {2.96125,6.00375,9.39625,12.78875}
    \node[art fat=#3,minimum height=.30cm,minimum width=.30cm] at (\xc,{(#1+#2)/2}) {};}
% Link title band: badge, question, and what both pipelines share.
\newcommand{\linkband}[5]{% center y, number, question, shared conditions, color
  \path[fill=#5!6,draw=#5!40,line width=.55pt,rounded corners=2mm] (.09,#1-.27) rectangle (16.01,#1+.27);
  \node[circle,fill=#5!85!black,text=white,inner sep=0pt,minimum size=.32cm,
    font=\sffamily\bfseries\fontsize{8}{8}\selectfont] at (.41,#1) {#2};
  \node[anchor=base west,inner sep=0pt,text=#5!85!black,font=\sffamily\bfseries\fontsize{9}{10}\selectfont] at (.67,#1-.07) {#3\vphantom{Ag}};
  \node[anchor=base east,inner sep=0pt,text=ArtSlate!85,font=\rmfamily\itshape\fontsize{8.2}{9.6}\selectfont] at (15.85,#1-.07) {#4\vphantom{Ag}};}
% Header
\shade[left color=ArtPurple,middle color=ArtTeal,right color=ArtAmber] (.09,-.03) rectangle (16.01,0);
\node[hlab] at (1.44,.20) {\faIcon{archive}\enspace Revision input};
\node[hlab] at (4.4825,.20) {\faIcon{pen-fancy}\enspace Improver};
\node[hlab] at (7.70,.20) {\textcolor{ArtAmber!90!black}{\faIcon{scroll}}\enspace Candidate procedure};
\node[hlab] at (11.0925,.20) {\textcolor{ArtBlue!85!black}{\faIcon{user-graduate}}\enspace Matched learner};
\node[hlab] at (14.485,.20) {\textcolor{ArtSage!90!black}{\faIcon{clipboard-check}}\enspace Outcome check};
% (1) First link: does what the improver learned from repairs improve the procedures it produces?
\linkband{-.35}{1}{First link (R1): does repair knowledge improve the improver?}{Same new failure histories, tools, objectives, budget}{ArtBlue}
\lrow{-.70}{-2.05}{ArtOld}{art_retain}{Earlier\\knowledge}{Earlier\\improver}{1}{}
\lrow{-2.15}{-3.50}{ArtRed}{art_retain}{Acquired\\knowledge}{Updated\\improver}{2}{}
% (2) Second link: does a self-revision proposed with that knowledge beat one proposed with it frozen?
\linkband{-3.91}{2}{Second link (R2): does the better improver help improve itself?}{Same new repair histories, tools, budget, interface}{ArtPurple}
\lrow{-4.26}{-5.61}{ArtOld}{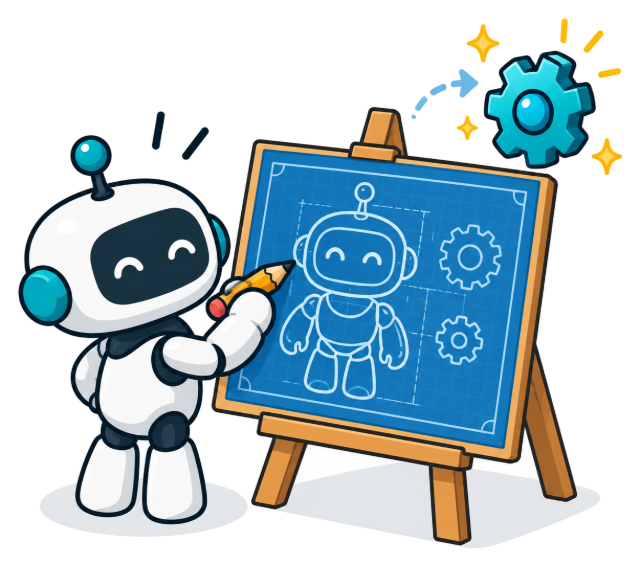}{Self-edit:\\frozen\\knowledge}{Control\\improver}{3}{}
\lrow{-5.71}{-7.06}{ArtRed}{art_selfrevise}{Self-edit:\\acquired\\knowledge}{Next\\improver}{4}{}
% What every comparison reports.
\path[fill=ArtRed!6,draw=ArtRed!40,line width=.55pt,rounded corners=2mm] (.09,-7.74) rectangle (16.01,-7.20);
\node[anchor=base west,inner sep=0pt,text=ArtRed!85!black,font=\sffamily\bfseries\fontsize{8.6}{10}\selectfont] at (.30,-7.54)
  {\faIcon{balance-scale}\enspace Compare physical outcomes\vphantom{Ag}};
\node[bitem] at (4.95,-7.54) {Four hardware-trained learners per seed};
\node[bitem] at (10.22,-7.54) {Improver update cost reported separately};
\end{tikzpicture}%
\caption{\textbf{A proposed test of the two links of recursive self-improvement (R1 and R2).} In R2, the updated improver revises itself with frozen or acquired knowledge, producing the control or next improver, respectively. Each then revises a learning procedure, which is tested through fresh matched learners under a common, protected outcome check; the next improver must beat both the control and the updated improver; for the updated improver, its learner from R1 serves as the baseline. The R2 control assumes that the acquired knowledge can be withheld without changing the parent improver, as with a memory snapshot; the count of four hardware-trained learners per seed assumes that one candidate per improver survives screening.}
\label{fig:recursive-test}
\end{figure}

%% file: sections/12_conclusion.tex
\sectionaccent{PAIInk}
\Needspace{16\baselineskip}
\section{Conclusion}
\label{sec:conclusion}

This survey examined how robots that improve from physical experience can move toward recursive self-improvement, in which a system improves not only its behavior but also the process by which it improves. We positioned RSI within robot learning: reinforcement learning, self-training, continual learning, meta-learning, and AutoML say how updates are computed or what they must cope with, whereas a ladder of claims states what an experiment shows to have improved. Along that ladder, we reviewed how physical experience improves policies, world models, and knowledge, the learning procedures that produce them, and the improver that revises those procedures. What limits a robot's next improvement (RQ1) is rarely visible in a terminal outcome; it is a hypothesis about missing experience, supervision, optimization, action generation, interfaces, or retention, to be tested by intervention rather than inferred from data volume or component accuracy. The changes that address it (RQ2) are kept in policies, world models, external knowledge, or the learning procedure, and they build on one another (RQ3) only under four conditions: feedback addresses remaining learning needs, the receiving component can use the change, its value survives changes in tasks or hardware, and continued practice remains safe and affordable. Physical constraints (RQ4) act through all four: incomplete sensing limits what a signal can distinguish; embodiment interfaces limit what a recipient can execute; wear, calibration drift, and hardware changes drive distribution shift; and reset, safety, and cost bound continued practice.

Existing studies, concentrated in manipulation, support behavioral improvement, later-learning benefits, and procedure improvement. Some systems also change their improver's knowledge or tools, but no physical study reviewed here has yet isolated a gain attributable to recursion, which requires a controlled comparison against a version of the improver whose acquired knowledge is frozen. These answers remain provisional, because the four conditions are overlapping diagnostic lenses rather than independently validated constructs and much of the evidence comes from recent preprints with few training runs or trials.

Digital agents already supply mechanisms for revising their own code and tools. Whether these mechanisms work with physical experience and under physical constraints depends on the open problems identified here: evaluation on tasks the revising system cannot tune against, with matched learners for claims beyond behavior; gains that keep accumulating as data stop teaching and bottlenecks move; practice that stays safe while the procedure governing it changes; and an improver that learns to find bottlenecks, choose experiments, and transfer repair knowledge. Physical recursive self-improvement will be demonstrated not when a robot changes many times, but when knowledge of why one repair worked causally improves how the next repair is made, and when that improved capability, applied to new evidence, helps improve itself again without losing the ability to keep learning. Even a robot that becomes measurably better at rewriting its own recipe for a single family of failures would give the first such evidence.

%% file: sections/appendix_landscape.tex
\sectionaccent{PAIInk}
\section{Extended Landscape of Physical Self-Improvement}
\label{app:landscape}

This appendix extends the coverage of the mechanisms reviewed in the main text. Appendix~\ref{app:reported-results} records selected numerical outcomes and their denominators, and Appendix~\ref{app:glossary} collects the terms used throughout.

Table~\ref{tab:human-system-roles} shows, stage by stage, what people and systems contribute to the reviewed loops. Table~\ref{tab:domain-scope} identifies the support for cross-domain interpretations. Manipulation supplies detailed component and downstream comparisons; navigation, locomotion, flight, and laboratories contribute particular reuse, adaptation, and experiment-selection mechanisms. Table~\ref{tab:broader-landscape} then extends the system coverage with policy post-training, autonomous resets, fleets, dynamics adaptation, reusable memory, continual learning, and contact-aware feedback. It includes both system-generated and expert-provided experience and distinguishes training from evaluation settings.
\input{tables/human_system_roles}
\FloatBarrier

\input{tables/domain_scope}
\FloatBarrier

\input{tables/broader_landscape}
\FloatBarrier

%% file: tables/human_system_roles.tex
\begin{table}[!htbp]
\centering
\caption{Human and system contributions at each stage of an improvement loop. Contributions can coexist.}
\label{tab:human-system-roles}
\begingroup\small
\setlength{\tabcolsep}{5pt}\renewcommand{\arraystretch}{1.15}\arrayrulecolor{PAIRule}
\begin{tabularx}{\linewidth}{@{}L{2.9cm}YY@{}}
\paitoprule
\pahead{Stage} & \pahead{Human contribution} & \pahead{System contribution}\\
\paimidrule
Experience & Demonstrations; resets & Rollouts; task or state selection\\\paihair
Supervision & Corrections; outcome labels & Predicted reward; progress estimates\\\paihair
Update construction & Training design; code edits & Parameter fitting; generated code edits\\\paihair
Selection & Checkpoint or candidate choice & Filtering; candidate ranking\\\paihair
Validation & Outcome audit; safety review & Instrumented test; contract check\\\paihair
Deployment & Approval; physical setup & Execution of an approved release or rollback\\
\paibottomrule
\end{tabularx}\endgroup
\end{table}

%% file: tables/domain_scope.tex
\begin{table}[!htbp]
\centering
\caption{Representative comparisons by robot domain. The marked conditions identify the focus of each comparison, not an exhaustive capability profile.}
\label{tab:domain-scope}
\begingroup\small
\setlength{\tabcolsep}{5pt}\renewcommand{\arraystretch}{1.18}
\begin{tabularx}{\linewidth}{@{}L{2.3cm}L{4.6cm}*{4}{C{0.62cm}}Y@{}}
\paitoprule
\pahead{Domain} & \pahead{Representative comparison} & \pahead{IF} & \pahead{U} & \pahead{DS} & \pahead{CP} & \pahead{Scope of support}\\
\paimidrule
Manipulation & FIERCE: refreshed vs frozen evaluator; DexTouch-WM: all-real vs half-generated data~\citep{arxiv260918651,arxiv260920649} & \condyes & \condyes & \condyes &  & Component and downstream tests on selected hardware tasks\\
\paihair
Navigation & AllDayNav: tests across rooms, episodes, and tasks~\citep{arxiv260610927} &  & \condyes & \condyes &  & Cross-room/episode/task tests of memory and policy\\
\paihair
Locomotion & Intelligent Trial and Error: behavior search after damage~\citep{cully2015adaptation} &  & \condyes & \condyes &  & Real damaged-robot adaptation; a precomputed repertoire\\
\paihair
Flight & Agile quadrotor flight: residual-model learning and in-flight policy updates~\citep{arxiv260210111} & \condyes &  &  & \condyes & Short-horizon adaptation on one quadrotor platform\\
\paihair
Robotic labs & Mobile chemist: measured outcomes guide the next experiment~\citep{burger2020chemist} & \condyes &  &  & \condyes & Experiment-selection precedent; a~supplied search algorithm\\
\paibottomrule
\end{tabularx}\endgroup
\end{table}

%% file: tables/broader_landscape.tex
\begingroup
\footnotesize
\setlength{\tabcolsep}{3pt}
\renewcommand{\arraystretch}{1.22}
\arrayrulecolor{PAIRule}
\begin{longtable}{@{}L{3.35cm}L{3.15cm}L{3.55cm}C{1.4cm}L{4.1cm}@{}}
\caption{Additional systems and learning settings. Icons identify reported experiments or transfer, not necessarily training.}\label{tab:broader-landscape}\\
\paitoprule
\pahead{Work} & \pahead{Experience} & \pahead{Component} & \pahead{Setting} & \pahead{Focus of evaluation} \\
\paimidrule
\endfirsthead
\multicolumn{5}{l}{\small\textit{Table~\thetable\ continued}}\\
\paitoprule
\pahead{Work} & \pahead{Experience} & \pahead{Component} & \pahead{Setting} & \pahead{Focus of evaluation} \\
\paimidrule
\endhead
\paimidrule\multicolumn{5}{r}{\footnotesize Continued on next page}\\\endfoot
\paibottomrule
\endlastfoot
\addlinespace[3pt]\multicolumn{5}{@{}l@{}}{\tablegroup{Policy post-training and correction}} \\*\paihair
\tableworkref{PA-RL}{arxiv241206685} & Offline data; online trials & Critic; policy & \envsim\,\envhw & Critic-guided policy fitting \\\paihair
\tableworkref{FLaRe}{arxiv240916578} & Simulated practice & Pretrained policy & \envsim\,\envhw & Long-horizon tasks; physical transfer \\\paihair
\tableworkref{VLA-RL}{arxiv250518719} & Rewarded rollouts & VLA policy & \envsim & Process-guided policy optimization \\\paihair
\tableworkref{SimpleVLA-RL}{arxiv250909674} & Outcome rewards & VLA policy & \envsim\,\envhw & Group-relative training; transfer \\\paihair
\tableworkref{ResiP}{arxiv240716677} & Demonstrations; trials & Residual controller & \envsim\,\envhw & Precise assembly after imitation \\\paihair
\tableworkref{DSRL}{arxiv250615799} & Online trials & Noise policy & \envsim\,\envhw & Fixed-generator improvement \\\paihair
\tableworkref{Real-Time EXPO-FT}{arxiv260918207} & Delayed-action trials & Fast editor; base policy & \envsim\,\envhw & Dynamic tasks under inference delay \\

\addlinespace[3pt]\multicolumn{5}{@{}l@{}}{\tablegroup{Fleet and deployment learning}} \\*\paihair
\tableworkref{Learning While Deploying}{arxiv260500416} & Fleet rollouts; interventions & Generalist policy & \envhw & Repeated updates across eight tasks \\\paihair
\tableworkref{OpenBot-Fleet}{arxiv240507515} & Household navigation & Cloud policy & \envsim\,\envhw & Varied homes; user-operated collection \\\paihair
\tableworkref{Runtime monitoring~+ imitation}{arxiv231017552} & Traces; corrections & Dynamics; classifier; policy & \envsim\,\envhw & Supervision requests near failure \\\paihair
\tableworkref{Leave No Trace}{arxiv171106782} & Task and reset trials & Forward/reset policies; values & \envsim & Autonomous resets; recoverability \\\paihair
\tableworkref{HIL-UMI}{arxiv260920659} & Handheld human actions & Advantage model; VLA & \envhw & Robot-free collection; hardware test \\

\addlinespace[3pt]\multicolumn{5}{@{}l@{}}{\tablegroup{Online dynamics adaptation}} \\*\paihair
\tableworkref{GrBAL}{arxiv180311347} & Recent transitions & Adaptive dynamics & \envsim\,\envhw & Adaptation to changed conditions \\\paihair
\tableworkref{Variational Neural Dynamics}{arxiv260627353} & Changing dynamics & Dynamics; policy & \envsim\,\envhw & Reuse under recurring conditions \\\paihair
\tableworkref{Self-adapting Robotic Agents}{arxiv260304029} & Residual-triggered trials & World model; policy & \envsim\,\envhw & Recovery from distribution shifts \\

\addlinespace[3pt]\multicolumn{5}{@{}l@{}}{\tablegroup{Memory and reusable recovery}} \\*\paihair
\tableworkref{LRLL}{arxiv240618746} & Tasks; teacher guidance & Code examples; skills & \envsim\,\envhw & Library growth; selected transfer \\\paihair
\tableworkref{Mem-World}{arxiv260618960} & Modeled rollouts & World model; memory; policy & \envhw & Scene persistence for manipulation \\\paihair
\tableworkref{RoboMemory}{arxiv250801415} & Interactive traces & Memory stores & \envsim\,\envhw & Accumulation on repeated tasks \\

\addlinespace[3pt]\multicolumn{5}{@{}l@{}}{\tablegroup{Continual learning from expert task streams}} \\*\paihair
\tableworkref{AtomicVLA}{arxiv260307648} & Demonstrations & Experts; router & \envsim\,\envhw & Old-skill retention during expansion \\\paihair
\tableworkref{CLARE}{arxiv260109512} & Expert datasets & Adapters; router & \envsim\,\envhw & Retention without stored exemplars \\\paihair
\tableworkref{LifelongVLA}{arxiv260714852} & Tasks; replay & Adapters; gate & \envsim\,\envhw & Five-task real-robot sequence \\\paihair
\tableworkref{iManip}{arxiv250307087} & Demonstrations & Prompts; replay & \envsim\,\envhw & Retention under incremental tasks \\\paihair
\tableworkref{MuSe}{arxiv260630988} & Force data; replay & Multisensory policy & \envhw & New contact tasks; old-task retention \\

\addlinespace[3pt]\multicolumn{5}{@{}l@{}}{\tablegroup{Feedback and contact-aware learning}} \\*\paihair
\tableworkref{PEBBLE}{arxiv210605091} & Preference queries & Reward; policy & \envsim & Updated rewards; replay relabeling \\\paihair
\tableworkref{ReWiND}{arxiv250510911} & Demos; online trials & Policy; prior reward & \envsim\,\envhw & New tasks without target demos \\\paihair
\tableworkref{RoboReward}{arxiv260100675} & Scored robot videos & Reward model & \envsim\,\envhw & Prediction and downstream RL \\\paihair
\tableworkref{TORL-VLA}{arxiv260609337} & Tactile trials; intervention & Refinement actor; critics (VLA frozen) & \envhw & Contact adaptation; rescue credit \\
\end{longtable}
\endgroup

%% file: sections/appendix_reported_results.tex
\section{Reported Outcomes and Experimental Context}
\label{app:reported-results}

Table~\ref{tab:reported-outcomes} collects numerical details that are secondary to the mechanism comparisons in the main text. Percentages retain each paper's denominator and metric where available. Approximate values are identified as such. The settings, supervision, and acquisition budgets differ across rows, so the table is not a performance ranking.

\begingroup\footnotesize
\setlength{\tabcolsep}{4pt}\renewcommand{\arraystretch}{1.20}\arrayrulecolor{PAIRule}
\begin{longtable}{@{}L{2.85cm}L{5.55cm}L{7.43cm}@{}}
\caption{Selected reported outcomes, baselines, and evaluation settings. Icons mark where each reported number was measured.}\label{tab:reported-outcomes}\\
\paitoprule
\pahead{System} & \pahead{Reported comparison} & \pahead{Setting and interpretation}\\
\paimidrule\endfirsthead
\multicolumn{3}{l}{\small\textit{Table~\thetable\ continued}}\\
\paitoprule
\pahead{System} & \pahead{Reported comparison} & \pahead{Setting and interpretation}\\
\paimidrule\endhead
\paimidrule\multicolumn{3}{r}{\footnotesize Continued on next page}\\\endfoot
\paibottomrule\endlastfoot

\systemcite{SOAR}{arxiv240720635} & \envhw\ Mean success 28\% $\to$ 57\%; released dataset about 30.5K trajectories & Ten evaluation scenes across five tabletop settings. The updated arXiv text (its Section~4 and Figure~5; the dataset size also appears in its abstract and Appendix~E) reports these values. The PMLR landing-page abstract instead states 25K trajectories.\\
\paihair
\systemcite{SI-EFM}{arxiv250915155} & \envhw\ LanguageTable approximately 62--63\% $\to$ 87--88\%; BananaTable approximately 63\% $\to$ 85\% & Approximate plotted rates in the real-robot experiments. BananaTable practice uses about eight hours on two robot stations. The learned feedback checkpoint stays fixed during policy improvement.\\
\paihair
\systemcite{HIL-SERL}{arxiv241021845} & \envhw\ RAM insertion: 100\% versus 29\% for HG-DAgger; cycle time 4.8 versus 8.3 seconds & 100 trials per condition. Matched imitation comparison in the main results table. Intervention, reward classification, and compliant control jointly define the learning system.\\
\paihair
\systemcite{RECAP}{arxiv251114759} & \envhw\ More than $2\times$ throughput on diverse laundry and espresso; roughly half the failure rate & Relative to supervised specialization of the same pretrained policy.\\
\paihair
\systemcite{RAPolicy}{arxiv260922888} & \envhw\ Joint-task success 52\% $\to$ 88\% & One joint five-task setting, alongside four single-task settings. Online training uses one to two hours per setting and includes human interventions.\\
\paihair
\systemcite{SARM2/\allowbreak SPIRAL}{arxiv260610305} & \envhw\ Crumpled-shorts folding: 4/12 $\to$ 8/12 with stage-aware reward versus 4/12 $\to$ 2/12 with sparse reward & Their Table~2; real bimanual arms. Both loops start from the same dense-reward checkpoint and run three rounds; rollout episodes differ, as expected for on-policy loops.\\
\paihair
\systemcite{GR-RL}{arxiv251201801} & \envhw\ Shoe lacing: 45.7\% $\to$ 83.3\%, against behavior cloning on all demonstrations & Bimanual mobile robot. An offline critic used as a progress estimator filters demonstrations, mirror augmentation is added, and a latent-noise predictor is then trained online.\\
\paihair
\systemcite{VLA-Precision}{arxiv260904355} & \envhw\ 98.3\% mean success over 180 trials on nine chemistry tasks, against 67.8\% for a $\pi_{0.5}$ baseline & The critic ensemble is calibrated from real experience; the baseline is trained with more demonstrations.\\
\paihair
\systemcite{VERITAS}{arxiv260618247} & \envsim\ SIMPLER Stack Blocks: 31.3\% $\to$ 59.2\% after fine-tuning & 656 verified successful rollouts across four simulated tasks are used for fine-tuning. Separate real-robot comparisons at matched data sizes test verifier-curated data against human teleoperation.\\
\paihair
\systemcite{SILVR}{arxiv250606658} & \envsim\ Meta-World unseen tasks: 26.8\% with the inverse-dynamics translator frozen versus 44.2\% when it is also fine-tuned & Their Table~A11: twelve unseen tasks, three seeds, fourth iteration; the video planner is fine-tuned in both conditions. The hardware setup keeps the translator fixed.\\
\paihair
\systemcite{RISE}{arxiv260211075} & \envhw\ Brick sorting, backpack packing, box closing: 85\%, 85\%, 95\% versus 50\%, 40\%, 60\% & Twenty hardware trials per task. The baseline is the authors' matched $\pi_{0.5}$ RECAP reimplementation.\\
\paihair
\systemcite{PlayWorld}{arxiv260309030} & \envhw\ Fold the Towel: 10\% $\to$ 75\% at the best checkpoint & Twenty hardware trials per checkpoint; two tasks overall. The 65-point absolute gain is followed by regression with further model-based training. Thirty hours of autonomous play precede the model/policy stages.\\
\paihair
\systemcite{World-Gymnast}{arxiv260202454} & \envhw\ Hardware: above SIMPLER-trained RL on three of four WidowX tasks (e.g., 72\% vs 32\%) & The VLA is trained in a learned video model with GPT-4o rewards; SIMPLER remains better on closing a drawer.\\
\paihair
\systemcite{DexTouch-WM}{arxiv260920649} & \envhw\ FTP-1 mean score: 0.73 all-real versus 0.69 with half generated by the human--robot model & Their Table~IV; four hardware tasks; means rounded to two decimals. Compare 200 real trajectories with 100 real plus 100 generated. Partial-credit scores. Data from the human--robot model lower all three policy-family means; data from the robot-only-adapted model give mixed results.\\
\paihair
\systemcite{PLD}{arxiv251100091} & \envhw\ Cube pickup: 30/30 for PLD data, 16/30 for RLPD data, 10/30 for human demonstrations & Two real-task distillation tests. Peg insertion is 30/30 for all three data conditions. PLD and RLPD each supply 200 successful episodes and all three datasets fine-tune $\pi_0$; the human-data size is not stated, and source collector acquisition differs.\\
\paihair
\systemcite{Robo-Cortex}{arxiv260518729} & \envsim\ SPL 24.03 $\to$ 39.33 with transferred heuristics; difference 15.30 & Heuristic-transfer ablation on Habitat's unseen image-goal split (their Table~3), without online updates, against Basic Pipeline. SPL is success weighted by path length. Hardware evidence is separate and preliminary: two indoor scenes with ten navigation goals.\\
\paihair
\systemcite{VASO}{arxiv260605395} & Formal compliance 97.2\% with handcrafted alignment, 96.8\% with automatic alignment; fewer than 100 optimization samples & Aggregate main table over generated plans; Jackal and PX4 executions are shown separately. The percentages are formal-specification scores, not measured probabilities of physical safety; the baseline task-completion comparison is evaluated in simulation.\\
\paihair
\systemcite{SHAPER}{arxiv260811350} & \envsim\ VLABench success 34.50\% versus 28.25\% for the seed agent & Four held-out simulation splits. Skill-only and harness-only controls test interacting components; the headline comparison changes both.\\
\paihair
\systemcite{Zetta}{arxiv260816590} & \envsim\ LIBERO-PRO macro-average 32.00\% $\to$ 71.13\%; RoboCasa 73.56\% $\to$ 93.56\% & Forty LIBERO-PRO task-setting pairs and eighteen RoboCasa tasks. RoboCasa spans four repair rounds. The abstract's 90.8\% LIBERO-PRO headline uses a different aggregation.\\
\paihair
\systemcite{RoboHarn-Evo}{arxiv260937583} & \envhw\ Physical feedback resolves more than 83\% of erroneous memory entries while retaining 95.8\% of valid ones & Entries are task- and action-level knowledge with stated applicability; zero-shot transfer of simulation-acquired knowledge is reported in Section~\ref{sec:memory}.\\
\paihair
\systemcite{Code transfer}{arxiv260919906} & \envsim\ Nine-task mean: 45.2\% without references, 41.5\% with initial code, 57.0\% with optimized code & Their Table~III: three independent generation runs, ten tests per task/run, 270 tests per condition in total. Same coding model and limits; privileged simulated observations and realized compute remain relevant.\\
\paihair
\systemcite{RATs}{arxiv260619419} & \envsim\ Held-out LIBERO-PRO success +20.6 percentage points over the CaP-Agent0 baseline & A code-skill library distilled from successful play.\\
\paihair
\systemcite{RoboSkill}{arxiv260937810} & \envhw\ Hard tasks: 80.0\% $\to$ 88.3\% & Real-robot evaluation of a reusable skill library.\\
\paihair
\systemcite{GLIDE}{arxiv260924996} & \envhw\ Teleoperated collection successes: 0, 1, and 0 of 10 without guardrails versus 7, 9, and 9 with refined guardrails & Their Table~1; three real tasks, ten trials each. Unrefined guardrails already reach 6 and 7 on two tasks.\\
\paihair
\systemcite{SEES}{arxiv260932698} & \envsim\ GR00T N1.5: 47.2\% $\to$ 55.7\% on RoboCasa365 atomic tasks and 10.2\% $\to$ 13.1\% on seen composite tasks & Three rounds. States before the most frequently stalling sub-skill are restored in simulation, and that skill family is trained with PPO against an LLM-written success check.\\
\paihair
\systemcite{REVOLVE}{arxiv260914633} & \envhw\ Judgment accuracy 72.0\% $\to$ 80.5\%; policy success 40.0\% $\to$ 58.5\% & Four real tasks across the reported history. Memory and policy co-evolve, supported by a task-specific reset policy and a classical correction controller configured before the loop; the two trends do not isolate a causal memory contribution, which the component test of Section~\ref{sec:component-tests} would.\\
\end{longtable}\endgroup

Read alongside the mechanism tables, these results distinguish isolated component effects from pipeline gains, hardware outcomes from simulated or formal scores, and acquisition from maintenance costs. Section~\ref{sec:cost-ledger} discusses the corresponding comparison and reporting requirements.

%% file: sections/appendix_evaluation.tex
\section{Glossary}
\label{app:glossary}

This glossary summarizes the terminology used in the review; section references point to the relevant discussion.

\begingroup\small
\setlength{\tabcolsep}{4pt}\renewcommand{\arraystretch}{1.18}\arrayrulecolor{PAIRule}
\begin{longtable}{@{}L{3.5cm}L{\dimexpr\linewidth-4.8cm-4\tabcolsep\relax}L{1.3cm}@{}}
\caption{Terms used in this review.}\label{tab:glossary}\\
\paitoprule
\pahead{Term} & \pahead{Meaning} & \pahead{Section}\\
\paimidrule\endfirsthead
\multicolumn{3}{l}{\small\textit{Table~\thetable\ continued}}\\
\paitoprule
\pahead{Term} & \pahead{Meaning} & \pahead{Section}\\
\paimidrule\endhead
\paimidrule\multicolumn{3}{r}{\footnotesize Continued on next page}\\\endfoot
\paibottomrule\endlastfoot
Action chunk & A short sequence of future actions that a policy outputs at once. & \ref{sec:primer}\\\paihair
Admission gate & A rule that turns protected evidence into a decision on whether a candidate update may run. & \ref{sec:admission}\\\paihair
Attempt, retry, round & An attempt is one try at a task; a retry is a further attempt on the same task instance within one episode; a learning round ends with an update from which later episodes start. & \ref{sec:selection}\\\paihair
Behavioral improvement & The revised system executes better under a common task test. & \ref{sec:key-terms}\\\paihair
Bottleneck & The component hypothesized to limit the next improvement, given the available repairs. & \ref{sec:agenda-diagnosis}\\\paihair
Continued practice & Whether another informative trial remains safe and affordable. & \ref{sec:four-uses}\\\paihair
Cost accounting & An account that separates setup and signal construction, acquisition of the revision, downstream training and evaluation, deployment, and maintenance. & \ref{sec:cost-ledger}\\\paihair
Evaluator & A component that supplies outcome judgments for learning: a person, a hand-written check, or a learned reward model such as a success detector. & \ref{sec:primer}\\\paihair
Improver & Whatever generates or selects revisions of the learning procedure: a person, a fixed rule, a search process, or a learned component. & \ref{sec:key-terms}\\\paihair
Informative feedback & Whether experience distinguishes the remaining learning need. & \ref{sec:four-uses}\\\paihair
Later-learning benefit & An acquired artifact (data, weights, a model, a skill, or code) helps a subsequent learner. & \ref{sec:key-terms}\\\paihair
Learner & The component that acquires task capabilities. & \ref{sec:key-terms}\\\paihair
Learning procedure & The decisions that turn experience into updates: what to collect, how to judge it, and how to construct the update. & \ref{sec:key-terms}\\\paihair
Level of support (R1, R2) & Mechanism only (machinery through which the improver could change), uncontrolled result (a comparison consistent with a link but without a control that withholds only the acquired knowledge), or controlled result (a frozen-knowledge or fixed-improver control with matched learners and budgets). & \ref{sec:key-terms}\\\paihair
Matched learner & A learner that differs from its comparison only in the component under test and shares its starting conditions, resources, and evaluation protocol, so that two inputs or procedures can be compared by what they teach. & \ref{sec:key-terms}\\\paihair
Payback period & The period within which the trials a reusable tool saves, minus its maintenance, must exceed what it cost to build. & \ref{sec:investment}\\\paihair
Procedure improvement & A revised collector, evaluator, editor, or updater teaches a matched learner better. & \ref{sec:key-terms}\\\paihair
Protected success evaluation & A success evaluation that a candidate update cannot rewrite and that remains informative about the relevant physical failure. & \ref{sec:evaluator-integrity}\\\paihair
Record of assumptions & A record, for each candidate change, of what changed, which safeguard depends on it, what was rechecked, and which hazards remain. & \ref{sec:risks}\\\paihair
Recursive self-improvement & Repair knowledge improves the improver (R1), and the improver, once improved, helps produce its own next improvement (R2). & \ref{sec:key-terms}\\\paihair
Repair & An update made in response to an observed failure or limitation. & \ref{sec:key-terms}\\\paihair
Repair knowledge & Knowledge retained from earlier repairs: which failure distinction motivated a change, under what conditions it applied, and what it improved downstream. & \ref{sec:key-terms}\\\paihair
Retained change & What experience leaves behind for later use: a policy, a model, a memory, a skill, software, or a change to the learning procedure. & \ref{sec:four-uses}\\\paihair
Robustness to distribution shift & Whether a change remains useful after the policy, model, task, or hardware changes, including changes that the improvement itself caused. & \ref{sec:four-uses}\\\paihair
Success evaluation & The person, sensor, or model that decides whether an attempt succeeded. & \ref{sec:physical-consequences}\\\paihair
Support (of a model) & The range of actions and contacts for which a model has comparable training examples. & \ref{sec:outgrowing}\\\paihair
Updater & A procedure that maps the learner's state, experience, and objective to an operation such as a parameter step, a loss function, a skill edit, or a code patch. & \ref{sec:construct-update}\\\paihair
Usability & Whether the component that must consume a change can represent, optimize, and execute it. & \ref{sec:four-uses}\\\paihair
Uses of retained change & Behavior; prediction and practice; reuse and recovery; and the learning procedure. Each names how a change is used next. & \ref{sec:four-uses}\\\paihair
Validity conditions & A statement attached to a retained change: its required inputs, action meanings, physical envelope, software dependencies, and the evidence under which it last worked. & \ref{sec:maintenance}\\
\end{longtable}\endgroup

%% file: sections/appendix_gallery.tex
% Appendix D: gallery of images released with reviewed systems.
% Layout: figures are placed by hand (two per page, minipage + \captionof) so the three gallery pages are full;
% every row is justified to the full text width (16.08 cm, 0.24 cm gaps); images are clipped, never padded.
\section{Gallery}
\label{app:gallery}

The figures in this appendix collect images released with several of the reviewed systems. They show the settings behind the results discussed in Section~\ref{sec:mechanisms} and Appendix~\ref{app:reported-results}: robots practising on hardware, fleets that pool experience, simulators and world models paired with real setups, agents that write the code that drives a robot, laboratories that run experiments on their own, and a wider range of robot bodies and settings, from legged and aerial robots to arms working in homes and outdoors. Each caption cites the work and names the source of the image.

\par\vfill
\noindent\begin{minipage}{\linewidth}
\centering
\begin{tikzpicture}[x=1cm,y=1cm,
  galbadge/.style={circle,fill=#1!88!black,draw=white,line width=.5pt,text=white,inner sep=0pt,minimum size=.4cm,font=\sffamily\bfseries\fontsize{7.6}{7.6}\selectfont},
  galcap/.style={anchor=north west,inner xsep=0pt,inner ysep=0pt,align=left,font=\fontsize{8.2}{9.6}\selectfont,text=ArtSlate}]
\path[use as bounding box] (0,0) rectangle (16.08,7.47);
\begin{scope}\clip[rounded corners=1.6mm] (0,4.413) rectangle ++(5.129,3.057);
  \node[inner sep=0pt] at (0.101,4.83) {\includegraphics[width=10.057cm]{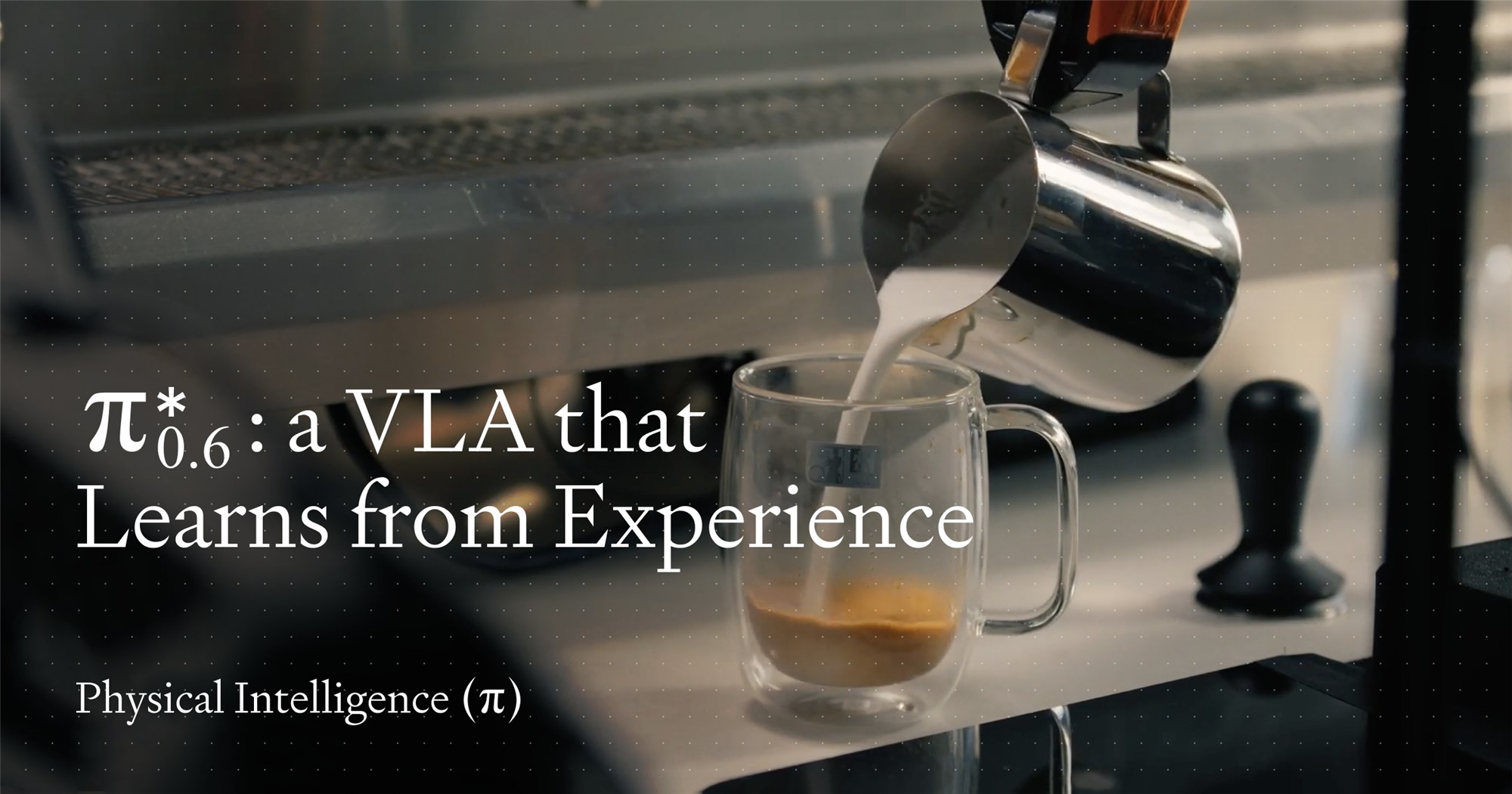}};
\end{scope}
\draw[ArtSlate!22,line width=.4pt,rounded corners=1.6mm] (0,4.413) rectangle ++(5.129,3.057);
\node[galbadge=ArtAmber] at (0.27,7.2) {a};
\node[galcap,text width=5.129cm] at (0,4.333) {\textbf{\boldmath$\pi^{*}_{0.6}$}\enspace pouring milk for a latte};
\begin{scope}\clip[rounded corners=1.6mm] (5.369,4.413) rectangle ++(4.351,3.057);
  \node[inner sep=0pt] at (8.704,6.185) {\includegraphics[width=6.668cm]{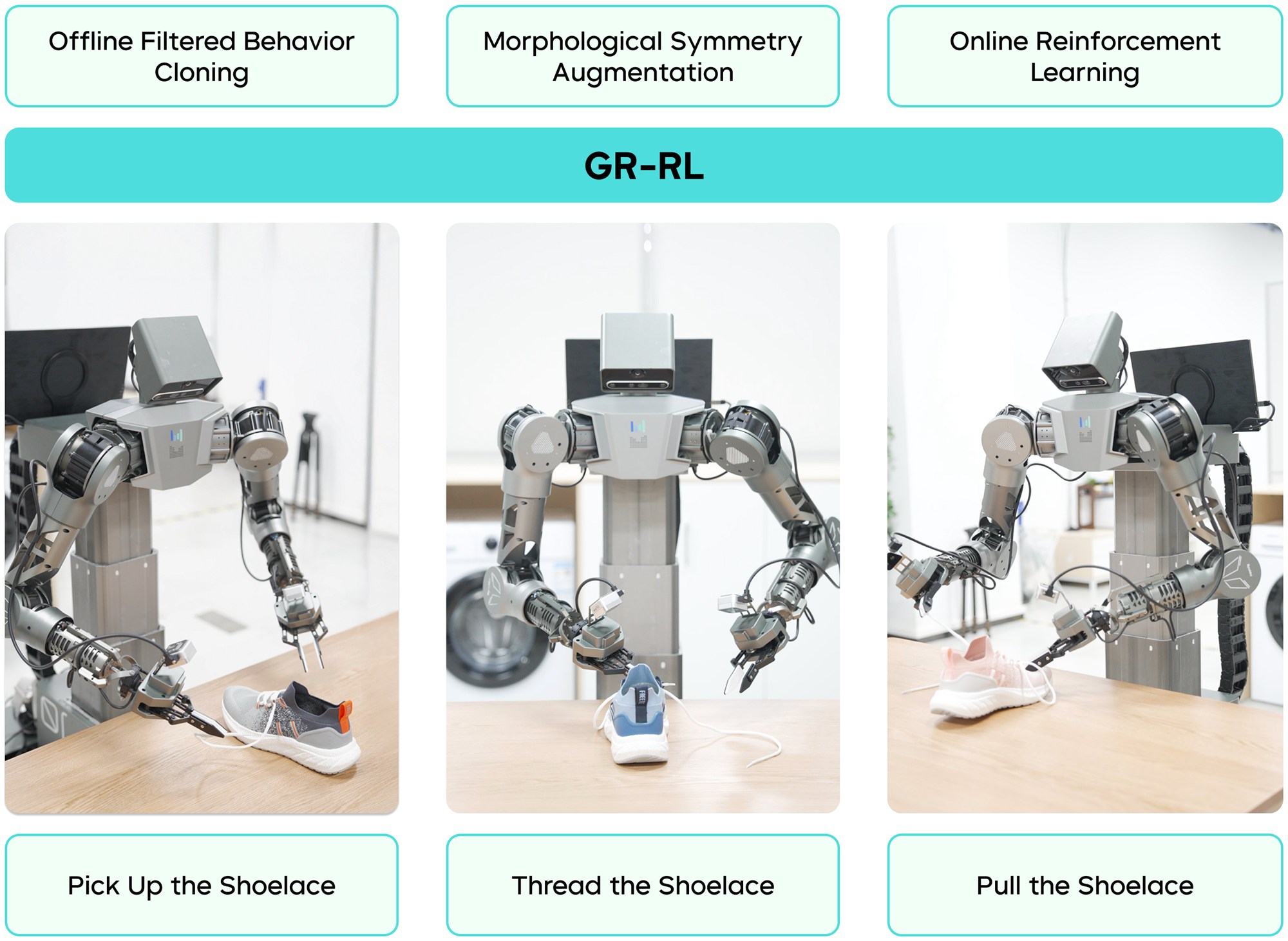}};
\end{scope}
\draw[ArtSlate!22,line width=.4pt,rounded corners=1.6mm] (5.369,4.413) rectangle ++(4.351,3.057);
\node[galbadge=ArtAmber] at (5.639,7.2) {b};
\node[galcap,text width=4.351cm] at (5.369,4.333) {\textbf{GR-RL}\enspace lacing a shoe};
\begin{scope}\clip[rounded corners=1.6mm] (9.96,4.413) rectangle ++(6.12,3.057);
  \node[inner sep=0pt] at (13.02,5.941) {\includegraphics[width=6.12cm]{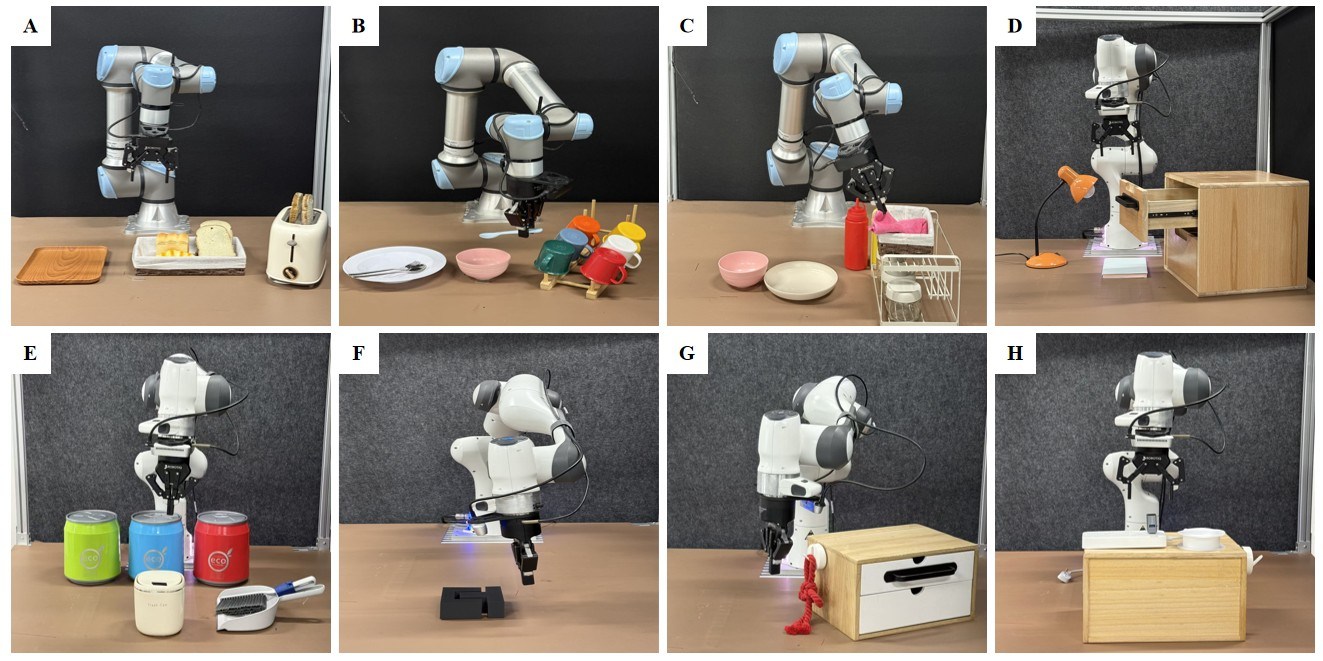}};
\end{scope}
\draw[ArtSlate!22,line width=.4pt,rounded corners=1.6mm] (9.96,4.413) rectangle ++(6.12,3.057);
\node[galbadge=ArtAmber] at (10.23,7.2) {c};
\node[galcap,text width=6.12cm] at (9.96,4.333) {\textbf{SiLRI}\enspace eight real-robot tasks};
\begin{scope}\clip[rounded corners=1.6mm] (0,0.538) rectangle ++(4.49,3.217);
  \node[inner sep=0pt] at (2.245,2.146) {\includegraphics[width=4.49cm]{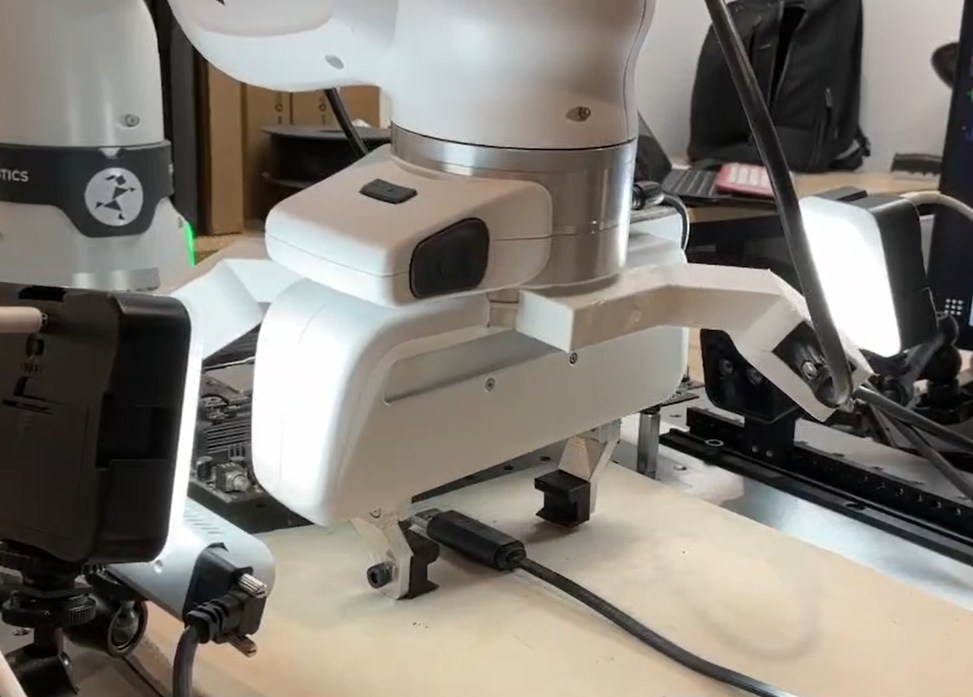}};
\end{scope}
\draw[ArtSlate!22,line width=.4pt,rounded corners=1.6mm] (0,0.538) rectangle ++(4.49,3.217);
\node[galbadge=ArtAmber] at (0.27,3.485) {d};
\node[galcap,text width=4.49cm] at (0,0.458) {\textbf{Max-Q}\enspace USB insertion};
\begin{scope}\clip[rounded corners=1.6mm] (4.73,0.538) rectangle ++(5.718,3.217);
  \node[inner sep=0pt] at (7.589,2.146) {\includegraphics[width=5.718cm]{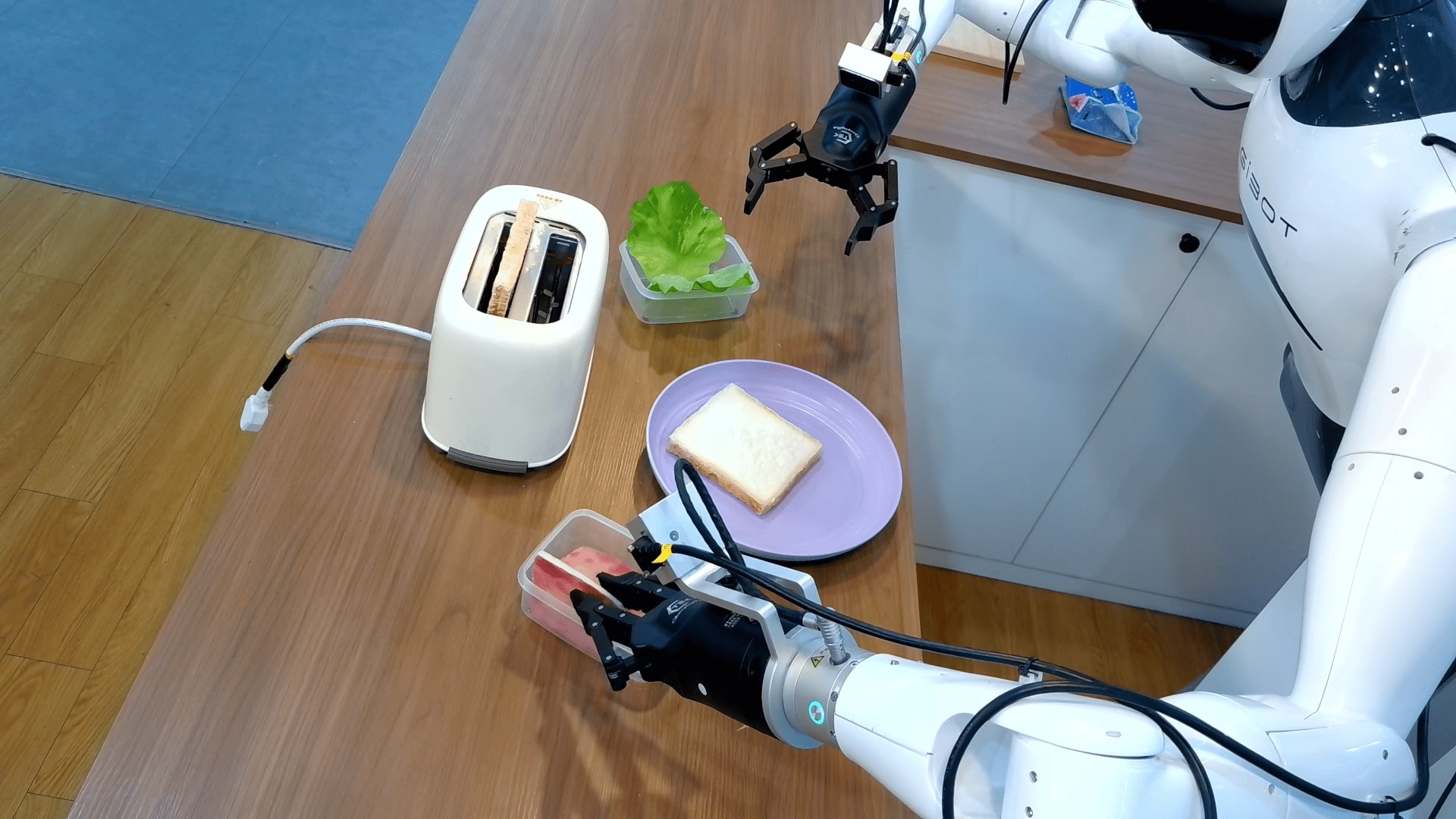}};
\end{scope}
\draw[ArtSlate!22,line width=.4pt,rounded corners=1.6mm] (4.73,0.538) rectangle ++(5.718,3.217);
\node[galbadge=ArtAmber] at (5,3.485) {e};
\node[galcap,text width=5.718cm] at (4.73,0.458) {\textbf{Genie Centurion}\enspace sandwich assembly};
\begin{scope}\clip[rounded corners=1.6mm] (10.689,0.538) rectangle ++(5.391,3.217);
  \node[inner sep=0pt] at (13.4,2.763) {\includegraphics[width=7.93cm]{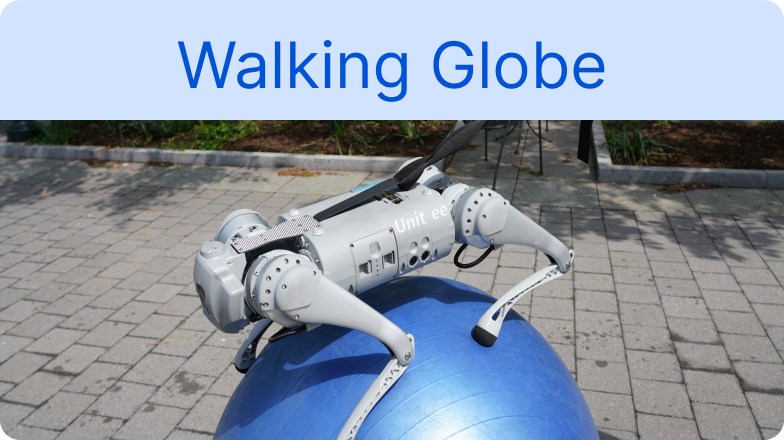}};
\end{scope}
\draw[ArtSlate!22,line width=.4pt,rounded corners=1.6mm] (10.689,0.538) rectangle ++(5.391,3.217);
\node[galbadge=ArtAmber] at (10.959,3.485) {f};
\node[galcap,text width=5.391cm] at (10.689,0.458) {\textbf{DrEureka}\enspace walking on a yoga ball};
\end{tikzpicture}
\captionof{figure}{\textbf{Gallery I: robots practising on real hardware.} (a) $\pi^{*}_{0.6}$~\citep{arxiv251114759}; (b) GR-RL~\citep{arxiv251201801}; (c) SiLRI~\citep{arxiv251224288}; (d) Max-Q~\citep{arxiv260815088}; (e) Genie Centurion~\citep{arxiv250518793}; (f) DrEureka~\citep{ma2024dreureka}. Image sources: (a) Physical Intelligence website; (b) ByteDance Seed website; (c, d, e) the papers, CC BY 4.0; (f) the DrEureka project page.}
\label{fig:gallery-hardware}
\end{minipage}
\par\vfill
\noindent\begin{minipage}{\linewidth}
\centering
\begin{tikzpicture}[x=1cm,y=1cm,
  galbadge/.style={circle,fill=#1!88!black,draw=white,line width=.5pt,text=white,inner sep=0pt,minimum size=.4cm,font=\sffamily\bfseries\fontsize{7.6}{7.6}\selectfont},
  galcap/.style={anchor=north west,inner xsep=0pt,inner ysep=0pt,align=left,font=\fontsize{8.2}{9.6}\selectfont,text=ArtSlate}]
\path[use as bounding box] (0,0) rectangle (16.08,7.858);
\begin{scope}\clip[rounded corners=1.6mm] (0,4.935) rectangle ++(5.205,2.924);
  \node[inner sep=0pt] at (2.603,6.396) {\includegraphics[width=5.205cm]{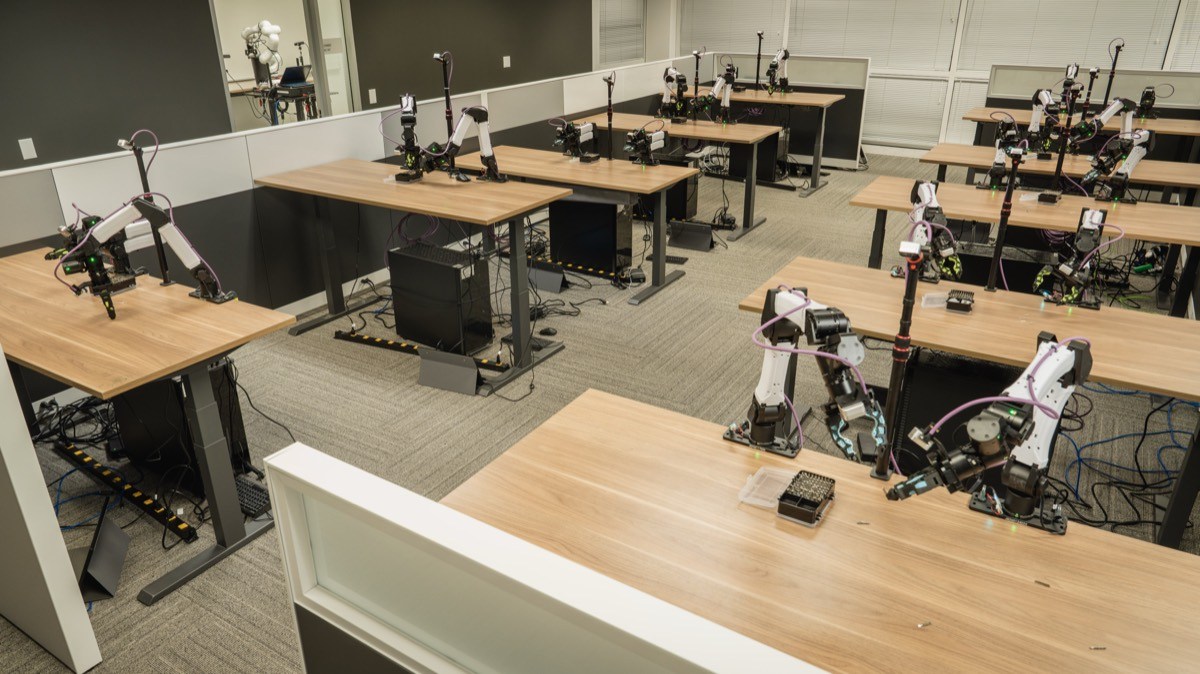}};
\end{scope}
\draw[ArtSlate!22,line width=.4pt,rounded corners=1.6mm] (0,4.935) rectangle ++(5.205,2.924);
\node[galbadge=ArtTeal] at (0.27,7.588) {a};
\node[galcap,text width=5.205cm] at (0,4.855) {\textbf{ENPIRE}\enspace a robot farm for agent-run experiments};
\begin{scope}\clip[rounded corners=1.6mm] (5.445,4.935) rectangle ++(5.197,2.924);
  \node[inner sep=0pt] at (8.044,6.396) {\includegraphics[width=5.197cm]{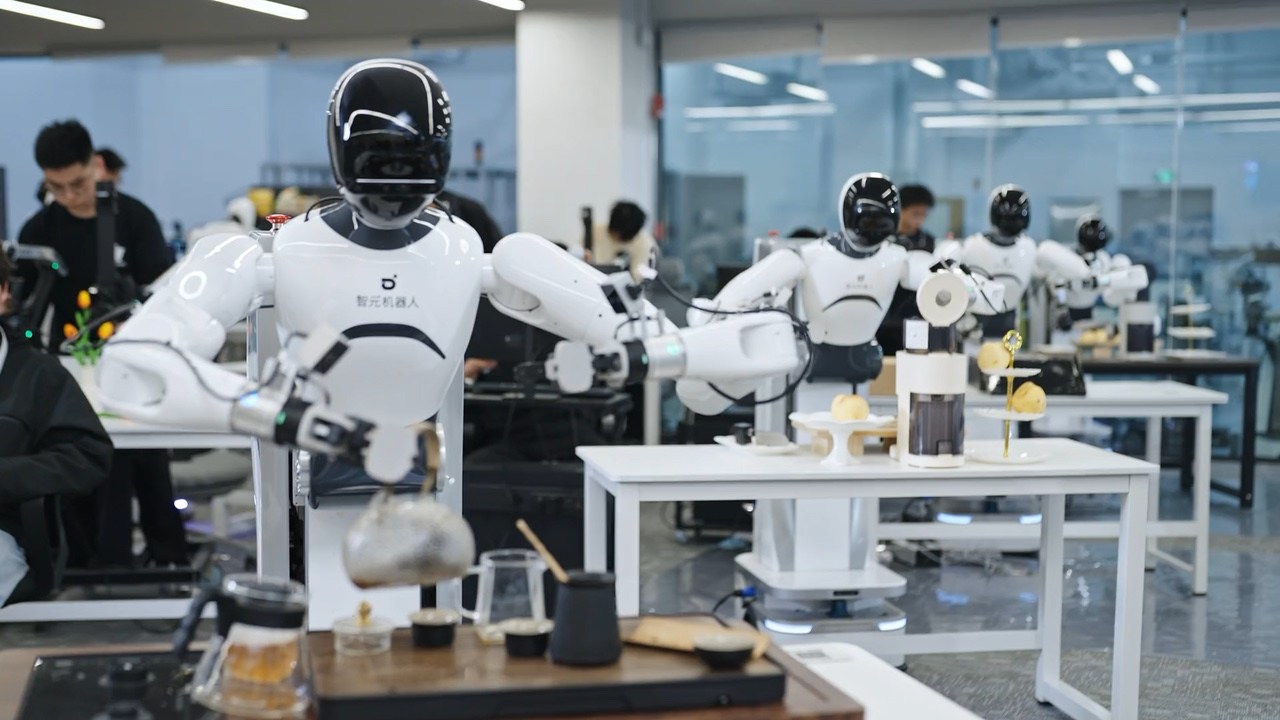}};
\end{scope}
\draw[ArtSlate!22,line width=.4pt,rounded corners=1.6mm] (5.445,4.935) rectangle ++(5.197,2.924);
\node[galbadge=ArtTeal] at (5.715,7.588) {b};
\node[galcap,text width=5.197cm] at (5.445,4.855) {\textbf{Learning While Deploying}\enspace a robot fleet};
\begin{scope}\clip[rounded corners=1.6mm] (10.883,4.935) rectangle ++(5.197,2.924);
  \node[inner sep=0pt] at (13.481,6.396) {\includegraphics[width=5.197cm]{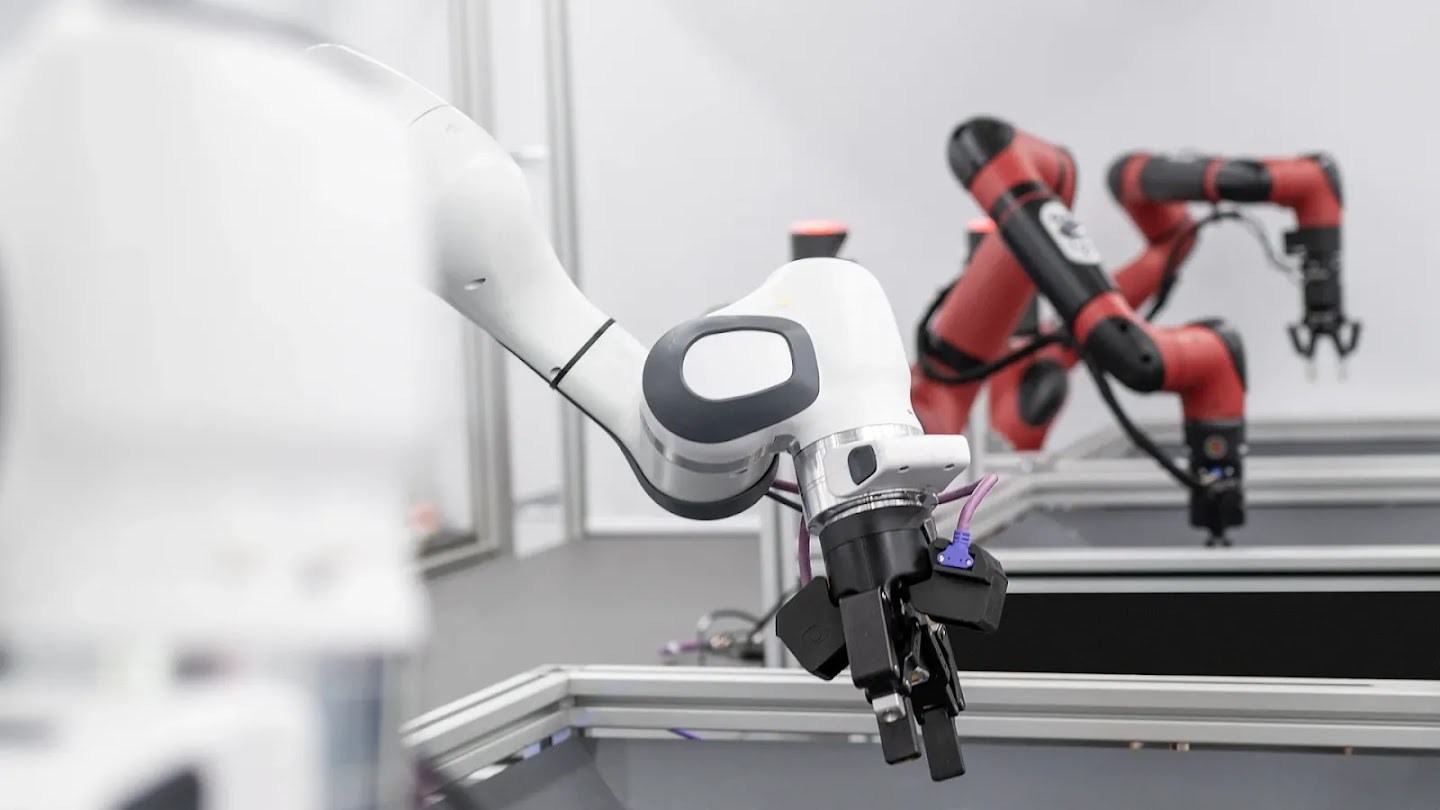}};
\end{scope}
\draw[ArtSlate!22,line width=.4pt,rounded corners=1.6mm] (10.883,4.935) rectangle ++(5.197,2.924);
\node[galbadge=ArtTeal] at (11.153,7.588) {c};
\node[galcap,text width=5.197cm] at (10.883,4.855) {\textbf{RoboCat}\enspace arms of different embodiments};
\begin{scope}\clip[rounded corners=1.6mm] (0,0.876) rectangle ++(4.537,3.063);
  \node[inner sep=0pt] at (2.269,2.407) {\includegraphics[width=4.537cm]{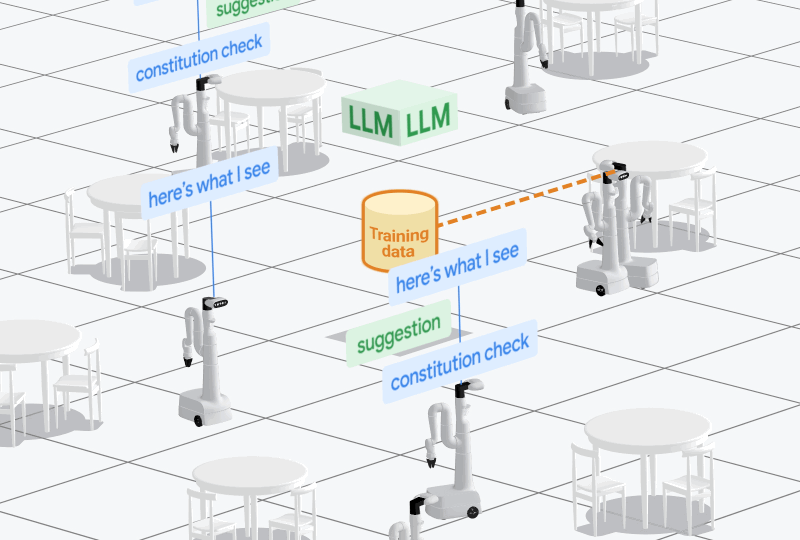}};
\end{scope}
\draw[ArtSlate!22,line width=.4pt,rounded corners=1.6mm] (0,0.876) rectangle ++(4.537,3.063);
\node[galbadge=ArtTeal] at (0.27,3.669) {d};
\node[galcap,text width=4.537cm] at (0,0.796) {\textbf{AutoRT}\enspace orchestrating mobile manipulators};
\begin{scope}\clip[rounded corners=1.6mm] (4.777,0.876) rectangle ++(4.869,3.063);
  \node[inner sep=0pt] at (7.212,2.407) {\includegraphics[width=9.738cm]{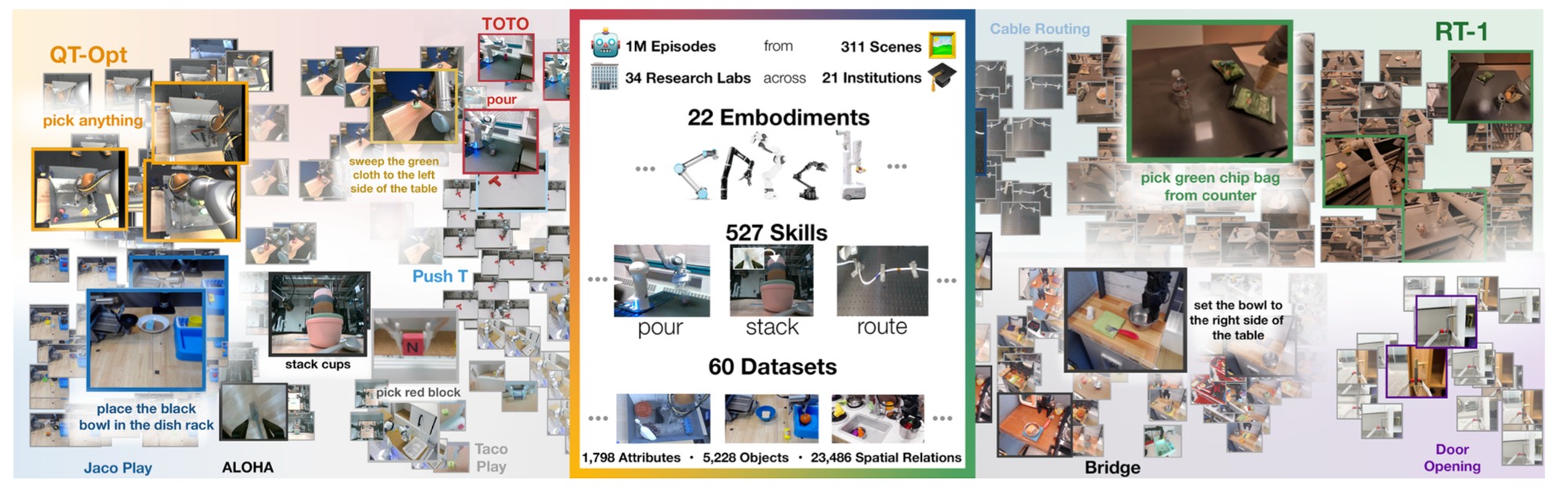}};
\end{scope}
\draw[ArtSlate!22,line width=.4pt,rounded corners=1.6mm] (4.777,0.876) rectangle ++(4.869,3.063);
\node[galbadge=ArtTeal] at (5.047,3.669) {e};
\node[galcap,text width=4.869cm] at (4.777,0.796) {\textbf{Open X-Embodiment}\enspace data from 22 embodiments};
\begin{scope}\clip[rounded corners=1.6mm] (9.886,0.876) rectangle ++(6.194,3.063);
  \node[inner sep=0pt] at (12.983,2.407) {\includegraphics[width=6.194cm]{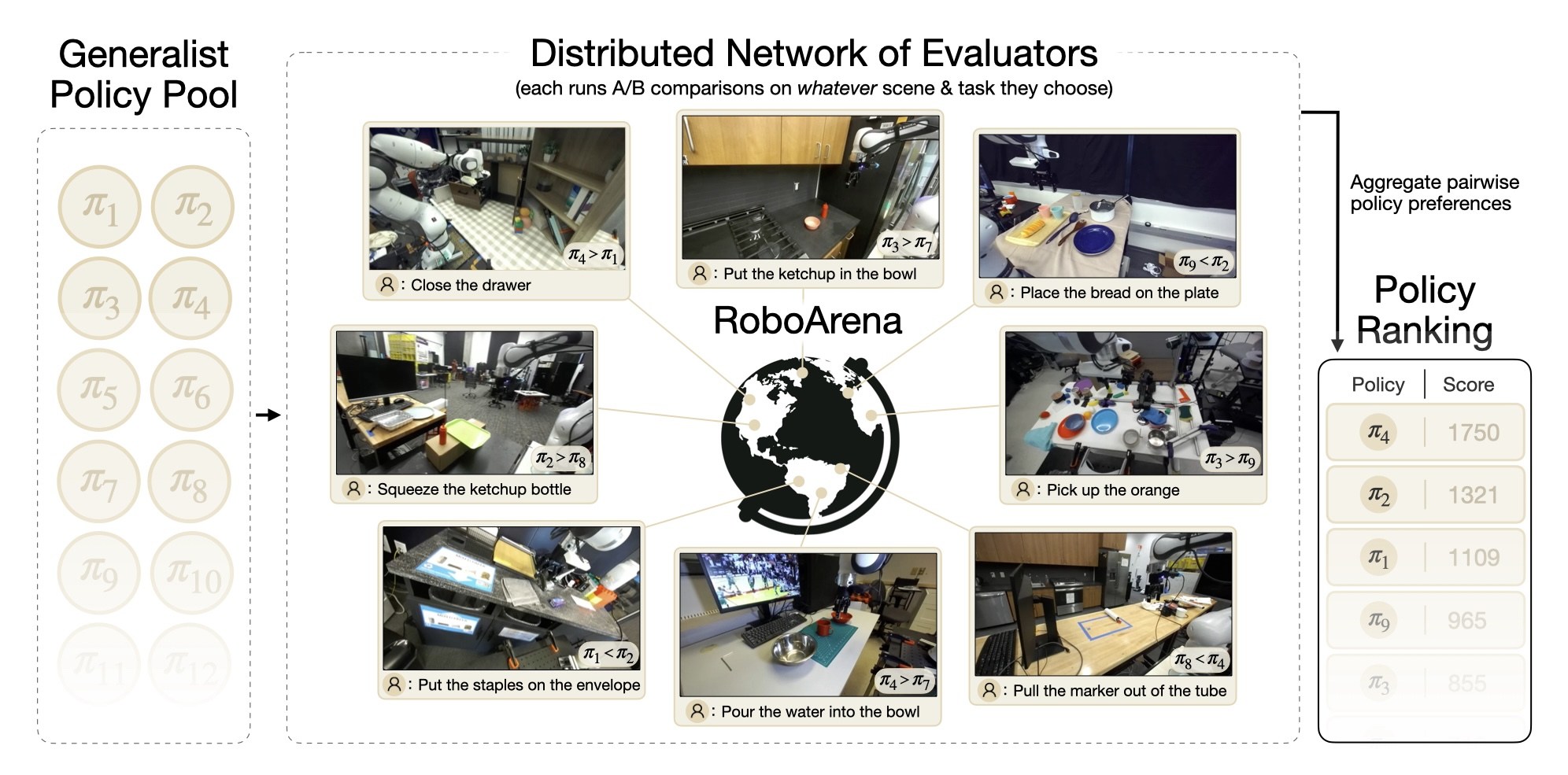}};
\end{scope}
\draw[ArtSlate!22,line width=.4pt,rounded corners=1.6mm] (9.886,0.876) rectangle ++(6.194,3.063);
\node[galbadge=ArtTeal] at (10.156,3.669) {f};
\node[galcap,text width=6.194cm] at (9.886,0.796) {\textbf{RoboArena}\enspace distributed real-world evaluation};
\end{tikzpicture}
\captionof{figure}{\textbf{Gallery II: fleets and scale.} (a) ENPIRE~\citep{arxiv260619980}; (b) Learning While Deploying~\citep{arxiv260500416}; (c) RoboCat~\citep{arxiv230611706}; (d) AutoRT~\citep{ahn2024autort}; (e) Open X-Embodiment~\citep{arxiv231008864}; (f) RoboArena~\citep{arxiv250618123}. Image sources: (a, b, d, e, f) the project pages, with (d) the first frame of an animation and (e) the central part of a collage; (c) Google DeepMind blog.}
\label{fig:gallery-fleets}
\end{minipage}
\clearpage
\noindent\begin{minipage}{\linewidth}
\centering
\begin{tikzpicture}[x=1cm,y=1cm,
  galbadge/.style={circle,fill=#1!88!black,draw=white,line width=.5pt,text=white,inner sep=0pt,minimum size=.4cm,font=\sffamily\bfseries\fontsize{7.6}{7.6}\selectfont},
  galcap/.style={anchor=north west,inner xsep=0pt,inner ysep=0pt,align=left,font=\fontsize{8.2}{9.6}\selectfont,text=ArtSlate}]
\path[use as bounding box] (0,0) rectangle (16.08,12.293);
\begin{scope}\clip[rounded corners=1.6mm] (0,8.953) rectangle ++(8.425,3.34);
  \node[inner sep=0pt] at (4.212,10.623) {\includegraphics[width=8.425cm]{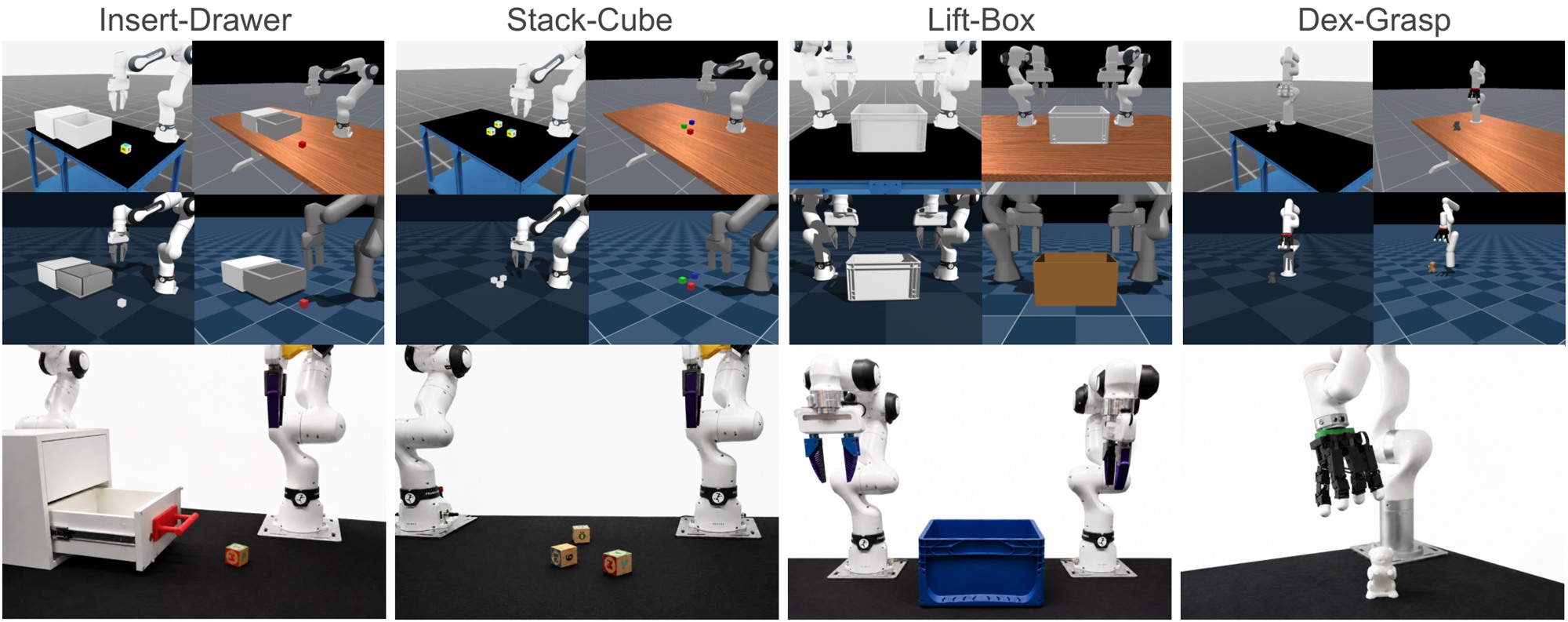}};
\end{scope}
\draw[ArtSlate!22,line width=.4pt,rounded corners=1.6mm] (0,8.953) rectangle ++(8.425,3.34);
\node[galbadge=ArtPurple] at (0.27,12.023) {a};
\node[galcap,text width=8.425cm] at (0,8.873) {\textbf{HARBOR}\enspace simulators and matching real setups};
\begin{scope}\clip[rounded corners=1.6mm] (8.665,8.953) rectangle ++(7.415,3.34);
  \node[inner sep=0pt] at (12.372,10.623) {\includegraphics[width=7.415cm]{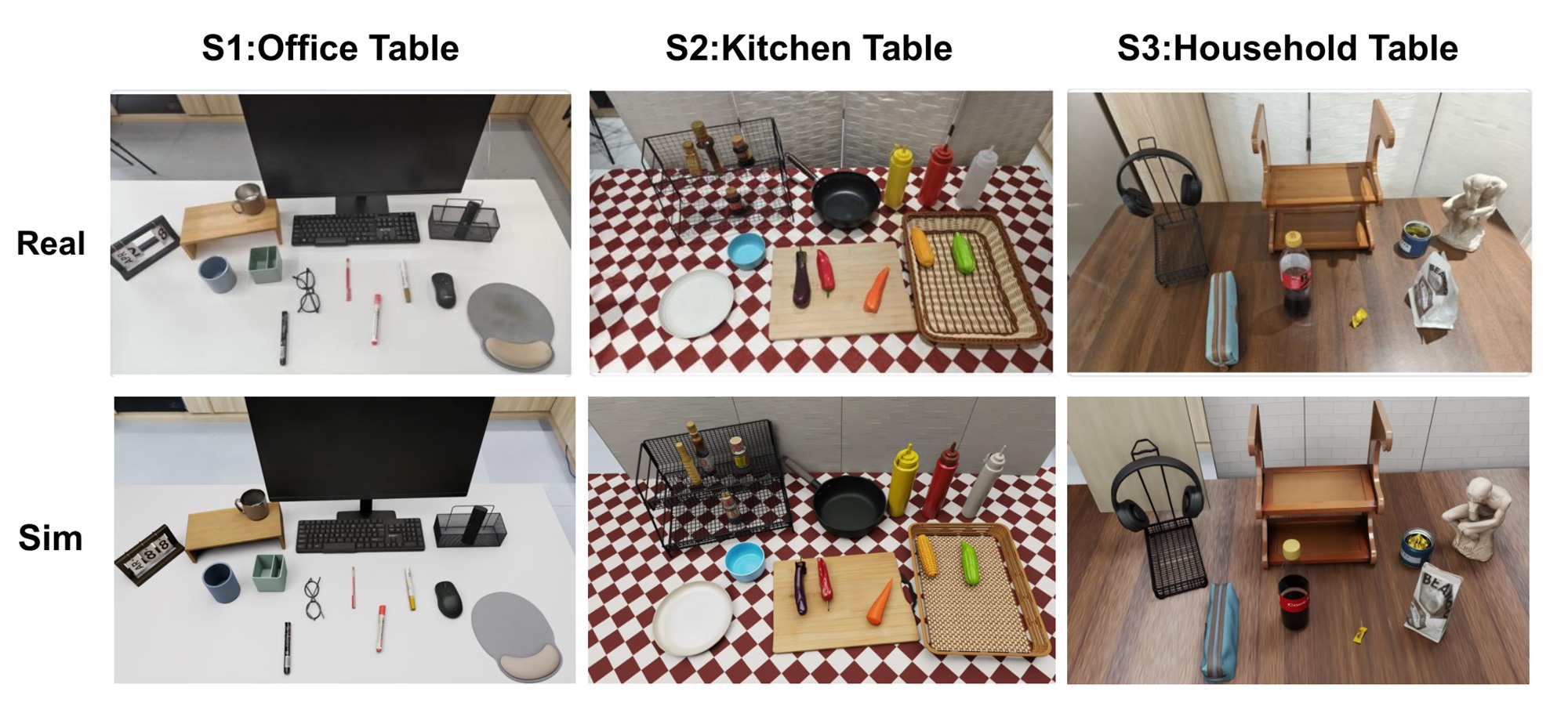}};
\end{scope}
\draw[ArtSlate!22,line width=.4pt,rounded corners=1.6mm] (8.665,8.953) rectangle ++(7.415,3.34);
\node[galbadge=ArtPurple] at (8.935,12.023) {b};
\node[galcap,text width=7.415cm] at (8.665,8.873) {\textbf{EmbodiRSI}\enspace real workspaces and simulated twins};
\begin{scope}\clip[rounded corners=1.6mm] (0,5.057) rectangle ++(5.755,3.237);
  \node[inner sep=0pt] at (2.878,6.676) {\includegraphics[width=5.755cm]{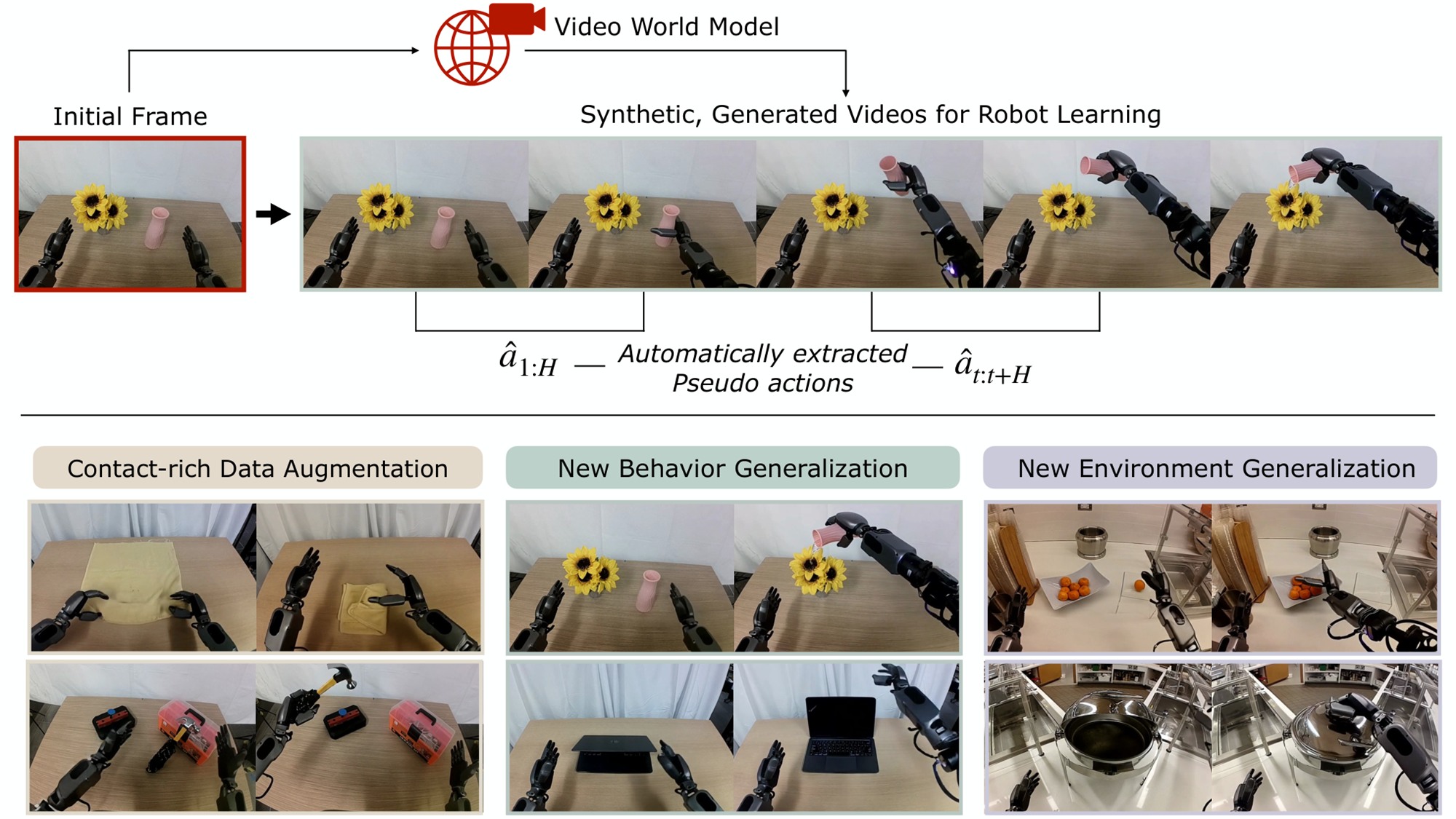}};
\end{scope}
\draw[ArtSlate!22,line width=.4pt,rounded corners=1.6mm] (0,5.057) rectangle ++(5.755,3.237);
\node[galbadge=ArtPurple] at (0.27,8.025) {c};
\node[galcap,text width=5.755cm] at (0,4.977) {\textbf{DreamGen}\enspace training data from a video world model};
\begin{scope}\clip[rounded corners=1.6mm] (5.995,5.057) rectangle ++(10.085,3.237);
  \node[inner sep=0pt] at (11.038,6.676) {\includegraphics[width=10.085cm]{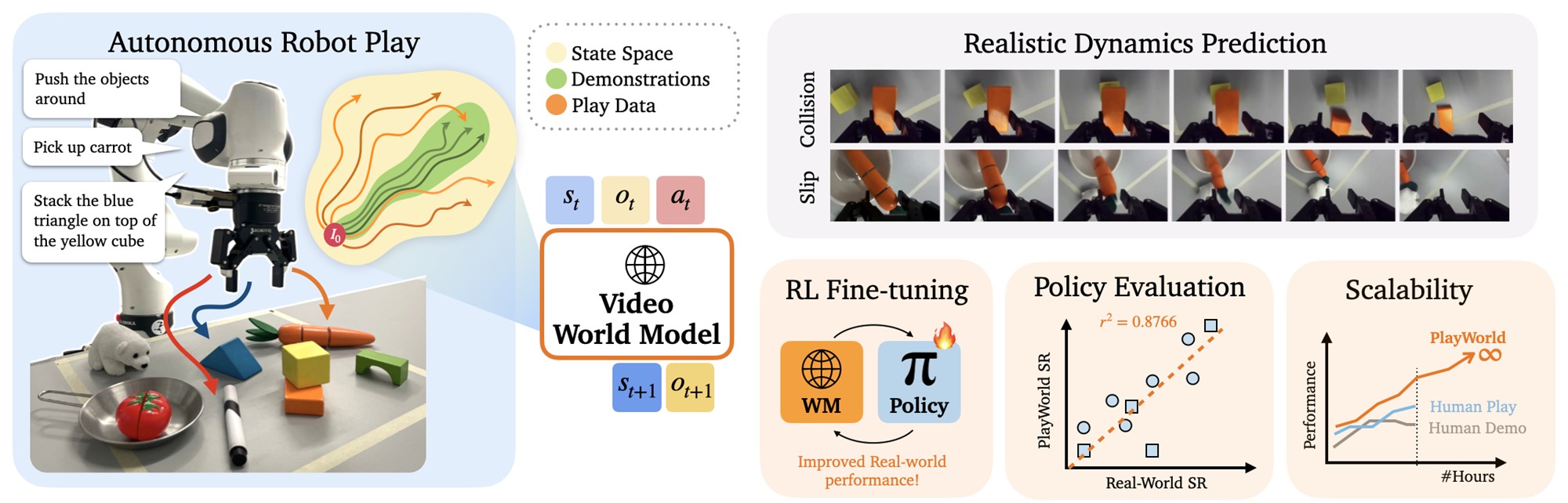}};
\end{scope}
\draw[ArtSlate!22,line width=.4pt,rounded corners=1.6mm] (5.995,5.057) rectangle ++(10.085,3.237);
\node[galbadge=ArtPurple] at (6.265,8.025) {d};
\node[galcap,text width=10.085cm] at (5.995,4.977) {\textbf{PlayWorld}\enspace autonomous play and a learned world model};
\begin{scope}\clip[rounded corners=1.6mm] (0,0.876) rectangle ++(10.177,3.185);
  \node[inner sep=0pt] at (5.089,2.469) {\includegraphics[width=10.177cm]{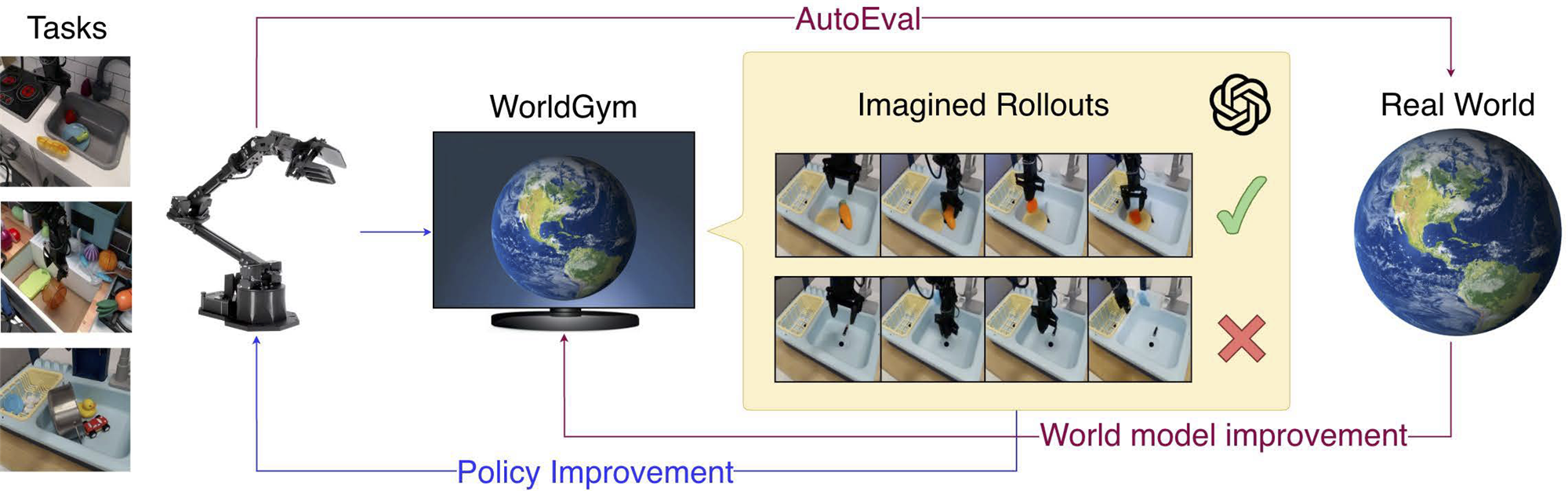}};
\end{scope}
\draw[ArtSlate!22,line width=.4pt,rounded corners=1.6mm] (0,0.876) rectangle ++(10.177,3.185);
\node[galbadge=ArtPurple] at (0.27,3.791) {e};
\node[galcap,text width=10.177cm] at (0,0.796) {\textbf{World-Gymnast}\enspace reinforcement learning inside a video world model};
\begin{scope}\clip[rounded corners=1.6mm] (10.417,0.876) rectangle ++(5.663,3.185);
  \node[inner sep=0pt] at (13.249,2.469) {\includegraphics[width=5.663cm]{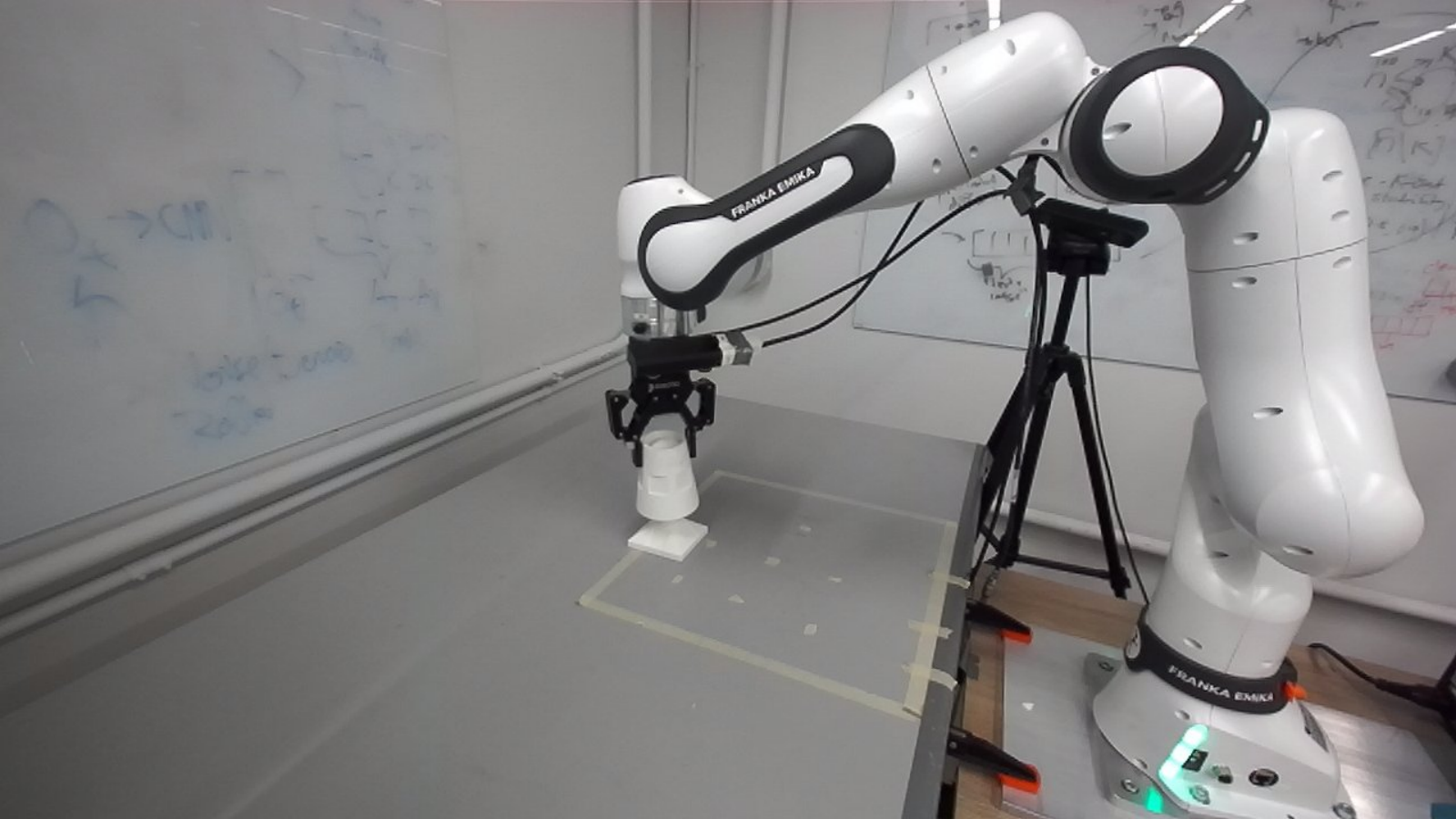}};
\end{scope}
\draw[ArtSlate!22,line width=.4pt,rounded corners=1.6mm] (10.417,0.876) rectangle ++(5.663,3.185);
\node[galbadge=ArtPurple] at (10.687,3.791) {f};
\node[galcap,text width=5.663cm] at (10.417,0.796) {\textbf{EmbodiedSWE}\enspace the real setup of a sim-to-real task};
\end{tikzpicture}
\captionof{figure}{\textbf{Gallery III: between simulation, world models, and reality.} (a) HARBOR~\citep{arxiv260608610}; (b) EmbodiRSI~\citep{arxiv260938905}; (c) DreamGen~\citep{jang2025dreamgen}; (d) PlayWorld~\citep{arxiv260309030}; (e) World-Gymnast~\citep{arxiv260202454}; (f) EmbodiedSWE~\citep{arxiv260927308}. Image sources: (a, c) the papers, CC BY 4.0; (b) the paper, CC BY-NC-SA 4.0; (d, e) the project pages; (f) Fig.~19(a) of the paper.}
\label{fig:gallery-simreal}
\end{minipage}
\par\vfill
\noindent\begin{minipage}{\linewidth}
\centering
\begin{tikzpicture}[x=1cm,y=1cm,
  galbadge/.style={circle,fill=#1!88!black,draw=white,line width=.5pt,text=white,inner sep=0pt,minimum size=.4cm,font=\sffamily\bfseries\fontsize{7.6}{7.6}\selectfont},
  galcap/.style={anchor=north west,inner xsep=0pt,inner ysep=0pt,align=left,font=\fontsize{8.2}{9.6}\selectfont,text=ArtSlate}]
\path[use as bounding box] (0,0) rectangle (16.08,8.934);
\begin{scope}\clip[rounded corners=1.6mm] (0,5.234) rectangle ++(8.835,3.7);
  \node[inner sep=0pt] at (4.417,7.084) {\includegraphics[width=8.835cm]{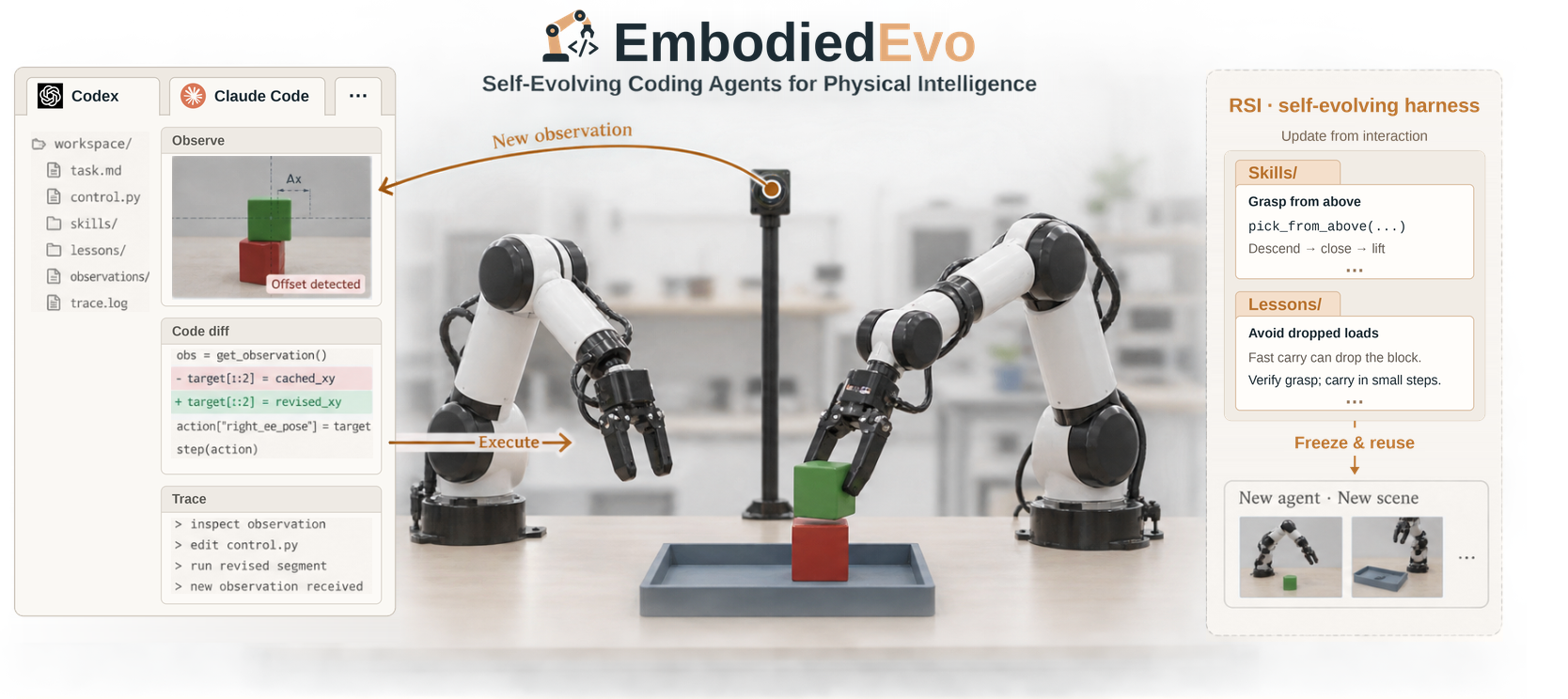}};
\end{scope}
\draw[ArtSlate!22,line width=.4pt,rounded corners=1.6mm] (0,5.234) rectangle ++(8.835,3.7);
\node[galbadge=PAISyn2] at (0.27,8.664) {a};
\node[galcap,text width=8.835cm] at (0,5.154) {\textbf{EmbodiedEvo}\enspace a self-evolving coding agent};
\begin{scope}\clip[rounded corners=1.6mm] (9.075,5.234) rectangle ++(7.005,3.7);
  \node[inner sep=0pt] at (12.577,7.084) {\includegraphics[width=7.005cm]{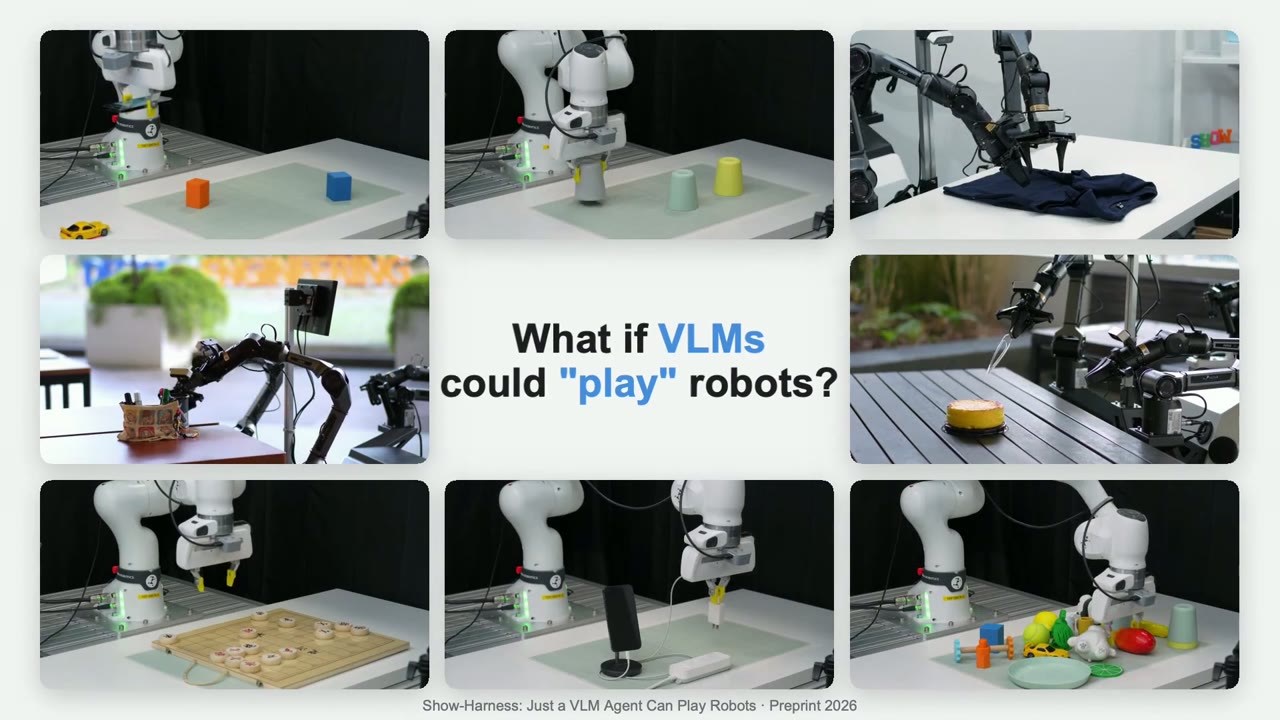}};
\end{scope}
\draw[ArtSlate!22,line width=.4pt,rounded corners=1.6mm] (9.075,5.234) rectangle ++(7.005,3.7);
\node[galbadge=PAISyn2] at (9.345,8.664) {b};
\node[galcap,text width=7.005cm] at (9.075,5.154) {\textbf{Show-Harness}\enspace VLM agents controlling robots};
\begin{scope}\clip[rounded corners=1.6mm] (0,0.876) rectangle ++(6.17,3.7);
  \node[inner sep=0pt] at (3.085,2.726) {\includegraphics[width=6.17cm]{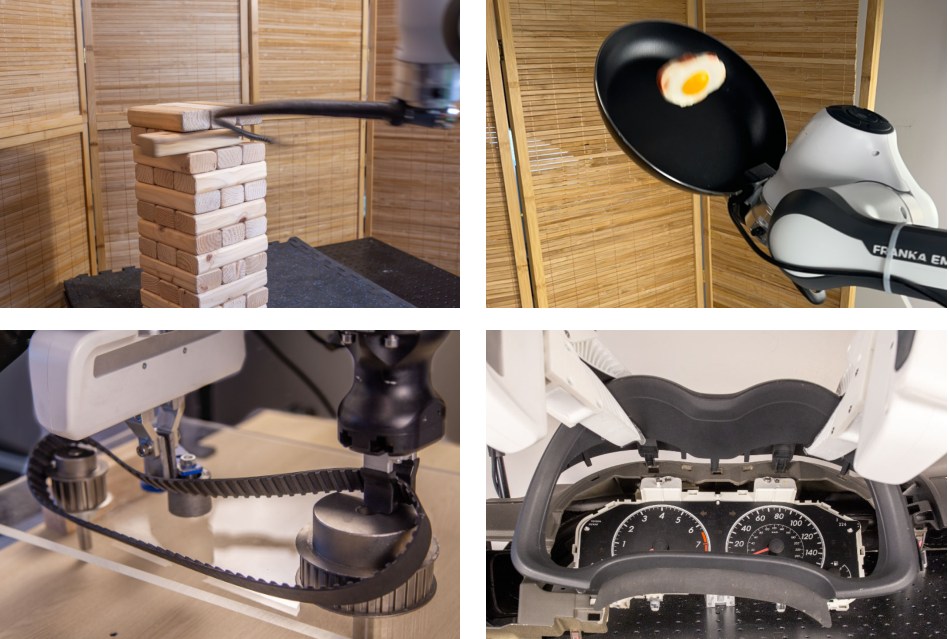}};
\end{scope}
\draw[ArtSlate!22,line width=.4pt,rounded corners=1.6mm] (0,0.876) rectangle ++(6.17,3.7);
\node[galbadge=PAISyn2] at (0.27,4.306) {c};
\node[galcap,text width=6.17cm] at (0,0.796) {\textbf{HIL-SERL}\enspace Jenga whipping, pan flipping, timing-belt and dashboard assembly};
\begin{scope}\clip[rounded corners=1.6mm] (6.41,0.876) rectangle ++(9.67,3.7);
  \node[inner sep=0pt] at (10.93,2.499) {\includegraphics[width=10.3cm]{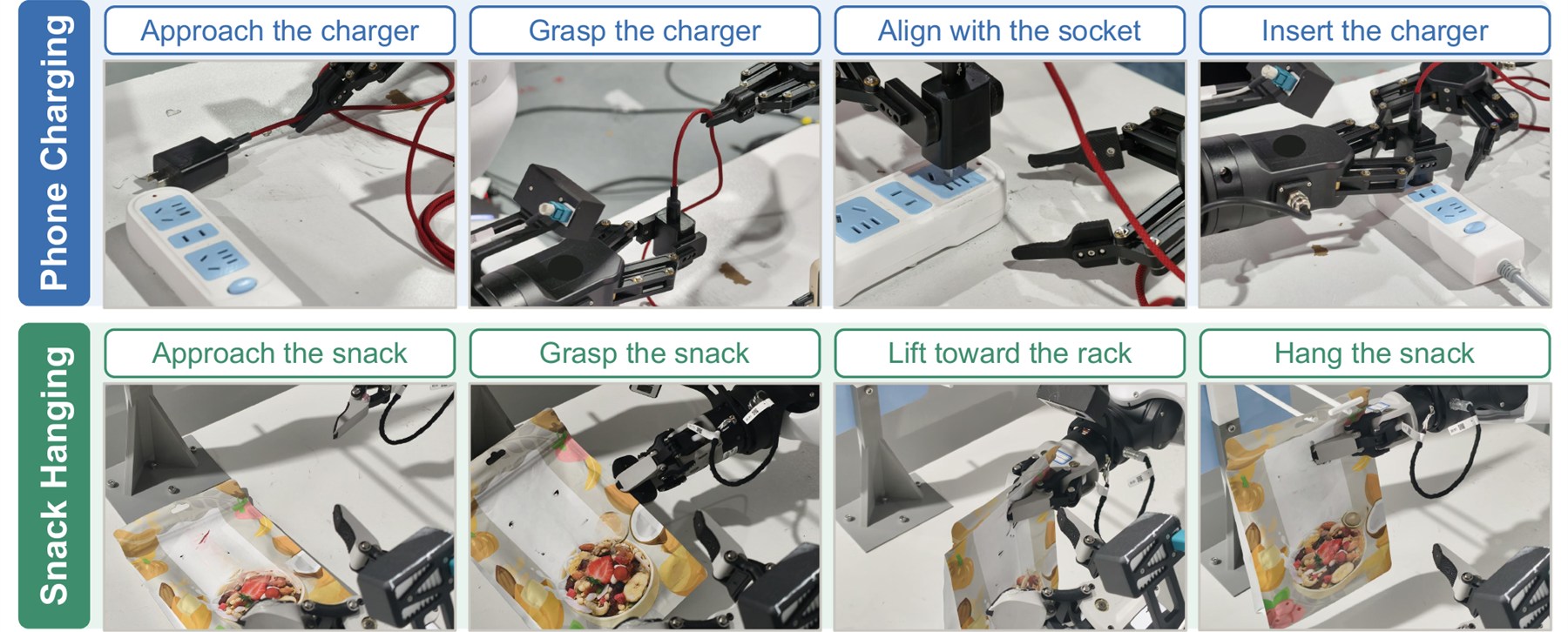}};
\end{scope}
\draw[ArtSlate!22,line width=.4pt,rounded corners=1.6mm] (6.41,0.876) rectangle ++(9.67,3.7);
\node[galbadge=PAISyn2] at (6.68,3.936) {d};
\node[galcap,text width=9.67cm] at (6.41,0.796) {\textbf{BEE}\enspace real-world phone charging and snack hanging};
\end{tikzpicture}
\captionof{figure}{\textbf{Gallery IV: agents that write robot code and learn online.} (a) EmbodiedEvo~\citep{arxiv261018867}; (b) Show-Harness~\citep{arxiv260910522}; (c) HIL-SERL~\citep{arxiv241021845}; (d) BEE~\citep{arxiv260927450}. Image sources: (a, b) the project pages; (c) top two rows of Fig.~1 and (d) the real-world rows of Fig.~2 of the papers, CC BY 4.0.}
\label{fig:gallery-agents}
\end{minipage}
\clearpage
\noindent\begin{minipage}{\linewidth}
\centering
\begin{tikzpicture}[x=1cm,y=1cm,
  galbadge/.style={circle,fill=#1!88!black,draw=white,line width=.5pt,text=white,inner sep=0pt,minimum size=.4cm,font=\sffamily\bfseries\fontsize{7.6}{7.6}\selectfont},
  galcap/.style={anchor=north west,inner xsep=0pt,inner ysep=0pt,align=left,font=\fontsize{8.2}{9.6}\selectfont,text=ArtSlate}]
\path[use as bounding box] (0,0) rectangle (16.08,4.891);
\begin{scope}\clip[rounded corners=1.6mm] (0,1.214) rectangle ++(7.048,3.677);
  \node[inner sep=0pt] at (3.524,3.052) {\includegraphics[width=7.048cm]{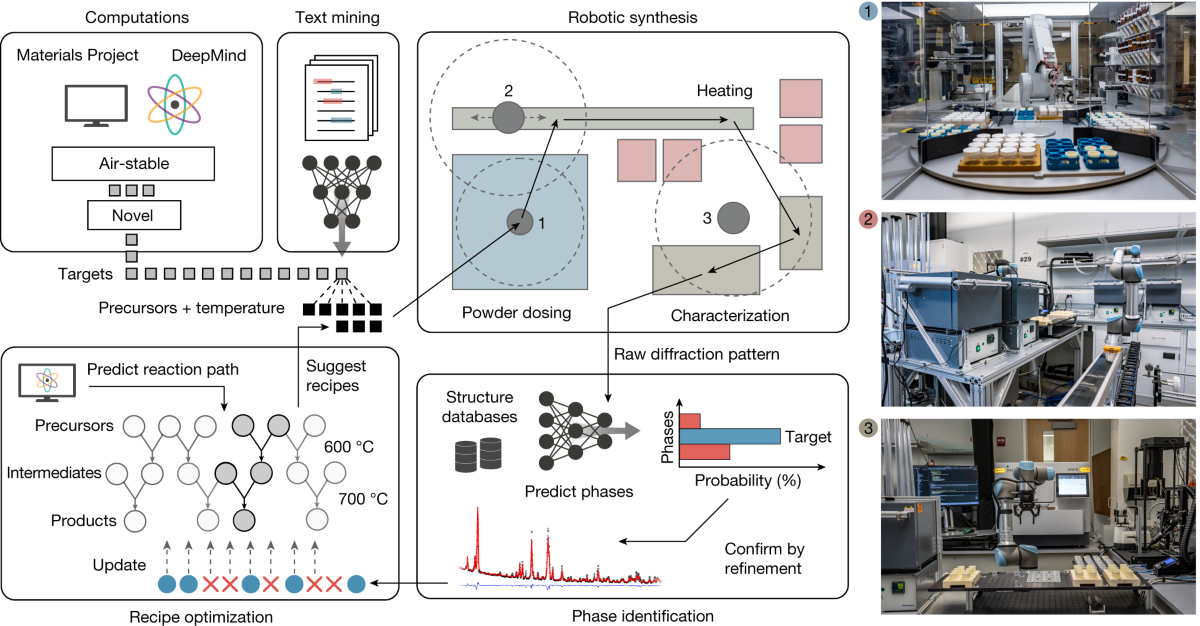}};
\end{scope}
\draw[ArtSlate!22,line width=.4pt,rounded corners=1.6mm] (0,1.214) rectangle ++(7.048,3.677);
\node[galbadge=ArtSage] at (0.27,4.621) {a};
\node[galcap,text width=7.048cm] at (0,1.134) {\textbf{A-Lab}\enspace autonomous materials synthesis};
\begin{scope}\clip[rounded corners=1.6mm] (7.288,1.214) rectangle ++(5.79,3.677);
  \node[inner sep=0pt] at (10.183,3.052) {\includegraphics[width=5.79cm]{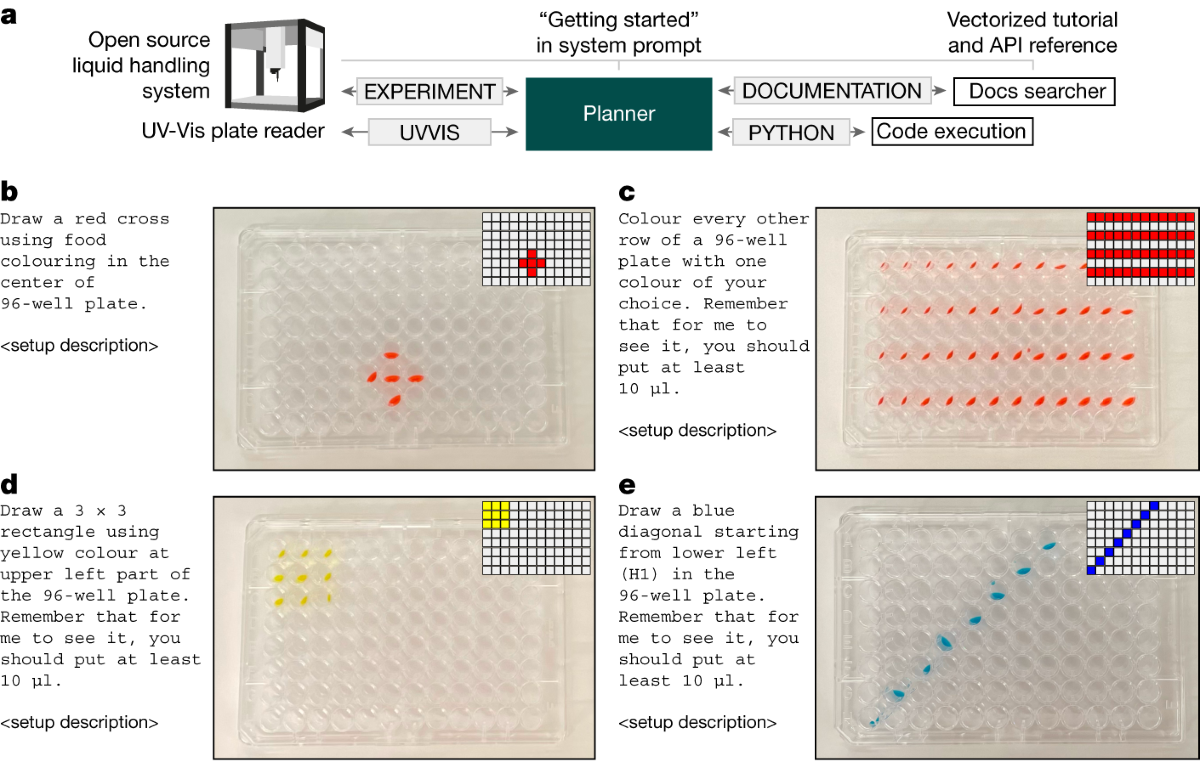}};
\end{scope}
\draw[ArtSlate!22,line width=.4pt,rounded corners=1.6mm] (7.288,1.214) rectangle ++(5.79,3.677);
\node[galbadge=ArtSage] at (7.558,4.621) {b};
\node[galcap,text width=5.79cm] at (7.288,1.134) {\textbf{Coscientist}\enspace an LLM agent running a liquid handler};
\begin{scope}\clip[rounded corners=1.6mm] (13.318,1.214) rectangle ++(2.762,3.677);
  \node[inner sep=0pt] at (15.686,2.23) {\includegraphics[width=4.736cm]{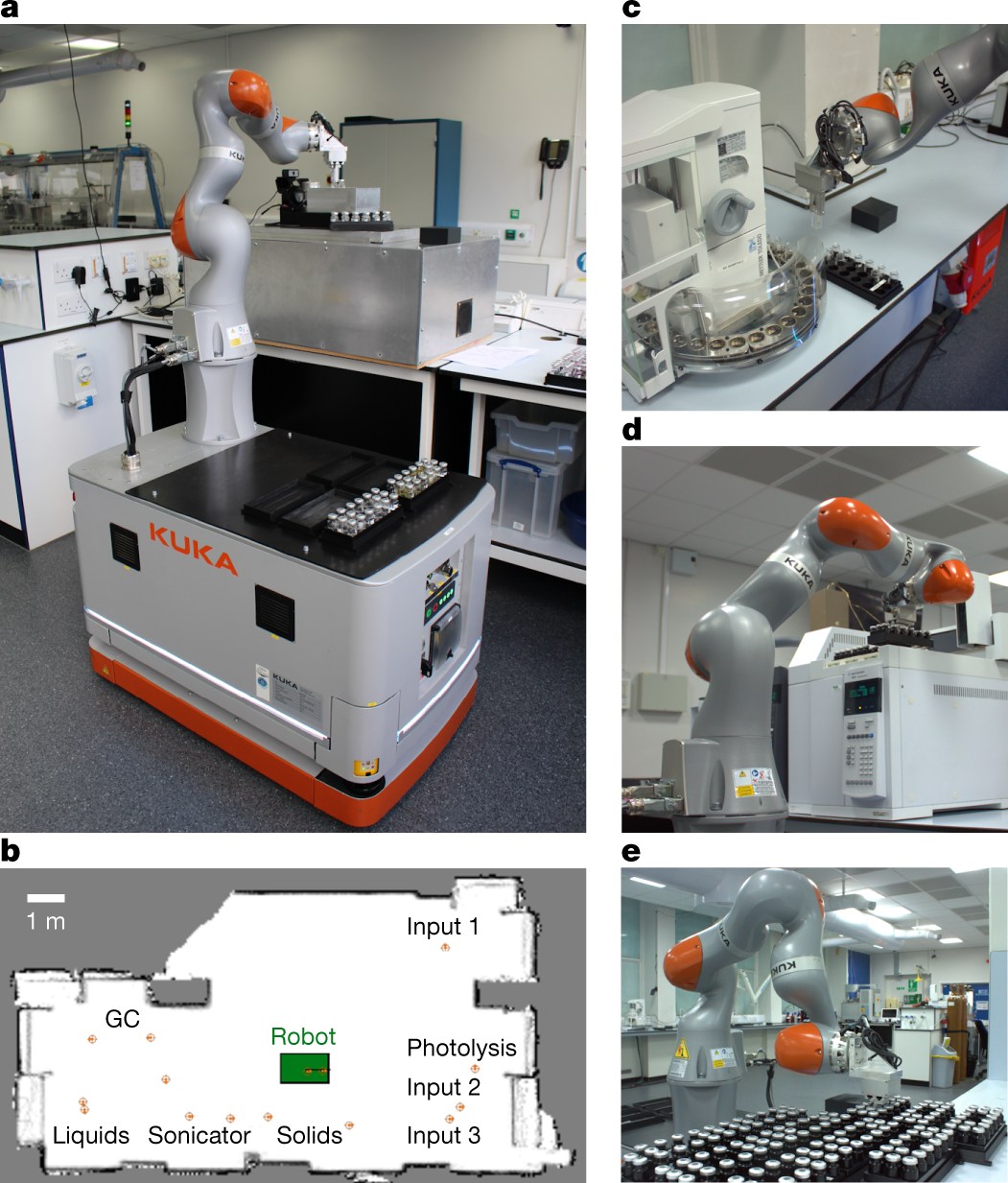}};
\end{scope}
\draw[ArtSlate!22,line width=.4pt,rounded corners=1.6mm] (13.318,1.214) rectangle ++(2.762,3.677);
\node[galbadge=ArtSage] at (13.588,4.621) {c};
\node[galcap,text width=2.762cm] at (13.318,1.134) {\textbf{Mobile robotic chemist}\enspace in a chemistry lab};
\end{tikzpicture}
\captionof{figure}{\textbf{Gallery V: self-driving laboratories.} (a) A-Lab~\citep{szymanski2023alab}; (b) Coscientist~\citep{boiko2023coscientist}; (c) the mobile robotic chemist~\citep{burger2020chemist}. Image sources: (a) Fig.~1 and (b) Fig.~4 of the papers, CC BY 4.0; (c) adapted from Fig.~1a of the paper, \textcopyright\ Springer Nature.}
\label{fig:gallery-labs}
\end{minipage}
\par\vfill
\noindent\begin{minipage}{\linewidth}
\centering
\begin{tikzpicture}[x=1cm,y=1cm,
  galbadge/.style={circle,fill=#1!88!black,draw=white,line width=.5pt,text=white,inner sep=0pt,minimum size=.4cm,font=\sffamily\bfseries\fontsize{7.6}{7.6}\selectfont},
  galcap/.style={anchor=north west,inner xsep=0pt,inner ysep=0pt,align=left,font=\fontsize{8.2}{9.6}\selectfont,text=ArtSlate}]
\path[use as bounding box] (0,0) rectangle (16.08,14.898);
\begin{scope}\clip[rounded corners=1.6mm] (0,11.398) rectangle ++(7.588,3.5);
  \node[inner sep=0pt] at (3.794,13.148) {\includegraphics[width=7.588cm]{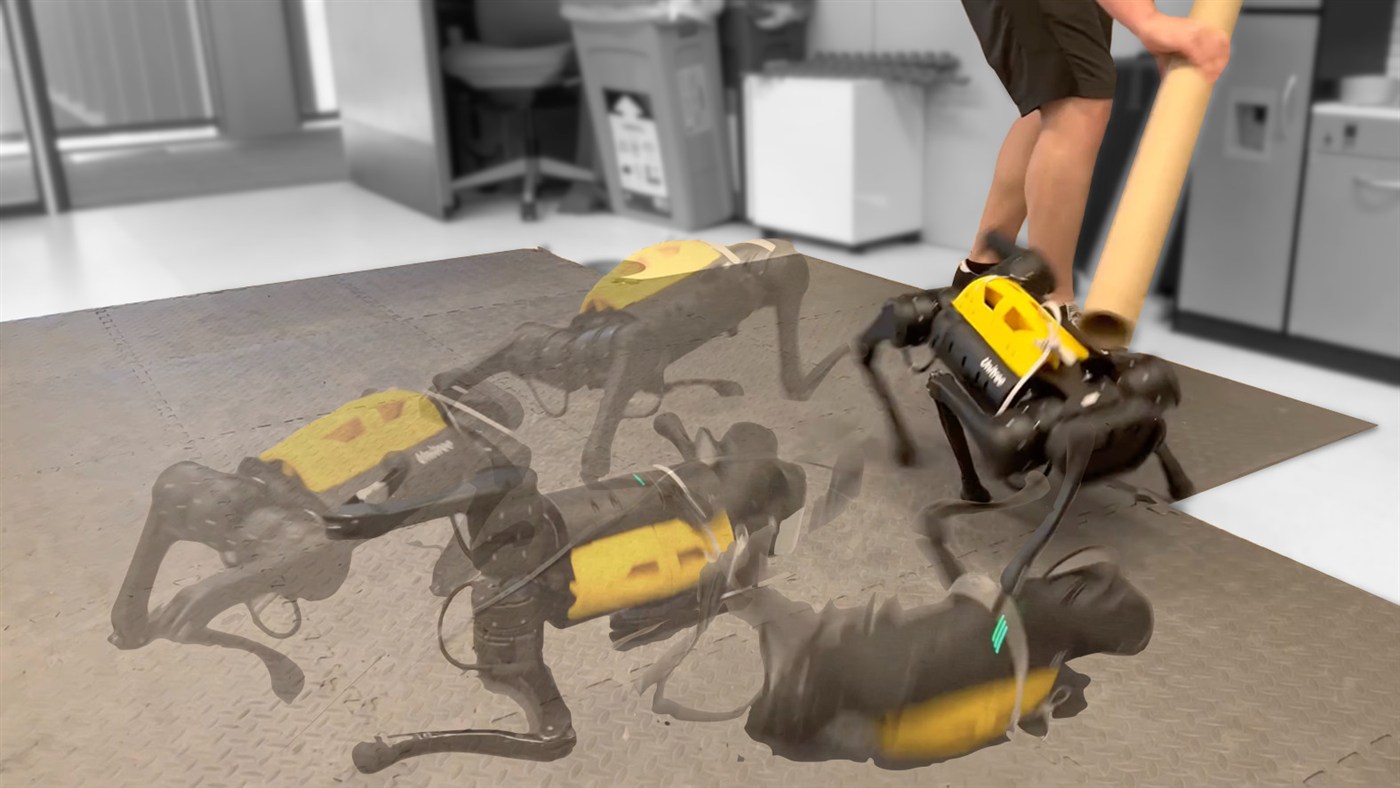}};
\end{scope}
\draw[ArtSlate!22,line width=.4pt,rounded corners=1.6mm] (0,11.398) rectangle ++(7.588,3.5);
\node[galbadge=PAISyn7] at (0.27,14.628) {a};
\node[galcap,text width=7.588cm] at (0,11.318) {\textbf{DayDreamer}\enspace a quadruped learning to withstand pushes};
\begin{scope}\clip[rounded corners=1.6mm] (7.828,11.398) rectangle ++(8.252,3.5);
  \node[inner sep=0pt] at (11.954,13.148) {\includegraphics[width=8.252cm]{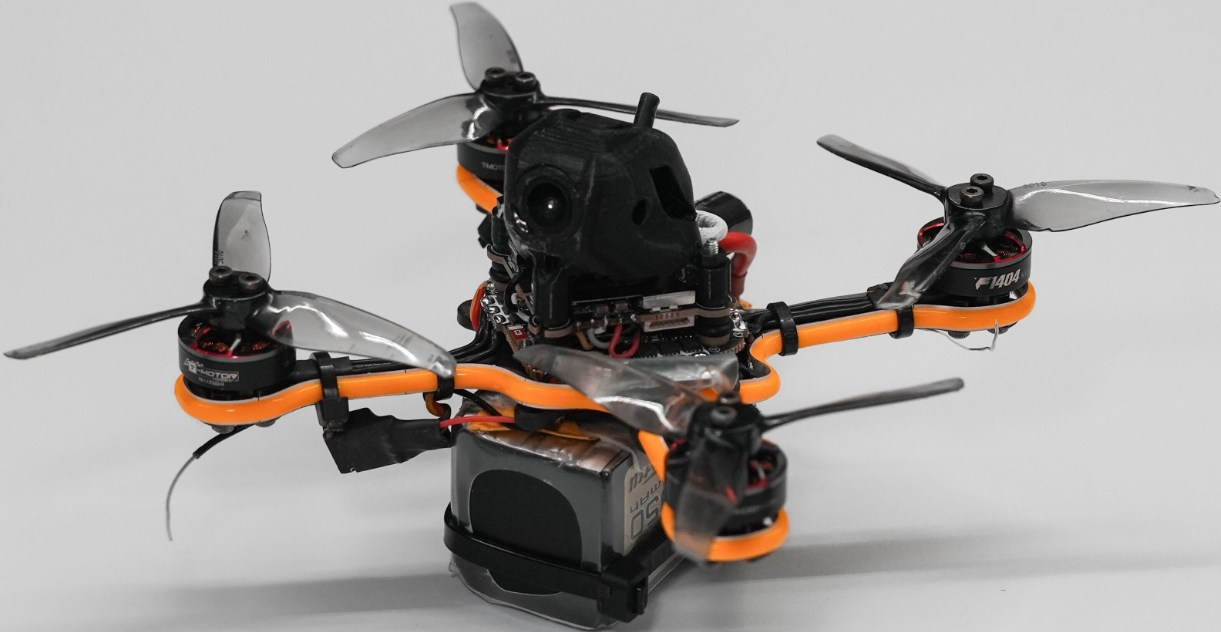}};
\end{scope}
\draw[ArtSlate!22,line width=.4pt,rounded corners=1.6mm] (7.828,11.398) rectangle ++(8.252,3.5);
\node[galbadge=PAISyn7] at (8.098,14.628) {b};
\node[galcap,text width=8.252cm] at (7.828,11.318) {\textbf{Agile flight}\enspace the quadrotor that learns in the real world};
\begin{scope}\clip[rounded corners=1.6mm] (0,7.815) rectangle ++(5.2,2.925);
  \node[inner sep=0pt] at (2.6,9.278) {\includegraphics[width=5.2cm]{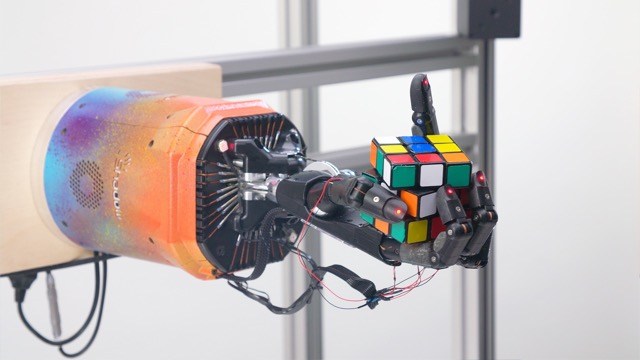}};
\end{scope}
\draw[ArtSlate!22,line width=.4pt,rounded corners=1.6mm] (0,7.815) rectangle ++(5.2,2.925);
\node[galbadge=PAISyn7] at (0.27,10.47) {c};
\node[galcap,text width=5.2cm] at (0,7.735) {\textbf{Robot hand}\enspace solving a Rubik's cube};
\begin{scope}\clip[rounded corners=1.6mm] (5.44,7.815) rectangle ++(5.2,2.925);
  \node[inner sep=0pt] at (8.04,9.278) {\includegraphics[width=5.2cm]{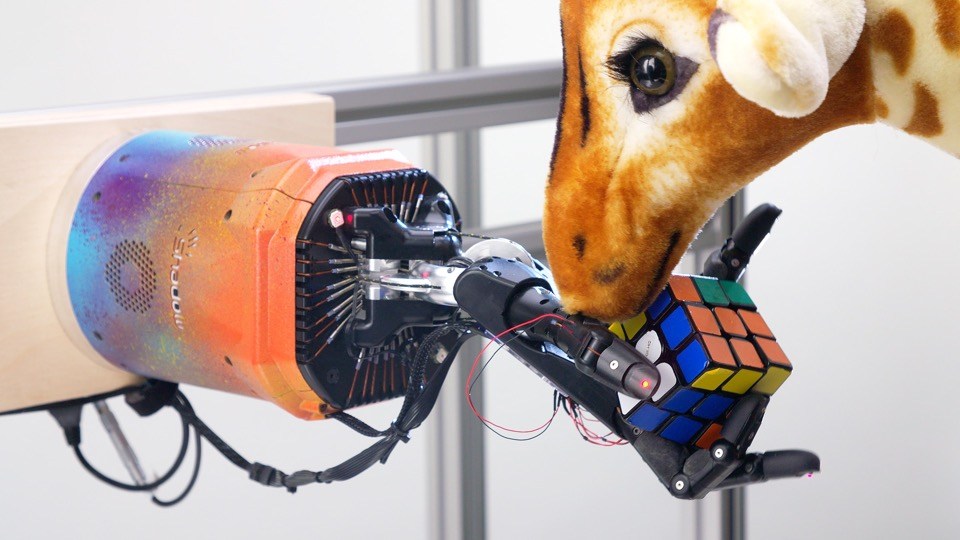}};
\end{scope}
\draw[ArtSlate!22,line width=.4pt,rounded corners=1.6mm] (5.44,7.815) rectangle ++(5.2,2.925);
\node[galbadge=PAISyn7] at (5.71,10.47) {d};
\node[galcap,text width=5.2cm] at (5.44,7.735) {\textbf{Robot hand}\enspace perturbed by a plush giraffe};
\begin{scope}\clip[rounded corners=1.6mm] (10.88,7.815) rectangle ++(5.2,2.925);
  \node[inner sep=0pt] at (13.48,9.278) {\includegraphics[width=5.2cm]{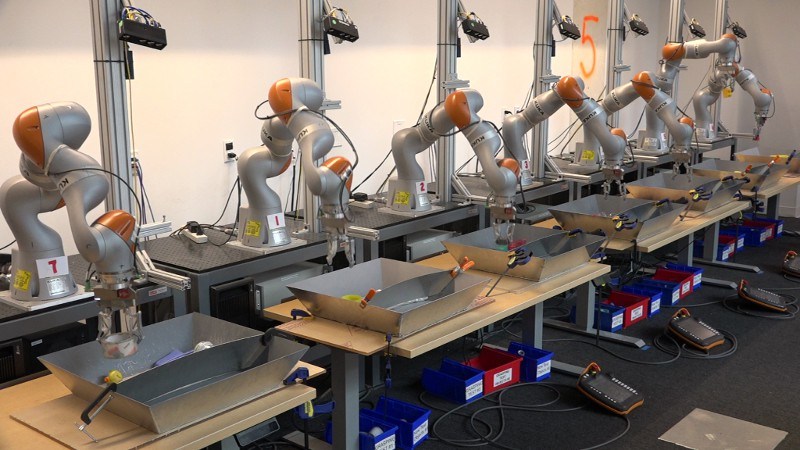}};
\end{scope}
\draw[ArtSlate!22,line width=.4pt,rounded corners=1.6mm] (10.88,7.815) rectangle ++(5.2,2.925);
\node[galbadge=PAISyn7] at (11.15,10.47) {e};
\node[galcap,text width=5.2cm] at (10.88,7.735) {\textbf{QT-Opt}\enspace seven arms collecting grasps};
\begin{scope}\clip[rounded corners=1.6mm] (0,4.104) rectangle ++(3.06,3.053);
  \node[inner sep=0pt] at (1.53,6.123) {\includegraphics[width=3.06cm]{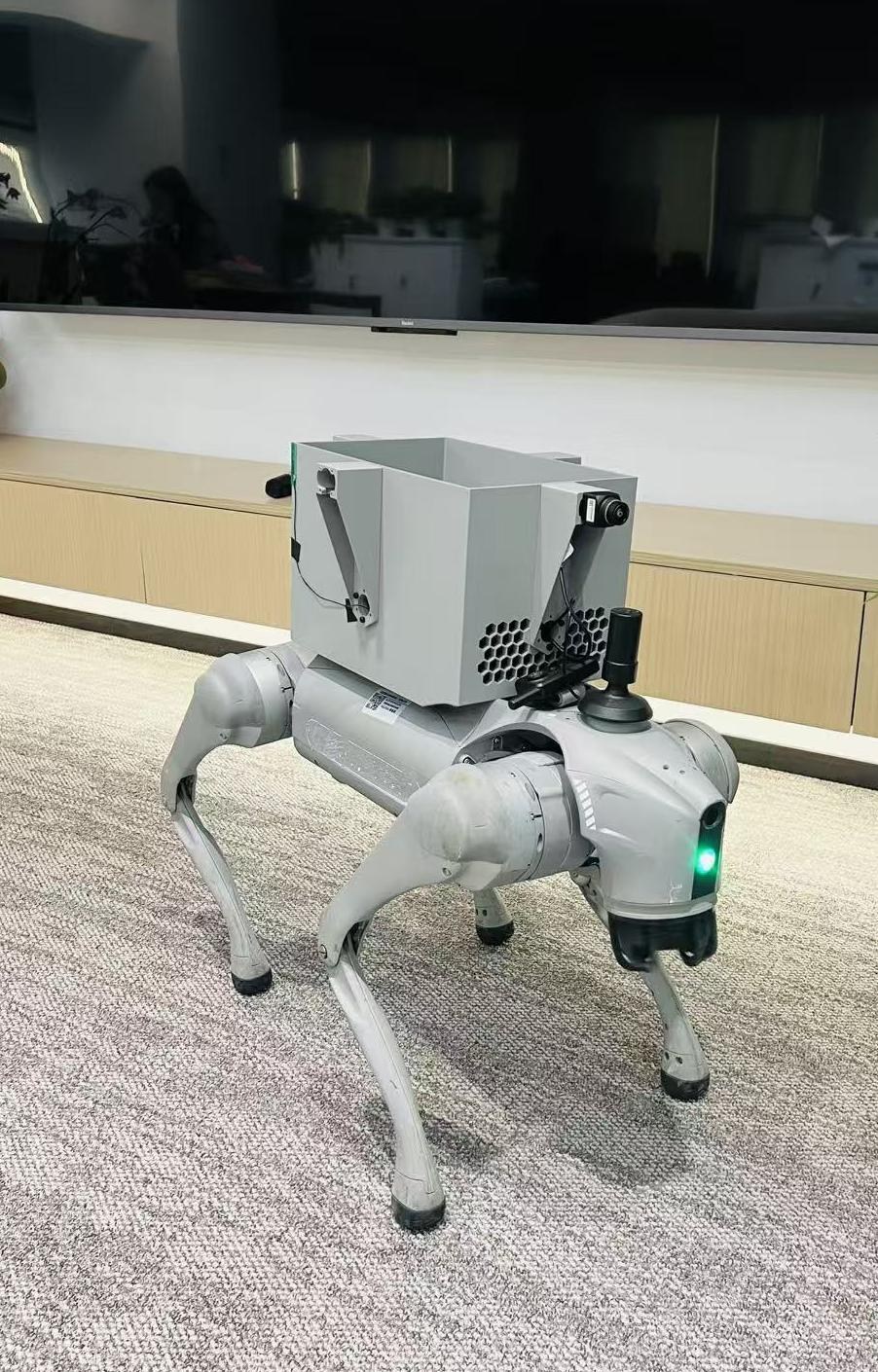}};
\end{scope}
\draw[ArtSlate!22,line width=.4pt,rounded corners=1.6mm] (0,4.104) rectangle ++(3.06,3.053);
\node[galbadge=PAISyn7] at (0.27,6.887) {f};
\node[galcap,text width=3.06cm] at (0,4.024) {\textbf{AllDayNav}\enspace a Go2 for lifelong navigation};
\begin{scope}\clip[rounded corners=1.6mm] (3.3,4.104) rectangle ++(3.392,3.053);
  \node[inner sep=0pt] at (4.995,5.631) {\includegraphics[width=3.392cm]{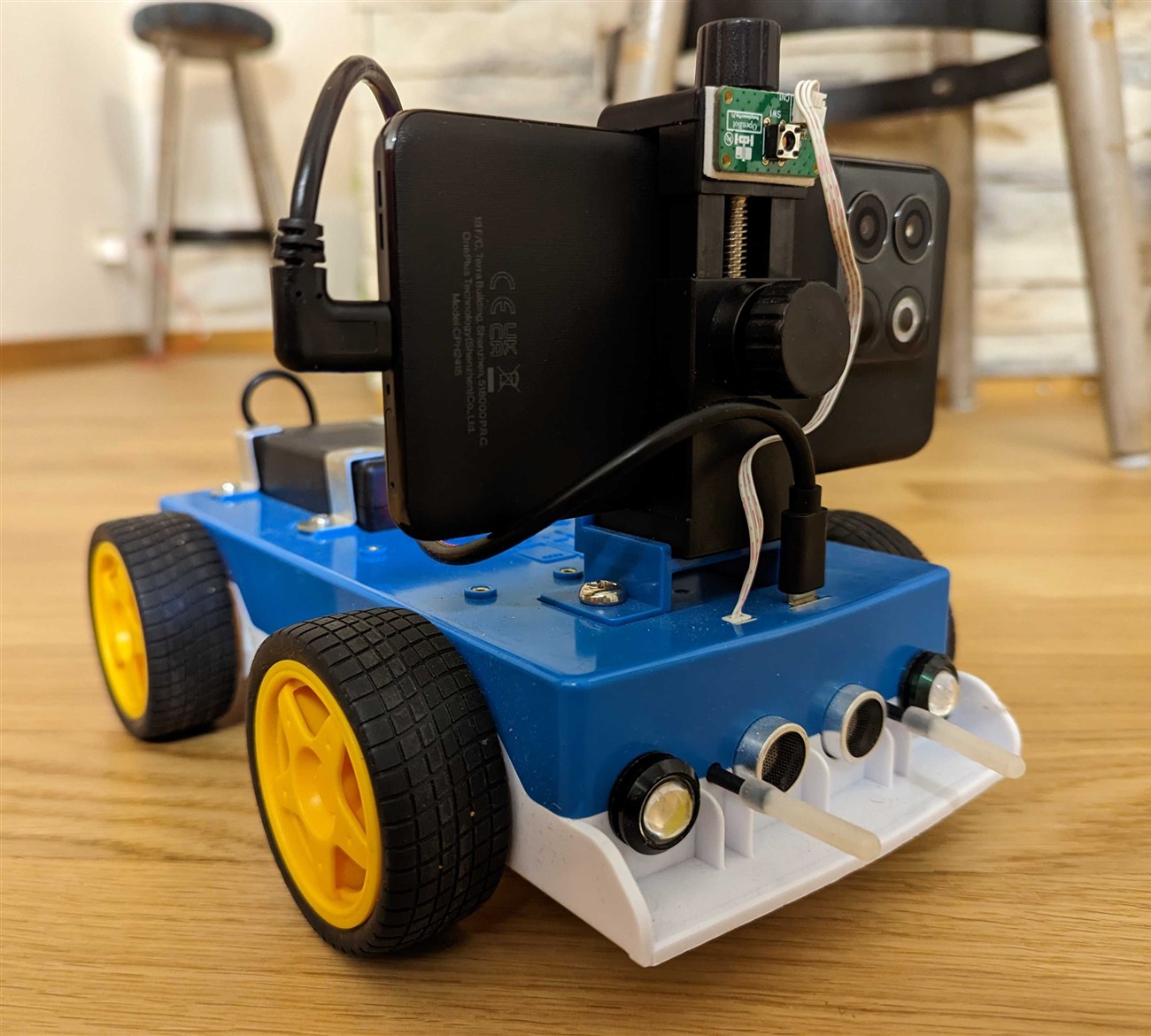}};
\end{scope}
\draw[ArtSlate!22,line width=.4pt,rounded corners=1.6mm] (3.3,4.104) rectangle ++(3.392,3.053);
\node[galbadge=PAISyn7] at (3.57,6.887) {g};
\node[galcap,text width=3.392cm] at (3.3,4.024) {\textbf{OpenBot-Fleet}\enspace a smartphone robot shipped to crowd workers};
\begin{scope}\clip[rounded corners=1.6mm] (6.931,4.104) rectangle ++(4.07,3.053);
  \node[inner sep=0pt] at (8.967,5.631) {\includegraphics[width=4.07cm]{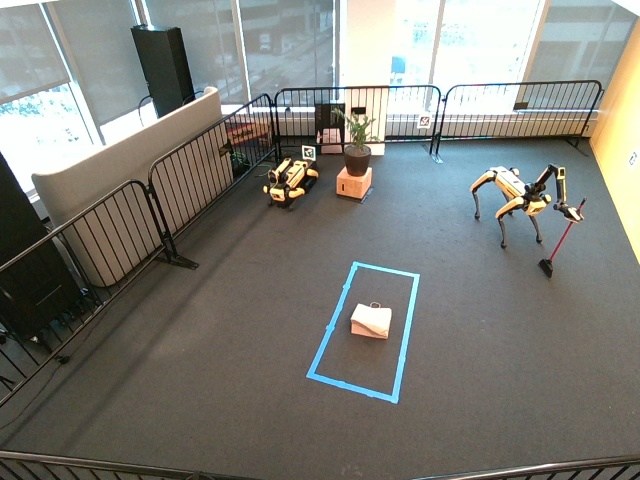}};
\end{scope}
\draw[ArtSlate!22,line width=.4pt,rounded corners=1.6mm] (6.931,4.104) rectangle ++(4.07,3.053);
\node[galbadge=PAISyn7] at (7.201,6.887) {h};
\node[galcap,text width=4.07cm] at (6.931,4.024) {\textbf{Spot}\enspace a sweeping task for autonomous RL};
\begin{scope}\clip[rounded corners=1.6mm] (11.242,4.104) rectangle ++(4.838,3.053);
  \node[inner sep=0pt] at (13.661,5.631) {\includegraphics[width=4.838cm]{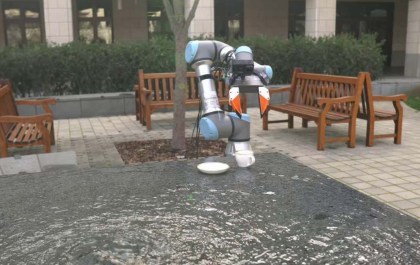}};
\end{scope}
\draw[ArtSlate!22,line width=.4pt,rounded corners=1.6mm] (11.242,4.104) rectangle ++(4.838,3.053);
\node[galbadge=PAISyn7] at (11.512,6.887) {i};
\node[galcap,text width=4.838cm] at (11.242,4.024) {\textbf{UMI}\enspace a cup task at a fountain};
\begin{scope}\clip[rounded corners=1.6mm] (0,0.538) rectangle ++(16.08,2.232);
  \node[inner sep=0pt] at (8.04,1.654) {\includegraphics[width=16.08cm]{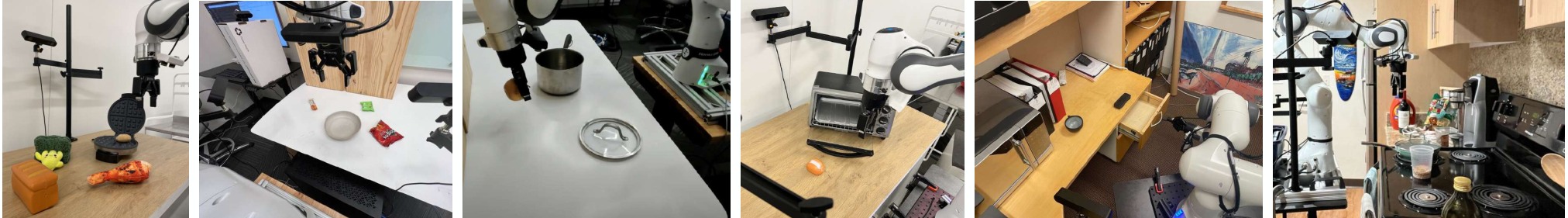}};
\end{scope}
\draw[ArtSlate!22,line width=.4pt,rounded corners=1.6mm] (0,0.538) rectangle ++(16.08,2.232);
\node[galbadge=PAISyn7] at (0.27,2.5) {j};
\node[galcap,text width=16.08cm] at (0,0.458) {\textbf{DROID}\enspace evaluation scenes in labs, an office and a home};
\end{tikzpicture}
\captionof{figure}{\textbf{Gallery VI: other bodies and settings.} (a) DayDreamer~\citep{wu2023daydreamer}; (b) agile quadrotor flight learned in the real world~\citep{arxiv260210111}; (c, d) solving a Rubik's cube with a robot hand~\citep{openai2019rubik}; (e) QT-Opt~\citep{kalashnikov2018qtopt}; (f) AllDayNav~\citep{arxiv260610927}; (g) OpenBot-Fleet~\citep{arxiv240507515}; (h) continual mobile manipulation with Spot~\citep{mendonca2024continuously}; (i) UMI~\citep{chi2024umi}; (j) DROID~\citep{arxiv240312945}. Image sources: (a) Fig.~8 and (j) Fig.~5 of the papers, CC BY 4.0; (b) the photograph used in Fig.~3, (c) Fig.~1, (d) Fig.~17(e), (e) Fig.~1, (f) Fig.~1, (g) Fig.~2, (h) Fig.~3(g) and (i) Fig.~9 of the papers.}
\label{fig:gallery-bodies}
\end{minipage}
\clearpage

%% file: main.bbl
\begin{thebibliography}{379}
\providecommand{\natexlab}[1]{#1}
\providecommand{\url}[1]{\texttt{#1}}
\expandafter\ifx\csname urlstyle\endcsname\relax
  \providecommand{\doi}[1]{doi: #1}\else
  \providecommand{\doi}{doi: \begingroup \urlstyle{rm}\Url}\fi

\bibitem[Achiam et~al.(2017)Achiam, Held, Tamar, and Abbeel]{achiam2017cpo}
Joshua Achiam, David Held, Aviv Tamar, and Pieter Abbeel.
\newblock {Constrained Policy Optimization}.
\newblock In \emph{Proceedings of the 34th International Conference on Machine Learning}, volume~70 of \emph{Proceedings of Machine Learning Research}, 2017.

\bibitem[Agia et~al.(2025)Agia, Sinha, Yang, Antonova, Pavone, Nishimura, Itkina, and Bohg]{arxiv250619121}
Christopher Agia, Rohan Sinha, Jingyun Yang, Rika Antonova, Marco Pavone, Haruki Nishimura, Masha Itkina, and Jeannette Bohg.
\newblock {CUPID: Curating Data your Robot Loves with Influence Functions}.
\newblock In \emph{Proceedings of The 9th Conference on Robot Learning (CoRL)}, volume 305 of \emph{Proceedings of Machine Learning Research}, pages 2907--2932, 2025.

\bibitem[Ahmed et~al.(2021)Ahmed, Tr{\"a}uble, Goyal, Neitz, Bengio, Sch{\"o}lkopf, W{\"u}thrich, and Bauer]{ahmed2021causalworld}
Ossama Ahmed, Frederik Tr{\"a}uble, Anirudh Goyal, Alexander Neitz, Yoshua Bengio, Bernhard Sch{\"o}lkopf, Manuel W{\"u}thrich, and Stefan Bauer.
\newblock {CausalWorld: A Robotic Manipulation Benchmark for Causal Structure and Transfer Learning}.
\newblock In \emph{International Conference on Learning Representations}, 2021.

\bibitem[Ahn et~al.(2024)Ahn, Dwibedi, Finn, {Gonzalez Arenas}, Gopalakrishnan, Hausman, Ichter, Irpan, Joshi, Julian, Kirmani, Leal, Lee, Levine, Lu, Maddineni, Rao, Sadigh, Sanketi, Sermanet, Vuong, Welker, Xia, Xiao, Xu, Xu, and Xu]{ahn2024autort}
Michael Ahn, Debidatta Dwibedi, Chelsea Finn, Montse {Gonzalez Arenas}, Keerthana Gopalakrishnan, Karol Hausman, Brian Ichter, Alex Irpan, Nikhil Joshi, Ryan Julian, Sean Kirmani, Isabel Leal, Edward Lee, Sergey Levine, Yao Lu, Sharath Maddineni, Kanishka Rao, Dorsa Sadigh, Pannag Sanketi, Pierre Sermanet, Quan Vuong, Stefan Welker, Fei Xia, Ted Xiao, Peng Xu, Steve Xu, and Zhuo Xu.
\newblock {AutoRT: Embodied Foundation Models for Large Scale Orchestration of Robotic Agents}.
\newblock arXiv preprint arXiv:2401.12963, 2024.

\bibitem[Alemohammad et~al.(2024)Alemohammad, Casco-Rodriguez, Luzi, Humayun, Babaei, LeJeune, Siahkoohi, and Baraniuk]{alemohammad2024mad}
Sina Alemohammad, Josue Casco-Rodriguez, Lorenzo Luzi, Ahmed~Imtiaz Humayun, Hossein Babaei, Daniel LeJeune, Ali Siahkoohi, and Richard~G. Baraniuk.
\newblock {Self-Consuming Generative Models Go MAD}.
\newblock In \emph{International Conference on Learning Representations}, 2024.

\bibitem[Alshiekh et~al.(2018)Alshiekh, Bloem, Ehlers, Könighofer, Niekum, and Topcu]{alshiekh2018shielding}
Mohammed Alshiekh, Roderick Bloem, Rüdiger Ehlers, Bettina Könighofer, Scott Niekum, and Ufuk Topcu.
\newblock {Safe Reinforcement Learning via Shielding}.
\newblock \emph{Proceedings of the AAAI Conference on Artificial Intelligence}, 32\penalty0 (1), 2018.
\newblock \doi{10.1609/aaai.v32i1.11797}.

\bibitem[Ames et~al.(2019)Ames, Coogan, Egerstedt, Notomista, Sreenath, and Tabuada]{ames2019cbf}
Aaron~D. Ames, Samuel Coogan, Magnus Egerstedt, Gennaro Notomista, Koushil Sreenath, and Paulo Tabuada.
\newblock {Control Barrier Functions: Theory and Applications}.
\newblock In \emph{European Control Conference}, pages 3420--3431, 2019.

\bibitem[Amodei et~al.(2016)Amodei, Olah, Steinhardt, Christiano, Schulman, and Mané]{amodei2016safety}
Dario Amodei, Chris Olah, Jacob Steinhardt, Paul Christiano, John Schulman, and Dan Mané.
\newblock {Concrete Problems in AI Safety}.
\newblock arXiv preprint arXiv:1606.06565, 2016.

\bibitem[Andrychowicz et~al.(2016)Andrychowicz, Denil, G{\'o}mez, Hoffman, Pfau, Schaul, Shillingford, and de~Freitas]{andrychowicz2016learning}
Marcin Andrychowicz, Misha Denil, Sergio G{\'o}mez, Matthew~W. Hoffman, David Pfau, Tom Schaul, Brendan Shillingford, and Nando de~Freitas.
\newblock {Learning to Learn by Gradient Descent by Gradient Descent}.
\newblock In \emph{Advances in Neural Information Processing Systems (NeurIPS)}, 2016.

\bibitem[Andrychowicz et~al.(2017)Andrychowicz, Wolski, Ray, Schneider, Fong, Welinder, McGrew, Tobin, Abbeel, and Zaremba]{arxiv170701495}
Marcin Andrychowicz, Filip Wolski, Alex Ray, Jonas Schneider, Rachel Fong, Peter Welinder, Bob McGrew, Josh Tobin, Pieter Abbeel, and Wojciech Zaremba.
\newblock {Hindsight Experience Replay}.
\newblock In \emph{Advances in Neural Information Processing Systems}, 2017.

\bibitem[Ankile et~al.(2025)Ankile, Simeonov, Shenfeld, Torne, and Agrawal]{arxiv240716677}
Lars Ankile, Anthony Simeonov, Idan Shenfeld, Marcel Torne, and Pulkit Agrawal.
\newblock {From Imitation to Refinement -- Residual RL for Precise Assembly}.
\newblock In \emph{IEEE International Conference on Robotics and Automation}, 2025.
\newblock \doi{10.1109/ICRA55743.2025.11127442}.

\bibitem[Anthony et~al.(2017)Anthony, Tian, and Barber]{anthony2017expert}
Thomas Anthony, Zheng Tian, and David Barber.
\newblock {Thinking Fast and Slow with Deep Learning and Tree Search}.
\newblock In \emph{Advances in Neural Information Processing Systems}, volume~30, 2017.

\bibitem[Argall et~al.(2009)Argall, Chernova, Veloso, and Browning]{argall2009survey}
Brenna~D. Argall, Sonia Chernova, Manuela Veloso, and Brett Browning.
\newblock {A Survey of Robot Learning from Demonstration}.
\newblock \emph{Robotics and Autonomous Systems}, 57\penalty0 (5):\penalty0 469--483, 2009.
\newblock \doi{10.1016/j.robot.2008.10.024}.

\bibitem[Assran et~al.(2025)Assran, Bardes, Fan, Garrido, Howes, Komeili, Muckley, Rizvi, Roberts, et~al.]{arxiv250609985}
Mido Assran, Adrien Bardes, David Fan, Quentin Garrido, Russell Howes, Mojtaba Komeili, Matthew Muckley, Ammar Rizvi, Claire Roberts, et~al.
\newblock {V-JEPA 2: Self-Supervised Video Models Enable Understanding, Prediction and Planning}.
\newblock arXiv preprint arXiv:2506.09985, 2025.

\bibitem[Atreya et~al.(2025)Atreya, Pertsch, Lee, Kim, Jain, Kuramshin, Eppner, Neary, Hu, Ramos, Tremblay, Arora, Ellis, Macesanu, Villasevil, Leonard, Cho, Aslan, Dass, Wang, Reger, Yuan, Yang, Gupta, Jayaraman, Berseth, Daniilidis, Martin-Martin, Lee, Liang, Finn, and Levine]{arxiv250618123}
Pranav Atreya, Karl Pertsch, Tony Lee, Moo~Jin Kim, Arhan Jain, Artur Kuramshin, Clemens Eppner, Cyrus Neary, Edward Hu, Fabio Ramos, Jonathan Tremblay, Kanav Arora, Kirsty Ellis, Luca Macesanu, Marcel~Torne Villasevil, Matthew Leonard, Meedeum Cho, Ozgur Aslan, Shivin Dass, Jie Wang, William Reger, Xingfang Yuan, Xuning Yang, Abhishek Gupta, Dinesh Jayaraman, Glen Berseth, Kostas Daniilidis, Roberto Martin-Martin, Youngwoon Lee, Percy Liang, Chelsea Finn, and Sergey Levine.
\newblock {RoboArena: Distributed Real-World Evaluation of Generalist Robot Policies}.
\newblock arXiv preprint arXiv:2506.18123, 2025.

\bibitem[Attarian et~al.(2026)Attarian, Vyse, Gerigk, Opryshko, Ruan, Almasri, Singh, Du, Gilitschenski, and Voelcker]{arxiv260310282}
Maria Attarian, Ian Vyse, Jasper Gerigk, Evgenii Opryshko, Yifan Ruan, Anas Almasri, Sumeet Singh, Yilun Du, Igor Gilitschenski, and Claas Voelcker.
\newblock {Update-Free On-Policy Steering via Verifiers}.
\newblock arXiv preprint arXiv:2603.10282, 2026.

\bibitem[Badithela et~al.(2026)Badithela, Snyder, Yin, Sceats, Pappas, Matni, Itkina, Nishimura, and Majumdar]{arxiv261014603}
Apurva Badithela, David Snyder, Tenny Yin, Jade Sceats, George Pappas, Nikolai Matni, Masha Itkina, Haruki Nishimura, and Anirudha Majumdar.
\newblock {AnyRank: Reliable Anytime Policy Ranking via Sequential Testing}.
\newblock arXiv preprint arXiv:2610.14603, 2026.
\newblock Accepted to CoRL 2026.

\bibitem[Bai et~al.(2026)Bai, Li, Huang, and Karlsson]{arxiv260830760}
Ziyi Bai, Siqi Li, Tinglei Huang, and B{\"o}rje~F. Karlsson.
\newblock {PRACTICE: From Experience to Expertise in Self-Evolving Embodied Agents}.
\newblock arXiv preprint arXiv:2608.30760, 2026.

\bibitem[Ball et~al.(2023)Ball, Smith, Kostrikov, and Levine]{ball2023rlpd}
Philip~J. Ball, Laura Smith, Ilya Kostrikov, and Sergey Levine.
\newblock {Efficient Online Reinforcement Learning with Offline Data}.
\newblock In \emph{International Conference on Machine Learning}, 2023.

\bibitem[Bao et~al.(2026)Bao, Huang, Lin, Feng, Li, Zhao, Li, He, Chi, and Zhang]{arxiv260937583}
Shifeng Bao, Fanding Huang, Yihan Lin, Youhe Feng, Guanlin Li, Chen Zhao, Yang Li, Jiawei He, Cheng Chi, and Jing Zhang.
\newblock {RoboHarn-Evo: Evolving Hierarchical Physical Knowledge for Self-Improving Robotic Manipulation}.
\newblock arXiv preprint arXiv:2609.37583, 2026.

\bibitem[Baranes and Oudeyer(2013)]{baranes2013goals}
Adrien Baranes and Pierre-Yves Oudeyer.
\newblock {Active Learning of Inverse Models with Intrinsically Motivated Goal Exploration in Robots}.
\newblock \emph{Robotics and Autonomous Systems}, 61\penalty0 (1):\penalty0 49--73, 2013.
\newblock \doi{10.1016/j.robot.2012.05.008}.

\bibitem[Bechtle et~al.(2021)Bechtle, Molchanov, Chebotar, Grefenstette, Righetti, Sukhatme, and Meier]{bechtle2021ml3}
Sarah Bechtle, Artem Molchanov, Yevgen Chebotar, Edward Grefenstette, Ludovic Righetti, Gaurav Sukhatme, and Franziska Meier.
\newblock {Meta-Learning via Learned Loss}.
\newblock In \emph{Proceedings of the International Conference on Pattern Recognition (ICPR)}, 2021.

\bibitem[Belkhale et~al.(2023)Belkhale, Cui, and Sadigh]{belkhale2023quality}
Suneel Belkhale, Yuchen Cui, and Dorsa Sadigh.
\newblock {Data Quality in Imitation Learning}.
\newblock In \emph{Advances in Neural Information Processing Systems}, volume~36, 2023.
\newblock \doi{10.52202/075280-3524}.

\bibitem[Bi et~al.(2026)Bi, Zhou, Tang, Pang, Huang, Liu, Wang, Huang, Wang, Cheng, Zhao, Li, Tan, Liu, Wan, Liu, Zhao, Bao, and Zhu]{arxiv260830237}
Hongzhe Bi, Zihao Zhou, Yihang Tang, Jingrui Pang, Shuhe Huang, Haitian Liu, Runqing Wang, Shuai Huang, Yichen Wang, Yiming Cheng, Ruowen Zhao, Zhenghua Li, Hengkai Tan, Xiaolong Liu, Jinhui Wan, Jiabao Liu, Min Zhao, Fan Bao, and Jun Zhu.
\newblock {Motus2: A Self-Evolving General World Model for Dexterous Manipulation}.
\newblock arXiv preprint arXiv:2608.30237, 2026.

\bibitem[Black et~al.(2025)Black, Brown, Driess, Esmail, Equi, Finn, Fusai, Groom, Hausman, Ichter, et~al.]{arxiv241024164}
Kevin Black, Noah Brown, Danny Driess, Adnan Esmail, Michael Equi, Chelsea Finn, Niccolo Fusai, Lachy Groom, Karol Hausman, Brian Ichter, et~al.
\newblock {$\pi_0$: A Vision-Language-Action Flow Model for General Robot Control}.
\newblock In \emph{Robotics: Science and Systems}, 2025.

\bibitem[Boiko et~al.(2023)Boiko, MacKnight, Kline, and Gomes]{boiko2023coscientist}
Daniil~A. Boiko, Robert MacKnight, Ben Kline, and Gabe Gomes.
\newblock {Autonomous chemical research with large language models}.
\newblock \emph{Nature}, 624\penalty0 (7992):\penalty0 570--578, 2023.
\newblock \doi{10.1038/s41586-023-06792-0}.

\bibitem[Bongard et~al.(2006)Bongard, Zykov, and Lipson]{bongard2006selfmodel}
Josh Bongard, Victor Zykov, and Hod Lipson.
\newblock {Resilient Machines Through Continuous Self-Modeling}.
\newblock \emph{Science}, 314\penalty0 (5802):\penalty0 1118--1121, 2006.
\newblock \doi{10.1126/science.1133687}.

\bibitem[Bousmalis et~al.(2024)Bousmalis, Vezzani, Rao, Devin, Lee, Bauza, et~al.]{arxiv230611706}
Konstantinos Bousmalis, Giulia Vezzani, Dushyant Rao, Coline Devin, Alex~X. Lee, Maria Bauza, et~al.
\newblock {RoboCat: A Self-Improving Generalist Agent for Robotic Manipulation}.
\newblock \emph{Transactions on Machine Learning Research}, 2024.
\newblock arXiv:2306.11706.

\bibitem[Brohan et~al.(2023{\natexlab{a}})Brohan, Brown, Carbajal, Chebotar, Chen, et~al.]{brohan2023rt2}
Anthony Brohan, Noah Brown, Justice Carbajal, Yevgen Chebotar, Xi~Chen, et~al.
\newblock {RT-2: Vision-Language-Action Models Transfer Web Knowledge to Robotic Control}.
\newblock In \emph{Conference on Robot Learning}, 2023{\natexlab{a}}.

\bibitem[Brohan et~al.(2023{\natexlab{b}})Brohan, Brown, Carbajal, Chebotar, Dabis, et~al.]{brohan2023rt1}
Anthony Brohan, Noah Brown, Justice Carbajal, Yevgen Chebotar, Joseph Dabis, et~al.
\newblock {RT-1: Robotics Transformer for Real-World Control at Scale}.
\newblock In \emph{Robotics: Science and Systems}, 2023{\natexlab{b}}.

\bibitem[Bruce et~al.(2024)Bruce, Dennis, Edwards, Parker-Holder, Shi, Hughes, Lai, Mavalankar, Steigerwald, Apps, et~al.]{bruce2024genie}
Jake Bruce, Michael Dennis, Ashley Edwards, Jack Parker-Holder, Yuge Shi, Edward Hughes, Matthew Lai, Aditi Mavalankar, Richie Steigerwald, Chris Apps, et~al.
\newblock {Genie: Generative Interactive Environments}.
\newblock In \emph{Proceedings of the 41st International Conference on Machine Learning}, volume 235 of \emph{Proceedings of Machine Learning Research}, pages 4603--4623, 2024.

\bibitem[Brunke et~al.(2022)Brunke, Greeff, Hall, Yuan, Zhou, Panerati, and Schoellig]{brunke2022safe}
Lukas Brunke, Melissa Greeff, Adam~W. Hall, Zhaocong Yuan, Siqi Zhou, Jacopo Panerati, and Angela~P. Schoellig.
\newblock {Safe Learning in Robotics: From Learning-Based Control to Safe Reinforcement Learning}.
\newblock \emph{Annual Review of Control, Robotics, and Autonomous Systems}, 5\penalty0 (1):\penalty0 411--444, 2022.
\newblock \doi{10.1146/annurev-control-042920-020211}.

\bibitem[Burger et~al.(2020)Burger, Maffettone, Gusev, Aitchison, Bai, Wang, Li, Alston, Li, Clowes, Rankin, Harris, Sprick, and Cooper]{burger2020chemist}
Benjamin Burger, Phillip~M. Maffettone, Vladimir~V. Gusev, Catherine~M. Aitchison, Yang Bai, Xiaoyan Wang, Xiaobo Li, Ben~M. Alston, Buyi Li, Rob Clowes, Nicola Rankin, Brandon Harris, Reiner~Sebastian Sprick, and Andrew~I. Cooper.
\newblock {A mobile robotic chemist}.
\newblock \emph{Nature}, 583\penalty0 (7815):\penalty0 237--241, 2020.
\newblock \doi{10.1038/s41586-020-2442-2}.

\bibitem[Cai et~al.(2026)Cai, Cao, Chen, Chua, Dai, Deng, Deng, Diao, Ding, Fan, Fu, Gong, Gu, He, Hou, Li, Li, Li, Li, Liao, Lin, Lin, Liu, Liu, Liu, Liu, Liu, Loy, Lu, Luo, Ma, Pan, Pan, Qian, Shang, Shao, Su, Tao, Tao, Wang, Wang, Wang, Wang, Wen, Wu, Wu, Wu, Wu, Wu, Xiao, Xu, Yamashita, Yang, Yang, Yin, Yin, Zhao, Zhong, Zhuo, and Zuo]{arxiv261022471}
Zhongang Cai, Ziang Cao, Tailai Chen, Han~Chong Chua, Linjun Dai, Hanming Deng, Hokin Deng, Haiwen Diao, Chenjing Ding, Weichen Fan, Xiao Fu, Zerui Gong, Chenyang Gu, Zexin He, Zhi Hou, Bo~Li, Qinyue Li, Quanzhou Li, Xiaowan Li, Zhichao Liao, Dahua Lin, Zilong Lin, Bo~Liu, Chunxiao Liu, Pengfei Liu, Tianqi Liu, Ziwei Liu, Chen~Change Loy, Lewei Lu, Jiapeng Luo, Yuexiao Ma, Liang Pan, Xingang Pan, Oscar Qian, Ziyi Shang, Yuyao Shao, Dawei Su, Dacheng Tao, Qingyi Tao, Quan Wang, Ruisi Wang, Xiaogang Wang, Ziyi Wang, Beichen Wen, Donghao Wu, Jiaxu Wu, Silei Wu, Xiaoming Wu, Zhonghua Wu, Hongcan Xiao, Weiye Xu, Atsushi Yamashita, Lei Yang, Zhe Yang, Baiqiao Yin, Wanqi Yin, Hanfei Zhao, Tianbo Zhong, Long Zhuo, and Jing Zuo.
\newblock {Has GPT-6-Astra Achieved Embodied Intelligence?}
\newblock arXiv preprint arXiv:2610.22471, 2026.

\bibitem[Chan et~al.(2026)Chan, Zhang, Liu, Cui, Zeng, Ding, Ren, Zhang, Chen, Liu, Dai, and Ju]{arxiv260518729}
Nga~Teng Chan, Yi~Zhang, Yechi Liu, Renwen Cui, Fanhu Zeng, Zeyuan Ding, Xiancong Ren, Zhang Zhang, Qifeng Chen, Jian Liu, Yong Dai, and Xiaozhu Ju.
\newblock {Robo-Cortex: A Self-Evolving Embodied Agent via Dual-Grain Cognitive Memory and Autonomous Knowledge Induction}.
\newblock arXiv preprint arXiv:2605.18729, 2026.

\bibitem[Chen et~al.(2025{\natexlab{a}})Chen, Lessing, Liu, and Finn]{arxiv250303707}
Annie~S. Chen, Alec~M. Lessing, Yuejiang Liu, and Chelsea Finn.
\newblock {Curating Demonstrations using Online Experience}.
\newblock In \emph{Proceedings of Robotics: Science and Systems (RSS)}, 2025{\natexlab{a}}.
\newblock \doi{10.15607/RSS.2025.XXI.071}.

\bibitem[Chen et~al.(2026{\natexlab{a}})Chen, Ding, Huang, Li, Wang, He, Li, Sun, Liu, Wu, and Cao]{arxiv260830880}
Fu~Chen, Xin Ding, Bingjia Huang, Xiangyu Li, Mingju Wang, Jiawei He, Kun Li, Wei Sun, Yunxin Liu, Hao Wu, and Ting Cao.
\newblock {Zeva: In-Context Causal Learning for Generalizable Embodied Manipulation}.
\newblock arXiv preprint arXiv:2608.30880, 2026{\natexlab{a}}.

\bibitem[Chen et~al.(2026{\natexlab{b}})Chen, Zhang, Schaefer, Chen, Chen, Cremers, Mees, and Leutenegger]{arxiv260621406}
Hanzhi Chen, Anran Zhang, Simon Schaefer, Kejia Chen, Shi Chen, Daniel Cremers, Oier Mees, and Stefan Leutenegger.
\newblock {Robot Self-Improvement via Human-Video Dynamics Models}.
\newblock arXiv preprint arXiv:2606.21406, 2026{\natexlab{b}}.

\bibitem[Chen et~al.(2026{\natexlab{c}})Chen, Feng, Ma, Yan, Lin, Zhang, Du, An, Yang, Liu, Li, Xu, Zhou, Su, Jin, Ma, Qi, Wang, Ma, Pu, Du, Song, Huang, Fu, Wang, Li, Chen, Xue, Li, Du, et~al.]{arxiv261000093}
Jiahao Chen, Zhou Feng, Oubo Ma, Yichen Yan, Ruixiao Lin, Hangtao Zhang, Linkang Du, Hengyu An, Yong Yang, Jun Liu, Junhao Li, Naen Xu, Chunyi Zhou, Yuan Su, Zehao Jin, Qianli Ma, Leyi Qi, Yiming Wang, Zhe Ma, Yuwen Pu, Mengyao Du, Yuanyi Song, Enhao Huang, Zhihui Fu, Jun Wang, Jinfeng Li, Yuefeng Chen, Hui Xue, Yiming Li, Tianyu Du, et~al.
\newblock {Safety in Self-Evolving Agents: A Survey}.
\newblock arXiv preprint arXiv:2610.00093, 2026{\natexlab{c}}.

\bibitem[Chen et~al.(2025{\natexlab{b}})Chen, Liu, Zhang, Guo, Xu, Lin, Zang, Li, Wei, Zhou, Zhang, Yu, Fan, Huang, Wang, and Yu]{arxiv251025889}
Kang Chen, Zhihao Liu, Tonghe Zhang, Zhen Guo, Si~Xu, Hao Lin, Hongzhi Zang, Xiang Li, Bingwen Wei, Jiakai Zhou, Quanlu Zhang, Zhaofei Yu, Guoliang Fan, Tiejun Huang, Yu~Wang, and Chao Yu.
\newblock {$\pi_{\texttt{RL}}$: Online RL Fine-tuning for Flow-based Vision-Language-Action Models}.
\newblock arXiv preprint arXiv:2510.25889, 2025{\natexlab{b}}.

\bibitem[Chen et~al.(2026{\natexlab{d}})Chen, Wang, and Qu]{arxiv260707663}
Mingguang Chen, Licheng Wang, and Bo~Qu.
\newblock {Recursive Self-Improvement in AI: From Bounded Self-Refinement to Autonomous Research Loops}.
\newblock arXiv preprint arXiv:2607.07663, 2026{\natexlab{d}}.

\bibitem[Chen et~al.(2026{\natexlab{e}})Chen, Zheng, Yu, Huang, Sun, Goldberg, Wen, Abbeel, Shentu, Wu, and Schwager]{arxiv260610305}
Qianzhong Chen, Hau Zheng, Justin Yu, Suning Huang, Jiankai Sun, Ken Goldberg, Chuan Wen, Pieter Abbeel, Yide Shentu, Philipp Wu, and Mac Schwager.
\newblock {SARM2: Multi-Task Stage Aware Reward Modeling for Self Improving Robotic Manipulation}.
\newblock arXiv preprint arXiv:2606.10305, 2026{\natexlab{e}}.

\bibitem[Chen et~al.(2026{\natexlab{f}})Chen, Liu, Wu, Guo, Wang, Qian, Chen, Cao, Tang, and Zhang]{arxiv260824885}
Sixiang Chen, Jiaming Liu, Jixian Wu, Yichen Guo, Tinghao Wang, Siyuan Qian, Hao Chen, Jiajun Cao, Jian Tang, and Shanghang Zhang.
\newblock {Do Robotic World Models Really Follow Actions? Diagnosing and Aligning Action-Conditioned Generation for Policy Learning}.
\newblock arXiv preprint arXiv:2608.24885, 2026{\natexlab{f}}.

\bibitem[Chen et~al.(2025{\natexlab{c}})Chen, Chen, Chen, Cai, Liu, et~al.]{chen2025robotwin2}
Tianxing Chen, Zanxin Chen, Baijun Chen, Zijian Cai, Yibin Liu, et~al.
\newblock {RoboTwin 2.0: A Scalable Data Generator and Benchmark with Strong Domain Randomization for Robust Bimanual Robotic Manipulation}.
\newblock arXiv preprint arXiv:2506.18088, 2025{\natexlab{c}}.

\bibitem[Chen et~al.(2026{\natexlab{g}})Chen, Chen, Li, Tang, Su, Lu, Wan, Chen, Liu, Yan, Su, Dou, Wang, Zhang, Liu, Qin, Liang, Wu, Lin, Lin, Wang, He, Wu, Wu, Zhou, Lei, Yu, Ji, Jin, Lin, Li, Xiong, Xu, Li, Chai, Xie, Wang, Mu, Dong, Matusik, Ding, Ding, Luo, and Tomizuka]{arxiv260704434}
Tianxing Chen, Yue Chen, Zixuan Li, Junyuan Tang, Kailun Su, Haoran Lu, Weijie Wan, Baijun Chen, Songling Liu, Haowen Yan, Honghao Su, Zhiyang Dou, Kaixuan Wang, Dandan Zhang, Yunze Liu, Yan Qin, Qiwei Liang, Qiwei Wu, Zijian Lin, Wenwei Lin, Yuran Wang, Minghua He, Tianshu Wu, Ruihai Wu, Jingquan Zhou, Kai-Chong Lei, Haibao Yu, Yuanfeng Ji, Weiyang Jin, Guanyu Lin, Xiaofan Li, Qi~Xiong, Renjing Xu, Zhongyu Li, Wenhao Chai, Enze Xie, Ziwei Wang, Yao Mu, Hao Dong, Wojciech Matusik, Mingyu Ding, Wenbo Ding, Ping Luo, and Masayoshi Tomizuka.
\newblock {RoboDojo: A Unified Sim-and-Real Benchmark for Comprehensive Evaluation of Generalist Robot Manipulation Policies}.
\newblock arXiv preprint arXiv:2607.04434, 2026{\natexlab{g}}.

\bibitem[Chen et~al.(2026{\natexlab{h}})Chen, Bai, Cao, Zeng, Lin, Lin, Liang, Ma, Huang, and Shou]{arxiv260910522}
Yanzhe Chen, Zechen Bai, Zhijun Cao, Wenzheng Zeng, Kevin~Qinghong Lin, Yiqi Lin, Guoqiang Liang, Kevin~Yuchen Ma, Qiming Huang, and Mike~Zheng Shou.
\newblock {Show-Harness: Just a VLM Agent Can Play Robots}.
\newblock arXiv preprint arXiv:2609.10522, 2026{\natexlab{h}}.

\bibitem[Chen et~al.(2026{\natexlab{i}})Chen, Yang, Xu, Ma, He, Li, Wang, He, Wu, Liu, Liu, Huang, and Wang]{arxiv260805799}
Yixiang Chen, Jiabing Yang, Yuan Xu, Qisen Ma, Keji He, Peiyan Li, Kai Wang, Ziheng He, Xiangnan Wu, Jing Liu, Nianfeng Liu, Yan Huang, and Liang Wang.
\newblock {XEWorld: Can Action-Conditioned World Models Generalize to Unseen Robot Embodiments?}
\newblock arXiv preprint arXiv:2608.05799, 2026{\natexlab{i}}.

\bibitem[Chen et~al.(2025{\natexlab{d}})Chen, Tian, Liu, Zhou, Li, and Zhao]{arxiv250205450}
Yuhui Chen, Shuai Tian, Shugao Liu, Yingting Zhou, Haoran Li, and Dongbin Zhao.
\newblock {ConRFT: A Reinforced Fine-tuning Method for VLA Models via Consistency Policy}.
\newblock arXiv preprint arXiv:2502.05450, 2025{\natexlab{d}}.

\bibitem[Chen et~al.(2026{\natexlab{j}})Chen, Liu, Li, and Wang]{arxiv260909941}
Zengjue Chen, Peidong Liu, Jiawei Li, and Qi~Wang.
\newblock {HaWMPO: Hallucination-Aware World Model-based Policy Optimization for Generalist Robot Policy}.
\newblock arXiv preprint arXiv:2609.09941, 2026{\natexlab{j}}.

\bibitem[Chen and Zhu(2026)]{arxiv260115533}
Zhikang Chen and Tingting Zhu.
\newblock {From Generative Engines to Actionable Simulators: The Imperative of Physical Grounding in World Models}.
\newblock arXiv preprint arXiv:2601.15533, 2026.

\bibitem[Cheng et~al.(2025)Cheng, Li, Huang, Yin, and Zou]{arxiv251213399}
Sitao Cheng, Tianle Li, Xuhan Huang, Xunjian Yin, and Difan Zou.
\newblock {Dual-Evolving Reinforcement Learning for Automated Reward Design}.
\newblock arXiv preprint arXiv:2512.13399, 2025.

\bibitem[Chi et~al.(2023)Chi, Feng, Du, Xu, Cousineau, Burchfiel, and Song]{arxiv230304137}
Cheng Chi, Siyuan Feng, Yilun Du, Zhenjia Xu, Eric Cousineau, Benjamin Burchfiel, and Shuran Song.
\newblock {Diffusion Policy: Visuomotor Policy Learning via Action Diffusion}.
\newblock In \emph{Robotics: Science and Systems}, 2023.

\bibitem[Chi et~al.(2024)Chi, Xu, Pan, Cousineau, Burchfiel, Feng, Tedrake, and Song]{chi2024umi}
Cheng Chi, Zhenjia Xu, Chuer Pan, Eric Cousineau, Benjamin Burchfiel, Siyuan Feng, Russ Tedrake, and Shuran Song.
\newblock {Universal Manipulation Interface: In-The-Wild Robot Teaching Without In-The-Wild Robots}.
\newblock In \emph{Robotics: Science and Systems}, 2024.

\bibitem[Chua et~al.(2018)Chua, Calandra, McAllister, and Levine]{arxiv180512114}
Kurtland Chua, Roberto Calandra, Rowan McAllister, and Sergey Levine.
\newblock {Deep Reinforcement Learning in a Handful of Trials using Probabilistic Dynamics Models}.
\newblock In \emph{Advances in Neural Information Processing Systems}, 2018.

\bibitem[Clark et~al.(2026)Clark, Wang, Gao, Hong, Choi, Cutkosky, Hou, and Song]{arxiv260630988}
Jaden Clark, Changhao Wang, Yihuai Gao, Seongheon Hong, Hojung Choi, Mark Cutkosky, Yifan Hou, and Shuran Song.
\newblock {Multisensory Continual Learning: Adapting Pretrained Visuomotor Policies to Force}.
\newblock arXiv preprint arXiv:2606.30988, 2026.

\bibitem[Cohn et~al.(1996)Cohn, Ghahramani, and Jordan]{cohn1996active}
David~A. Cohn, Zoubin Ghahramani, and Michael~I. Jordan.
\newblock {Active Learning with Statistical Models}.
\newblock \emph{Journal of Artificial Intelligence Research}, 4:\penalty0 129--145, 1996.
\newblock \doi{10.1613/jair.295}.

\bibitem[Colas et~al.(2022)Colas, Karch, Sigaud, and Oudeyer]{colas2022autotelic}
Cédric Colas, Tristan Karch, Olivier Sigaud, and Pierre-Yves Oudeyer.
\newblock {Autotelic Agents with Intrinsically Motivated Goal-Conditioned Reinforcement Learning: A Short Survey}.
\newblock \emph{Journal of Artificial Intelligence Research}, 74:\penalty0 1159--1199, 2022.
\newblock \doi{10.1613/jair.1.13554}.

\bibitem[Cully et~al.(2015)Cully, Clune, Tarapore, and Mouret]{cully2015adaptation}
Antoine Cully, Jeff Clune, Danesh Tarapore, and Jean-Baptiste Mouret.
\newblock {Robots That Can Adapt Like Animals}.
\newblock \emph{Nature}, 521:\penalty0 503--507, 2015.
\newblock \doi{10.1038/nature14422}.

\bibitem[Dasari et~al.(2020)Dasari, Ebert, Tian, Nair, Bucher, Schmeckpeper, Singh, Levine, and Finn]{dasari2020robonet}
Sudeep Dasari, Frederik Ebert, Stephen Tian, Suraj Nair, Bernadette Bucher, Karl Schmeckpeper, Siddharth Singh, Sergey Levine, and Chelsea Finn.
\newblock {RoboNet: Large-Scale Multi-Robot Learning}.
\newblock In \emph{Proceedings of the Conference on Robot Learning}, volume 100 of \emph{Proceedings of Machine Learning Research}, pages 885--897, 2020.

\bibitem[Deisenroth and Rasmussen(2011)]{deisenroth2011pilco}
Marc~Peter Deisenroth and Carl~Edward Rasmussen.
\newblock {PILCO: A Model-Based and Data-Efficient Approach to Policy Search}.
\newblock In \emph{International Conference on Machine Learning}, pages 465--472, 2011.

\bibitem[Deng et~al.(2026)Deng, Liu, Zhang, Yan, Cao, and Wang]{arxiv260940306}
Haoyuan Deng, Jiebin Liu, Tengxiao Zhang, Langning Yan, Hongye Cao, and Ziwei Wang.
\newblock {DynaHarness: A Dynamic Physical Harness for Self-Evolving Robot Agents}.
\newblock arXiv preprint arXiv:2609.40306, 2026.

\bibitem[Ding et~al.(2026)Ding, Mi, Huang, Wang, Zhang, Hao, Chen, Li, Zheng, Guo, Wang, Li, Wu, Liu, and Cao]{arxiv260816590}
Xin Ding, Liang Mi, Mingzhe Huang, Zixuan Wang, Chao Zhang, Zixu Hao, Fu~Chen, Xiangyu Li, Yikai Zheng, Yaoyu Guo, Weijun Wang, Kun Li, Hao Wu, Yunxin Liu, and Ting Cao.
\newblock {Zetta $\zeta$: An Efficient Closed-Loop Embodied Harness for Self-Evolving Physical Intelligence}.
\newblock arXiv preprint arXiv:2608.16590, 2026.

\bibitem[Domberg and Schildbach(2026)]{arxiv260304029}
Fabian Domberg and Georg Schildbach.
\newblock {Self-adapting Robotic Agents through Online Continual Reinforcement Learning with World Model Feedback}.
\newblock arXiv preprint arXiv:2603.04029, 2026.

\bibitem[Dong et~al.(2026)Dong, Hung, Sadigh, and Finn]{arxiv260918207}
Perry Dong, Kuo-Han Hung, Dorsa Sadigh, and Chelsea Finn.
\newblock {Reinforcement Learning for Real-Time Vision-Language-Action Policies}.
\newblock arXiv preprint arXiv:2609.18207, 2026.

\bibitem[Du et~al.(2026)Du, Sun, Xu, Shah, Itkina, and Song]{arxiv260826545}
Maximilian Du, Zhanyi Sun, Chen Xu, Paarth Shah, Masha Itkina, and Shuran Song.
\newblock {Memory Anchors for Continual Robot Learning}.
\newblock arXiv preprint arXiv:2608.26545, 2026.

\bibitem[Du et~al.(2023)Du, Yang, Dai, Dai, Nachum, Tenenbaum, Schuurmans, and Abbeel]{du2023unipi}
Yilun Du, Mengjiao Yang, Bo~Dai, Hanjun Dai, Ofir Nachum, Joshua~B. Tenenbaum, Dale Schuurmans, and Pieter Abbeel.
\newblock {Learning Universal Policies via Text-Guided Video Generation}.
\newblock In \emph{Advances in Neural Information Processing Systems}, volume~36, 2023.
\newblock \doi{10.52202/075280-0403}.

\bibitem[Duan et~al.(2025)Duan, Pumacay, Kumar, Wang, Tian, Yuan, Krishna, Fox, Mandlekar, and Guo]{duan2025aha}
Jiafei Duan, Wilbert Pumacay, Nishanth Kumar, Yi~Ru Wang, Shulin Tian, Wentao Yuan, Ranjay Krishna, Dieter Fox, Ajay Mandlekar, and Yijie Guo.
\newblock {AHA: A Vision-Language-Model for Detecting and Reasoning Over Failures in Robotic Manipulation}.
\newblock In \emph{International Conference on Learning Representations}, pages 45493--45517, 2025.

\bibitem[Duan et~al.(2016)Duan, Schulman, Chen, Bartlett, Sutskever, and Abbeel]{duan2016rl2}
Yan Duan, John Schulman, Xi~Chen, Peter~L. Bartlett, Ilya Sutskever, and Pieter Abbeel.
\newblock {RL$^2$: Fast Reinforcement Learning via Slow Reinforcement Learning}.
\newblock arXiv preprint arXiv:1611.02779, 2016.

\bibitem[Duan et~al.(2026)Duan, Liu, Tang, Chen, Zhou, Liu, Xu, Wu, Chen, Zhou, Wang, Yu, Han, Zhu, Zhou, Lu, Zhou, Liu, Fang, Tian, Chen, Li, Zuo, Zhang, Yang, Wang, Qiu, He, Li, Zhou, Liu, Wen, Kang, Zhou, and Wu]{arxiv260911873}
Yi~Duan, Ying Liu, Zirui Tang, Haodong Chen, Jun Zhou, Yumou Liu, Bangrui Xu, Yukai Wu, Sidi Chen, Yuhan Zhou, Haoyu Wang, Xiaoyou Yu, Shaokun Han, Xuzhou Zhu, Le~Zhou, Bolin Lu, Wei Zhou, Jiachen Liu, Nuozhou Fang, Jiaxin Tian, Ruoyu Chen, Yuxuan Li, Kai Zuo, Kaiyan Zhang, Qianyu Yang, Zijie Wang, Jiantao Qiu, Conghui He, Guoliang Li, Bowen Zhou, Zhiyuan Liu, Zhoufutu Wen, Jihua Kang, Xuanhe Zhou, and Fan Wu.
\newblock {The Last AI Built by Humans: Toward Genuine Recursive Self-Improvement}.
\newblock arXiv preprint arXiv:2609.11873, 2026.

\bibitem[Dwork et~al.(2015)Dwork, Feldman, Hardt, Pitassi, Reingold, and Roth]{dwork2015holdout}
Cynthia Dwork, Vitaly Feldman, Moritz Hardt, Toniann Pitassi, Omer Reingold, and Aaron Roth.
\newblock {Generalization in Adaptive Data Analysis and Holdout Reuse}.
\newblock In \emph{Advances in Neural Information Processing Systems}, volume~28, 2015.

\bibitem[Ebert et~al.(2017)Ebert, Finn, Lee, and Levine]{ebert2017visualplanning}
Frederik Ebert, Chelsea Finn, Alex~X. Lee, and Sergey Levine.
\newblock {Self-Supervised Visual Planning with Temporal Skip Connections}.
\newblock In \emph{Proceedings of the 1st Annual Conference on Robot Learning}, volume~78 of \emph{Proceedings of Machine Learning Research}, pages 344--356, 2017.

\bibitem[Everitt et~al.(2021)Everitt, Hutter, Kumar, and Krakovna]{everitt2021tampering}
Tom Everitt, Marcus Hutter, Ramana Kumar, and Victoria Krakovna.
\newblock {Reward tampering problems and solutions in reinforcement learning: a causal influence diagram perspective}.
\newblock \emph{Synthese}, 198\penalty0 (S27):\penalty0 6435--6467, 2021.
\newblock \doi{10.1007/s11229-021-03141-4}.

\bibitem[Eysenbach et~al.(2018)Eysenbach, Gu, Ibarz, and Levine]{arxiv171106782}
Benjamin Eysenbach, Shixiang Gu, Julian Ibarz, and Sergey Levine.
\newblock {Leave no Trace: Learning to Reset for Safe and Autonomous Reinforcement Learning}.
\newblock In \emph{International Conference on Learning Representations}, 2018.

\bibitem[Fang et~al.(2026)Fang, Li, Tong, Liu, and Chalvatzaki]{arxiv260932069}
Yuan Fang, Zechu Li, Haolei Tong, Puze Liu, and Georgia Chalvatzaki.
\newblock {Find Something You Can't Do: Agentic Real-World Reinforcement Learning for Self-Improving VLA Models}.
\newblock arXiv preprint arXiv:2609.32069, 2026.

\bibitem[Farahmand et~al.(2017)Farahmand, Barreto, and Nikovski]{farahmand2017value}
Amir-Massoud Farahmand, Andre Barreto, and Daniel Nikovski.
\newblock {Value-Aware Loss Function for Model-based Reinforcement Learning}.
\newblock In \emph{Proceedings of the 20th International Conference on Artificial Intelligence and Statistics}, volume~54 of \emph{Proceedings of Machine Learning Research}, pages 1486--1494, 2017.

\bibitem[Fei et~al.(2025)Fei, Wang, Shi, Dai, Cai, Qian, Ji, He, Zhang, Fei, Fu, Gong, and Qiu]{arxiv251013626}
Senyu Fei, Siyin Wang, Junhao Shi, Zihao Dai, Jikun Cai, Pengfang Qian, Li~Ji, Xinzhe He, Shiduo Zhang, Zhaoye Fei, Jinlan Fu, Jingjing Gong, and Xipeng Qiu.
\newblock {LIBERO-Plus: In-depth Robustness Analysis of Vision-Language-Action Models}.
\newblock arXiv preprint arXiv:2510.13626, 2025.

\bibitem[Feng et~al.(2026)Feng, Wang, and Zhu]{arxiv260204411}
Tongtong Feng, Xin Wang, and Wenwu Zhu.
\newblock {Self-evolving Embodied AI}.
\newblock arXiv preprint arXiv:2602.04411, 2026.

\bibitem[Fernando et~al.(2024)Fernando, Banarse, Michalewski, Osindero, and Rocktäschel]{fernando2024promptbreeder}
Chrisantha Fernando, Dylan Banarse, Henryk Michalewski, Simon Osindero, and Tim Rocktäschel.
\newblock {Promptbreeder: Self-Referential Self-Improvement Via Prompt Evolution}.
\newblock In \emph{Proceedings of the 41st International Conference on Machine Learning}, volume 235 of \emph{Proceedings of Machine Learning Research}, pages 13481--13544, 2024.

\bibitem[Finn et~al.(2017)Finn, Abbeel, and Levine]{finn2017maml}
Chelsea Finn, Pieter Abbeel, and Sergey Levine.
\newblock {Model-Agnostic Meta-Learning for Fast Adaptation of Deep Networks}.
\newblock In \emph{Proceedings of the 34th International Conference on Machine Learning}, volume~70 of \emph{Proceedings of Machine Learning Research}, pages 1126--1135, 2017.

\bibitem[Gao et~al.(2026)Gao, Zhou, Zheng, Ge, Zhu, Wang, Zeng, Shangguan, Wu, He, Jia, Wu, and He]{arxiv260935432}
Hongcheng Gao, Jingjing Zhou, Zelin Zheng, Shijia Ge, Jay Zhu, Yazhe Wang, Jianshu Zeng, Xuan Shangguan, Di~Wu, Lingyu He, Zhiqi Jia, Sihang Wu, and Xiao He.
\newblock {Self-Evolving Coding Agents: From Digital Programs to Physical-World Intelligence}.
\newblock arXiv preprint arXiv:2609.35432, 2026.

\bibitem[Gao et~al.(2025)Gao, Geng, Hua, Hu, Juan, Liu, Liu, Qiu, Qi, Wu, Wang, Xiao, Zhou, Zhang, Zhang, Xiang, Fang, Zhao, Liu, Ren, Qian, Wang, Hu, Wang, Wu, Ji, and Wang]{arxiv250721046}
Huan-ang Gao, Jiayi Geng, Wenyue Hua, Mengkang Hu, Xinzhe Juan, Hongzhang Liu, Shilong Liu, Jiahao Qiu, Xuan Qi, Yiran Wu, Hongru Wang, Han Xiao, Yuhang Zhou, Shaokun Zhang, Jiayi Zhang, Jinyu Xiang, Yixiong Fang, Qiwen Zhao, Dongrui Liu, Qihan Ren, Cheng Qian, Zhenhailong Wang, Minda Hu, Huazheng Wang, Qingyun Wu, Heng Ji, and Mengdi Wang.
\newblock {A Survey of Self-Evolving Agents: What, When, How, and Where to Evolve on the Path to Artificial Super Intelligence}.
\newblock arXiv preprint arXiv:2507.21046, 2025.
\newblock Published in Transactions on Machine Learning Research (01/2026).

\bibitem[García and Fernández(2015)]{garcia2015safe}
Javier García and Fernando Fernández.
\newblock {A Comprehensive Survey on Safe Reinforcement Learning}.
\newblock \emph{Journal of Machine Learning Research}, 16:\penalty0 1437--1480, 2015.

\bibitem[Ge et~al.(2026)Ge, Zhou, Zeng, Wan, Wu, Zheng, Wang, Jia, Shangguan, Zhu, Liu, He, Wu, He, and Gao]{arxiv260939018}
Shijia Ge, Alex Zhou, Jianshu Zeng, Yexing Wan, Di~Wu, Zelin Zheng, Yazhe Wang, Zhiqi Jia, Xuan Shangguan, Jay Zhu, Yijun Liu, Lingyu He, Sihang Wu, Xiao He, and Hongcheng Gao.
\newblock {Make Code as Policy Great Again: Frontier Agents Write, Call, and Evolve Robot Tools}.
\newblock arXiv preprint arXiv:2609.39018, 2026.

\bibitem[Ghasemipour et~al.(2025)Ghasemipour, Wahid, Tompson, Sanketi, and Mordatch]{arxiv250915155}
Seyed Kamyar~Seyed Ghasemipour, Ayzaan Wahid, Jonathan Tompson, Pannag Sanketi, and Igor Mordatch.
\newblock {Self-Improving Embodied Foundation Models}.
\newblock In \emph{Advances in Neural Information Processing Systems}, 2025.
\newblock arXiv:2509.15155.

\bibitem[{GigaBrain Team} et~al.(2026){GigaBrain Team}, Wang, Li, Ni, Huang, Zhao, Li, Li, Lv, Liu, et~al.]{arxiv260212099}
{GigaBrain Team}, Boyuan Wang, Bohan Li, Chaojun Ni, Guan Huang, Guosheng Zhao, Hao Li, Jie Li, Jindi Lv, Jingyu Liu, et~al.
\newblock {GigaBrain-0.5M*: a VLA That Learns From World Model-Based Reinforcement Learning}.
\newblock arXiv preprint arXiv:2602.12099, 2026.

\bibitem[Giridhar et~al.(2026)Giridhar, Khandelwal, Collins, Georgiev, and Garg]{arxiv260821204}
Varun Giridhar, Anant Khandelwal, Jeremy~A. Collins, Ignat Georgiev, and Animesh Garg.
\newblock {Beyond Imitation: Self-Improving Robot Policies via Off-Policy Q-Planning}.
\newblock arXiv preprint arXiv:2608.21204, 2026.

\bibitem[Good(1966)]{good1966speculations}
Irving~John Good.
\newblock {Speculations Concerning the First Ultraintelligent Machine}.
\newblock In \emph{Advances in Computers}, volume~6, pages 31--88. Elsevier, 1966.
\newblock \doi{10.1016/s0065-2458(08)60418-0}.

\bibitem[Grimm et~al.(2020)Grimm, Barreto, Singh, and Silver]{grimm2020equivalence}
Christopher Grimm, André Barreto, Satinder Singh, and David Silver.
\newblock {The Value Equivalence Principle for Model-Based Reinforcement Learning}.
\newblock In \emph{Advances in Neural Information Processing Systems}, volume~33, 2020.

\bibitem[Guo et~al.(2025)Guo, Shi, Chen, and Finn]{arxiv251010125}
Yanjiang Guo, Lucy~Xiaoyang Shi, Jianyu Chen, and Chelsea Finn.
\newblock {Ctrl-World: A Controllable Generative World Model for Robot Manipulation}.
\newblock arXiv preprint arXiv:2510.10125, 2025.
\newblock Published at ICLR 2026.

\bibitem[Guo et~al.(2026)Guo, Lee, Shi, Chen, Liang, and Finn]{arxiv260212063}
Yanjiang Guo, Tony Lee, Lucy~Xiaoyang Shi, Jianyu Chen, Percy Liang, and Chelsea Finn.
\newblock {VLAW: Iterative Co-Improvement of Vision-Language-Action Policy and World Model}.
\newblock In \emph{International Conference on Machine Learning}, 2026.

\bibitem[Ha and Schmidhuber(2018)]{ha2018worldmodels}
David Ha and J{\"u}rgen Schmidhuber.
\newblock {Recurrent World Models Facilitate Policy Evolution}.
\newblock In \emph{Advances in Neural Information Processing Systems}, volume~31, 2018.

\bibitem[Hafner et~al.(2020)Hafner, Lillicrap, Ba, and Norouzi]{hafner2020dreamer}
Danijar Hafner, Timothy Lillicrap, Jimmy Ba, and Mohammad Norouzi.
\newblock {Dream to Control: Learning Behaviors by Latent Imagination}.
\newblock In \emph{International Conference on Learning Representations}, 2020.

\bibitem[Hafner et~al.(2025)Hafner, Pasukonis, Ba, and Lillicrap]{hafner2025dreamerv3}
Danijar Hafner, Jurgis Pasukonis, Jimmy Ba, and Timothy Lillicrap.
\newblock {Mastering Diverse Control Tasks Through World Models}.
\newblock \emph{Nature}, 640:\penalty0 647--653, 2025.
\newblock \doi{10.1038/s41586-025-08744-2}.

\bibitem[Han et~al.(2026)Han, Zeng, Zhang, Zhao, Wang, Jin, Li, Chen, Huang, Li, Shen, and Dong]{arxiv260920659}
Zimu Han, Yiming Zeng, Jiyao Zhang, Zihao Zhao, Yuanfei Wang, Yixiang Jin, Shiqi Li, Shuangben Chen, Wei Huang, Ruodai Li, Hui Shen, and Hao Dong.
\newblock {HIL-UMI: Bringing Human-in-the-Loop Post-Training of Vision-Language-Action Models to Universal Manipulation Interface}.
\newblock arXiv preprint arXiv:2609.20659, 2026.

\bibitem[Hansen et~al.(2024)Hansen, Su, and Wang]{hansen2024tdmpc2}
Nicklas Hansen, Hao Su, and Xiaolong Wang.
\newblock {TD-MPC2: Scalable, Robust World Models for Continuous Control}.
\newblock In \emph{International Conference on Learning Representations}, 2024.

\bibitem[Hao et~al.(2026)Hao, Syed, Ichnowski, and Schneider]{arxiv260701111}
Haoran Hao, Shahram~Najam Syed, Jeffrey Ichnowski, and Jeff Schneider.
\newblock {FAR: Failure-Aware Retry for Test-Time Recovery and Continual Policy Improvement}.
\newblock arXiv preprint arXiv:2607.01111, 2026.

\bibitem[He et~al.(2026{\natexlab{a}})He, Tian, Fan, Pi, Lu, Su, Shi, and Feng]{arxiv260728251}
Linxuan He, Yuying Tian, Lingxiang Fan, Jiaqi Pi, Yinqiao Lu, Shang Su, Mengkai Shi, and Shuo Feng.
\newblock {Policy-Level Recursive Self-Improvement for Embodied AI with a Criticality World Model}.
\newblock arXiv preprint arXiv:2607.28251, 2026{\natexlab{a}}.

\bibitem[He et~al.(2026{\natexlab{b}})He, Sun, Liang, Li, and Cong]{arxiv260714852}
Yao He, Gan Sun, Wenqi Liang, Fazeng Li, and Yang Cong.
\newblock {Towards Human-like Physical Intelligence: Lifelong Vision-Language-Action Learning for Robotic Manipulation}.
\newblock arXiv preprint arXiv:2607.14852, 2026{\natexlab{b}}.

\bibitem[He et~al.(2026{\natexlab{c}})He, Ai, Mu, Liu, Wan, Fu, Du, Christensen, and Su]{arxiv251101177}
Zihao He, Bo~Ai, Tongzhou Mu, Yulin Liu, Weikang Wan, Jiawei Fu, Yilun Du, Henrik~I. Christensen, and Hao Su.
\newblock {Scaling Cross-Embodiment World Models for Dexterous Manipulation}.
\newblock In \emph{IEEE/RSJ International Conference on Intelligent Robots and Systems}, 2026{\natexlab{c}}.
\newblock arXiv:2511.01177.

\bibitem[Hirose et~al.(2024)Hirose, Shah, Stachowicz, Sridhar, and Levine]{arxiv240300991}
Noriaki Hirose, Dhruv Shah, Kyle Stachowicz, Ajay Sridhar, and Sergey Levine.
\newblock {SELFI: Autonomous Self-Improvement with Reinforcement Learning for Social Navigation}.
\newblock arXiv preprint arXiv:2403.00991, 2024.

\bibitem[Hong et~al.(2026{\natexlab{a}})Hong, Zhu, Sun, Xu, He, Jin, Yuan, Zeng, and Chen]{arxiv260921358}
Yijun Hong, Jiarun Zhu, Xiaoquan Sun, Le~Xu, Qijun He, Xin Jin, Mingqi Yuan, Wenjun Zeng, and Jiayu Chen.
\newblock {FAN: Foresight Action Normalization for Continual Adaptation of Vision-Language-Action Models}.
\newblock arXiv preprint arXiv:2609.21358, 2026{\natexlab{a}}.

\bibitem[Hong et~al.(2026{\natexlab{b}})Hong, Yu, Li, and Choi]{arxiv251102239}
Youngjin Hong, Houjian Yu, Mingen Li, and Changhyun Choi.
\newblock {LACY: A Vision-Language Model-based Language-Action Cycle for Self-Improving Robotic Manipulation}.
\newblock In \emph{IEEE International Conference on Robotics and Automation (ICRA)}, 2026{\natexlab{b}}.

\bibitem[Hoque et~al.(2022{\natexlab{a}})Hoque, Balakrishna, Novoseller, Wilcox, Brown, and Goldberg]{hoque2021thriftydagger}
Ryan Hoque, Ashwin Balakrishna, Ellen Novoseller, Albert Wilcox, Daniel~S. Brown, and Ken Goldberg.
\newblock {ThriftyDAgger: Budget-Aware Novelty and Risk Gating for Interactive Imitation Learning}.
\newblock In \emph{Proceedings of the 5th Conference on Robot Learning}, volume 164 of \emph{Proceedings of Machine Learning Research}, pages 598--608, 2022{\natexlab{a}}.

\bibitem[Hoque et~al.(2022{\natexlab{b}})Hoque, Chen, Sharma, Dharmarajan, Thananjeyan, Abbeel, and Goldberg]{arxiv220614349}
Ryan Hoque, Lawrence~Yunliang Chen, Satvik Sharma, Karthik Dharmarajan, Brijen Thananjeyan, Pieter Abbeel, and Ken Goldberg.
\newblock {Fleet-DAgger: Interactive Robot Fleet Learning with Scalable Human Supervision}.
\newblock In \emph{Conference on Robot Learning}, 2022{\natexlab{b}}.
\newblock arXiv:2206.14349.

\bibitem[Hou et~al.(2026)Hou, Li, Jia, An, Guo, Leng, Geng, Ze, Harada, Torr, Mees, Pollefeys, Liu, Wu, Abbeel, Malik, Du, and Yang]{arxiv260500080}
Bohan Hou, Gen Li, Jindou Jia, Tuo An, Xinying Guo, Sicong Leng, Haoran Geng, Yanjie Ze, Tatsuya Harada, Philip Torr, Oier Mees, Marc Pollefeys, Zhuang Liu, Jiajun Wu, Pieter Abbeel, Jitendra Malik, Yilun Du, and Jianfei Yang.
\newblock {World Model for Robot Learning: A Comprehensive Survey}.
\newblock arXiv preprint arXiv:2605.00080, 2026.

\bibitem[Houthooft et~al.(2018)Houthooft, Chen, Isola, Stadie, Wolski, Ho, and Abbeel]{houthooft2018epg}
Rein Houthooft, Richard~Y. Chen, Phillip Isola, Bradly~C. Stadie, Filip Wolski, Jonathan Ho, and Pieter Abbeel.
\newblock {Evolved Policy Gradients}.
\newblock In \emph{Advances in Neural Information Processing Systems}, volume~31, 2018.

\bibitem[Hu et~al.(2026{\natexlab{a}})Hu, Huang, Lee, and Kanoulas]{arxiv260829772}
Dong Hu, Chao Huang, Carman K.~M. Lee, and Dimitrios Kanoulas.
\newblock {Self-Aware Active Learning Enables Continual Improvement in Autonomous Driving}.
\newblock arXiv preprint arXiv:2608.29772, 2026{\natexlab{a}}.

\bibitem[Hu et~al.(2025{\natexlab{a}})Hu, Hendrix, Farhadi, Kembhavi, Martin-Martin, Stone, Zeng, and Ehsani]{arxiv240916578}
Jiaheng Hu, Rose Hendrix, Ali Farhadi, Aniruddha Kembhavi, Roberto Martin-Martin, Peter Stone, Kuo-Hao Zeng, and Kiana Ehsani.
\newblock {FLaRe: Achieving Masterful and Adaptive Robot Policies with Large-Scale Reinforcement Learning Fine-Tuning}.
\newblock In \emph{IEEE International Conference on Robotics and Automation}, pages 3617--3624, 2025{\natexlab{a}}.
\newblock \doi{10.1109/ICRA55743.2025.11127934}.

\bibitem[Hu et~al.(2026{\natexlab{b}})Hu, Hu, Hong, Chen, and Liu]{arxiv261012369}
Kairui Hu, Siyuan Hu, Fangzhou Hong, Zhaoxi Chen, and Ziwei Liu.
\newblock {Embodied Turing Machines: Stateful Code for Robot Recursive Self-Improvement}.
\newblock arXiv preprint arXiv:2610.12369, 2026{\natexlab{b}}.

\bibitem[Hu et~al.(2025{\natexlab{b}})Hu, Lu, and Clune]{hu2025adas}
Shengran Hu, Cong Lu, and Jeff Clune.
\newblock {Automated Design of Agentic Systems}.
\newblock In \emph{International Conference on Learning Representations}, 2025{\natexlab{b}}.

\bibitem[Huang et~al.(2023)Huang, Xia, Xiao, Chan, Liang, Florence, Zeng, Tompson, Mordatch, Chebotar, Sermanet, Jackson, Brown, Luu, Levine, Hausman, and Ichter]{huang2023inner}
Wenlong Huang, Fei Xia, Ted Xiao, Harris Chan, Jacky Liang, Pete Florence, Andy Zeng, Jonathan Tompson, Igor Mordatch, Yevgen Chebotar, Pierre Sermanet, Tomas Jackson, Noah Brown, Linda Luu, Sergey Levine, Karol Hausman, and Brian Ichter.
\newblock {Inner Monologue: Embodied Reasoning through Planning with Language Models}.
\newblock In \emph{Proceedings of the 6th Conference on Robot Learning}, volume 205 of \emph{Proceedings of Machine Learning Research}, pages 1769--1782, 2023.
\newblock arXiv:2207.05608.

\bibitem[Hutter et~al.(2019)Hutter, Kotthoff, and Vanschoren]{hutter2019automl}
Frank Hutter, Lars Kotthoff, and Joaquin Vanschoren, editors.
\newblock \emph{{Automated Machine Learning: Methods, Systems, Challenges}}.
\newblock The Springer Series on Challenges in Machine Learning. Springer, Cham, 2019.
\newblock \doi{10.1007/978-3-030-05318-5}.

\bibitem[Iacob et~al.(2026)Iacob, Jovanovi{\'c}, Shen, Burkhardt, Kurmanji, Tastan, Sani, Venanzi, Nie, Odonnat, Cao, Marino, Qiu, Antonova, and Lane]{arxiv260626294}
Alex Iacob, Andrej Jovanovi{\'c}, William~F. Shen, Daniel Burkhardt, Meghdad Kurmanji, Nurbek Tastan, Lorenzo Sani, Niccol{\`o} Alberto~Elia Venanzi, Jiayi Nie, Ambroise Odonnat, Zeyu Cao, Bill Marino, Xinchi Qiu, Rika Antonova, and Nicholas~D. Lane.
\newblock {The Red Queen G{\"o}del Machine: Co-Evolving Agents and Their Evaluators}.
\newblock arXiv preprint arXiv:2606.26294, 2026.

\bibitem[Ichter et~al.(2023)Ichter, Brohan, Chebotar, Finn, Hausman, et~al.]{ahn2023saycan}
Brian Ichter, Anthony Brohan, Yevgen Chebotar, Chelsea Finn, Karol Hausman, et~al.
\newblock {Do As I Can, Not As I Say: Grounding Language in Robotic Affordances}.
\newblock In \emph{Proceedings of the 6th Conference on Robot Learning (CoRL 2022)}, volume 205 of \emph{Proceedings of Machine Learning Research}, pages 287--318, 2023.

\bibitem[Irie et~al.(2022)Irie, Schlag, Csord{\'a}s, and Schmidhuber]{irie2022modern}
Kazuki Irie, Imanol Schlag, R{\'o}bert Csord{\'a}s, and J{\"u}rgen Schmidhuber.
\newblock {A Modern Self-Referential Weight Matrix That Learns to Modify Itself}.
\newblock In \emph{Proceedings of the International Conference on Machine Learning (ICML)}, 2022.

\bibitem[Jang et~al.(2025)Jang, Ye, Lin, Xiang, Bjorck, Fang, Hu, Huang, Kundalia, Lin, et~al.]{jang2025dreamgen}
Joel Jang, Seonghyeon Ye, Zongyu Lin, Jiannan Xiang, Johan Bjorck, Yu~Fang, Fengyuan Hu, Spencer Huang, Kaushil Kundalia, Yen-Chen Lin, et~al.
\newblock {DreamGen: Unlocking Generalization in Robot Learning through Video World Models}.
\newblock In \emph{Proceedings of the 9th Conference on Robot Learning}, volume 305 of \emph{Proceedings of Machine Learning Research}, pages 5170--5194, 2025.

\bibitem[Janner et~al.(2019)Janner, Fu, Zhang, and Levine]{arxiv190608253}
Michael Janner, Justin Fu, Marvin Zhang, and Sergey Levine.
\newblock {When to Trust Your Model: Model-Based Policy Optimization}.
\newblock In \emph{Advances in Neural Information Processing Systems}, 2019.

\bibitem[Jena et~al.(2026)Jena, Wanaskar, Jain, Chadha, Sharma, and Das]{arxiv260801851}
Gaytri Jena, Kapil Wanaskar, Vinija Jain, Aman Chadha, Vasu Sharma, and Amitava Das.
\newblock {Weights or Skills? A Survey of Robot-Learning Techniques: from Action-Predicting Weights to Robots that Write their Own Skills}.
\newblock arXiv preprint arXiv:2608.01851, 2026.

\bibitem[Jia et~al.(2026)Jia, Lin, Zhang, Zhang, Liu, and Jiang]{arxiv260912541}
Mengzhao Jia, Yang Lin, Xixin Zhang, Zhihan Zhang, Xiaobai Liu, and Meng Jiang.
\newblock {Agent as Policy for Robotic Manipulation}.
\newblock arXiv preprint arXiv:2609.12541, 2026.

\bibitem[Jiang et~al.(2026)Jiang, Zhou, Jiang, Huang, Wei, Chen, Zhou, Guo, Lin, Zhang, Wang, Li, Yu, and Zhao]{arxiv260213977}
Zhennan Jiang, Shangqing Zhou, Yutong Jiang, Zefang Huang, Mingjie Wei, Yuhui Chen, Tianxing Zhou, Zhen Guo, Hao Lin, Quanlu Zhang, Yu~Wang, Haoran Li, Chao Yu, and Dongbin Zhao.
\newblock {WoVR: World Models as Reliable Simulators for Post-Training VLA Policies with RL}.
\newblock arXiv preprint arXiv:2602.13977, 2026.

\bibitem[Kalashnikov et~al.(2018)Kalashnikov, Irpan, Pastor, Ibarz, Herzog, Jang, Quillen, Holly, Kalakrishnan, Vanhoucke, and Levine]{kalashnikov2018qtopt}
Dmitry Kalashnikov, Alex Irpan, Peter Pastor, Julian Ibarz, Alexander Herzog, Eric Jang, Deirdre Quillen, Ethan Holly, Mrinal Kalakrishnan, Vincent Vanhoucke, and Sergey Levine.
\newblock {Scalable Deep Reinforcement Learning for Vision-Based Robotic Manipulation}.
\newblock In \emph{Proceedings of the 2nd Conference on Robot Learning}, volume~87 of \emph{Proceedings of Machine Learning Research}, pages 651--673, 2018.

\bibitem[Kalashnikov et~al.(2022)Kalashnikov, Varley, Chebotar, Swanson, Jonschkowski, Finn, Levine, and Hausman]{kalashnikov2022mtopt}
Dmitry Kalashnikov, Jake Varley, Yevgen Chebotar, Benjamin Swanson, Rico Jonschkowski, Chelsea Finn, Sergey Levine, and Karol Hausman.
\newblock {Scaling Up Multi-Task Robotic Reinforcement Learning}.
\newblock In \emph{Proceedings of the 5th Conference on Robot Learning}, volume 164 of \emph{Proceedings of Machine Learning Research}, pages 557--575, 2022.

\bibitem[Kelly et~al.(2019)Kelly, Sidrane, Driggs-Campbell, and Kochenderfer]{kelly2019hgdagger}
Michael Kelly, Chelsea Sidrane, Katherine Driggs-Campbell, and Mykel~J. Kochenderfer.
\newblock {HG-DAgger: Interactive Imitation Learning with Human Experts}.
\newblock In \emph{IEEE International Conference on Robotics and Automation}, pages 8077--8083, 2019.
\newblock \doi{10.1109/ICRA.2019.8793698}.

\bibitem[Khandelwal and Gupta(2026)]{arxiv260327416}
Nimesh Khandelwal and Shakti~S. Gupta.
\newblock {Agent-Driven Autonomous Reinforcement Learning Research: Iterative Policy Improvement for Quadruped Locomotion}.
\newblock arXiv preprint arXiv:2603.27416, 2026.

\bibitem[Khazatsky et~al.(2024)Khazatsky, Pertsch, Nair, Balakrishna, Dasari, Karamcheti, Nasiriany, Srirama, Chen, Ellis, et~al.]{arxiv240312945}
Alexander Khazatsky, Karl Pertsch, Suraj Nair, Ashwin Balakrishna, Sudeep Dasari, Siddharth Karamcheti, Soroush Nasiriany, Mohan~Kumar Srirama, Lawrence~Yunliang Chen, Kirsty Ellis, et~al.
\newblock {DROID: A Large-Scale In-The-Wild Robot Manipulation Dataset}.
\newblock In \emph{Robotics: Science and Systems}, 2024.

\bibitem[Kim et~al.(2026)Kim, Lee, Narang, and Guo]{arxiv261018596}
Changyeon Kim, Kimin Lee, Yashraj Narang, and Yijie Guo.
\newblock {ACCRUE: Continual Off-Policy Reinforcement Learning for Robot Foundation Models}.
\newblock arXiv preprint arXiv:2610.18596, 2026.

\bibitem[Kim et~al.(2024)Kim, Pertsch, Karamcheti, Xiao, Balakrishna, Nair, Rafailov, Foster, Sanketi, Vuong, Kollar, Burchfiel, Tedrake, Sadigh, Levine, Liang, and Finn]{arxiv240609246}
Moo~Jin Kim, Karl Pertsch, Siddharth Karamcheti, Ted Xiao, Ashwin Balakrishna, Suraj Nair, Rafael Rafailov, Ethan Foster, Pannag Sanketi, Quan Vuong, Thomas Kollar, Benjamin Burchfiel, Russ Tedrake, Dorsa Sadigh, Sergey Levine, Percy Liang, and Chelsea Finn.
\newblock {OpenVLA: An Open-Source Vision-Language-Action Model}.
\newblock In \emph{Conference on Robot Learning}, 2024.

\bibitem[Kober et~al.(2013)Kober, Bagnell, and Peters]{kober2013rl}
Jens Kober, J.~Andrew Bagnell, and Jan Peters.
\newblock {Reinforcement Learning in Robotics: A Survey}.
\newblock \emph{The International Journal of Robotics Research}, 32\penalty0 (11):\penalty0 1238--1274, 2013.
\newblock \doi{10.1177/0278364913495721}.

\bibitem[Krantz et~al.(2020)Krantz, Wijmans, Majumdar, Batra, and Lee]{krantz2020vlnce}
Jacob Krantz, Erik Wijmans, Arjun Majumdar, Dhruv Batra, and Stefan Lee.
\newblock {Beyond the Nav-Graph: Vision-and-Language Navigation in Continuous Environments}.
\newblock In \emph{European Conference on Computer Vision}, 2020.

\bibitem[Kress-Gazit et~al.(2024)Kress-Gazit, Hashimoto, Kuppuswamy, Shah, Horgan, Richardson, Feng, and Burchfiel]{arxiv240909491}
Hadas Kress-Gazit, Kunimatsu Hashimoto, Naveen Kuppuswamy, Paarth Shah, Phoebe Horgan, Gordon Richardson, Siyuan Feng, and Benjamin Burchfiel.
\newblock {Robot Learning as an Empirical Science: Best Practices for Policy Evaluation}.
\newblock arXiv preprint arXiv:2409.09491, 2024.

\bibitem[Kumar et~al.(2021)Kumar, Fu, Pathak, and Malik]{arxiv210704034}
Ashish Kumar, Zipeng Fu, Deepak Pathak, and Jitendra Malik.
\newblock {RMA: Rapid Motor Adaptation for Legged Robots}.
\newblock In \emph{Robotics: Science and Systems}, 2021.
\newblock arXiv:2107.04034.

\bibitem[Kumar et~al.(2024)Kumar, Silver, McClinton, Zhao, Proulx, Lozano-P{\'e}rez, Kaelbling, and Barry]{arxiv240215025}
Nishanth Kumar, Tom Silver, Willie McClinton, Linfeng Zhao, Stephen Proulx, Tom{\'a}s Lozano-P{\'e}rez, Leslie~Pack Kaelbling, and Jennifer~L. Barry.
\newblock {Practice Makes Perfect: Planning to Learn Skill Parameter Policies}.
\newblock In \emph{Proceedings of Robotics: Science and Systems (RSS)}, 2024.
\newblock \doi{10.15607/RSS.2024.XX.040}.

\bibitem[Kuroki and Tang(2026)]{arxiv260919906}
So~Kuroki and Yujin Tang.
\newblock {Learning and Transferring Closed-Loop Robot Software}.
\newblock arXiv preprint arXiv:2609.19906, 2026.

\bibitem[Kwon et~al.(2023)Kwon, Xie, Bullard, and Sadigh]{kwon2023reward}
Minae Kwon, Sang~Michael Xie, Kalesha Bullard, and Dorsa Sadigh.
\newblock {Reward Design with Language Models}.
\newblock In \emph{International Conference on Learning Representations}, 2023.

\bibitem[Lambert et~al.(2020)Lambert, Amos, Yadan, and Calandra]{lambert2020mismatch}
Nathan Lambert, Brandon Amos, Omry Yadan, and Roberto Calandra.
\newblock {Objective Mismatch in Model-based Reinforcement Learning}.
\newblock In \emph{Proceedings of the 2nd Conference on Learning for Dynamics and Control}, volume 120 of \emph{Proceedings of Machine Learning Research}, pages 761--770, 2020.

\bibitem[Lang et~al.(2026)Lang, Lu, Sang, Luo, Chen, Li, Yu, Shi, and Wang]{arxiv260938905}
Haoran Lang, Haotao Lu, Shiyu Sang, Haoyang Luo, Guo Chen, Qun Li, Jingyi Yu, Ye~Shi, and Jingya Wang.
\newblock {EmbodiRSI: Recursive Self-Improvement for Data-Efficient Robot Adaptation}.
\newblock arXiv preprint arXiv:2609.38905, 2026.

\bibitem[Laroche et~al.(2019)Laroche, Trichelair, and des Combes]{laroche2019spibb}
Romain Laroche, Paul Trichelair, and R{\'e}mi~Tachet des Combes.
\newblock {Safe Policy Improvement with Baseline Bootstrapping}.
\newblock In \emph{Proceedings of the 36th International Conference on Machine Learning}, volume~97 of \emph{Proceedings of Machine Learning Research}, 2019.

\bibitem[Lee et~al.(2026{\natexlab{a}})Lee, Xu, Seely, Lee, Sojoudi, Zaharia, and Tang]{arxiv261012176}
Hyunin Lee, Jinglue Xu, Jeffrey Seely, Donghyun Lee, Somayeh Sojoudi, Matei Zaharia, and Yujin Tang.
\newblock {Recursive Self-Improvement through Multi-Agent Self-Supervision}.
\newblock arXiv preprint arXiv:2610.12176, 2026{\natexlab{a}}.

\bibitem[Lee et~al.(2021)Lee, Smith, and Abbeel]{arxiv210605091}
Kimin Lee, Laura Smith, and Pieter Abbeel.
\newblock {PEBBLE: Feedback-Efficient Interactive Reinforcement Learning via Relabeling Experience and Unsupervised Pre-training}.
\newblock In \emph{International Conference on Machine Learning}, 2021.

\bibitem[Lee et~al.(2026{\natexlab{b}})Lee, Wagenmaker, Pertsch, Liang, Levine, and Finn]{arxiv260100675}
Tony Lee, Andrew Wagenmaker, Karl Pertsch, Percy Liang, Sergey Levine, and Chelsea Finn.
\newblock {RoboReward: General-Purpose Vision-Language Reward Models for Robotics}.
\newblock arXiv preprint arXiv:2601.00675, 2026{\natexlab{b}}.

\bibitem[Lei et~al.(2025{\natexlab{a}})Lei, Li, Yu, Wei, Guo, Jiang, Wang, Liang, and Xu]{arxiv251014830}
Kun Lei, Huanyu Li, Dongjie Yu, Zhenyu Wei, Lingxiao Guo, Zhennan Jiang, Ziyu Wang, Shiyu Liang, and Huazhe Xu.
\newblock {RL-100: Performant Robotic Manipulation with Real-World Reinforcement Learning}.
\newblock arXiv preprint arXiv:2510.14830, 2025{\natexlab{a}}.

\bibitem[Lei et~al.(2025{\natexlab{b}})Lei, Cai, Yang, Wu, Ren, Cui, Tan, Hong, Hu, Zhu, Jiang, Wang, Tan, Wan, Li, Li, Cui, Zhao, and Han]{arxiv250801415}
Mingcong Lei, Honghao Cai, Yuyuan Yang, Yimou Wu, Jinke Ren, Zezhou Cui, Liangchen Tan, Junkun Hong, Gehan Hu, Shuangyu Zhu, Shaohan Jiang, Ge~Wang, Junyuan Tan, Zhenglin Wan, Zheng Li, Zhen Li, Shuguang Cui, Yiming Zhao, and Yatong Han.
\newblock {RoboMemory: A Brain-inspired Multi-memory Agentic Framework for Interactive Environmental Learning in Physical Embodied Systems}.
\newblock arXiv preprint arXiv:2508.01415, 2025{\natexlab{b}}.

\bibitem[Lei et~al.(2026)Lei, Liu, Zhang, Liu, Cai, Chen, Wu, Wang, Li, Wen, Zhao, Zhang, Lian, and Chen]{arxiv260121570}
Zixing Lei, Genjia Liu, Yuanshuo Zhang, Qipeng Liu, Yuzhu Cai, Sixiang Chen, Jixian Wu, Yunhong Wang, Weixin Li, Chuan Wen, Bo~Zhao, Shanghang Zhang, Wenzhao Lian, and Siheng Chen.
\newblock {From Digital to Physical: Digital Agents as Autonomous Coaches for Physical Intelligence}.
\newblock arXiv preprint arXiv:2601.21570, 2026.

\bibitem[Li et~al.(2026{\natexlab{a}})Li, Mishra, Liu, Tao, Kumar, Gray, Wickramarachchi, Francis, Scherer, and Silver]{arxiv260618328}
Bowen Li, Mayank Mishra, Y.~Isabel Liu, Stone Tao, Nishanth Kumar, Alexander~G. Gray, Ruwan Wickramarachchi, Jonathan Francis, Sebastian Scherer, and Tom Silver.
\newblock {Recover, Discover, Plan: Learning Skills and Concepts from Robot Failures}.
\newblock arXiv preprint arXiv:2606.18328, 2026{\natexlab{a}}.

\bibitem[Li et~al.(2025{\natexlab{a}})Li, Michel, Pan, Liu, Roberts, and Xie]{arxiv250309595}
Chenyu Li, Oscar Michel, Xichen Pan, Sainan Liu, Mike Roberts, and Saining Xie.
\newblock {PISA Experiments: Exploring Physics Post-Training for Video Diffusion Models by Watching Stuff Drop}.
\newblock In \emph{International Conference on Machine Learning}, volume 267 of \emph{Proceedings of Machine Learning Research}, pages 35685--35709, 2025{\natexlab{a}}.

\bibitem[Li et~al.(2026{\natexlab{b}})Li, You, Su, and Guibas]{arxiv260220323}
Haoyang Li, Yang You, Hao Su, and Leonidas Guibas.
\newblock {PhysMem: Scaling Test-Time Memory for Embodied Physical Reasoning}.
\newblock arXiv preprint arXiv:2602.20323, 2026{\natexlab{b}}.

\bibitem[Li et~al.(2026{\natexlab{c}})Li, Zuo, Yu, Zhang, Yang, Zhang, Zhu, Zhang, Chen, Cui, et~al.]{arxiv250909674}
Haozhan Li, Yuxin Zuo, Jiale Yu, Yuhao Zhang, Zhaohui Yang, Kaiyan Zhang, Xuekai Zhu, Yuchen Zhang, Tianxing Chen, Ganqu Cui, et~al.
\newblock {SimpleVLA-RL: Scaling VLA Training via Reinforcement Learning}.
\newblock In \emph{International Conference on Learning Representations}, 2026{\natexlab{c}}.

\bibitem[Li et~al.(2026{\natexlab{d}})Li, Xu, Xu, Du, Yan, Wang, Yan, Yang, Liu, Shan, and Yao]{arxiv260821899}
Houlin Li, Minghui Xu, Guo Xu, Xuan Du, Xiaohan Yan, Chun Wang, Yuxiang Yan, Shukai Yang, Yongcheng Liu, Wei Shan, and Maoqing Yao.
\newblock {CIDER: Continual Interactive Distillation for Embodied Reinforcement Learning}.
\newblock arXiv preprint arXiv:2608.21899, 2026{\natexlab{d}}.

\bibitem[Li et~al.(2026{\natexlab{e}})Li, Zhou, Zhu, Chen, Wang, Wang, Hu, Yu, Jiang, Su, Ma, He, Shen, Yang, Ren, Yao, Wang, and Mu]{arxiv260311558}
Ruiying Li, Yunlang Zhou, YuYao Zhu, Kylin Chen, Jingyuan Wang, Sukai Wang, Kongtao Hu, Minhui Yu, Bowen Jiang, Zhan Su, Jiayao Ma, Xin He, Yongjian Shen, Yang Yang, Guanghui Ren, Maoqing Yao, Wenhao Wang, and Yao Mu.
\newblock {RoboClaw: An Agentic Framework for Scalable Long-Horizon Robotic Tasks}.
\newblock arXiv preprint arXiv:2603.11558, 2026{\natexlab{e}}.

\bibitem[Li et~al.(2026{\natexlab{f}})Li, Gu, Gong, Wang, Zhou, Wang, Zhu, Jiang, Li, Wu, Zhang, Qiao, Zhang, and Wang]{arxiv261015782}
Songze Li, Haoming Gu, Ziyang Gong, Zhitong Wang, Gengze Zhou, Zun Wang, Yuhan Zhu, Tianxiang Jiang, Juanzi Li, Qi~Wu, Weinan Zhang, Yu~Qiao, Wenwei Zhang, and Yi~Wang.
\newblock {Data-Centric Recursive Self-Improvement for Navigation via Policy-Synchronized Language-Action Dual Bootstrapping}.
\newblock arXiv preprint arXiv:2610.15782, 2026{\natexlab{f}}.

\bibitem[Li et~al.(2026{\natexlab{g}})Li, Chi, Zheng, Liu, Wang, Gao, Jiang, Hong, Xiao, Yang, and Na]{arxiv261018867}
Wenyi Li, Yizhe Chi, Youjie Zheng, Chunyu Liu, Xiaoqiu Wang, Mingju Gao, Zheng Jiang, Deyao Hong, Calvin Xiao, Kaisen Yang, and Qinhuai Na.
\newblock {EmbodiedEvo: Self-Evolving Coding Agent For Physical Intelligence}.
\newblock arXiv preprint arXiv:2610.18867, 2026{\natexlab{g}}.

\bibitem[Li et~al.(2026{\natexlab{h}})Li, Qin, Zhou, Wang, Guo, and Zhang]{arxiv261016517}
Wenzhan Li, Yiran Qin, Heng Zhou, Huirui Wang, Yulan Guo, and Ruimao Zhang.
\newblock {Trust-Calibrated VLA Policy Post-Training with Evolving Imagination}.
\newblock arXiv preprint arXiv:2610.16517, 2026{\natexlab{h}}.

\bibitem[Li et~al.(2024)Li, Hsu, Gu, Pertsch, Mees, Walke, Fu, Lunawat, Sieh, Kirmani, Levine, Wu, Finn, Su, Vuong, and Xiao]{arxiv240505941}
Xuanlin Li, Kyle Hsu, Jiayuan Gu, Karl Pertsch, Oier Mees, Homer~Rich Walke, Chuyuan Fu, Ishikaa Lunawat, Isabel Sieh, Sean Kirmani, Sergey Levine, Jiajun Wu, Chelsea Finn, Hao Su, Quan Vuong, and Ted Xiao.
\newblock {Evaluating Real-World Robot Manipulation Policies in Simulation}.
\newblock In \emph{Conference on Robot Learning}, 2024.

\bibitem[Li et~al.(2025{\natexlab{b}})Li, Ma, Xu, Cui, Cui, Han, Huang, Kong, Liu, Niu, Peng, Qiao, Ren, Shi, Su, Tian, Xiao, Zhang, Zheng, Li, and Wu]{arxiv251201801}
Yunfei Li, Xiao Ma, Jiafeng Xu, Yu~Cui, Zhongren Cui, Zhigang Han, Liqun Huang, Tao Kong, Yuxiao Liu, Hao Niu, Wanli Peng, Jingchao Qiao, Zeyu Ren, Haixin Shi, Zhi Su, Jiawen Tian, Yuyang Xiao, Shenyu Zhang, Liwei Zheng, Hang Li, and Yonghui Wu.
\newblock {GR-RL: Going Dexterous and Precise for Long-Horizon Robotic Manipulation}.
\newblock arXiv preprint arXiv:2512.01801, 2025{\natexlab{b}}.

\bibitem[Li et~al.(2026{\natexlab{i}})Li, Jin, Liu, Liu, Prasad, D'Eramo, and Chalvatzaki]{arxiv260608610}
Zechu Li, Yufeng Jin, Xiaoyang Liu, Puze Liu, Vignesh Prasad, Carlo D'Eramo, and Georgia Chalvatzaki.
\newblock {HARBOR: A Harness Framework for Agentic Robot Reinforcement Learning}.
\newblock arXiv preprint arXiv:2606.08610, 2026{\natexlab{i}}.

\bibitem[Li et~al.(2026{\natexlab{j}})Li, Yu, Zhang, Niu, Qi, Zheng, Chen, Liu, Zhang, Wang, and Yi]{arxiv261026848}
Zekai Li, Jiayi Yu, Jinlu Zhang, Luyang Niu, Zekun Qi, Jun Zheng, Xuchuan Chen, Dairu Liu, Yintianrun Zhang, He~Wang, and Li~Yi.
\newblock {Where is Physical RSI capability from? An Observational Scaling Law of LLM-as-policy}.
\newblock arXiv preprint arXiv:2610.26848, 2026{\natexlab{j}}.

\bibitem[Li et~al.(2026{\natexlab{k}})Li, Zhang, Huang, Zhou, Li, Wang, Chang, Wei, Zhang, Chen, Zhou, Li, and Cheng]{arxiv260929394}
Zexi Li, Yehang Zhang, Haojian Huang, Bohan Zhou, Wenqian Li, Chenxu Wang, Yifan Chang, Yangkai Wei, Tianyi Zhang, Ying-Cong Chen, Kaiwen Zhou, Yinchuan Li, and James Cheng.
\newblock {RACaP: Agentic Reasoning, Acting, and Coding as Policies for Evolvable Robot Learning}.
\newblock arXiv preprint arXiv:2609.29394, 2026{\natexlab{k}}.

\bibitem[Li et~al.(2026{\natexlab{l}})Li, Zhang, Ren, Huang, Lin, Wang, Gao, Guo, Karray, Liu, Russell, and Gong]{arxiv260932698}
Ziwen Li, Hanlue Zhang, Zhenyang Ren, Tianyu Huang, Runqi Lin, Haoyu Wang, Zhengqing Gao, Yandong Guo, Fakhri Karray, Tongliang Liu, Chris Russell, and Mingming Gong.
\newblock {SEES: A Self-Evolving Embodied System via Failure-Guided VLA Policy Adaptation}.
\newblock arXiv preprint arXiv:2609.32698, 2026{\natexlab{l}}.

\bibitem[Liang et~al.(2023)Liang, Huang, Xia, Xu, Hausman, Ichter, Florence, and Zeng]{liang2023code}
Jacky Liang, Wenlong Huang, Fei Xia, Peng Xu, Karol Hausman, Brian Ichter, Pete Florence, and Andy Zeng.
\newblock {Code as Policies: Language Model Programs for Embodied Control}.
\newblock In \emph{IEEE International Conference on Robotics and Automation}, pages 9493--9500, 2023.
\newblock \doi{10.1109/ICRA48891.2023.10160591}.
\newblock arXiv:2209.07753.

\bibitem[Liang et~al.(2024)Liang, Xia, Yu, Zeng, Attarian, Villalonga, Bennice, Bewley, Dostmohamed, Fu, Gileadi, Giustina, Gopalakrishnan, Hasenclever, Humplik, Hsu, Joshi, Jyenis, Kew, Kirmani, Lee, Lee, Michaely, Moore, Oslund, Rao, Ren, Tabanpour, Vuong, Wahid, Xiao, Xu, Zhuang, Xu, Frey, Caluwaerts, Zhang, Ichter, Tompson, Takayama, Vanhoucke, Shafran, Mataric, Sadigh, Heess, Rao, Stewart, Tan, and Parada]{arxiv240211450}
Jacky Liang, Fei Xia, Wenhao Yu, Andy Zeng, Maria Attarian, Maria~Bauza Villalonga, Matthew Bennice, Alex Bewley, Adil Dostmohamed, Chuyuan Fu, Nimrod Gileadi, Marissa Giustina, Keerthana Gopalakrishnan, Leonard Hasenclever, Jan Humplik, Jasmine Hsu, Nikhil~J. Joshi, Ben Jyenis, J.~Chase Kew, Sean Kirmani, Tsang-Wei~Edward Lee, Kuang-Huei Lee, Assaf~Hurwitz Michaely, Joss Moore, Kenneth Oslund, Dushyant Rao, Allen~Z. Ren, Baruch Tabanpour, Quan Vuong, Ayzaan Wahid, Ted Xiao, Ying Xu, Vincent Zhuang, Peng Xu, Erik Frey, Ken Caluwaerts, Tingnan Zhang, Brian Ichter, Jonathan Tompson, Leila Takayama, Vincent Vanhoucke, Izhak Shafran, Maja Mataric, Dorsa Sadigh, Nicolas Heess, Kanishka Rao, Nik Stewart, Jie Tan, and Carolina Parada.
\newblock {Learning to Learn Faster from Human Feedback with Language Model Predictive Control}.
\newblock In \emph{Robotics: Science and Systems}, 2024.

\bibitem[Liang et~al.(2026)Liang, Liao, Zhang, Hou, Cai, Deng, He, Huang, Tan, Wang, Yu, Tian, Sartoretti, Luo, Mu, Wu, Li, and Lv]{arxiv260932862}
Jingsong Liang, Shuhao Liao, Shizhe Zhang, Diyuan Hou, Yuxin Cai, Xinjian Deng, Chengyang He, Wenhui Huang, Runjia Tan, Zhidong Wang, Lan Yu, Xuesong Tian, Guillaume Sartoretti, Jie Luo, Yao Mu, Wenjun Wu, Wanhua Li, and Chen Lv.
\newblock {RoboFoundry: System-as-Policy Evolution for Self-Learning Embodied Agents}.
\newblock arXiv preprint arXiv:2609.32862, 2026.

\bibitem[Liang et~al.(2025)Liang, Wang, Wang, Bastani, Jayaraman, and Ma]{liang2025eurekaverse}
William Liang, Sam Wang, Hung-Ju Wang, Osbert Bastani, Dinesh Jayaraman, and Yecheng~Jason Ma.
\newblock {Environment Curriculum Generation via Large Language Models}.
\newblock In \emph{Proceedings of the 8th Conference on Robot Learning (CoRL 2024)}, volume 270 of \emph{Proceedings of Machine Learning Research}, pages 433--454, 2025.

\bibitem[Lin et~al.(2026)Lin, Cao, Luo, Chai, Lee, and Hsu]{arxiv261007753}
Hongzhan Lin, Shidong Cao, Ziyang Luo, Wenhao Chai, Mong-Li Lee, and Wynne Hsu.
\newblock {From Evidence to Action: How Tool-Using Agents Fail}.
\newblock arXiv preprint arXiv:2610.07753, 2026.

\bibitem[Liu et~al.(2023{\natexlab{a}})Liu, Zhu, Gao, Feng, Liu, Zhu, and Stone]{arxiv230603310}
Bo~Liu, Yifeng Zhu, Chongkai Gao, Yihao Feng, Qiang Liu, Yuke Zhu, and Peter Stone.
\newblock {LIBERO: Benchmarking Knowledge Transfer for Lifelong Robot Learning}.
\newblock In \emph{Advances in Neural Information Processing Systems}, 2023{\natexlab{a}}.

\bibitem[Liu et~al.(2026{\natexlab{a}})Liu, Zhang, Lin, Wang, Wang, Wang, and Qi]{arxiv261003476}
Chenzhi Liu, Yue Zhang, Jiehong Lin, Jianan Wang, Bo~Wang, Zhongrui Wang, and Xiaojuan Qi.
\newblock {MobiAgent: Dual-Loop Recursive Policy Self-Improvement for Long-Horizon Mobile Manipulation}.
\newblock arXiv preprint arXiv:2610.03476, 2026{\natexlab{a}}.
\newblock Accepted to CoRL 2026.

\bibitem[Liu et~al.(2026{\natexlab{b}})Liu, Li, Ding, Wen, Ren, Ma, Jian, Wang, Yu, Li, and Song]{arxiv260914633}
Hanyu Liu, Qian Li, Yizhu Ding, Jiayi Wen, Keqiang Ren, Yunsheng Ma, Tao Jian, Zhihua Wang, Zhuofan Yu, Xinran Li, and Zhigong Song.
\newblock {REVOLVE: An Automated Closed-Loop Framework for Evolving Robot Manipulation with Minimal Human Intervention}.
\newblock arXiv preprint arXiv:2609.14633, 2026{\natexlab{b}}.

\bibitem[Liu et~al.(2023{\natexlab{b}})Liu, Nasiriany, Zhang, Bao, and Zhu]{arxiv221108416}
Huihan Liu, Soroush Nasiriany, Lance Zhang, Zhiyao Bao, and Yuke Zhu.
\newblock {Robot Learning on the Job: Human-in-the-Loop Autonomy and Learning During Deployment}.
\newblock In \emph{Robotics: Science and Systems}, 2023{\natexlab{b}}.
\newblock arXiv:2211.08416.

\bibitem[Liu et~al.(2024)Liu, Dass, Mart{\'i}n-Mart{\'i}n, and Zhu]{arxiv231017552}
Huihan Liu, Shivin Dass, Roberto Mart{\'i}n-Mart{\'i}n, and Yuke Zhu.
\newblock {Model-Based Runtime Monitoring with Interactive Imitation Learning}.
\newblock In \emph{IEEE International Conference on Robotics and Automation}, 2024.
\newblock arXiv:2310.17552.

\bibitem[Liu et~al.(2026{\natexlab{c}})Liu, Kim, Liu, Liu, and Zhu]{arxiv260303818}
Huihan Liu, Changyeon Kim, Bo~Liu, Minghuan Liu, and Yuke Zhu.
\newblock {Pretrained Vision-Language-Action Models are Surprisingly Resistant to Forgetting in Continual Learning}.
\newblock arXiv preprint arXiv:2603.03818, 2026{\natexlab{c}}.

\bibitem[Liu et~al.(2026{\natexlab{d}})Liu, Chen, Liu, Zhang, Chen, Xie, Yao, Xu, Cui, and Zhang]{arxiv260939685}
Jiajun Liu, Yifan Chen, Yichao Liu, Jiayi Zhang, Ruoqu Chen, Shaoxuan Xie, Guocai Yao, Mengdi Xu, Sen Cui, and Changshui Zhang.
\newblock {RoboCoach: World Models as Active Coaches for Compositional Robot Skills}.
\newblock arXiv preprint arXiv:2609.39685, 2026{\natexlab{d}}.

\bibitem[Liu et~al.(2026{\natexlab{e}})Liu, Wu, Wen, You, Zhang, Hao, Lin, and Su]{arxiv260924749}
Shuaijun Liu, Chengyu Wu, Qifu Wen, Feiyang You, Chenglong Zhang, Shuyang Hao, Xi~Lin, and Ningxin Su.
\newblock {D-JEPA: A Decision-Aligned Latent World Model}.
\newblock arXiv preprint arXiv:2609.24749, 2026{\natexlab{e}}.

\bibitem[Liu et~al.(2026{\natexlab{f}})Liu, Wu, Liu, Cui, and Zhang]{arxiv260825572}
Xiang Liu, Kunwei Wu, Miao Liu, Sen Cui, and Changshui Zhang.
\newblock {ConfAL-WM: Confidence-Guided Active Learning for Action-Conditioned World Models}.
\newblock arXiv preprint arXiv:2608.25572, 2026{\natexlab{f}}.

\bibitem[Liu et~al.(2026{\natexlab{g}})Liu, Bai, Ci, Ma, and Shou]{arxiv260206508}
Xiaokang Liu, Zechen Bai, Hai Ci, Kevin~Yuchen Ma, and Mike~Zheng Shou.
\newblock {World-VLA-Loop: Closed-Loop Learning of Video World Model and VLA Policy}.
\newblock arXiv preprint arXiv:2602.06508, 2026{\natexlab{g}}.

\bibitem[Liu et~al.(2026{\natexlab{h}})Liu, Chen, Song, Pu, Mo, Bai, Chen, Sun, Zhong, Shen, and Lin]{arxiv260716636}
Yang Liu, Weixing Chen, Xinshuai Song, Tao Pu, Siwen Mo, Yongjie Bai, Zihao Chen, Qianran Sun, Liruo Zhong, Ying Shen, and Liang Lin.
\newblock {PhyAgentOS: A Self-Evolving Operating System for Embodied Agents with Decoupled Cognitive Planning and Physical Execution}.
\newblock arXiv preprint arXiv:2607.16636, 2026{\natexlab{h}}.

\bibitem[Liu et~al.(2023{\natexlab{c}})Liu, Bahety, and Song]{liu2023reflect}
Zeyi Liu, Arpit Bahety, and Shuran Song.
\newblock {REFLECT: Summarizing Robot Experiences for Failure Explanation and Correction}.
\newblock In \emph{Conference on Robot Learning}, volume 229 of \emph{Proceedings of Machine Learning Research}, pages 3468--3484, 2023{\natexlab{c}}.

\bibitem[Lopez-Paz and Ranzato(2017)]{lopezpaz2017gem}
David Lopez-Paz and Marc'Aurelio Ranzato.
\newblock {Gradient Episodic Memory for Continual Learning}.
\newblock In \emph{Advances in Neural Information Processing Systems}, volume~30, 2017.

\bibitem[Lou et~al.(2026)Lou, Zeng, Wang, and Si]{arxiv260928530}
Zeyu Lou, Yanhong Zeng, Yong Wang, and Chenyang Si.
\newblock {Know Your Body: A Harness for Direct and Self-Improving Robot Control with VLMs}.
\newblock arXiv preprint arXiv:2609.28530, 2026.

\bibitem[Lu et~al.(2024)Lu, Lu, Lange, Foerster, Clune, and Ha]{lu2024aiscientist}
Chris Lu, Cong Lu, Robert~Tjarko Lange, Jakob Foerster, Jeff Clune, and David Ha.
\newblock {The AI Scientist: Towards Fully Automated Open-Ended Scientific Discovery}.
\newblock arXiv preprint arXiv:2408.06292, 2024.

\bibitem[Lu et~al.(2025)Lu, Guo, Zhang, Zhou, Jiang, Gao, Tang, and Wang]{arxiv250518719}
Guanxing Lu, Wenkai Guo, Chubin Zhang, Yuheng Zhou, Haonan Jiang, Zifeng Gao, Yansong Tang, and Ziwei Wang.
\newblock {VLA-RL: Towards Masterful and General Robotic Manipulation with Scalable Reinforcement Learning}.
\newblock arXiv preprint arXiv:2505.18719, 2025.

\bibitem[Lu et~al.(2026{\natexlab{a}})Lu, Wu, Kou, Fu, Xiao, Mandlekar, Xu, Shi, Goldberg, Chen, Chowdhury, Zhu, Fan, and Wang]{arxiv260700272}
Runyu Lu, Yubo Wu, Ethan Kou, Letian Fu, Wenli Xiao, Ajay Mandlekar, Yinzhen Xu, Guanya Shi, Ken Goldberg, Ang Chen, Mosharaf Chowdhury, Yuke Zhu, Linxi~"Jim" Fan, and Guanzhi Wang.
\newblock {ASPIRE: Agentic /Skills Discovery for Robotics}.
\newblock arXiv preprint arXiv:2607.00272, 2026{\natexlab{a}}.

\bibitem[Lu et~al.(2026{\natexlab{b}})Lu, Zhai, Wang, Zeng, Yang, Liu, Cheng, Zhang, Qiu, Cheng, Laptev, Xu, Riviere, Loianno, Xing, and Zuo]{arxiv260916074}
Zuxing Lu, Hongjia Zhai, Guanzhi Wang, Huajian Zeng, Jiaqi Yang, Jingyu Liu, Lei Cheng, Yuantai Zhang, Yuheng Qiu, Zezhou Cheng, Ivan Laptev, Danfei Xu, Benjamin Riviere, Giuseppe Loianno, Eric Xing, and Xingxing Zuo.
\newblock {World-Action Models for Robot Learning and Control: A Survey}.
\newblock arXiv preprint arXiv:2609.16074, 2026{\natexlab{b}}.

\bibitem[Luo et~al.(2026{\natexlab{a}})Luo, Zeng, Jia, Du, and Sun]{arxiv250606658}
Calvin Luo, Zilai Zeng, Mingxi Jia, Yilun Du, and Chen Sun.
\newblock {Self-Improving Loops for Visual Robotic Planning}.
\newblock In \emph{International Conference on Learning Representations}, 2026{\natexlab{a}}.
\newblock arXiv:2506.06658.

\bibitem[Luo et~al.(2024)Luo, Hu, Xu, Tan, Berg, Sharma, Schaal, Finn, Gupta, and Levine]{luo2024serl}
Jianlan Luo, Zheyuan Hu, Charles Xu, You~Liang Tan, Jacob Berg, Archit Sharma, Stefan Schaal, Chelsea Finn, Abhishek Gupta, and Sergey Levine.
\newblock {SERL: A Software Suite for Sample-Efficient Robotic Reinforcement Learning}.
\newblock In \emph{IEEE International Conference on Robotics and Automation}, pages 16961--16969, 2024.

\bibitem[Luo et~al.(2025)Luo, Xu, Wu, and Levine]{arxiv241021845}
Jianlan Luo, Charles Xu, Jeffrey Wu, and Sergey Levine.
\newblock {Precise and Dexterous Robotic Manipulation via Human-in-the-Loop Reinforcement Learning}.
\newblock \emph{Science Robotics}, 2025.
\newblock \doi{10.1126/scirobotics.ads5033}.

\bibitem[Luo et~al.(2026{\natexlab{b}})Luo, Li, Li, Lin, Zhou, Ju, Zhang, Jin, Lee, and Hsu]{arxiv260916679}
Meng Luo, Yanlin Li, Hao Li, Hongzhan Lin, Pengfei Zhou, Tianjie Ju, Ran Zhang, Yeying Jin, Mong-Li Lee, and Wynne Hsu.
\newblock {AI for Games in the Foundation Model Era}.
\newblock arXiv preprint arXiv:2609.16679, 2026{\natexlab{b}}.

\bibitem[Ma et~al.(2026)Ma, Gao, Qiang, Dai, and Li]{arxiv260934210}
Haitong Ma, Chenxiao Gao, Rushi Qiang, Bo~Dai, and Na~Li.
\newblock {RLE-Bench: A Qualifying Exam for Coding Agents as Robot Learning Engineers}.
\newblock arXiv preprint arXiv:2609.34210, 2026.

\bibitem[Ma and Wu(2026)]{arxiv260910873}
Qinzhen Ma and Ruihai Wu.
\newblock {When Validation Stops Learning: Auditing Update Admission for Continual Embodied Agents}.
\newblock arXiv preprint arXiv:2609.10873, 2026.

\bibitem[Ma et~al.(2023{\natexlab{a}})Ma, Kumar, Zhang, Bastani, and Jayaraman]{ma2023liv}
Yecheng~Jason Ma, Vikash Kumar, Amy Zhang, Osbert Bastani, and Dinesh Jayaraman.
\newblock {LIV: Language-Image Representations and Rewards for Robotic Control}.
\newblock In \emph{International Conference on Machine Learning}, volume 202 of \emph{Proceedings of Machine Learning Research}, pages 23301--23320, 2023{\natexlab{a}}.

\bibitem[Ma et~al.(2023{\natexlab{b}})Ma, Sodhani, Jayaraman, Bastani, Kumar, and Zhang]{ma2023vip}
Yecheng~Jason Ma, Shagun Sodhani, Dinesh Jayaraman, Osbert Bastani, Vikash Kumar, and Amy Zhang.
\newblock {VIP: Towards Universal Visual Reward and Representation via Value-Implicit Pre-Training}.
\newblock In \emph{International Conference on Learning Representations}, 2023{\natexlab{b}}.

\bibitem[Ma et~al.(2024{\natexlab{a}})Ma, Liang, Wang, Huang, Bastani, Jayaraman, Zhu, Fan, and Anandkumar]{ma2024eureka}
Yecheng~Jason Ma, William Liang, Guanzhi Wang, De-An Huang, Osbert Bastani, Dinesh Jayaraman, Yuke Zhu, Linxi Fan, and Anima Anandkumar.
\newblock {Eureka}: Human-level reward design via coding large language models.
\newblock In \emph{International Conference on Learning Representations}, 2024{\natexlab{a}}.

\bibitem[Ma et~al.(2024{\natexlab{b}})Ma, Liang, Wang, Zhu, Fan, Bastani, and Jayaraman]{ma2024dreureka}
Yecheng~Jason Ma, William Liang, Hung-Ju Wang, Yuke Zhu, Linxi Fan, Osbert Bastani, and Dinesh Jayaraman.
\newblock {DrEureka}: Language model guided sim-to-real transfer.
\newblock In \emph{Robotics: Science and Systems}, 2024{\natexlab{b}}.
\newblock \doi{10.15607/RSS.2024.XX.094}.

\bibitem[Mark et~al.(2024)Mark, Gao, Sampaio, Srirama, Sharma, Finn, and Kumar]{arxiv241206685}
Max~Sobol Mark, Tian Gao, Georgia~Gabriela Sampaio, Mohan~Kumar Srirama, Archit Sharma, Chelsea Finn, and Aviral Kumar.
\newblock {Policy Agnostic RL: Offline RL and Online RL Fine-Tuning of Any Class and Backbone}.
\newblock arXiv preprint arXiv:2412.06685, 2024.

\bibitem[Meier et~al.(2018)Meier, Kappler, and Schaal]{meier2018online}
Franziska Meier, Daniel Kappler, and Stefan Schaal.
\newblock {Online Learning of a Memory for Learning Rates}.
\newblock In \emph{Proceedings of the IEEE International Conference on Robotics and Automation (ICRA)}, 2018.

\bibitem[Mendonca et~al.(2023)Mendonca, Bahl, and Pathak]{arxiv230206604}
Russell Mendonca, Shikhar Bahl, and Deepak Pathak.
\newblock {ALAN: Autonomously Exploring Robotic Agents in the Real World}.
\newblock In \emph{IEEE International Conference on Robotics and Automation}, 2023.
\newblock arXiv:2302.06604.

\bibitem[Mendonca et~al.(2024)Mendonca, Panov, Bucher, Wang, and Pathak]{mendonca2024continuously}
Russell Mendonca, Emmanuel Panov, Bernadette Bucher, Jiuguang Wang, and Deepak Pathak.
\newblock {Continuously Improving Mobile Manipulation with Autonomous Real-World RL}.
\newblock In \emph{Proceedings of the Conference on Robot Learning (CoRL)}, 2024.

\bibitem[Meng et~al.(2026{\natexlab{a}})Meng, Du, Chen, Zhao, Lu, Hu, and Shieh]{arxiv260725886}
Fanqing Meng, Lingxiao Du, Qiguang Chen, Ziqi Zhao, Haocheng Lu, Mengkang Hu, and Michael~Qizhe Shieh.
\newblock {RSIBench-Data: Benchmarking Data-Centric Research for Recursive Self-Improvement}.
\newblock arXiv preprint arXiv:2607.25886, 2026{\natexlab{a}}.

\bibitem[Meng et~al.(2026{\natexlab{b}})Meng, Du, Lu, Chen, Zhao, Wu, Zhuo, Hu, and Shieh]{arxiv261010310}
Fanqing Meng, Lingxiao Du, Haocheng Lu, Qiguang Chen, Ziqi Zhao, Zijian Wu, Jiayuan Zhuo, Mengkang Hu, and Michael~Qizhe Shieh.
\newblock {RSIGym: A Flexible Environment for Recursive Self-Improvement}.
\newblock arXiv preprint arXiv:2610.10310, 2026{\natexlab{b}}.

\bibitem[Metz et~al.(2021)Metz, Freeman, Maheswaranathan, and Sohl-Dickstein]{metz2021training}
Luke Metz, C.~Daniel Freeman, Niru Maheswaranathan, and Jascha Sohl-Dickstein.
\newblock {Training Learned Optimizers with Randomly Initialized Learned Optimizers}.
\newblock arXiv preprint arXiv:2101.07367, 2021.

\bibitem[Mirchandani et~al.(2025)Mirchandani, Belkhale, Hejna, Choi, Islam, and Sadigh]{arxiv241101813}
Suvir Mirchandani, Suneel Belkhale, Joey Hejna, Evelyn Choi, Md~Sazzad Islam, and Dorsa Sadigh.
\newblock {So You Think You Can Scale Up Autonomous Robot Data Collection?}
\newblock In \emph{Proceedings of the 8th Conference on Robot Learning (CoRL 2024)}, volume 270 of \emph{Proceedings of Machine Learning Research}, pages 2791--2810, 2025.

\bibitem[{MMLab \& PhysicalRSI contributors}(2026)]{physicalrsi2026}
{MMLab \& PhysicalRSI contributors}.
\newblock {Physical RSI 1.0: Recursive Self-Harness for Scaling Embodied Skills}.
\newblock Project website and open-source software, \url{https://mmlab.hk/research/PhysicalRSI}, 2026.
\newblock Preview release.

\bibitem[Moldovan and Abbeel(2012)]{moldovan2012safe}
Teodor~Mihai Moldovan and Pieter Abbeel.
\newblock {Safe Exploration in Markov Decision Processes}.
\newblock In \emph{International Conference on Machine Learning}, 2012.

\bibitem[M{\"u}ller et~al.(2024)M{\"u}ller, Brahmbhatt, Deka, Leboutet, Hafner, and Koltun]{arxiv240507515}
Matthias M{\"u}ller, Samarth Brahmbhatt, Ankur Deka, Quentin Leboutet, David Hafner, and Vladlen Koltun.
\newblock {OpenBot-Fleet: A System for Collective Learning with Real Robots}.
\newblock In \emph{IEEE International Conference on Robotics and Automation}, 2024.
\newblock arXiv:2405.07515.

\bibitem[Nagabandi et~al.(2019)Nagabandi, Clavera, Liu, Fearing, Abbeel, Levine, and Finn]{arxiv180311347}
Anusha Nagabandi, Ignasi Clavera, Simin Liu, Ronald~S. Fearing, Pieter Abbeel, Sergey Levine, and Chelsea Finn.
\newblock {Learning to Adapt in Dynamic, Real-World Environments Through Meta-Reinforcement Learning}.
\newblock In \emph{International Conference on Learning Representations}, 2019.

\bibitem[Nakamoto et~al.(2023)Nakamoto, Zhai, Singh, Mark, Ma, Finn, Kumar, and Levine]{nakamoto2023calql}
Mitsuhiko Nakamoto, Simon Zhai, Anikait Singh, Max~Sobol Mark, Yi~Ma, Chelsea Finn, Aviral Kumar, and Sergey Levine.
\newblock {Cal-QL: Calibrated Offline RL Pre-Training for Efficient Online Fine-Tuning}.
\newblock In \emph{Advances in Neural Information Processing Systems}, volume~36, pages 62244--62269, 2023.

\bibitem[Nakamoto et~al.(2024)Nakamoto, Mees, Kumar, and Levine]{arxiv241013816}
Mitsuhiko Nakamoto, Oier Mees, Aviral Kumar, and Sergey Levine.
\newblock {Steering Your Generalists: Improving Robotic Foundation Models via Value Guidance}.
\newblock In \emph{Conference on Robot Learning}, 2024.

\bibitem[Nasiriany et~al.(2024)Nasiriany, Maddukuri, Zhang, Parikh, Lo, Joshi, Mandlekar, and Zhu]{arxiv240602523}
Soroush Nasiriany, Abhiram Maddukuri, Lance Zhang, Adeet Parikh, Aaron Lo, Abhishek Joshi, Ajay Mandlekar, and Yuke Zhu.
\newblock {RoboCasa: Large-Scale Simulation of Everyday Tasks for Generalist Robots}.
\newblock In \emph{Robotics: Science and Systems}, 2024.

\bibitem[Navasardyan et~al.(2026)Navasardyan, Danielyan, and Davtyan]{arxiv260903611}
Zaruhi Navasardyan, Tatul Danielyan, and Hrant Davtyan.
\newblock {FailBench: How Reliable are VLMs at Judging Robot Task Success?}
\newblock arXiv preprint arXiv:2609.03611, 2026.

\bibitem[Ng et~al.(2026)Ng, Zhang, Wu, Lui, Zhao, Gao, and Ren]{arxiv261013376}
Chi~Kit Ng, Yidong Zhang, Tianchun Wu, Siu~Hing Lui, Hanlin Zhao, Huxin Gao, and Hongliang Ren.
\newblock {EndoRSI: Physical-Evidence-Driven Recursive Improvement of Endoscopic Robot Policies}.
\newblock arXiv preprint arXiv:2610.13376, 2026.

\bibitem[Nguyen et~al.(2026)Nguyen, Raha, Le, Huh, Niyato, and Hong]{arxiv260913236}
Loc~X. Nguyen, Avi~Deb Raha, Huy~Q. Le, Eui-Nam Huh, Dusit Niyato, and Choong~Seon Hong.
\newblock {Self-Evolving AI for Humanoids: Mechanisms, Safety, and Evaluation of Post-Deployment Self-Improvement}.
\newblock arXiv preprint arXiv:2609.13236, 2026.

\bibitem[Novikov et~al.(2025)Novikov, Vũ, Eisenberger, Dupont, Huang, Wagner, Shirobokov, Kozlovskii, Ruiz, Mehrabian, Kumar, See, Chaudhuri, Holland, Davies, Nowozin, Kohli, and Balog]{arxiv250613131}
Alexander Novikov, Ngân Vũ, Marvin Eisenberger, Emilien Dupont, Po-Sen Huang, Adam~Zsolt Wagner, Sergey Shirobokov, Borislav Kozlovskii, Francisco J.~R. Ruiz, Abbas Mehrabian, M.~Pawan Kumar, Abigail See, Swarat Chaudhuri, George Holland, Alex Davies, Sebastian Nowozin, Pushmeet Kohli, and Matej Balog.
\newblock {AlphaEvolve: A Coding Agent for Scientific and Algorithmic Discovery}.
\newblock arXiv preprint arXiv:2506.13131, 2025.

\bibitem[{NVIDIA} et~al.(2025){NVIDIA}, Agarwal, Ali, Bala, Balaji, Barker, Cai, Chattopadhyay, Chen, Cui, et~al.]{nvidia2025cosmos}
{NVIDIA}, Niket Agarwal, Arslan Ali, Maciej Bala, Yogesh Balaji, Erik Barker, Tiffany Cai, Prithvijit Chattopadhyay, Yongxin Chen, Yin Cui, et~al.
\newblock {Cosmos World Foundation Model Platform for Physical AI}.
\newblock arXiv preprint arXiv:2501.03575, 2025.

\bibitem[{Octo Model Team} et~al.(2024){Octo Model Team}, Ghosh, Walke, Pertsch, Black, et~al.]{octo2024}
{Octo Model Team}, Dibya Ghosh, Homer Walke, Karl Pertsch, Kevin Black, et~al.
\newblock {Octo: An Open-Source Generalist Robot Policy}.
\newblock In \emph{Robotics: Science and Systems}, 2024.

\bibitem[Oh et~al.(2020)Oh, Hessel, Czarnecki, Xu, van Hasselt, Singh, and Silver]{oh2020lpg}
Junhyuk Oh, Matteo Hessel, Wojciech~M. Czarnecki, Zhongwen Xu, Hado van Hasselt, Satinder Singh, and David Silver.
\newblock {Discovering Reinforcement Learning Algorithms}.
\newblock In \emph{Advances in Neural Information Processing Systems}, volume~33, 2020.

\bibitem[Oh et~al.(2025)Oh, Kemaev, Farquhar, Calian, Hessel, Zintgraf, Singh, van Hasselt, and Silver]{oh2025discorl}
Junhyuk Oh, Iurii Kemaev, Greg Farquhar, Dan~A. Calian, Matteo Hessel, Luisa Zintgraf, Satinder Singh, Hado van Hasselt, and David Silver.
\newblock {Discovering State-of-the-Art Reinforcement Learning Algorithms}.
\newblock \emph{Nature}, 2025.

\bibitem[{Open X-Embodiment Collaboration} et~al.(2024)]{arxiv231008864}
{Open X-Embodiment Collaboration} et~al.
\newblock {Open X-Embodiment: Robotic Learning Datasets and RT-X Models}.
\newblock In \emph{IEEE International Conference on Robotics and Automation}, pages 6892--6903, 2024.
\newblock \doi{10.1109/ICRA57147.2024.10611477}.

\bibitem[{OpenAI} et~al.(2019){OpenAI}, Akkaya, Andrychowicz, Chociej, Litwin, McGrew, Petron, Paino, Plappert, Powell, Ribas, Schneider, Tezak, Tworek, Welinder, Weng, Yuan, Zaremba, and Zhang]{openai2019rubik}
{OpenAI}, Ilge Akkaya, Marcin Andrychowicz, Maciek Chociej, Mateusz Litwin, Bob McGrew, Arthur Petron, Alex Paino, Matthias Plappert, Glenn Powell, Raphael Ribas, Jonas Schneider, Nikolas Tezak, Jerry Tworek, Peter Welinder, Lilian Weng, Qiming Yuan, Wojciech Zaremba, and Lei Zhang.
\newblock {Solving Rubik's Cube with a Robot Hand}.
\newblock arXiv preprint arXiv:1910.07113, 2019.

\bibitem[Orseau and Armstrong(2016)]{orseau2016interruptible}
Laurent Orseau and Stuart Armstrong.
\newblock {Safely Interruptible Agents}.
\newblock In \emph{Proceedings of the Thirty-Second Conference on Uncertainty in Artificial Intelligence}, 2016.

\bibitem[Oudeyer et~al.(2007)Oudeyer, Kaplan, and Hafner]{oudeyer2007intrinsic}
Pierre-Yves Oudeyer, Fr{\'e}d{\'e}ric Kaplan, and Verena~V. Hafner.
\newblock {Intrinsic Motivation Systems for Autonomous Mental Development}.
\newblock \emph{IEEE Transactions on Evolutionary Computation}, 11\penalty0 (2):\penalty0 265--286, 2007.
\newblock \doi{10.1109/tevc.2006.890271}.

\bibitem[Pan et~al.(2026)Pan, Feng, Zhang, Li, Song, Qu, Wang, Li, Xiong, Chen, Liu, and Luo]{arxiv260103044}
Mingjie Pan, Siyuan Feng, Qinglin Zhang, Xinchen Li, Jianheng Song, Chendi Qu, Yi~Wang, Chuankang Li, Ziyu Xiong, Zhi Chen, Yi~Liu, and Jianlan Luo.
\newblock {SOP: A Scalable Online Post-Training System for Vision-Language-Action Models}.
\newblock arXiv preprint arXiv:2601.03044, 2026.

\bibitem[Parakh et~al.(2024)Parakh, Fong, Simeonov, Chen, Gupta, and Agrawal]{arxiv230914321}
Meenal Parakh, Alisha Fong, Anthony Simeonov, Tao Chen, Abhishek Gupta, and Pulkit Agrawal.
\newblock {Lifelong Robot Learning with Human Assisted Language Planners}.
\newblock In \emph{IEEE International Conference on Robotics and Automation}, 2024.
\newblock arXiv:2309.14321.

\bibitem[Parisi et~al.(2019)Parisi, Kemker, Part, Kanan, and Wermter]{parisi2019continual}
German~I. Parisi, Ronald Kemker, Jose~L. Part, Christopher Kanan, and Stefan Wermter.
\newblock {Continual Lifelong Learning with Neural Networks: A Review}.
\newblock \emph{Neural Networks}, 113:\penalty0 54--71, 2019.
\newblock \doi{10.1016/j.neunet.2019.01.012}.

\bibitem[Phan et~al.(2020)Phan, Grosu, Jansen, Paoletti, Smolka, and Stoller]{phan2020nsa}
Dung~T. Phan, Radu Grosu, Nils Jansen, Nicola Paoletti, Scott~A. Smolka, and Scott~D. Stoller.
\newblock {Neural Simplex Architecture}.
\newblock In \emph{NASA Formal Methods Symposium}, 2020.

\bibitem[{Physical Intelligence} et~al.(2025){Physical Intelligence}, Black, Brown, Darpinian, Dhabalia, et~al.]{black2025pi05}
{Physical Intelligence}, Kevin Black, Noah Brown, James Darpinian, Karan Dhabalia, et~al.
\newblock {$\pi_{0.5}$: a Vision-Language-Action Model with Open-World Generalization}.
\newblock In \emph{Proceedings of the 9th Conference on Robot Learning}, volume 305 of \emph{Proceedings of Machine Learning Research}, pages 17--40, 2025.

\bibitem[{Physical Intelligence} et~al.(2026){Physical Intelligence}, Amin, Aniceto, Balakrishna, Black, Conley, Connors, Darpinian, Dhabalia, DiCarlo, et~al.]{arxiv251114759}
{Physical Intelligence}, Ali Amin, Raichelle Aniceto, Ashwin Balakrishna, Kevin Black, Ken Conley, Grace Connors, James Darpinian, Karan Dhabalia, Jared DiCarlo, et~al.
\newblock {$\pi^{*}_{0.6}$: a VLA That Learns From Experience}.
\newblock In \emph{Robotics: Science and Systems}, 2026.
\newblock \doi{10.15607/RSS.2026.XXII.087}.

\bibitem[Pinto and Gupta(2016)]{pinto2016supersizing}
Lerrel Pinto and Abhinav Gupta.
\newblock {Supersizing Self-Supervision: Learning to Grasp from 50K Tries and 700 Robot Hours}.
\newblock In \emph{IEEE International Conference on Robotics and Automation}, pages 3406--3413, 2016.
\newblock \doi{10.1109/ICRA.2016.7487517}.

\bibitem[Portelas et~al.(2020)Portelas, Colas, Weng, Hofmann, and Oudeyer]{portelas2020curriculum}
Rémy Portelas, Cédric Colas, Lilian Weng, Katja Hofmann, and Pierre-Yves Oudeyer.
\newblock {Automatic Curriculum Learning For Deep RL: A Short Survey}.
\newblock In \emph{Proceedings of the 29th International Joint Conference on Artificial Intelligence}, 2020.
\newblock \doi{10.24963/ijcai.2020/671}.

\bibitem[Qin et~al.(2026)Qin, Chen, Lin, Liu, Lyu, Su, Jin, Yu, Luo, Ding, Chen, and Xu]{arxiv260920649}
Yan Qin, Yue Chen, Wenwei Lin, Shujia Liu, Chuqiao Lyu, Kailun Su, Weiyang Jin, Chenze Yu, Ping Luo, Wenbo Ding, Tianxing Chen, and Renjing Xu.
\newblock {DexTouch-WM: Learning Action-Conditioned Tactile World Models from Human Touch for Dexterous Robot Manipulation}.
\newblock arXiv preprint arXiv:2609.20649, 2026.

\bibitem[Qu et~al.(2025)Qu, Lan, Zurbr{\"u}gg, Chen, Mower, Bou-Ammar, and Hutter]{arxiv250716713}
Kaixian Qu, Guowei Lan, Ren{\'e} Zurbr{\"u}gg, Changan Chen, Christopher~E. Mower, Haitham Bou-Ammar, and Marco Hutter.
\newblock {A Pragmatist Robot: Learning to Plan Tasks by Experiencing the Real World}.
\newblock arXiv preprint arXiv:2507.16713, 2025.

\bibitem[Raghavan and Singh(2026)]{arxiv260902811}
Dharini Raghavan and Amritpal Singh.
\newblock {Do Better Imagined Rollouts Mean Better Robot Control? A Controlled Study of World-Model Evaluation Under Feedback}.
\newblock arXiv preprint arXiv:2609.02811, 2026.

\bibitem[Real et~al.(2020)Real, Liang, So, and Le]{real2020automlzero}
Esteban Real, Chen Liang, David~R. So, and Quoc~V. Le.
\newblock {AutoML-Zero: Evolving Machine Learning Algorithms From Scratch}.
\newblock In \emph{Proceedings of the 37th International Conference on Machine Learning}, volume 119 of \emph{Proceedings of Machine Learning Research}, pages 8007--8019, 2020.

\bibitem[Ren et~al.(2025)Ren, Lidard, Ankile, Simeonov, Agrawal, Majumdar, Burchfiel, Dai, and Simchowitz]{ren2025dppo}
Allen~Z. Ren, Justin Lidard, Lars~L. Ankile, Anthony Simeonov, Pulkit Agrawal, Anirudha Majumdar, Benjamin Burchfiel, Hongkai Dai, and Max Simchowitz.
\newblock {Diffusion Policy Policy Optimization}.
\newblock In \emph{International Conference on Learning Representations}, pages 77288--77329, 2025.

\bibitem[Ren et~al.(2026)Ren, Zhu, Xing, and Scaramuzza]{arxiv260210111}
Yunfan Ren, Zhiyuan Zhu, Jiaxu Xing, and Davide Scaramuzza.
\newblock {Learning Agile Quadrotor Flight in the Real World}.
\newblock In \emph{Robotics: Science and Systems}, 2026.
\newblock arXiv:2602.10111.

\bibitem[Riedmiller et~al.(2018)Riedmiller, Hafner, Lampe, Neunert, Degrave, de~Wiele, Mnih, Heess, and Springenberg]{riedmiller2018sacx}
Martin Riedmiller, Roland Hafner, Thomas Lampe, Michael Neunert, Jonas Degrave, Tom~Van de~Wiele, Volodymyr Mnih, Nicolas Heess, and Jost~Tobias Springenberg.
\newblock {Learning by Playing---Solving Sparse Reward Tasks from Scratch}.
\newblock In \emph{Proceedings of the 35th International Conference on Machine Learning}, volume~80 of \emph{Proceedings of Machine Learning Research}, pages 4344--4353, 2018.

\bibitem[Robeyns et~al.(2025)Robeyns, Szummer, and Aitchison]{arxiv250415228}
Maxime Robeyns, Martin Szummer, and Laurence Aitchison.
\newblock {A Self-Improving Coding Agent}.
\newblock In \emph{ICLR 2025 Workshop on Self-Improving Foundation Models Without Human Supervision}, 2025.

\bibitem[Rocamonde et~al.(2024)Rocamonde, Montesinos, Nava, Perez, and Lindner]{rocamonde2024vlmrm}
Juan Rocamonde, Victoriano Montesinos, Elvis Nava, Ethan Perez, and David Lindner.
\newblock {Vision-Language Models are Zero-Shot Reward Models for Reinforcement Learning}.
\newblock In \emph{International Conference on Learning Representations}, pages 28446--28463, 2024.

\bibitem[R{\"o}mer et~al.(2026{\natexlab{a}})R{\"o}mer, Seeliger, Liu, Sturgis, Bagatella, Marta, Krause, and Schoellig]{arxiv260618043}
Ralf R{\"o}mer, Maximilian Seeliger, Saida Liu, Ben Sturgis, Marco Bagatella, Daniel Marta, Andreas Krause, and Angela~P. Schoellig.
\newblock {Uncertainty Quantification for Flow-Based Generalist Robot Policies}.
\newblock arXiv preprint arXiv:2606.18043, 2026{\natexlab{a}}.

\bibitem[R{\"o}mer et~al.(2026{\natexlab{b}})R{\"o}mer, Zhang, Li, and Schoellig]{arxiv260109512}
Ralf R{\"o}mer, Yi~Zhang, Yuming Li, and Angela~P. Schoellig.
\newblock {CLARE: Continual Learning for Vision-Language-Action Models via Autonomous Adapter Routing and Expansion}.
\newblock \emph{IEEE Robotics and Automation Letters}, 2026{\natexlab{b}}.
\newblock \doi{10.1109/LRA.2026.3693992}.

\bibitem[Ross et~al.(2011)Ross, Gordon, and Bagnell]{ross2011dagger}
Stephane Ross, Geoffrey Gordon, and Drew Bagnell.
\newblock {A Reduction of Imitation Learning and Structured Prediction to No-Regret Online Learning}.
\newblock In \emph{Proceedings of the Fourteenth International Conference on Artificial Intelligence and Statistics}, volume~15 of \emph{Proceedings of Machine Learning Research}, pages 627--635, 2011.

\bibitem[Sadeghi et~al.(2026)Sadeghi, Seidenschwarz, Allardice, Srinivasan, Graham, and Hawke]{arxiv260816829}
Jonathan Sadeghi, Jenny Seidenschwarz, Jesse Allardice, Sirish Srinivasan, Benjamin Graham, and Jeffrey Hawke.
\newblock {CaliBench: Are the Stochastic Dynamics of Video World Models Physically Calibrated?}
\newblock \emph{Transactions on Machine Learning Research}, 2026.

\bibitem[Savva et~al.(2019)Savva, Kadian, Maksymets, Zhao, Wijmans, Jain, Straub, Liu, Koltun, Malik, Parikh, and Batra]{savva2019habitat}
Manolis Savva, Abhishek Kadian, Oleksandr Maksymets, Yili Zhao, Erik Wijmans, Bhavana Jain, Julian Straub, Jia Liu, Vladlen Koltun, Jitendra Malik, Devi Parikh, and Dhruv Batra.
\newblock {Habitat: A Platform for Embodied AI Research}.
\newblock In \emph{IEEE/CVF International Conference on Computer Vision}, 2019.

\bibitem[Schmidhuber(2003)]{schmidhuber2003godel}
Juergen Schmidhuber.
\newblock {Goedel Machines: Self-Referential Universal Problem Solvers Making Provably Optimal Self-Improvements}.
\newblock arXiv preprint arXiv:cs/0309048, 2003.

\bibitem[Sharma et~al.(2026)Sharma, Sun, Lu, Zhang, Liu, and Yang]{arxiv260202454}
Ansh~Kumar Sharma, Yixiang Sun, Ninghao Lu, Yunzhe Zhang, Jiarao Liu, and Sherry Yang.
\newblock {World-Gymnast: Training Robots with Reinforcement Learning in a World Model}.
\newblock arXiv preprint arXiv:2602.02454, 2026.

\bibitem[Sharma et~al.(2023)Sharma, Ahmed, Ahmad, and Finn]{sharma2023selfimproving}
Archit Sharma, Ahmed~M. Ahmed, Rehaan Ahmad, and Chelsea Finn.
\newblock {Self-Improving Robots: End-to-End Autonomous Visuomotor Reinforcement Learning}.
\newblock In \emph{Proceedings of the 7th Conference on Robot Learning}, volume 229 of \emph{Proceedings of Machine Learning Research}, pages 3292--3308, 2023.

\bibitem[Shen et~al.(2026)Shen, You, Liu, Zheng, Zha, Yamazaki, Zhang, Huang, Sun, Chen, He, Liu, Chang, Fragkiadaki, Shah, Schwager, Henderson, Abraham, and Xu]{arxiv260927308}
Zeyu Shen, Haoxiang You, Yilang Liu, Zhicheng Zheng, Lihan Zha, Kashu Yamazaki, Mingtong Zhang, Suning Huang, Jiankai Sun, Qianzhong Chen, Lucy He, Kaiyuan Liu, Haoran Chang, Katerina Fragkiadaki, Dhruv Shah, Mac Schwager, Peter Henderson, Ian Abraham, and Canwen Xu.
\newblock {EmbodiedSWE: Coding Agents for Long Horizon Dexterous Robotics}.
\newblock arXiv preprint arXiv:2609.27308, 2026.

\bibitem[Shinn et~al.(2023)Shinn, Cassano, Gopinath, Narasimhan, and Yao]{arxiv230311366}
Noah Shinn, Federico Cassano, Ashwin Gopinath, Karthik Narasimhan, and Shunyu Yao.
\newblock {Reflexion: Language Agents with Verbal Reinforcement Learning}.
\newblock In \emph{Advances in Neural Information Processing Systems}, volume~36, 2023.
\newblock \doi{10.52202/075280-0377}.

\bibitem[Shumailov et~al.(2024)Shumailov, Shumaylov, Zhao, Papernot, Anderson, and Gal]{shumailov2024collapse}
Ilia Shumailov, Zakhar Shumaylov, Yiren Zhao, Nicolas Papernot, Ross Anderson, and Yarin Gal.
\newblock {AI models collapse when trained on recursively generated data}.
\newblock \emph{Nature}, 631\penalty0 (8022):\penalty0 755--759, 2024.
\newblock \doi{10.1038/s41586-024-07566-y}.

\bibitem[Si et~al.(2026)Si, Bi, Yang, Ni, Huang, Wang, Jiang, Zhang, Ruan, Zhang, Wang, Li, Feng, Dong, and Zhou]{arxiv261001939}
Ruiyang Si, Jianxin Bi, Shunyu Yang, Rui Ni, Wenbo Huang, Qiang Wang, Shulong Jiang, Yintianrun Zhang, Feixiang Ruan, Kewei Zhang, Duomin Wang, Xiuyu Li, Haiwen Feng, Zhen Dong, and Daquan Zhou.
\newblock {Fewer Tokens, Better Action: GPT-6 Astra Robot Agents with 14\% Higher Success Rate but 65\% Fewer Tokens}.
\newblock arXiv preprint arXiv:2610.01939, 2026.

\bibitem[Song et~al.(2026{\natexlab{a}})Song, Chen, Zeng, Tang, Hou, Jing, Xie, and Zhang]{arxiv261013829}
Jingwei Song, Zhisheng Chen, Bingfan Zeng, Jing Tang, Yubo Hou, Jinhao Jing, Zhengwei Xie, and Linfeng Zhang.
\newblock {MC-EvoBench: Auditing Experience-Driven Self-Improvement in Minecraft Agents}.
\newblock arXiv preprint arXiv:2610.13829, 2026{\natexlab{a}}.

\bibitem[Song et~al.(2026{\natexlab{b}})Song, Liang, Fu, Wang, Yang, and Liu]{arxiv261010498}
Python Song, Zhixuan Liang, Kelsey Fu, Mengdi Wang, Junfeng Yang, and Shilong Liu.
\newblock {EmbodiedRSI: Active Continual Robot Learning Through Hypothesis-Guided Co-Evolution}.
\newblock arXiv preprint arXiv:2610.10498, 2026{\natexlab{b}}.

\bibitem[Song et~al.(2026{\natexlab{c}})Song, Mittal, and Jain]{arxiv260924996}
Yuchen Song, Aditya Mittal, and Unnat Jain.
\newblock {Learning Beyond What Humans Can Demonstrate}.
\newblock arXiv preprint arXiv:2609.24996, 2026{\natexlab{c}}.
\newblock Accepted at CoRL 2026; proceedings forthcoming.

\bibitem[Sontakke et~al.(2023)Sontakke, Zhang, Arnold, Pertsch, B{\i}y{\i}k, Sadigh, Finn, and Itti]{sontakke2023roboclip}
Sumedh Sontakke, Jesse Zhang, S{\'e}bastien Arnold, Karl Pertsch, Erdem B{\i}y{\i}k, Dorsa Sadigh, Chelsea Finn, and Laurent Itti.
\newblock {RoboCLIP: One Demonstration is Enough to Learn Robot Policies}.
\newblock In \emph{Advances in Neural Information Processing Systems}, volume~36, pages 55681--55693, 2023.

\bibitem[Srikanth et~al.(2026)Srikanth, Zhao, Xu, Wu, and Jiang]{arxiv260926457}
Dhruv Srikanth, Bingchen Zhao, Dixing Xu, Yuxiang Wu, and Zhengyao Jiang.
\newblock {Recursive self-improvement of AI research agents}.
\newblock arXiv preprint arXiv:2609.26457, 2026.

\bibitem[Su et~al.(2026)Su, Shen, Qian, Qian, Zhang, Yan, Chen, Zhang, Wang, Cong, and Shang]{arxiv260904355}
Chenyu Su, Zhaolong Shen, Yuan Qian, Chen Qian, Rui Zhang, Feng Yan, Weixing Chen, Fei Zhang, Jiamin Wang, Shuang Cong, and Weiwei Shang.
\newblock {VLA-Precision: Asymmetric Co-Bootstrapping for Efficient Real-World Online RL of Vision-Language-Action Models}.
\newblock arXiv preprint arXiv:2609.04355, 2026.

\bibitem[Sun et~al.(2022)Sun, Orbik, Devin, Yang, Gupta, Berseth, and Levine]{sun2022relmm}
Charles Sun, J{\k{e}}drzej Orbik, Coline~Manon Devin, Brian~H. Yang, Abhishek Gupta, Glen Berseth, and Sergey Levine.
\newblock {Fully Autonomous Real-World Reinforcement Learning with Applications to Mobile Manipulation}.
\newblock In \emph{Proceedings of the 5th Conference on Robot Learning}, volume 164 of \emph{Proceedings of Machine Learning Research}, pages 308--319, 2022.

\bibitem[Sun et~al.(2024)Sun, Liu, Lyu, Yang, Zhang, and Li]{arxiv241014660}
Shengjie Sun, Runze Liu, Jiafei Lyu, Jing-Wen Yang, Liangpeng Zhang, and Xiu Li.
\newblock {A Large Language Model-Driven Reward Design Framework via Dynamic Feedback for Reinforcement Learning}.
\newblock arXiv preprint arXiv:2410.14660, 2024.

\bibitem[Sutton(1990)]{sutton1990dyna}
Richard~S. Sutton.
\newblock {Integrated Architectures for Learning, Planning, and Reacting Based on Approximating Dynamic Programming}.
\newblock In \emph{International Conference on Machine Learning}, pages 216--224, 1990.
\newblock \doi{10.1016/B978-1-55860-141-3.50030-4}.

\bibitem[Sutton and Barto(2018)]{sutton2018rl}
Richard~S. Sutton and Andrew~G. Barto.
\newblock \emph{{Reinforcement Learning: An Introduction}}.
\newblock MIT Press, 2nd edition, 2018.

\bibitem[Sutton et~al.(1999)Sutton, Precup, and Singh]{sutton1999options}
Richard~S. Sutton, Doina Precup, and Satinder Singh.
\newblock {Between MDPs and Semi-MDPs: A Framework for Temporal Abstraction in Reinforcement Learning}.
\newblock \emph{Artificial Intelligence}, 112\penalty0 (1--2):\penalty0 181--211, 1999.
\newblock \doi{10.1016/S0004-3702(99)00052-1}.

\bibitem[Sutton et~al.(2011)Sutton, Modayil, Delp, Degris, Pilarski, White, and Precup]{sutton2011horde}
Richard~S. Sutton, Joseph Modayil, Michael Delp, Thomas Degris, Patrick~M. Pilarski, Adam White, and Doina Precup.
\newblock {Horde: A Scalable Real-Time Architecture for Learning Knowledge from Unsupervised Sensorimotor Interaction}.
\newblock In \emph{Proceedings of the 10th International Conference on Autonomous Agents and Multiagent Systems}, pages 761--768, 2011.

\bibitem[Sygkounas et~al.(2026)Sygkounas, Aregbede, Loutfi, and Persson]{arxiv260722832}
Alkis Sygkounas, Victor Aregbede, Amy Loutfi, and Andreas Persson.
\newblock {MEMENTO: Memory-Guided Memetic Code-as-Policy Evolution}.
\newblock arXiv preprint arXiv:2607.22832, 2026.

\bibitem[Szymanski et~al.(2023)Szymanski, Rendy, Fei, Kumar, He, Milsted, McDermott, Gallant, Cubuk, Merchant, Kim, Jain, Bartel, Persson, Zeng, and Ceder]{szymanski2023alab}
Nathan~J. Szymanski, Bernardus Rendy, Yuxing Fei, Rishi~E. Kumar, Tanjin He, David Milsted, Matthew~J. McDermott, Max Gallant, Ekin~Dogus Cubuk, Amil Merchant, Haegyeom Kim, Anubhav Jain, Christopher~J. Bartel, Kristin Persson, Yan Zeng, and Gerbrand Ceder.
\newblock {An autonomous laboratory for the accelerated synthesis of inorganic materials}.
\newblock \emph{Nature}, 624\penalty0 (7990):\penalty0 86--91, 2023.
\newblock \doi{10.1038/s41586-023-06734-w}.
\newblock Title and article corrected in 2026.

\bibitem[Tan et~al.(2026{\natexlab{a}})Tan, Xu, Xu, Tao, Li, Hong, Feng, Du, and Wang]{arxiv260727017}
Kaizhen Tan, Sizhe Xu, Xin Xu, Siru Tao, Yixiao Li, Hanzhe Hong, Yang Feng, Heqing Du, and Zhaonan Wang.
\newblock {What Can Latent World Models Know? Physical Information in Multimodal Predictive Representations}.
\newblock arXiv preprint arXiv:2607.27017, 2026{\natexlab{a}}.

\bibitem[Tan et~al.(2026{\natexlab{b}})Tan, Tu, Yan, Yu, Tian, and Lv]{arxiv260918651}
Runjia Tan, Yuang Tu, Yujie Yan, Lan Yu, Xuesong Tian, and Chen Lv.
\newblock {FIERCE: From Generalist Robot Policies to Fast Specialists via Progress-Failure Feedback}.
\newblock arXiv preprint arXiv:2609.18651, 2026{\natexlab{b}}.

\bibitem[Tang et~al.(2026)Tang, Ma, Li, and Yuan]{arxiv260913406}
Hongyao Tang, Yi~Ma, Pengyi Li, and Yifu Yuan.
\newblock {Generalized Agent Iteration: One Formal Framework for Iterative Policy Improvement and Recursive Self-Improvement}.
\newblock arXiv preprint arXiv:2609.13406, 2026.

\bibitem[Thananjeyan et~al.(2021)Thananjeyan, Balakrishna, Nair, Luo, Srinivasan, Hwang, Gonzalez, Ibarz, Finn, and Goldberg]{thananjeyan2021recovery}
Brijen Thananjeyan, Ashwin Balakrishna, Suraj Nair, Michael Luo, Krishnan Srinivasan, Minho Hwang, Joseph~E. Gonzalez, Julian Ibarz, Chelsea Finn, and Ken Goldberg.
\newblock {Recovery RL: Safe Reinforcement Learning With Learned Recovery Zones}.
\newblock \emph{IEEE Robotics and Automation Letters}, 6\penalty0 (3):\penalty0 4915--4922, 2021.

\bibitem[Tharwat et~al.(2026)Tharwat, Nematollahi, Alles, Zhang, Chandra, and Valada]{arxiv261013816}
Bahey Tharwat, Iman Nematollahi, Marvin Alles, Xingyuan Zhang, Akshay~L Chandra, and Abhinav Valada.
\newblock {Dream to Remember: Continual Steering of Generalist Robot Policies with World Models}.
\newblock arXiv preprint arXiv:2610.13816, 2026.

\bibitem[Thomas et~al.(2015)Thomas, Theocharous, and Ghavamzadeh]{thomas2015hcpi}
Philip Thomas, Georgios Theocharous, and Mohammad Ghavamzadeh.
\newblock {High Confidence Policy Improvement}.
\newblock In \emph{Proceedings of the 32nd International Conference on Machine Learning}, volume~37 of \emph{Proceedings of Machine Learning Research}, pages 2380--2388, 2015.

\bibitem[Thrun and Mitchell(1995)]{thrun1995lifelong}
Sebastian Thrun and Tom~M. Mitchell.
\newblock {Lifelong Robot Learning}.
\newblock \emph{Robotics and Autonomous Systems}, 15\penalty0 (1--2):\penalty0 25--46, 1995.
\newblock \doi{10.1016/0921-8890(95)00004-Y}.

\bibitem[Thrun and Pratt(1998)]{thrun1998learning}
Sebastian Thrun and Lorien Pratt, editors.
\newblock \emph{{Learning to Learn}}.
\newblock Springer US, 1998.
\newblock \doi{10.1007/978-1-4615-5529-2}.

\bibitem[Tian et~al.(2026)Tian, Zhang, Liu, Liu, Liu, and Wang]{arxiv261008995}
Wenqing Tian, Zeyu Zhang, Zhaocheng Liu, Fengwei Liu, Qiang Liu, and Liang Wang.
\newblock {PhysEvo: Astra Can Act, Let It}.
\newblock arXiv preprint arXiv:2610.08995, 2026.

\bibitem[Tziafas and Kasaei(2024)]{arxiv240618746}
Georgios Tziafas and Hamidreza Kasaei.
\newblock {Lifelong Robot Library Learning: Bootstrapping Composable and Generalizable Skills for Embodied Control with Language Models}.
\newblock In \emph{IEEE International Conference on Robotics and Automation}, 2024.

\bibitem[Wagenmaker et~al.(2025)Wagenmaker, Zhang, Nakamoto, Park, Yagoub, Nagabandi, Gupta, and Levine]{arxiv250615799}
Andrew Wagenmaker, Yunchu Zhang, Mitsuhiko Nakamoto, Seohong Park, Waleed Yagoub, Anusha Nagabandi, Abhishek Gupta, and Sergey Levine.
\newblock {Steering Your Diffusion Policy with Latent Space Reinforcement Learning}.
\newblock In \emph{Conference on Robot Learning}, 2025.

\bibitem[Waibel et~al.(2011)Waibel, Beetz, Civera, D'Andrea, Elfring, G{\'a}lvez-L{\'o}pez, H{\"a}ussermann, Janssen, Montiel, Perzylo, Schiessle, Tenorth, Zweigle, and de~Molengraft]{waibel2011roboearth}
Markus Waibel, Michael Beetz, Javier Civera, Raffaello D'Andrea, Jos Elfring, Dorian G{\'a}lvez-L{\'o}pez, Kai H{\"a}ussermann, Rob Janssen, J.~M.~M. Montiel, Alexander Perzylo, Bj{\"o}rn Schiessle, Moritz Tenorth, Oliver Zweigle, and M.~J. G. Ren{\'e}~Van de~Molengraft.
\newblock {RoboEarth---A World Wide Web for Robots}.
\newblock \emph{IEEE Robotics \& Automation Magazine}, 18\penalty0 (2):\penalty0 69--82, 2011.
\newblock \doi{10.1109/MRA.2011.941632}.

\bibitem[Wan et~al.(2024)Wan, Zhu, Shah, and Zhu]{arxiv231102058}
Weikang Wan, Yifeng Zhu, Rutav Shah, and Yuke Zhu.
\newblock {LOTUS: Continual Imitation Learning for Robot Manipulation Through Unsupervised Skill Discovery}.
\newblock In \emph{IEEE International Conference on Robotics and Automation}, 2024.
\newblock arXiv:2311.02058.

\bibitem[Wan et~al.(2026)Wan, Yue, Liu, Chu, Wang, Chen, Jiang, Zhu, Dong, and Zhu]{arxiv260909250}
Yang Wan, Xihang Yue, Zhirui Liu, Ziyuan Chu, Shuxun Wang, Yuhan Chen, Xiaonan Jiang, Xukun Zhu, Yubo Dong, and Linchao Zhu.
\newblock {No Free Checker: A Survey of Verifiers for Robot Policies}.
\newblock arXiv preprint arXiv:2609.09250, 2026.

\bibitem[Wang et~al.(2026{\natexlab{a}})Wang, Wang, Pei, Zhang, Liang, Hu, Li, Wu, Han, Zhang, Qi, Wu, Zhang, Zheng, Pan, Navarro-Alarcon, Liu, and Zhou]{arxiv260600113}
Fangyuan Wang, Ziyuan Wang, Guorui Pei, Mengshi Zhang, Canxi Liang, Jun Hu, Zhongxuan Li, Jinsong Wu, Ning Han, Zeqing Zhang, Jiaming Qi, Hongmin Wu, Shiyao Zhang, Pai Zheng, Jia Pan, David Navarro-Alarcon, Sichao Liu, and Peng Zhou.
\newblock {World Models for Robotic Manipulation: A Survey}.
\newblock arXiv preprint arXiv:2606.00113, 2026{\natexlab{a}}.

\bibitem[Wang et~al.(2024{\natexlab{a}})Wang, Xie, Jiang, Mandlekar, Xiao, Zhu, Fan, and Anandkumar]{arxiv230516291}
Guanzhi Wang, Yuqi Xie, Yunfan Jiang, Ajay Mandlekar, Chaowei Xiao, Yuke Zhu, Linxi Fan, and Anima Anandkumar.
\newblock {Voyager: An Open-Ended Embodied Agent with Large Language Models}.
\newblock \emph{Transactions on Machine Learning Research}, 2024{\natexlab{a}}.

\bibitem[Wang et~al.(2016)Wang, Kurth-Nelson, Tirumala, Soyer, Leibo, Munos, Blundell, Kumaran, and Botvinick]{wang2016learningrl}
Jane~X Wang, Zeb Kurth-Nelson, Dhruva Tirumala, Hubert Soyer, Joel~Z Leibo, Remi Munos, Charles Blundell, Dharshan Kumaran, and Matt Botvinick.
\newblock {Learning to reinforcement learn}.
\newblock arXiv preprint arXiv:1611.05763, 2016.

\bibitem[Wang(2026)]{arxiv260931760}
Jiaming Wang.
\newblock {What Stops Recursive Self-Improvement in Robotics? Lessons from 123 Rounds of Agentic Skill Discovery}.
\newblock arXiv preprint arXiv:2609.31760, 2026.

\bibitem[Wang et~al.(2025{\natexlab{a}})Wang, Ma, Cao, Zheng, Feng, Zhang, Pang, and Liang]{arxiv250308153}
Jing Wang, Ao~Ma, Ke~Cao, Jun Zheng, Jiasong Feng, Zhanjie Zhang, Wanyuan Pang, and Xiaodan Liang.
\newblock {WISA: World Simulator Assistant for Physics-Aware Text-to-Video Generation}.
\newblock In \emph{Advances in Neural Information Processing Systems}, volume~38, pages 6144--6172, 2025{\natexlab{a}}.
\newblock \doi{10.52202/085713-0191}.

\bibitem[Wang et~al.(2026{\natexlab{b}})Wang, Jie, Yan, Heng, and Li]{arxiv260927612}
Kailin Wang, Haoxiang Jie, Yaoyuan Yan, Zhiyou Heng, and Zhaosong Li.
\newblock {RegenHarness: A Robot Agent Harness with Evidence-Gated Recursive Self-Improvement}.
\newblock arXiv preprint arXiv:2609.27612, 2026{\natexlab{b}}.

\bibitem[Wang et~al.(2024{\natexlab{b}})Wang, Ling, Yuan, Shridhar, Bao, Qin, Wang, Xu, and Wang]{wang2024gensim}
Lirui Wang, Yiyang Ling, Zhecheng Yuan, Mohit Shridhar, Chen Bao, Yuzhe Qin, Bailin Wang, Huazhe Xu, and Xiaolong Wang.
\newblock {GenSim: Generating Robotic Simulation Tasks via Large Language Models}.
\newblock In \emph{International Conference on Learning Representations}, 2024{\natexlab{b}}.

\bibitem[Wang et~al.(2026{\natexlab{c}})Wang, Osterberg, Tian, Shorinwa, Bowkett, Pastor, Wu, and Schwager]{arxiv260624884}
Maggie Wang, Lars Osterberg, Stephen Tian, Ola Shorinwa, Joseph Bowkett, Daniel Pastor, Jiajun Wu, and Mac Schwager.
\newblock {InSight: Self-Guided Skill Acquisition via Steerable VLAs}.
\newblock arXiv preprint arXiv:2606.24884, 2026{\natexlab{c}}.

\bibitem[Wang et~al.(2026{\natexlab{d}})Wang, Chang, Ma, Yang, Feng, Liu, Yang, Cao, Wang, Su, and Wang]{arxiv261020474}
Peidong Wang, Ying Chang, Zhiming Ma, Xiaocui Yang, Shi Feng, Daqi Liu, Yuzhe Yang, Xu~Cao, Chuang Wang, Jian Su, and Daling Wang.
\newblock {AdaRSI: On-Demand Recursive Self-Improvement for Embodied Agents}.
\newblock arXiv preprint arXiv:2610.20474, 2026{\natexlab{d}}.

\bibitem[Wang et~al.(2026{\natexlab{e}})Wang, Ma, Chang, Luo, Zhang, Wang, Yang, Feng, Yang, and Li]{arxiv260811350}
Peidong Wang, Zhiming Ma, Ying Chang, Xufang Luo, Yiqun Zhang, Zihan Wang, Xiaocui Yang, Shi Feng, Yuqing Yang, and Dongsheng Li.
\newblock {Self-Evolving Embodied Agents via Skill-Harness Evolution}.
\newblock arXiv preprint arXiv:2608.11350, 2026{\natexlab{e}}.

\bibitem[Wang et~al.(2020)Wang, Lehman, Rawal, Zhi, Li, Clune, and Stanley]{wang2020poet}
Rui Wang, Joel Lehman, Aditya Rawal, Jiale Zhi, Yulun Li, Jeff Clune, and Kenneth~O. Stanley.
\newblock {Enhanced POET: Open-Ended Reinforcement Learning through Unbounded Invention of Learning Challenges and Their Solutions}.
\newblock In \emph{Proceedings of the 37th International Conference on Machine Learning}, volume 119 of \emph{Proceedings of Machine Learning Research}, pages 9940--9951, 2020.

\bibitem[Wang et~al.(2025{\natexlab{b}})Wang, Song, Liu, Ma, Feng, Wang, Jiang, Chen, Zhan, Wang, Meng, Shi, He, Ren, Yang, and Yao]{arxiv250518793}
Wenhao Wang, Jianheng Song, Chiming Liu, Jiayao Ma, Siyuan Feng, Jingyuan Wang, Yuxin Jiang, Kylin Chen, Sikang Zhan, Yi~Wang, Tong Meng, Modi Shi, Xindong He, Guanghui Ren, Yang Yang, and Maoqing Yao.
\newblock {Genie Centurion: Accelerating Scalable Real-World Robot Training with Human Rewind-and-Refine Guidance}.
\newblock arXiv preprint arXiv:2505.18793, 2025{\natexlab{b}}.

\bibitem[Wang et~al.(2026{\natexlab{f}})Wang, Pi{\k{e}}kos, Nanbo, Laakom, Chen, Ostaszewski, Zhuge, and Schmidhuber]{arxiv251021614}
Wenyi Wang, Piotr Pi{\k{e}}kos, Li~Nanbo, Firas Laakom, Yimeng Chen, Mateusz Ostaszewski, Mingchen Zhuge, and J{\"u}rgen Schmidhuber.
\newblock {Huxley-G{\"o}del Machine: Human-Level Coding Agent Development by an Approximation of the Optimal Self-Improving Machine}.
\newblock In \emph{International Conference on Learning Representations (ICLR)}, 2026{\natexlab{f}}.

\bibitem[Wang et~al.(2026{\natexlab{g}})Wang, Wu, Zhang, Wang, Shao, Luan, Li, Li, Gu, Zhou, Shi, Ni, Lu, Meng, Li, Wu, Zhao, Guo, Li, Tang, Zhang, Xie, and Qu]{arxiv260929166}
Xin Wang, Wenhao Wu, Menghao Zhang, Zhi Wang, Kun Shao, Jian Luan, Yang Li, Qing Li, Shangding Gu, Huichi Zhou, Shuqing Shi, Fei Ni, Shuo Lu, Weicheng Meng, Kang Li, Jin Wu, Kang Zhao, Shangmin Guo, Gen Li, Yongqiang Tang, Zhizhong Zhang, Yuan Xie, and Heng Qu.
\newblock {HarnessPAI: An Evolving Harness for Physical AI}.
\newblock arXiv preprint arXiv:2609.29166, 2026{\natexlab{g}}.

\bibitem[Wang et~al.(2026{\natexlab{h}})Wang, Li, Wang, Zhang, Cui, Tang, Buch, and Xu]{arxiv261018792}
Xinyi Wang, Bangzheng Li, Hanyu Wang, Lin Zhang, Andrea~Yaoyun Cui, Zhenggang Tang, Shyamal Buch, and Dejia Xu.
\newblock {SEAR: Self-Evolving Agentic Robotics}.
\newblock arXiv preprint arXiv:2610.18792, 2026{\natexlab{h}}.

\bibitem[Wang et~al.(2026{\natexlab{i}})Wang, Jiang, Deng, Xue, Ye, Duan, Haghtalab, Sastry, Abbeel, and Qi]{arxiv261002204}
Yen-Jen Wang, Haozhe Jiang, Shuying Deng, Haoru Xue, Weirui Ye, Rocky Duan, Nika Haghtalab, S.~Shankar Sastry, Pieter Abbeel, and Haozhi Qi.
\newblock {Reconstruct, Practice, Go Real: Guided Self-Improvement for Embodied Agents}.
\newblock arXiv preprint arXiv:2610.02204, 2026{\natexlab{i}}.

\bibitem[Wang et~al.(2026{\natexlab{j}})Wang, Li, Xie, Yang, Nie, Cai, Zhang, Qu, Wu, Song, Ren, Huang, Pan, Feng, Chen, and Luo]{arxiv260500416}
Yi~Wang, Xinchen Li, Pengwei Xie, Pu~Yang, Buqing Nie, Yunuo Cai, Qinglin Zhang, Chendi Qu, Jeffrey Wu, Jianheng Song, Xinlin Ren, Jingshun Huang, Mingjie Pan, Siyuan Feng, Zhi Chen, and Jianlan Luo.
\newblock {Learning While Deploying: Fleet-Scale Reinforcement Learning for Generalist Robot Policies}.
\newblock arXiv preprint arXiv:2605.00416, 2026{\natexlab{j}}.

\bibitem[Wang et~al.(2024{\natexlab{c}})Wang, Sun, Zhang, Xian, B{\i}y{\i}k, Held, and Erickson]{wang2024rlvlmf}
Yufei Wang, Zhanyi Sun, Jesse Zhang, Zhou Xian, Erdem B{\i}y{\i}k, David Held, and Zackory Erickson.
\newblock {RL-VLM-F: Reinforcement Learning from Vision Language Foundation Model Feedback}.
\newblock In \emph{International Conference on Machine Learning}, volume 235 of \emph{Proceedings of Machine Learning Research}, pages 51484--51501, 2024{\natexlab{c}}.

\bibitem[Wang et~al.(2024{\natexlab{d}})Wang, Xian, Chen, Wang, Wang, Fragkiadaki, Erickson, Held, and Gan]{wang2024robogen}
Yufei Wang, Zhou Xian, Feng Chen, Tsun-Hsuan Wang, Yian Wang, Katerina Fragkiadaki, Zackory Erickson, David Held, and Chuang Gan.
\newblock {RoboGen: Towards Unleashing Infinite Data for Automated Robot Learning via Generative Simulation}.
\newblock In \emph{Proceedings of the 41st International Conference on Machine Learning}, 2024{\natexlab{d}}.

\bibitem[Wang et~al.(2026{\natexlab{k}})Wang, Ge, Wang, Wang, Xu, Frandsen, Fu, Wilson, and Reddy]{arxiv261002616}
Zekai Wang, Yingqiang Ge, Zekun Wang, Hai Wang, Yuhui Xu, Joshua Frandsen, Shancong Fu, Ashia~C. Wilson, and Chandan~K. Reddy.
\newblock {VERSE: Verified Self-Evolving Optimizer for Agent Harnesses}.
\newblock arXiv preprint arXiv:2610.02616, 2026{\natexlab{k}}.

\bibitem[Wang and Wang(2026)]{arxiv260815088}
Zihang Wang and Yishan Wang.
\newblock {Max-Q Selective Imitation for Human-in-the-Loop Online Robot Learning with Monte Carlo Q-Chunk Critics}.
\newblock arXiv preprint arXiv:2608.15088, 2026.

\bibitem[Wei et~al.(2026{\natexlab{a}})Wei, Bai, Yue, Wang, Guo, Chen, Pu, Wu, Yue, Li, Qiu, Liu, Ge, Zhou, Chen, and Ge]{arxiv260917372}
Jiacheng Wei, Jerry Bai, Xiaoyu Yue, Zidong Wang, Xiaoyang Guo, Cheng Chen, Fanqi Pu, Fan Wu, Zhixu Yue, Yizhuo Li, Feng Qiu, Bo~Liu, Yuying Ge, Hui Zhou, Chenyi Chen, and Yixiao Ge.
\newblock {XPACE: Joint World and Action Modeling from Heterogeneous Experience}.
\newblock arXiv preprint arXiv:2609.17372, 2026{\natexlab{a}}.

\bibitem[Wei et~al.(2026{\natexlab{b}})Wei, Tan, and Wang]{arxiv260939304}
Zhijie Wei, Ferris Tan, and Jinghui Wang.
\newblock {Scale and Selection: What Makes Automatic Harness Evolution Work for Visual-Interface Robot Agents}.
\newblock arXiv preprint arXiv:2609.39304, 2026{\natexlab{b}}.

\bibitem[Wen and Wang(2026)]{arxiv261017114}
Xuan Wen and Baichao Wang.
\newblock {D-World: An Explicit World-Language-Action Model with Recursive Self-Improvement under Dual Control}.
\newblock arXiv preprint arXiv:2610.17114, 2026.

\bibitem[Wen et~al.(2026)Wen, Chen, Cao, Chen, Zou, Yu, Kuang, Xue, Lv, Wen, and Lu]{arxiv261012424}
Zimo Wen, Yijin Chen, Yuxuan Cao, Wendi Chen, Yanwen Zou, Wenye Yu, Fuhang Kuang, Han Xue, Jun Lv, Chuan Wen, and Cewu Lu.
\newblock {RoboRSI: Stable, Efficient, and Reusable Robot Self-Evolution in Complex Real-World Environments}.
\newblock arXiv preprint arXiv:2610.12424, 2026.

\bibitem[Wu et~al.(2026)Wu, Wang, Lu, Wang, Xiong, Wang, Yu, Ma, and Wang]{arxiv260933653}
Duo Wu, Haifeng Wang, Rongwei Lu, Jinghe Wang, Tianyi Xiong, Zhimin Wang, Chao Yu, Shuai Ma, and Zhi Wang.
\newblock {Demonstration-Free Success-Probability Reward Learning for Generalist Robot Policies}.
\newblock arXiv preprint arXiv:2609.33653, 2026.

\bibitem[Wu et~al.(2023)Wu, Escontrela, Hafner, Abbeel, and Goldberg]{wu2023daydreamer}
Philipp Wu, Alejandro Escontrela, Danijar Hafner, Pieter Abbeel, and Ken Goldberg.
\newblock {DayDreamer: World Models for Physical Robot Learning}.
\newblock In \emph{Proceedings of the 6th Conference on Robot Learning}, volume 205 of \emph{Proceedings of Machine Learning Research}, pages 2226--2240, 2023.

\bibitem[Wu et~al.(2025)Wu, Lu, Chen, Liu, Zhuang, and Hu]{arxiv250204692}
Zhenwei Wu, Jinxiong Lu, Yuxiao Chen, Yunxin Liu, Yueting Zhuang, and Luhui Hu.
\newblock {STRIDE: Automating Reward Design, Deep Reinforcement Learning Training and Feedback Optimization in Humanoid Robotics Locomotion}.
\newblock arXiv preprint arXiv:2502.04692, 2025.

\bibitem[Xiang et~al.(2025)Xiang, Zhang, Huang, He, Liu, Tang, Zhou, Luo, Wen, Chen, Lin, Han, Xu, Li, Dong, and Liang]{arxiv251004978}
Kun Xiang, Terry~Jingchen Zhang, Yinya Huang, Jixi He, Zirong Liu, Yueling Tang, Ruizhe Zhou, Lijing Luo, Youpeng Wen, Xiuwei Chen, Bingqian Lin, Jianhua Han, Hang Xu, Hanhui Li, Bin Dong, and Xiaodan Liang.
\newblock {Aligning Perception, Reasoning, Modeling and Interaction: A Survey on Physical AI}.
\newblock arXiv preprint arXiv:2510.04978, 2025.

\bibitem[Xiao et~al.(2026{\natexlab{a}})Xiao, Lin, Peng, Xue, He, Luo, Xie, Hu, Fan, Shi, and Zhu]{arxiv251100091}
Wenli Xiao, Haotian Lin, Andy Peng, Haoru Xue, Tairan He, Zhengyi Luo, Yuqi Xie, Fengyuan Hu, Linxi~"Jim" Fan, Guanya Shi, and Yuke Zhu.
\newblock {Self-Improving Vision-Language-Action Models with Data Generation via Residual RL}.
\newblock In \emph{International Conference on Learning Representations}, 2026{\natexlab{a}}.

\bibitem[Xiao et~al.(2026{\natexlab{b}})Xiao, Xie, Zhang, Lin, Fu, Xue, Lu, Yang, Dai, Wang, Wu, Wang, Sastry, Goldberg, Fan, Zhu, and Shi]{arxiv260619980}
Wenli Xiao, Jia Xie, Tonghe Zhang, Haotian Lin, Letian~"Max" Fu, Haoru Xue, Jalen Lu, Yi~Yang, Cunxi Dai, Zi~Wang, Jimmy Wu, Guanzhi Wang, S.~Shankar Sastry, Ken Goldberg, Linxi~"Jim" Fan, Yuke Zhu, and Guanya Shi.
\newblock {ENPIRE: Agentic Robot Policy Self-Improvement in the Real World}.
\newblock arXiv preprint arXiv:2606.19980, 2026{\natexlab{b}}.
\newblock Accepted at CoRL 2026; proceedings forthcoming.

\bibitem[Xie and Finn(2022)]{xie2022lifelong}
Annie Xie and Chelsea Finn.
\newblock {Lifelong Robotic Reinforcement Learning by Retaining Experiences}.
\newblock In \emph{Proceedings of the 1st Conference on Lifelong Learning Agents}, volume 199 of \emph{Proceedings of Machine Learning Research}, pages 838--855, 2022.

\bibitem[Xie et~al.(2026{\natexlab{a}})Xie, Ruan, Wang, Zhang, Yang, Jin, and Shi]{arxiv261008640}
Jing Xie, Shouwei Ruan, Yubin Wang, Yuxiang Zhang, Haitao Yang, Songchang Jin, and Dianxi Shi.
\newblock {RIWANav: Recursive World--Action Models with Self-Improvement for Urban Navigation}.
\newblock arXiv preprint arXiv:2610.08640, 2026{\natexlab{a}}.

\bibitem[Xie et~al.(2020)Xie, Luong, Hovy, and Le]{xie2020noisystudent}
Qizhe Xie, Minh-Thang Luong, Eduard Hovy, and Quoc~V. Le.
\newblock {Self-training with Noisy Student improves ImageNet classification}.
\newblock In \emph{Proceedings of the IEEE/CVF Conference on Computer Vision and Pattern Recognition}, 2020.

\bibitem[Xie et~al.(2026{\natexlab{b}})Xie, Zhang, Wang, and Chen]{arxiv260302623}
Senwei Xie, Yuntian Zhang, Ruiping Wang, and Xilin Chen.
\newblock {Uni-Skill: Building Self-Evolving Skill Repository for Generalizable Robotic Manipulation}.
\newblock In \emph{IEEE International Conference on Robotics and Automation}, 2026{\natexlab{b}}.
\newblock arXiv:2603.02623.

\bibitem[Xie et~al.(2026{\natexlab{c}})Xie, Chen, Cao, Lu, Wu, and Jiang]{arxiv260937810}
Sicheng Xie, Yitong Chen, Haidong Cao, Shunlin Lu, Zuxuan Wu, and Yu-Gang Jiang.
\newblock {Explore, Execute, Evolve: A Skill Acquisition and Reuse Loop for Embodied Agents}.
\newblock arXiv preprint arXiv:2609.37810, 2026{\natexlab{c}}.

\bibitem[Xie et~al.(2024)Xie, Zhao, Wu, Liu, Luo, Zhong, Yang, and Yu]{xie2024text2reward}
Tianbao Xie, Siheng Zhao, Chen~Henry Wu, Yitao Liu, Qian Luo, Victor Zhong, Yanchao Yang, and Tao Yu.
\newblock {Text2Reward}: Reward shaping with language models for reinforcement learning.
\newblock In \emph{International Conference on Learning Representations}, 2024.

\bibitem[Xie et~al.(2026{\natexlab{d}})Xie, Xu, Jiang, and Zhao]{arxiv261013991}
Zixuan Xie, Chang Xu, Yuhua Jiang, and Li~Zhao.
\newblock {Predict Before You Modify: Learning Improvement Dynamics for Self-Evolving Agents}.
\newblock arXiv preprint arXiv:2610.13991, 2026{\natexlab{d}}.

\bibitem[Xing et~al.(2026)Xing, Zhu, Ren, Geles, Zhai, Reiter, and Scaramuzza]{arxiv260627353}
Jiaxu Xing, Zhiyuan Zhu, Yunfan Ren, Ismail Geles, Yifan Zhai, Rudolf Reiter, and Davide Scaramuzza.
\newblock {Continual Robot Policy Learning via Variational Neural Dynamics}.
\newblock arXiv preprint arXiv:2606.27353, 2026.

\bibitem[Xiong et~al.(2026)Xiong, Hu, and Clune]{arxiv260207755}
Yiming Xiong, Shengran Hu, and Jeff Clune.
\newblock {Learning to Continually Learn via Meta-learning Agentic Memory Designs}.
\newblock arXiv preprint arXiv:2602.07755, 2026.

\bibitem[Xu et~al.(2026{\natexlab{a}})Xu, Shang, and Ferrara]{arxiv260816889}
Bingxin Xu, Yuzhang Shang, and Emilio Ferrara.
\newblock {Don't Drop the Baton: Long-Horizon Robot Manipulation via Agentic Subtask Exploration and Transition-Aware Memory}.
\newblock arXiv preprint arXiv:2608.16889, 2026{\natexlab{a}}.

\bibitem[Xu et~al.(2026{\natexlab{b}})Xu, Springenberg, Equi, Amin, Esmail, Levine, and Ke]{arxiv260423073}
Charles Xu, Jost~Tobias Springenberg, Michael Equi, Ali Amin, Adnan Esmail, Sergey Levine, and Liyiming Ke.
\newblock {RL Token: Bootstrapping Online RL with Vision-Language-Action Models}.
\newblock arXiv preprint arXiv:2604.23073, 2026{\natexlab{b}}.

\bibitem[Xu et~al.(2026{\natexlab{c}})Xu, Hu, Chen, Gao, Ye, and Shi]{arxiv261009134}
Jiuyi Xu, Xiao Hu, Meida Chen, Peng Gao, Yang Ye, and Yangming Shi.
\newblock {World Models Dream of Success: Diagnosing and Repairing Failure Insensitivity in Robot World Models}.
\newblock arXiv preprint arXiv:2610.09134, 2026{\natexlab{c}}.

\bibitem[Xu et~al.(2026{\natexlab{d}})Xu, Cheng, Wang, Huang, Xiao, Chen, Zhong, Liu, Li, Li, Lv, Zhang, Jiang, and Lin]{arxiv261016491}
Xin Xu, Ke~Cheng, Hao Wang, Yangru Huang, Yanghao Xiao, Zhichao Chen, Zhisheng Zhong, Xun Liu, Yi~Li, Qian Li, Haibin Lv, Jun Zhang, Jie Jiang, and Zhouchen Lin.
\newblock {Drafting Agents, Not Just Tokens: Speculative Recursive Self-Improvement for Agentic Systems}.
\newblock arXiv preprint arXiv:2610.16491, 2026{\natexlab{d}}.

\bibitem[Xu et~al.(2018)Xu, van Hasselt, and Silver]{xu2018metagradient}
Zhongwen Xu, Hado van Hasselt, and David Silver.
\newblock {Meta-Gradient Reinforcement Learning}.
\newblock In \emph{Advances in Neural Information Processing Systems}, volume~31, 2018.

\bibitem[Xue et~al.(2026{\natexlab{a}})Xue, Ding, Shen, Wang, Yin, Wu, Chen, Wang, and Yang]{arxiv260804003}
Shuhan Xue, Zixin Ding, Yichen Shen, Yinjie Wang, Zhenfei Yin, Yingcheng Wu, Yuxin Chen, Mengdi Wang, and Ling Yang.
\newblock {PAST-Bench: Benchmarking the Foundations of Recursive Self-Improvement in Personal Agents}.
\newblock arXiv preprint arXiv:2608.04003, 2026{\natexlab{a}}.

\bibitem[Xue et~al.(2026{\natexlab{b}})Xue, Xu, Liu, Wu, Gu, Song, Jia, and Wang]{arxiv260702431}
Yuquan Xue, Le~Xu, Zeyi Liu, Zhenyu Wu, Zhengyi Gu, Xinyang Song, Bofang Jia, and Ziwei Wang.
\newblock {WorldSample: Closed-loop Real-robot RL with World Modelling}.
\newblock arXiv preprint arXiv:2607.02431, 2026{\natexlab{b}}.

\bibitem[Yadav et~al.(2026)Yadav, Zhou, Wagenmaker, Pertsch, and Levine]{arxiv251208333}
Yajat Yadav, Zhiyuan Zhou, Andrew Wagenmaker, Karl Pertsch, and Sergey Levine.
\newblock {Robust Finetuning of Vision-Language-Action Robot Policies via Parameter Merging}.
\newblock In \emph{International Conference on Learning Representations}, 2026.

\bibitem[Yan et~al.(2026)Yan, Zhu, Jia, Yin, He, Zhong, Li, Lu, Li, Zhang, Chen, Song, Chen, Gao, and Li]{arxiv260806799}
Haodong Yan, Jiaguan Zhu, Mingyuan Jia, Ruiqing Yin, Junjie He, Zhide Zhong, Junfeng Li, Jinxuan Lu, Hengtao Li, Tianran Zhang, Jiayi Chen, Wenxuan Song, Wen Chen, Yuxiang Gao, and Haoang Li.
\newblock {Is Forward Prediction Enough? Physical State Grounding for JEPA World Models}.
\newblock arXiv preprint arXiv:2608.06799, 2026.

\bibitem[Yang et~al.(2026{\natexlab{a}})Yang, Zhang, Zhu, Chen, and Jiang]{arxiv260922888}
Jiarui Yang, Jiajin Zhang, Bin Zhu, Jingjing Chen, and Yu-Gang Jiang.
\newblock {Stable and Efficient Real-World Online VLA Post-Training via Asynchronous Replay-Anchored Policy Improvement}.
\newblock arXiv preprint arXiv:2609.22888, 2026{\natexlab{a}}.

\bibitem[Yang et~al.(2026{\natexlab{b}})Yang, Lin, Li, Zhang, Lin, Wu, et~al.]{arxiv260211075}
Jiazhi Yang, Kunyang Lin, Jinwei Li, Wencong Zhang, Tianwei Lin, Longyan Wu, et~al.
\newblock {RISE: Self-Improving Robot Policy with Compositional World Model}.
\newblock In \emph{Robotics: Science and Systems}, 2026{\natexlab{b}}.
\newblock arXiv:2602.11075.

\bibitem[Yang et~al.(2024{\natexlab{a}})Yang, Mark, Vu, Sharma, Bohg, and Finn]{yang2023robofume}
Jingyun Yang, Max~Sobol Mark, Brandon Vu, Archit Sharma, Jeannette Bohg, and Chelsea Finn.
\newblock {Robot Fine-Tuning Made Easy: Pre-Training Rewards and Policies for Autonomous Real-World Reinforcement Learning}.
\newblock In \emph{IEEE International Conference on Robotics and Automation}, pages 4804--4811, 2024{\natexlab{a}}.
\newblock \doi{10.1109/ICRA57147.2024.10610421}.

\bibitem[Yang et~al.(2026{\natexlab{c}})Yang, Kong, and Jo]{arxiv260937968}
Jungwoo Yang, Injin Kong, and Yohan Jo.
\newblock {SelfSearch: Reward-Free Search for Self-Improving Agents}.
\newblock arXiv preprint arXiv:2609.37968, 2026{\natexlab{c}}.

\bibitem[Yang et~al.(2025)Yang, Chen, Zhang, Zhao, Qian, et~al.]{yang2025embodiedbench}
Rui Yang, Hanyang Chen, Junyu Zhang, Mark Zhao, Cheng Qian, et~al.
\newblock {EmbodiedBench: Comprehensive Benchmarking Multi-modal Large Language Models for Vision-Driven Embodied Agents}.
\newblock In \emph{International Conference on Machine Learning}, 2025.

\bibitem[Yang et~al.(2024{\natexlab{b}})Yang, Du, Ghasemipour, Tompson, Kaelbling, Schuurmans, and Abbeel]{yang2024unisim}
Sherry Yang, Yilun Du, Kamyar Ghasemipour, Jonathan Tompson, Leslie Kaelbling, Dale Schuurmans, and Pieter Abbeel.
\newblock {Learning Interactive Real-World Simulators}.
\newblock In \emph{International Conference on Learning Representations}, 2024{\natexlab{b}}.

\bibitem[Yang et~al.(2026{\natexlab{d}})Yang, Liao, Mei, Wang, Yang, Wen, Zhang, Li, Lv, Chen, Shi, Liu, Yan, and Lee]{arxiv260308403}
Yu~Yang, Yue Liao, Jianbiao Mei, Baisen Wang, Xuemeng Yang, Licheng Wen, Jiangning Zhang, Xiangtai Li, Liang Lv, Hanlin Chen, Botian Shi, Yong Liu, Shuicheng Yan, and Gim~Hee Lee.
\newblock {SPIRAL: Self-Evolving Action-Conditioned Video Generation via Reflective Planning Agents}.
\newblock arXiv preprint arXiv:2603.08403, 2026{\natexlab{d}}.

\bibitem[Yang et~al.(2026{\natexlab{e}})Yang, Bhatt, Wang, Tetteh, Wang, and Topcu]{arxiv260605395}
Yunhao Yang, Neel~P. Bhatt, Kevin Wang, Samuel Tetteh, Zhangyang Wang, and Ufuk Topcu.
\newblock {VASO: Formally Verifiable Self-Evolving Skills for Physical AI Agents}.
\newblock arXiv preprint arXiv:2606.05395, 2026{\natexlab{e}}.

\bibitem[Yin et~al.(2026{\natexlab{a}})Yin, Liang, Zhang, Liu, Li, Zhang, and Wang]{arxiv260610927}
Hang Yin, Yinan Liang, Jiazhao Zhang, Jiahang Liu, Minghan Li, Zhizheng Zhang, and He~Wang.
\newblock {AllDayNav: Lifelong Navigation via Real-World Reinforcement Learning}.
\newblock arXiv preprint arXiv:2606.10927, 2026{\natexlab{a}}.

\bibitem[Yin et~al.(2026{\natexlab{b}})Yin, Mei, Zheng, Yamane, Wang, Sceats, Bateman, Zha, Badithela, Shorinwa, and Majumdar]{arxiv260309030}
Tenny Yin, Zhiting Mei, Zhonghe Zheng, Miyu Yamane, David Wang, Jade Sceats, Samuel~M. Bateman, Lihan Zha, Apurva Badithela, Ola Shorinwa, and Anirudha Majumdar.
\newblock {PlayWorld: Learning Robot World Models from Autonomous Play}.
\newblock arXiv preprint arXiv:2603.09030, 2026{\natexlab{b}}.

\bibitem[Yin et~al.(2025)Yin, Wang, Pan, Lin, Wan, and Wang]{arxiv241004444}
Xunjian Yin, Xinyi Wang, Liangming Pan, Li~Lin, Xiaojun Wan, and William~Yang Wang.
\newblock {G\"odel Agent: A Self-Referential Agent Framework for Recursively Self-Improvement}.
\newblock In \emph{Proceedings of the 63rd Annual Meeting of the Association for Computational Linguistics (Volume 1: Long Papers)}, pages 27890--27913, 2025.

\bibitem[Yu(2026)]{arxiv260807555}
Hang Yu.
\newblock {You Don't Need To Stay in The Loop: An Agentic Robotics Loop for Robot-Policy Improvement}.
\newblock arXiv preprint arXiv:2608.07555, 2026.

\bibitem[Yu et~al.(2018)Yu, Finn, Dasari, Xie, Zhang, Abbeel, and Levine]{yu2018daml}
Tianhe Yu, Chelsea Finn, Sudeep Dasari, Annie Xie, Tianhao Zhang, Pieter Abbeel, and Sergey Levine.
\newblock {One-Shot Imitation from Observing Humans via Domain-Adaptive Meta-Learning}.
\newblock In \emph{Robotics: Science and Systems}, 2018.

\bibitem[Yu et~al.(2019)Yu, Quillen, He, Julian, et~al.]{yu2019metaworld}
Tianhe Yu, Deirdre Quillen, Zhanpeng He, Ryan Julian, et~al.
\newblock {Meta-World: A Benchmark and Evaluation for Multi-Task and Meta Reinforcement Learning}.
\newblock In \emph{Conference on Robot Learning}, 2019.

\bibitem[Yu et~al.(2023)Yu, Gileadi, Fu, Kirmani, Lee, {Gonzalez Arenas}, Chiang, Erez, Hasenclever, Humplik, Ichter, Xiao, Xu, Zeng, Zhang, Heess, Sadigh, Tan, Tassa, and Xia]{yu2023l2r}
Wenhao Yu, Nimrod Gileadi, Chuyuan Fu, Sean Kirmani, Kuang-Huei Lee, Montserrat {Gonzalez Arenas}, Hao-Tien~Lewis Chiang, Tom Erez, Leonard Hasenclever, Jan Humplik, Brian Ichter, Ted Xiao, Peng Xu, Andy Zeng, Tingnan Zhang, Nicolas Heess, Dorsa Sadigh, Jie Tan, Yuval Tassa, and Fei Xia.
\newblock {Language to Rewards for Robotic Skill Synthesis}.
\newblock In \emph{Conference on Robot Learning}, volume 229 of \emph{Proceedings of Machine Learning Research}, pages 374--404, 2023.

\bibitem[Yu et~al.(2026{\natexlab{a}})Yu, Zhu, Li, Wang, Zhou, and Chen]{arxiv261015814}
Yue Yu, Bin Zhu, Junhui Li, Jiayu Wang, Shijie Zhou, and Jingjing Chen.
\newblock {ReTrace: Cross-Stage Failure Attribution for Self-Improving Robotic Manipulation}.
\newblock arXiv preprint arXiv:2610.15814, 2026{\natexlab{a}}.

\bibitem[Yu et~al.(2026{\natexlab{b}})Yu, Wang, Liu, Ren, Yu, Bai, Zhang, Jia, Jia, Zhang, Zhang, Shang, Chen, Zhou, Wang, and Yang]{arxiv260935575}
Zhuoyuan Yu, Jiacheng Wang, Tianle Liu, Yihua Ren, Peng Yu, Chen Bai, Ziheng Zhang, Yufei Jia, Jindou Jia, Yuhang Zhang, Xinrui Zhang, Yujing Shang, Yuxiang Chen, Chuhao Zhou, Tiancai Wang, and Jianfei Yang.
\newblock {F4R: Failure-Driven Recognition, Reconstruction, Refinement, and Redeployment for Continual Robot Self-Improvement}.
\newblock arXiv preprint arXiv:2609.35575, 2026{\natexlab{b}}.

\bibitem[Yuan et~al.(2024)Yuan, Pang, Cho, Li, Sukhbaatar, Xu, and Weston]{yuan2024selfrewarding}
Weizhe Yuan, Richard~Yuanzhe Pang, Kyunghyun Cho, Xian Li, Sainbayar Sukhbaatar, Jing Xu, and Jason Weston.
\newblock {Self-Rewarding Language Models}.
\newblock In \emph{Proceedings of the 41st International Conference on Machine Learning}, volume 235 of \emph{Proceedings of Machine Learning Research}, pages 57905--57923, 2024.

\bibitem[Zelikman et~al.(2022)Zelikman, Wu, Mu, and Goodman]{zelikman2022star}
Eric Zelikman, Yuhuai Wu, Jesse Mu, and Noah~D. Goodman.
\newblock {STaR: Bootstrapping Reasoning With Reasoning}.
\newblock In \emph{Advances in Neural Information Processing Systems}, volume~35, 2022.
\newblock \doi{10.52202/068431-1126}.

\bibitem[Zelikman et~al.(2024)Zelikman, Lorch, Mackey, and Kalai]{zelikman2024stop}
Eric Zelikman, Eliana Lorch, Lester Mackey, and Adam~Tauman Kalai.
\newblock {Self-Taught Optimizer (STOP): Recursively Self-Improving Code Generation}.
\newblock In \emph{Conference on Language Modeling}, 2024.

\bibitem[Zhai et~al.(2025)Zhai, Zhang, Zhang, Huang, Zhang, Zhou, Zhang, Liu, Lin, and Pang]{arxiv250915937}
Shaopeng Zhai, Qi~Zhang, Tianyi Zhang, Fuxian Huang, Haoran Zhang, Ming Zhou, Shengzhe Zhang, Litao Liu, Sixu Lin, and Jiangmiao Pang.
\newblock {A Vision-Language-Action-Critic Model for Robotic Real-World Reinforcement Learning}.
\newblock arXiv preprint arXiv:2509.15937, 2025.

\bibitem[Zhang et~al.(2025{\natexlab{a}})Zhang, Zhou, and Vasudevan]{arxiv250421486}
Bohao Zhang, Zichang Zhou, and Ram Vasudevan.
\newblock {Provably-Safe, Online System Identification}.
\newblock In \emph{Robotics: Science and Systems}, 2025{\natexlab{a}}.
\newblock \doi{10.15607/RSS.2025.XXI.121}.

\bibitem[Zhang et~al.(2026{\natexlab{a}})Zhang, Zhao, Cheng, Pan, and Zeng]{arxiv260903681}
Chenhao Zhang, Hanyu Zhao, Hang Cheng, Tengfei Pan, and Long Zeng.
\newblock {WISE: World-model-guided Imagination Scheduling for Efficient Post-training of Vision-Language-Action Models}.
\newblock arXiv preprint arXiv:2609.03681, 2026{\natexlab{a}}.

\bibitem[Zhang et~al.(2026{\natexlab{b}})Zhang, Ito, Hoshino, Ikehata, and Sato]{arxiv260902159}
Chuhan Zhang, Seiji Ito, Kenta Hoshino, Satoshi Ikehata, and Ikuro Sato.
\newblock {World-Coherent Decoding: Self-Verifying Test-Time Planning for World Action Models}.
\newblock arXiv preprint arXiv:2609.02159, 2026{\natexlab{b}}.

\bibitem[Zhang et~al.(2024)Zhang, Lehman, Stanley, and Clune]{zhang2024omni}
Jenny Zhang, Joel Lehman, Kenneth Stanley, and Jeff Clune.
\newblock {OMNI: Open-endedness via Models of human Notions of Interestingness}.
\newblock In \emph{International Conference on Learning Representations}, 2024.

\bibitem[Zhang et~al.(2026{\natexlab{c}})Zhang, Hu, Lu, Lange, and Clune]{zhang2025dgm}
Jenny Zhang, Shengran Hu, Cong Lu, Robert Lange, and Jeff Clune.
\newblock {Darwin G{\"o}del Machine: Open-Ended Evolution of Self-Improving Agents}.
\newblock In \emph{International Conference on Learning Representations}, 2026{\natexlab{c}}.

\bibitem[Zhang et~al.(2026{\natexlab{d}})Zhang, Zhao, Yang, Foerster, Clune, Jiang, Devlin, and Shavrina]{arxiv260319461}
Jenny Zhang, Bingchen Zhao, Wannan Yang, Jakob Foerster, Jeff Clune, Minqi Jiang, Sam Devlin, and Tatiana Shavrina.
\newblock {HyperAgents}.
\newblock arXiv preprint arXiv:2603.19461, 2026{\natexlab{d}}.

\bibitem[Zhang et~al.(2025{\natexlab{b}})Zhang, Huang, Gu, Ma, and Zhang]{arxiv251120633}
Jiahui Zhang, Ze~Huang, Chun Gu, Zipei Ma, and Li~Zhang.
\newblock {Reinforcing Action Policies by Prophesying}.
\newblock arXiv preprint arXiv:2511.20633, 2025{\natexlab{b}}.

\bibitem[Zhang et~al.(2025{\natexlab{c}})Zhang, Luo, Anwar, Sontakke, Lim, Thomason, Biyik, and Zhang]{arxiv250510911}
Jiahui Zhang, Yusen Luo, Abrar Anwar, Sumedh~Anand Sontakke, Joseph~J Lim, Jesse Thomason, Erdem Biyik, and Jesse Zhang.
\newblock {ReWiND: Language-Guided Rewards Teach Robot Policies without New Demonstrations}.
\newblock In \emph{Conference on Robot Learning}, 2025{\natexlab{c}}.

\bibitem[Zhang et~al.(2026{\natexlab{e}})Zhang, Zhang, Song, Zhou, Song, and Wang]{arxiv260932416}
Jiawei Zhang, Xiangrong Zhang, Rui Song, Huanbin Zhou, Chengye Song, and Hongzhou Wang.
\newblock {RE-0: Verified Recursive Improvement of Embodied Code-as-Policy Agents through Local On-Policy Distillation}.
\newblock arXiv preprint arXiv:2609.32416, 2026{\natexlab{e}}.

\bibitem[Zhang et~al.(2026{\natexlab{f}})Zhang, Ge, Yoo, Fu, Yang, Liu, Saravanan, Yin, Yu, Niu, Wang, Herzig, Goldberg, Bai, Chan, Stoica, Kanazawa, Lei, Feng, and Darrell]{arxiv260619419}
Junyi Zhang, Jiaxin Ge, Hanjun Yoo, Letian Fu, Zihan Yang, Yaowei Liu, Raj Saravanan, Shaofeng Yin, Justin Yu, Dantong Niu, Zirui Wang, Roei Herzig, Ken Goldberg, Yutong Bai, David~M. Chan, Ion Stoica, Angjoo Kanazawa, Jiahui Lei, Haiwen Feng, and Trevor Darrell.
\newblock {Playful Agentic Robot Learning}.
\newblock arXiv preprint arXiv:2606.19419, 2026{\natexlab{f}}.

\bibitem[Zhang et~al.(2026{\natexlab{g}})Zhang, Tang, Zhan, Chen, Chen, Han, Zhu, Xu, Xu, Wu, Lin, and Liang]{arxiv260307648}
Likui Zhang, Tao Tang, Zhihao Zhan, Xiuwei Chen, Zisheng Chen, Jianhua Han, Jiangtong Zhu, Pei Xu, Hang Xu, Hefeng Wu, Liang Lin, and Xiaodan Liang.
\newblock {AtomicVLA: Unlocking the Potential of Atomic Skill Learning in Robots}.
\newblock In \emph{IEEE/CVF Conference on Computer Vision and Pattern Recognition}, pages 20743--20754, 2026{\natexlab{g}}.

\bibitem[Zhang and Shah(2026)]{arxiv260618247}
Mingtong Zhang and Dhruv Shah.
\newblock {Visual Verification Enables Inference-time Steering and Autonomous Policy Improvement}.
\newblock In \emph{Robotics: Science and Systems}, 2026.
\newblock \doi{10.15607/RSS.2026.XXII.079}.

\bibitem[Zhang et~al.(2025{\natexlab{d}})Zhang, Xu, Liu, Yu, Li, et~al.]{zhang2024vlabench}
Shiduo Zhang, Zhe Xu, Peiju Liu, Xiaopeng Yu, Yuan Li, et~al.
\newblock {VLABench: A Large-Scale Benchmark for Language-Conditioned Robotics Manipulation with Long-Horizon Reasoning Tasks}.
\newblock In \emph{IEEE/CVF International Conference on Computer Vision}, pages 11142--11152, 2025{\natexlab{d}}.
\newblock \doi{10.1109/ICCV51701.2025.01037}.

\bibitem[Zhang et~al.(2026{\natexlab{h}})Zhang, Cheng, Gao, Peng, Zheng, Tu, Wang, Pan, Hu, and Liu]{arxiv260803872}
Shuoqin Zhang, Tongtong Cheng, Xiru Gao, Jinzhuo Peng, Bin Zheng, Jiahao Tu, Ke~Wang, Jia Pan, Zhe Hu, and Kai Liu.
\newblock {EvoHIL: Self-Evolving Reward and Flow-Matched Policy Optimization for Robust Human-in-the-Loop Reinforcement Learning}.
\newblock arXiv preprint arXiv:2608.03872, 2026{\natexlab{h}}.

\bibitem[Zhang et~al.(2026{\natexlab{i}})Zhang, Wang, Bauer, Yu, Huang, Wu, and Pavone]{arxiv261009228}
Xilun Zhang, Maggie Wang, Erik Bauer, Hong-Xing Yu, Huang Huang, Jiajun Wu, and Marco Pavone.
\newblock {Co-Evolving Robot Orchestrators and Policies through Deployment}.
\newblock arXiv preprint arXiv:2610.09228, 2026{\natexlab{i}}.

\bibitem[Zhang et~al.(2026{\natexlab{j}})Zhang, Zhang, Gao, Li, Liu, Nie, Zhu, Qiu, Yan, Liu, Tang, Rao, Fang, Wei, Wang, Ding, and Yu]{arxiv260708448}
Yixian Zhang, Huanming Zhang, Feng Gao, Xiao Li, Zhihao Liu, Yi~Nie, Chunyang Zhu, Jiaxing Qiu, Yuchen Yan, Jiyuan Liu, Wenhao Tang, Jiaji Rao, Zhengru Fang, Changxu Wei, Yu~Wang, Wenbo Ding, and Chao Yu.
\newblock {Harness VLA: Steering Frozen VLAs into Reliable Manipulation Primitives via Memory-Guided Agents}.
\newblock arXiv preprint arXiv:2607.08448, 2026{\natexlab{j}}.

\bibitem[Zhang et~al.(2026{\natexlab{k}})Zhang, Liu, and Zhang]{arxiv261005151}
Yuzhi Zhang, Xinyu Liu, and Yu~Zhang.
\newblock {Recursive Self-Improvement of Visuomotor Policies through Local Recovery Supervision}.
\newblock arXiv preprint arXiv:2610.05151, 2026{\natexlab{k}}.

\bibitem[Zhao et~al.(2026{\natexlab{a}})Zhao, Chen, Hsu, Hwang, Srikanth, Liang, Saran, Tabrez, Nikolaidis, and Seita]{arxiv261014415}
Shihan Zhao, Henry Chen, Ya-Chuan Hsu, Minjune Hwang, Siddharth Srikanth, Freddie Liang, Akanksha Saran, Aaquib Tabrez, Stefanos Nikolaidis, and Daniel Seita.
\newblock {Red-Teaming of Vision-Language-Action Models via Quality Diversity Environment Generation}.
\newblock arXiv preprint arXiv:2610.14415, 2026{\natexlab{a}}.

\bibitem[Zhao et~al.(2026{\natexlab{b}})Zhao, Yan, Wan, Du, Chi, Mao, Wu, Yang, Li, Yang, Wu, Yan, Liu, Li, Ren, Shan, and Yao]{arxiv260927450}
Weihui Zhao, Xiaohan Yan, Zunian Wan, Xuan Du, Zhaozhan Chi, Jianbo Mao, Ruipu Wu, Rushuai Yang, Houlin Li, Shukai Yang, Jing Wu, Yuxiang Yan, Yongcheng Liu, Chuankang Li, Guanghui Ren, Wei Shan, and Maoqing Yao.
\newblock {BEE: Intervention-Adaptive Real-World Reinforcement Learning with Vision-Language-Action Models}.
\newblock arXiv preprint arXiv:2609.27450, 2026{\natexlab{b}}.

\bibitem[Zhao et~al.(2024)Zhao, Weber, and Wermter]{arxiv240515019}
Xufeng Zhao, Cornelius Weber, and Stefan Wermter.
\newblock {Agentic Skill Discovery}.
\newblock arXiv preprint arXiv:2405.15019, 2024.

\bibitem[Zhao et~al.(2025)Zhao, Jin, Jiang, Zhang, Wu, Ren, Xu, Che, Sun, Wu, Liu, and Tang]{arxiv251224288}
Yinuo Zhao, Huiqian Jin, Lechun Jiang, Xinyi Zhang, Kun Wu, Pei Ren, Zhiyuan Xu, Zhengping Che, Lei Sun, Dapeng Wu, Chi~Harold Liu, and Jian Tang.
\newblock {Real-world Reinforcement Learning from Suboptimal Interventions}.
\newblock arXiv preprint arXiv:2512.24288, 2025.

\bibitem[Zheng et~al.(2026{\natexlab{a}})Zheng, Yang, Ma, Xu, Xie, Li, Wang, Ma, Liu, Mao, and Liu]{arxiv260609337}
Huaihang Zheng, Yi~Yang, Kai Ma, Shenglin Xu, Tian Xie, Guozheng Li, Xiangyu Wang, Yiren Ma, Si~Liu, Yinian Mao, and Baoxu Liu.
\newblock {TORL-VLA: Tactile Guided Online Reinforcement Learning for Contact-Rich Manipulation}.
\newblock arXiv preprint arXiv:2606.09337, 2026{\natexlab{a}}.

\bibitem[Zheng et~al.(2025)Zheng, Cai, Wu, Wei, Tang, Wu, and Zheng]{arxiv250307087}
Zexin Zheng, Jia-Feng Cai, Xiao-Ming Wu, Yi-Lin Wei, Yu-Ming Tang, Ancong Wu, and Wei-Shi Zheng.
\newblock {iManip: Skill-Incremental Learning for Robotic Manipulation}.
\newblock In \emph{IEEE/CVF International Conference on Computer Vision}, pages 13890--13900, 2025.
\newblock arXiv:2503.07087.

\bibitem[Zheng et~al.(2026{\natexlab{b}})Zheng, Yu, Peng, shi, Li, Zhang, Li, Wang, Lu, and Jia]{arxiv260618960}
Zirui Zheng, Jiaqian Yu, Xiongfeng Peng, jun shi, Mingyi Li, Chao Zhang, Weiming Li, Dong Wang, Huchuan Lu, and Xu~Jia.
\newblock {Mem-World: Memory-Augmented Action-Conditioned World Models for Persistent Robot Manipulation}.
\newblock arXiv preprint arXiv:2606.18960, 2026{\natexlab{b}}.
\newblock Accepted at CoRL 2026; proceedings forthcoming.

\bibitem[Zhou et~al.(2025{\natexlab{a}})Zhou, Pan, LeCun, and Pinto]{zhou2025dinowm}
Gaoyue Zhou, Hengkai Pan, Yann LeCun, and Lerrel Pinto.
\newblock {DINO-WM: World Models on Pre-trained Visual Features enable Zero-shot Planning}.
\newblock In \emph{International Conference on Machine Learning}, volume 267 of \emph{Proceedings of Machine Learning Research}, pages 79115--79135, 2025{\natexlab{a}}.

\bibitem[Zhou et~al.(2026{\natexlab{a}})Zhou, Yang, Chen, Zhang, Fu, Li, Zhu, Liu, Yang, Meng, Wang, Han, Sun, Ye, Wang, Li, Zhou, Li, Jin, Shao, Chen, Chen, Ma, Zhou, Spanos, Hu, Schwager, Zhang, Feng, and Gu]{arxiv261018842}
Hai Zhou, Fan Yang, Zeyi Chen, Lejun Zhang, Chenyang Fu, Jianping Li, Jiayi Zhu, Xianyang Liu, Yingxuan Yang, Lili Meng, Lu~Wang, Xinyang Han, Jiankai Sun, Weirui Ye, Guanchu Wang, Wanhua Li, Ming Zhou, Boyi Li, Ming Jin, Kun Shao, Muhao Chen, Guang Chen, Lei Ma, Joey~Tianyi Zhou, Costas Spanos, Xia Hu, Mac Schwager, Weinan Zhang, Qian Feng, and Shangding Gu.
\newblock {Who Improves the Robot? Toward Physical Recursive Self-Improvement}.
\newblock arXiv preprint arXiv:2610.18842, 2026{\natexlab{a}}.

\bibitem[Zhou et~al.(2025{\natexlab{b}})Zhou, Xu, Tie, Chen, Zhang, Chu, Zhou, and Sun]{arxiv251003827}
Xueyang Zhou, Yangming Xu, Guiyao Tie, Yongchao Chen, Guowen Zhang, Duanfeng Chu, Pan Zhou, and Lichao Sun.
\newblock {LIBERO-PRO: Towards Robust and Fair Evaluation of Vision-Language-Action Models Beyond Memorization}.
\newblock arXiv preprint arXiv:2510.03827, 2025{\natexlab{b}}.

\bibitem[Zhou et~al.(2026{\natexlab{b}})Zhou, Luo, Cao, Xiao, Peng, Li, Tian, Lian, Wang, Ma, Ivanovic, Pavone, and Ding]{arxiv261008761}
Zewei Zhou, Rachel Luo, Yulong Cao, Chaowei Xiao, Chensheng Peng, Boyi Li, Thomas Tian, Zheng Lian, Yan Wang, Jiaqi Ma, Boris Ivanovic, Marco Pavone, and Wenhao Ding.
\newblock {VeriFine: Scaling Verification for Self-Improvement in Embodied Reasoning}.
\newblock arXiv preprint arXiv:2610.08761, 2026{\natexlab{b}}.

\bibitem[Zhou et~al.(2025{\natexlab{c}})Zhou, Atreya, Lee, Walke, Mees, and Levine]{arxiv240720635}
Zhiyuan Zhou, Pranav Atreya, Abraham Lee, Homer Walke, Oier Mees, and Sergey Levine.
\newblock {Autonomous Improvement of Instruction Following Skills via Foundation Models}.
\newblock In \emph{Proceedings of the 8th Conference on Robot Learning (CoRL 2024)}, volume 270 of \emph{Proceedings of Machine Learning Research}, pages 4805--4825, 2025{\natexlab{c}}.

\bibitem[Zhou et~al.(2025{\natexlab{d}})Zhou, Atreya, Tan, Pertsch, and Levine]{arxiv250324278}
Zhiyuan Zhou, Pranav Atreya, You~Liang Tan, Karl Pertsch, and Sergey Levine.
\newblock {AutoEval: Autonomous Evaluation of Generalist Robot Manipulation Policies in the Real World}.
\newblock arXiv preprint arXiv:2503.24278, 2025{\natexlab{d}}.

\bibitem[Zhu et~al.(2025{\natexlab{a}})Zhu, Tanaka, Goldberg, and Hong]{arxiv250602507}
Alvin Zhu, Yusuke Tanaka, Andrew Goldberg, and Dennis Hong.
\newblock {AURA: Autonomous Upskilling with Retrieval-Augmented Agents}.
\newblock arXiv preprint arXiv:2506.02507, 2025{\natexlab{a}}.

\bibitem[Zhu et~al.(2025{\natexlab{b}})Zhu, Yan, Hong, Shou, Ma, and Guo]{arxiv251109515}
Fangqi Zhu, Zhengyang Yan, Zicong Hong, Quanxin Shou, Xiao Ma, and Song Guo.
\newblock {WMPO: World Model-based Policy Optimization for Vision-Language-Action Models}.
\newblock arXiv preprint arXiv:2511.09515, 2025{\natexlab{b}}.

\bibitem[Zhu et~al.(2020)Zhu, Yu, Gupta, Shah, Hartikainen, Singh, Kumar, and Levine]{zhu2020ingredients}
Henry Zhu, Justin Yu, Abhishek Gupta, Dhruv Shah, Kristian Hartikainen, Avi Singh, Vikash Kumar, and Sergey Levine.
\newblock {The Ingredients of Real-World Robotic Reinforcement Learning}.
\newblock In \emph{International Conference on Learning Representations (ICLR)}, 2020.

\bibitem[Zoph and Le(2017)]{zoph2017nas}
Barret Zoph and Quoc~V. Le.
\newblock {Neural Architecture Search with Reinforcement Learning}.
\newblock In \emph{International Conference on Learning Representations}, 2017.

\end{thebibliography}
